\RequirePackage{fix-cm}
\documentclass[twocolumn]{svjour3}          
\smartqed  
\usepackage{graphicx}
\usepackage{mathptmx}      
\usepackage{latexsym}
\usepackage[T2A,T3,T1]{fontenc}
\usepackage[utf8]{inputenc}
\usepackage{newtxtext,newtxmath}
\usepackage[russian,english]{babel}
\usepackage{float}
\usepackage[noenc]{tipa}
\usepackage{tipx}
\usepackage{hyperref}
\usepackage{makecell}
\usepackage[table]{xcolor}
\usepackage{minibox}
\hypersetup{
	colorlinks=true,
	linkcolor=blue,
	filecolor=blue,      
	urlcolor=blue,
}

\begin{document}

\title{FST Morphological Analyser and Generator for \emph{Mapudüngun} 
}

\author{Andrés Chandía\\
	Universitat Pompeu Fabra\\
	Translation and Language Sciences Faculty\\
	\email{\href{mailto:andres@chandia.net}{andres@chandia.net} / \href{andres.chandia@upf.edu}{andres.chandia@upf.edu}}
}

\authorrunning{Andrés Chandía \and \href{mailto:andres@chandia.net}{andres@chandia.net}} 

\institute{Andrés Chandía
	\at Universitat Pompeu Fabra, Translation and Language Sciences Faculty,  \email{\href{mailto:andres@chandia.net}{andres@chandia.net} / \href{andres.chandia@upf.edu}{andres.chandia@upf.edu}}  
}

\date{Received: date / Accepted: date}

\maketitle

\begin{abstract}
Following the \emph{Mapuche} grammar by Smeets, this article describes the main morphophonological aspects of \emph{Mapudüngun}, explaining what triggers them and the contexts where they arise. We present a computational approach producing a finite state morphological analyser (and generator) capable of classifying and appropriately tagging all the components (roots and suffixes) that interact in a \emph{Mapuche} word form. The bulk of the article focuses on presenting details about the morphology of \emph{Mapudüngun} verb and its formalisation using FOMA. A system evaluation process and its results are also present in this article.

\end{abstract}

\section{Introduction} \label{intro}
This article explains the morphophonological aspects of \emph{Mapudüngun} which have to be taken into account when developing a rule based morphological analyser, which is our purpose.

The implementation we have chosen is by means of Finite State Transducers (FST). The language that feed these machines (the FST) is made out of complex regular expressions which need to encode from the language of interest, \emph{Mapudüngun} in this case, the way the different elements (roots and suffixes) interact, the conditions they have to fulfil in doing so, and the changes this very same interaction produces among the elements.

So, this work tells how we have translated \emph{Mapudüngun}’s morphophonological behaviour into regular expressions, which have to be as accurate as possible in order to obtain optimal results by means of an FST analyser\footnote{The system user interface is is available on\\ \href{http://www.chandia.net/dungupeyem}{http://www.chandia.net/dungupeyem} and the code is on\\ \href{http://www.chandia.net/dungupeyem/repositorio}{http://www.chandia.net/dungupeyem/repositorio}}.

Along section \ref{sec:01} and \ref{sec:04} we present the \emph{Mapuche} language, its typology and morphology are the central topics, which include the suffixes of this language and how the verbs are formed, from the stem to the final form, along with some exceptions and particularities.

Section \ref{sec:23}, p. \pageref{sec:23}, is centred in the computational technology we use and the specific tool to achieve our goal. We begin by explaining what computational morphology implies, and how it can be handled by Finite State Transducers (FST). FOMA is the FST compiling program we use to generate our tools, so we do a review of it. And finally, we refer to the first steps in the incorporation of the \emph{Mapudüngun} elements into the computational flow of work.

Section \ref{sec:38}, p. \pageref{sec:38} presents the embodiment of the processes and phenomena explained in the section describing \emph{Mapudüngun} into the code the compiler is capable of interpret and process. The techniques applied in order to encode the different parts of \emph{Mapudüngun} are also described, and the rules that manage their interaction and changes derived from it; and how the different aspects of \emph{Mapudüngun} morphology are treated from the computational point of view. We will explain in detail the stem typology, and the strategies to manage them; the interaction of suffixes after the stem, verb paradigms and verb nominalization. The mobility of some suffixes and the special behaviour of some verb roots are also presented in this section.

In section \ref{sec:53}, p. \pageref{sec:53}, we do account of some \emph{Mapudüngun} realizations that come from other sources and dialects, different from Smeets' work which is the base of our development. We explain how and why we have incorporated them into our system.

A brief count on the FST analyser comes in section \ref{sec:64}, p. \pageref{sec:64}, where we display data on the amount of lexicon, suffixes, and rules, besides the compilation values.

Section \ref{sec:65}, p. \pageref{sec:65} brings up the subject of assessment. We explain how our machine has been evaluated, the parameters taken into account. We introduce some other machines to compare to, and also another similar system; all for the sake of an accurate comparison and subsequent evaluation of the outcome the system produces. In this section there is also a comprehensive analysis of the forms that were not recognized by the system, and the reasons for that.

The final section \ref{sec:80}, p. \pageref{sec:80} (before the conclusions, p. \pageref{sec:80}), is just a brief account of the web interfaces we have developed to access our tools. We simply show the elements found on these interfaces and how to operate them.

\section{\emph{Mapudüngun}, the \emph{Mapuche} language} \label{sec:01}

\paragraph{} \label{tp:1} Along this section we present \emph{Mapudüngun}, its location, typology and a basic description of its conformation, which includes the phonemes it presents and their graphic representation. Morphology comes next, where we present some information about \emph{Mapudüngun} suffixes and how the verbs are formed. Finally, we introduce the stems formation and its particularities. Mainly, we present the morphophonological aspect of some specific phenomena of \emph{Mapudüngun}. A complete description of the language is found in the book we base our analyser upon: "A Grammar of \emph{Mapuche}" by Ineke Smeets \cite{RefB:21}.

\emph{Mapudüngun} is an isolated language\footnote{The relationship between \emph{Mapudüngun} and others aboriginal American languages has not yet been established.} spoken actively by approximately 144,000 people in Chile [Zúñiga 2006] \cite{RefB:24}, as well as by some 8,400 people in Argentina [Instituto Nacional de Estadísticas y Censos 2005], virtually all of whom are bilingual in Spanish [Sadowsky, S. 2013: 87-96] \cite{RefB:18}.

The word \emph{Mapudüngun} is a compound of two nominal roots, \emph{mapu} meaning 'soil, land, earth, ground, country'; \emph{düngu} meaning 'language, matter, subject, tongue (as in mother tongue)'. \emph{Mapudüngun} is usually translated as 'the language of the land, the speaking of earth' or 'lengua/habla de la tierra' in Spanish.

\subsection{\emph{Mapudüngun}: polysynthetic and agglutinative} \label{sec:02}

\paragraph{} \label{tp:2} Polysynthesis means that there are many elements or morphemes in (verb) forms, which is typical of the Native American languages, \emph{Mapudüngun} among them.

In agglutinative languages a series of concepts are distributed in several morphemes [Zúñiga 2006: 199] \cite{RefB:24}. Agglutination is when morphemes are inside words, not altering their own form and being identifiable in different contexts, as in Basque, Turkish, Quechua and \emph{Mapudüngun}. The original meaning of the stem is modified by the affixes attached to it.

In \emph{Mapudüngun}, verbs may contain many morphemes, e.g., \emph{di-tu-l-me-tu-a-fi-ñ} 'I will reach it, I will find it'. This word has eight significant elements.

"The fact that the language is polysynthetic means that it is rich, in terms of the ability to create new words " [Zúñiga 2006: 202] \cite{RefB:24}.

In \emph{Mapudüngun}, nominal forms are simple, while verbal ones are extremely complex, presenting a good number of derivative and inflectional morphemes. They can realize as univalent, only one actant as subject; bivalent or mono-transitive verbs, two actants, subject and object; and trivalent or bi-transitive verbs, three actants, subject, primary object and secondary object with the semantic roles of agent (A), human receiver (R), and inanimate patient or theme (T), respectively.

"In verbal phrases there are morphemes that behave as verbal derivatives (verbalizer, causative, transitivizer, benefactive/malefactive, modal, locative and directional, manner; and affixes that are part of non-finite verb forms), also obligatory verbal inflectional suffixes (time, mode, person and number) and facultative (negation, aspect, passive, reflexive/reciprocal/medial and mediative)" [Fernández-Garay \& Malvestitti 2002: 36-37] \cite{RefB:07}.

\subsection{The \emph{Mapuche} alphabet} \label{sec:03}

\paragraph{} \label{tp:3} Smeets states that 19 consonants [table \ref{tab:01}] and 6 vowels [table \ref{tab:02}] form the \emph{Mapuche} phonemic system.

\begin{table}[htb]
	\caption{Consonants [Smeets, I. 2008: 23] \cite{RefB:21}}
	\label{tab:01}
	\begin{tabular}{|c|c|c|c|c|c|c|}
		\hline\noalign{\smallskip}
		& Labial & \makecell{Interdental\\alveolar} & Palatal & Retroflex & Velar \\
		\noalign{\smallskip}\hline\noalign{\smallskip}
		Plosives & p & t & ch & tr & k \\
		\noalign{\smallskip}\hline\noalign{\smallskip}
		Fricatives & f & \textcrd~ ~~~|~~~ s & sh &  & \\
		\noalign{\smallskip}\hline\noalign{\smallskip}
		Glides & w &  & y & r & q \\
		\noalign{\smallskip}\hline\noalign{\smallskip}
		Nasals & m & n & ñ &  & ng \\
		\noalign{\smallskip}\hline\noalign{\smallskip}
		Laterals &  & l & ll &  & \\
		\noalign{\smallskip}\hline
	\end{tabular}
\end{table}

\begin{table}[htb]
	\caption{Vowels [Smeets, I. 2008: 25] \cite{RefB:21}}
	\label{tab:02}
	\begin{tabular}{|c|c|c|c|}
		\hline\noalign{\smallskip}
		& Front & Central & Back\\
		\noalign{\smallskip}\hline\noalign{\smallskip}
		High & i & ü & u\\
		\noalign{\smallskip}\hline\noalign{\smallskip}
		Mid & e &  & o\\
		\noalign{\smallskip}\hline\noalign{\smallskip}
		Low & & a &\\
		\noalign{\smallskip}\hline
	\end{tabular}
\end{table}

Smeets also adds loaned sounds from Spanish b, d, g (as in Spanish '\textbf{b}ote', '\textbf{d}uen\textbf{d}e' and '\textbf{gu}erra' respectively) and the voiceless fricative x (as in Spanish "\textbf{j}efe"). She does not include the interdental series present in some \emph{Mapudüngun} variants, usually represented as \emph{l'}, \emph{n'}, \emph{t'}; because the dialect she studied did not present it, and her data, "in agreement with Croese's findings, do not call for a distinction between the interdental" and the alveolar series \emph{l}, \emph{n}, \emph{t}. "A tentative conclusion might be that the distinction is dying out" [Smeets, I. 2008: 31] \cite{RefB:21}.

\section{\emph{Mapudüngun} morphology} \label{sec:04}

\paragraph{} \label{tp:4} In this section, and along this article, we mainly refer to the verb morphology because it is the more complex part of \emph{Mapudüngun}, and virtually all morphophonological changes are found inside the verb form. Other categories of words are mentioned because they occur as verbal stems together with a verbalizing suffix. In other cases they are used to bringing up a special case, or because a specific suffix also interacts with nouns, adjectives, adverbs, etc. in a nonverbal form. This does not mean that our work only covers the \emph{Mapuche} verb; but for making this article not too extensive we do not expand the topic to all parts of speech (categories).

\subsection{Verb suffixes} \label{sec:05}

\paragraph{} \label{tp:5} We begin by exposing suffixes because they occur in almost all verb stems, the only stems where a suffix do not occur are those formed by a single verbal root. But in \emph{Mapudüngun}, adjective, adverb, noun and other roots need a suffix to become verbal stems.

\begin{quote} \label{note:01}
    {\small To simplify, we call verb stem to any form, simple or complex, to which suffixes are bonded in order to form a complete verb predication. A simple stem is made of one root only, a complex stem may imply two roots, a root with some suffixes, or a combination of them all. It may be argued that these are lemmas instead, but as we say, to keep it simple, we call all these forms, verb stems. More details are found in sections \ref{sec:13} \nameref{sec:13}, p. \pageref{sec:13}; \ref{sec:16} \nameref{sec:16}, p. \pageref{sec:16} and \ref{sec:45} \nameref{sec:45}, p. \pageref{sec:45}.}
\end{quote}

After the stem, in a \emph{Mapuche} verb form, suffixes "occur in a more or less fixed position relative to one another" [Smeets, I. 2008: 17] \cite{RefB:21}. But also there are quite a few incidental factors that shape the \emph{Mapuche} complex verb form.

Verb suffixes are located on one of the thirty-six slots assigned to the verb form on the basis of their relative position and function. Slot 1 occupies word final position and slot 36 is next to the root. The order of these slots determine the morphotactics of the verb forms. Some slots host a few mutually excluding affixes, some of them may present variation in their form and some others may be zero markers. Some suffixes may exclude others from different slots for grammatical or semantic reasons.

Even though it is not rare to find up to seven or eight suffixes following the root (see E\ref{ex:1}), verbs usually contain between four and six suffixes in spontaneous speech.

In the following lines we try to graphically represent three different \emph{Mapuche} verb forms. \textbf{S} represents the stem. Every dot represents a slot; the leftmost dot is slot 36, the rightmost dot is slot 1. \textbf{X} is a suffix occurrence in a slot. \textbf{Ø} is also a suffix occurrence but with a null morpheme, which is a morpheme that has no phonemic or graphic realization.
\\

Minimal intransitive verb 2\textsuperscript{nd} person plural \\
S . . . . . . . . . . . . . . . . . . . . . . . . . . . . . . X X X . . .  \\

Minimal transitive verb 2\textsuperscript{nd} → 1\textsuperscript{st} persons plural\\
S . . . . . . . . . . . X . . . . . . . . . . . . . . . . . . X Ø X . . .  \\

Representation of example E\ref{ex:1}\\
S . . . . X . . . . . X . . . . . . . X X X . . . . . . . . . X . X Ø X X  \\

\begin{example} \label{ex:1}\ Verb with 10 suffixes [Smeets, I. 2008: 443 (76)] \cite{RefB:21}\\
	\emph{nü-nie-ñma-r-pu-tu-e-y-iñ-mu}\\
	'they continued to take it away from us'\\
	Root: \emph{nü-} \texttt{-TV.nü\_tomar}
	\begin{enumerate}
		\item[] Suffixes:
		\item \emph{-nie-} Progressive persistent (\texttt{+PRPS.nie32})
		\item \emph{-ñma-} Indirect object (\texttt{+IO.ñma26})
		\item \emph{-r-} Interruptive (\texttt{+ITR.r18})
		\item \emph{-pu-} Locative (\texttt{+LOC.pu17})
		\item \emph{-tu-} Iterative/restorative (\texttt{+RE.tu16})
		\item \emph{-e-} Internal direct object (\texttt{+IDO.e6})
		\item \emph{-y-} Indicative (\texttt{+IND.y4})
		\item \emph{-Ø-} First person (\texttt{+1.Ø3})
		\item \emph{-iñ-} Plural (\texttt{+PL.iñ2})
		\item \emph{-mu} Dative subject(\texttt{+DS3A.mew1})
	\end{enumerate}
\end{example}

\begin{quote} \label{note:02}
	{\small In example E\ref{ex:1} the root and suffixes are displayed as items to better identify them, but the analyser output is visualized linearly, as follows:\newline  \texttt{-TV.nü\_tomar+PRPS.nie32+IO.ñma26+ITR.r18+LOC.pu17 +RE.tu16+IDO.e6+IND.y4+1.Ø3+PL.iñ2+DS3A.mew1}\newline Analysis tags express, starting from the left, the abbreviated name of the part of speech (PoS) or suffix.\newline PoS are introduced by a \textbf{-} (minus) sign, suffixes, by a \textbf{+} (plus) sign. \texttt{-TV} is 'transitive verb', \texttt{-IV} is 'intransitive verb', \texttt{-N} is 'noun', etc.\newline Concerning suffixes, \texttt{+PRPS} is 'progressive persistent', \texttt{+IDO} is 'internal direct object', \texttt{+PL} is 'plural', etc. A complete list of tags meaning is found in annex \ref{anx:01} \nameref{anx:01}, p. \pageref{anx:01}.\newline After the abbreviated name of the PoS or suffix, separated by a dot, it is the root or suffix standard form. Roots are followed by their meaning in Spanish with an underscore \_ as separator: \texttt{.nü\_tomar} in E\ref{ex:1}. For the already mentioned suffixes, the forms are \texttt{.nie}, \texttt{.e} and \texttt{.iñ}, respectively.\newline The number at the end of each tag indicates the slot (the position in the verb chain) where the verb suffix is located.\newline For instance:\newline \texttt{-TV.nü\_tomar}: "the transitive verb root \emph{nü} which means 'tomar' in Spanish ('take')"\newline \texttt{+PRPS.nie32}:  "the progressive persistent suffix, which form is \emph{nie}, is located in slot 32".\newline \texttt{+IDO.e6}: "the internal direct object suffix, which form is \emph{e}, is located in slot 6, etc.\newline These items will be the result from the analyser we have built, and similar structure and information will be shown in all the examples.}
\end{quote}

Slots 1 to 15 hold inflectional suffixes in fixed positions. Slots 16 to 27 hold derivational suffixes, some of which are mobile. Slots 28 to 36 hold derivational suffixes in fixed positions, except for the rather mobile suffix \emph{‑uw-}, which usually fits in slot 31 and marks reflexivity/reciprocity. Mobile suffixes are assigned to their most usual position. "A difference in order of the suffixes does not always result in a semantic difference" [Smeets, I. 2008: 177] \cite{RefB:21} (see sections \ref{sec:10}, p \pageref{sec:10} and \ref{sec:50}, p. \pageref{sec:50}).

\subsubsection{Verbalizers (slot 36)} \label{sec:06}

\paragraph{} \label{tp:6} Nouns, adjectives, adverbs and numerals (roots) "can be changed into verbs by means of suffixation" [Smeets, I. 2008: 304] \cite{RefB:21}. "There are six verbalizing suffixes. They immediately follow the root and fill slot 36" [Smeets, I. 2008: 121] \cite{RefB:21}.

\begin{enumerate} \label{it:01}
	\item Suffix \emph{-Ø-} indicates the verbalization of a noun, adjective, numeral and a number of adverb roots.
	\item Suffix \emph{-l-} verbalizes noun, adverb, numeral roots and the interrogative pronoun \emph{tunte-} 'how much'.
	\item Suffix \emph{-nge-} can verbalize noun, adjective, numeral roots and the interrogative element \emph{chum-} 'how'. A verb formed with \emph{-nge-} is intransitive.
	\item Suffix \emph{-ntu-} verbalizes adjective roots.
	\item Suffix \emph{-tu-} verbalizes noun roots.
	\item Suffix \emph{-ye-} verbalizes noun roots.
\end{enumerate}

\begin{example} \label{ex:2} \\
	\emph{mapu-che} 'person of the land (\emph{Mapuche} person)'\\
	\texttt{-NN.mapu\_tierra-NN.che\_persona} \medskip \\
	\emph{mapu-che-nge-n} 'I am a person of the land (a \emph{Mapuche})'\\
	\texttt{-NN.mapu\_tierra-NN.che\_persona\\ +VRB.nge36-IV+IND1SG.n3}\\
\end{example}

\subsubsection{Stem formative (slot 36)} \label{sec:07}

\paragraph{} \label{tp:7} Reduplication is another resource in \emph{Mapudüngun}, and reduplicated roots are also used to form verbs, but for doing so, they are obligatorily followed by a verbalizing suffix, even when it is a reduplicated verb root, in this case Smeets calls these suffixes "stem formative in reduplicated roots (SFR)" and they are also assigned in slot 36. There are four stem formatives:

\begin{enumerate} \label{it:02}
	\item Suffix \emph{-Ø-} occurs when the reduplicated root is an onomatopoeia or a verb. The resulting verb is intransitive.
	\item Suffix \emph{-nge-} is added to reduplicated verb roots, the resulting verb is intransitive.
	\item Suffix \emph{-tu-} is added to reduplicated noun or verb roots. The resulting verb of a reduplicated verb root has the same valence as the single form.
	\item Suffix \emph{-ye-} is added to reduplicated verb roots, the resulting verb is transitive.
\end{enumerate}

\begin{example} \label{ex:3} [Smeets, I. 2008: 481 (33)] \cite{RefB:21} \\
	\emph{ñiwa-ñiwa-tu-fu-n} 'I always did my best'\\
	\texttt{-IV.ñiwa\_esforzar-RVBR+SFR.tu36+IPD.fu8\\ +IND1SG.n3}
\end{example}

\subsubsection{Derivational suffixes (slots 16 to 35)} \label{sec:08}

\paragraph{} \label{tp:8} From slot 16 to 27 the suffixes mostly act as semantic modifiers. From slot 28 to 35, they have an aspectual or valency function. Suffixes most commonly used are:

\begin{itemize} \label{it:03}
	\item[] Causatives \emph{-l-} (e.r.\footnote{e.r. stands for "examples references". In the referred section there is a list of the examples that contain the suffix being mentioned. The entire list is under annex \ref{anx:08} \nameref{anx:08}, p. \pageref{anx:08}.} \ref{tp:133}) and \emph{-m-} (e.r. \ref{tp:134}), slot 34. These suffixes make the event denoted by the stem to be actually applied or happen, in this sense, they operate as transitivizers also. \medskip
	\item[] Factitive \emph{-ka-} (e.r. \ref{tp:135}) and transitivizer \emph{-tu-} (e.r. \ref{tp:136}), slot 33. It indicates that the agent causes the event denoted by the verb to take place, often it also adds intensive value. \medskip
	\item[] Reflexive/reciprocal \emph{-w-} (e.r. \ref{tp:139}), slot 31. It indicates reflexivity when combined with a singular subject. The reflexive morpheme -(u)w- indicates reflexivity or reciprocity when it combines with a dual or plural subject. \medskip
	\item[] Stative \emph{-le-} (e.r. \ref{tp:142}), slot 28. It denotes a state which may or may not involve agentivity on the part of the subject. With a few verbs, it may denote either an ongoing event or the resulting state. It may be used to indicate a quality or characteriztic that is not permanent or intrinsic. \medskip
	\item[] Beneficiary \emph{-el-} (e.r. \ref{tp:143}), slot 27. It makes the (animate) patient become the beneficiary of the event. \medskip
	\item[] Passive \emph{-nge-} (e.r. \ref{tp:147}), slot 23. It indicates that a participant, a 3\textsuperscript{rd} person with the role of agent, is not found in the situation described by the sentence, but outside the speech act. \medskip
	\item[] 1\textsuperscript{st} person agent \emph{-w-} (e.r. \ref{tp:148}), slot 23. It indicates a non-declared participant to be determined by the context. Which is a 1\textsuperscript{st} person non-singular, the agent, and implicitly includes the listener who is the patient. \medskip
	\item[] Thither \emph{-me-} (e.r. \ref{tp:153}), slot 20. It indicates that the denoted situation involves motion away from the speaker or another orientation point, with a connotation of temporariness. \medskip
	\item[] Persistence \emph{-we-} (e.r. \ref{tp:154}), slot 19. It indicates a situation which persists after a previous event has taken place. \medskip
	\item[] Hither \emph{-pa-} (e.r. \ref{tp:157}), slot 17. It indicates that the denoted situation either involves a movement towards the speaker or takes place at a location near the speaker. It may indicate a development towards the present. \medskip
	\item[] Locative \emph{-pu-} (e.r. \ref{tp:158}), slot 17. It indicates that the event takes place away from the speaker. It does not imply motion and indicates a permanent situation.
\end{itemize}

\subsubsection{Inflectional suffixes (slots 5 to 15)} \label{sec:09}

\paragraph{} \label{tp:9} Among these suffixes are those that indicate aspect, tense, negation and truth value:

\begin{itemize} \label{it:04}
	\item[] Pluperfect \emph{-wye-}, slot 15. Indicates that the event takes place before the past or future orientation moment (see following example and E\ref{ex:5}).
	\begin{example} \label{ex:4} [Smeets, I. 2008: 69 (62)] \cite{RefB:21} \\
		\emph{tripa-\textbf{wye}-y} 'he had left'\\
		\texttt{-IV.tripa\_salir+PLPF.wye15+IND.y4+3.Ø3}
	\end{example} \medskip
	\item[] Constant feature \emph{-ke-} (e.r. \ref{tp:162}), slot 14. Indicates a constant or characteriztic feature of the subject. \medskip
	\item[] Proximity \emph{-pe-} (e.r. \ref{tp:163}), slot 13. It seems to indicate an event or a feature in the recent past, a strong probability and doubt. \medskip
	\item[] Reportative \emph{-rke-}, slot 12. It indicates that the situation has not been directly witnessed; the speaker has been informed by others, has heard rumours, or has deduced it (see following example). 
	\begin{example} \label{ex:5} [Smeets, I. 2008: 254 (1)] \cite{RefB:21} \\
		\emph{füta-nge-wye-\textbf{rke}-y} 'she had been married, they say'\\
		\texttt{-NN.füta\_marido+VRB.nge36-IV+PLPF.wye15\\+REP.rke12+IND.y4+3.Ø3}
	\end{example} \medskip
	\item[] Affirmative \emph{-lle-}, slot 11. It adds emphasis (E\ref{ex:59}). \medskip
	\item[] Non-realized situation \emph{-a-} (e.r. \ref{tp:169}), slot 9. It denotes a non-actual fact. The situation will take place after the orientation moment. \medskip
	\item[] Impeditive \emph{-fu-} (e.r. \ref{tp:170}), slot 8. It denotes that the event does not conclude as expected or that it can not be completed. \medskip
	\item[] Pluperfect \emph{-mu-}(e.r. \ref{tp:171}), slot 7. It indicates that an event is realized before an orientation moment in the past. It occurs in complementary distribution with the pluperfect \emph{-wye-}, slot 15. \medskip
	\item[] Constant feature \emph{-ye-}, slot 5. As suffix \emph{-ke-}, slot 14, it also denotes a characteriztic or constant feature, and they appear in complementary distribution (E\ref{ex:162}).
\end{itemize}

\subsubsection{Suffix mobility} \label{sec:10}

\paragraph{} \label{tp:10} Smeets identify suffixes from slots 28 to 36 as fixed suffixes, and from slots 16 to 27 as mobile, later ones appear in non-common positions respect to other suffixes. A detailed list of the mobile suffixes with their usual position (slot) follows:

\begin{itemize} \label{it:05}
	\item[] Repetitive/Restorative \emph{-tu-} (e.r. \ref{tp:160}), slot 16. It indicates that a situation is repeated or restored. \medskip
	\item[] Hither \emph{-pa-} (e.r. \ref{tp:157}), slot 17. (Explained in \ref{it:03}). \medskip
	\item[] Persistence \emph{-we-} (e.r. \ref{tp:154}), slot 19. (Explained in \ref{it:03}). \medskip
	\item[] Thither \emph{-me-} (e.r. \ref{tp:153}), slot 20. (Explained in \ref{it:03}). \medskip
	\item[] Immediate \emph{-fem-}, slot 21. It denotes immediate action (see following example). \begin{example} \label{ex:6} [Smeets, I. 2008: 271 (20)] \cite{RefB:21} \\
		\emph{ye-nge-\textbf{fem}-üy} 'it was brought immediately'\\
		\texttt{-TV.ye\_traer+PASS.nge23+IMM.fem21\\+IND.y4+3.Ø3}
	\end{example} \medskip
	\item[] Sudden \emph{-rume-}, slot 21. (E\ref{ex:162}).	It denotes sudden action. \medskip
	\item[] Play \emph{-kantu-}, slot 22 (E\ref{ex:200}). It denotes an action performed in jest, for fun, not in earnest, or just to pretend to be doing. \medskip
	\item[] Simulative \emph{-faluw-}, slot 22. It indicates simulation, not real intention to do something (see following example). \begin{example} \label{ex:7} [Smeets, I. 2008: 265 (9)] \cite{RefB:21} \\
		\emph{illku-le-\textbf{faluw}-ün} 'I pretended to be angry'\\
		\texttt{--IV.illku\_enojar+ST.le28+SIM.faluw22\\+IND1SG.n3}
	\end{example} \medskip
	\item[] Passive \emph{-nge-} (e.r. \ref{tp:146}), slot 23. (Explained in \ref{it:03}). \medskip
	\item[] Pluralizer \emph{-ye-}, slot 24 (E\ref{ex:91}). It is especially used with intransitive verbs which take a 3\textsuperscript{rd} person subject. With a 1\textsuperscript{st} or 2\textsuperscript{nd} person plural subject, it indicates a numerous subject. With transitive verbs, it indicates that numerous patients of the event.\medskip
	\item[] Force \emph{-fal-}, slot 25 (E\ref{ex:155}). It indicates either that there is a necessity or obligation for the subject to perform the action, or that the subject orders someone else to perform the action. \medskip
	\item[] Beneficiary \emph{-el-} (e.r. \ref{tp:143}), slot 27. (Explained in \ref{it:03}). \medskip
	\item[] Stative \emph{-le-} (e.r. \ref{tp:142}), slot 28. (Explained in \ref{it:03}). \medskip
	\item[] Reflexive/Reciprocal \emph{-w-} (e.r. \ref{tp:139}), slot 31. (Explained in \ref{it:03}). \medskip
	\item[] Transitivizer \emph{-tu-} (e.r. \ref{tp:136}), slot 33. It may be added to intransitive and transitive verbs, and it adds an object. With intransitive verbs, the form has one object. With transitive verbs, the form has two objects.
\end{itemize}

Mobility does not imply a semantic change, and as more suffixes a verb presents less displacement occurs. See the following examples:

\begin{example} \label{ex:8} [Smeets, I. 2008: 270 (19)] \cite{RefB:21}\\
	\emph{ngilla-l-\textbf{me}-mu-y-iñ} 'you went to buy for us'\\
	\texttt{‑TV.ngilla\_comprar+BEN.el27+TH.me20+2A.mu23\\+IND.y4+1.Ø3+PL.iñ2}
\end{example}

\begin{example} \label{ex:9} [Smeets, I. 2008: 263 (11)] \cite{RefB:21}\\
	\emph{i-\textbf{me}-we-ke-la-y} 'he no longer always eats there'\\ \texttt{‑TV.i\_comer+TH.me20+PS.we19+CF.ke14+NEG.la10\\+IND.y4+3.Ø3}
\end{example}

\begin{example} \label{ex:10} [Smeets, I. 2008: 421 (62)] \cite{RefB:21}\\
	\emph{pütu-yekü-\textbf{me}-tu-y-ng-ün} 'they drank all the time'\\ \texttt{‑TV.püto\_beber+ITR.yekü18+TH.me20+RE.tu16\\+IND.y4+3.ng3+PL.ün2}
\end{example}

In the examples above, the thither suffix \emph{-me-} presents three different positions respect to the other suffixes. In E\ref{ex:8} it is close to slot 27, displaced beyond slot 23. In E\ref{ex:9} it occurs in its usual position, just before the suffix \emph{-we-}, slot 19. Finally, in E\ref{ex:10}, it appears between suffixes of slots 18 and 16, to the right of its usual position.

\subsubsection{Verb paradigms} \label{sec:11}

\paragraph{} \label{tp:11} In the previous section we have skipped suffixes from slots 10 and 6, they take part in the transitive verb paradigm, we include them here. Suffixes of slot 23 are also included in this paradigm together with those of mood, person, number and dative subject of slots 4, 3, 2 and 1 respectively.

Negation, positioned in slot 10, may actually be part of transitive and intransitive forms. There are three negation morphemes, one per mood, reason to show them in the verb paradigms.

The simplest verb form is intransitive, less suffixes than in transitive forms are mandatory: mood (slot 4), person (slot 3) and number (slot 2). See examples below:

\begin{example} \label{ex:11}\ \\
	\emph{küpa-y-m-i} 'you (sg) came'\\
	\texttt{‑IV.küpa\_venir+IND.y4+2.m3+SG.i2}
\end{example}

\begin{example} \label{ex:12}\ \\
	\emph{küpa-la-y-m-u} 'you two did not come'\\
	\texttt{‑IV.küpa\_venir+NEG.la10+IND.y4+2.m3+DL.u2}
\end{example}

\begin{example} \label{ex:13}\ \\
	\emph{küpa-no-l-m-ün} 'if you (pl) do not come'\\
	\texttt{‑IV.küpa\_venir+NEG.no10+CND.l4+2.m3+PL.ün2}
\end{example}

The imperative mood have forms for 1\textsuperscript{st} person singular; 2\textsuperscript{nd} person singular, dual and plural; and for 3\textsuperscript{rd} person (undefined number). Indicative forms of 1\textsuperscript{st} person dual and plural may be used adhortatively. Negation suffix for imperative is \emph{-ki-} (slot 10), which always co-occur with the conditional marker \emph{-l-} (slot 4). Negation of the adhortative forms, which are indicative, is accomplished by the \emph{-ki-l-} combination of imperative negation and conditional mood mark when the intention is imperative (adhortative). See examples below (the complete conjugation of the intransitive verb \emph{küpa-} 'to come' is in annex \ref{anx:10}, table \ref{tab:13}):

\begin{example} \label{ex:14}\ \\
	\emph{küpa-m-u} 'come, you both!'\\
	\texttt{‑IV.küpa\_venir+IMP.Ø4+2.m3+DL.u2}
\end{example}

\begin{example} \label{ex:15}\ \\
	\emph{küpa-ki-l-chi} 'I better not come'\\
	\texttt{‑IV.küpa\_venir+NEG.ki10+CNI\footnote{Even though the form is the same, we have labelled it \texttt{+CND} 'conditional' and \texttt{+CNI} 'conditional marker in imperative forms, to better distinguish them'.}.l4+IMP1SG.chi3}
\end{example}

\begin{example} \label{ex:16}\ \\
	\emph{küpa-y-u} ind: 'we both came' imp: 'let we both come'\\
	\texttt{‑IV.küpa\_venir+IND.y4+1.Ø3+DL.u2}
\end{example}

\begin{example} \label{ex:17}\ \\
	\emph{küpa-ki-l-y-u} 'let we both not come'\\
	\texttt{‑IV.küpa\_venir+NEG.ki10+CNI.l4+1.y3+DL.u2}
\end{example}

The transitive paradigm demands more suffixes to reflect the relations between agent, patient and object. Interacting also the suffixes \emph{-w-} 1\textsuperscript{st} person agent, and \emph{-mu-} 2\textsuperscript{nd} person agent, slot 23; \emph{-e-} internal direct object and \emph{-fi-} external direct object, slot 6; and \emph{-Ø-} dative subject, 1\textsuperscript{st} or 2\textsuperscript{nd} person agent, and \emph{-mew-} $\sim$ \emph{-ew-} dative subject, 3\textsuperscript{rd} person agent (no number), slot 1. A complete explanation of the transitive paradigm is in chapter 26 "Slots" of Smeets 2008 \cite{RefB:21}. See examples below (the complete conjugation of the transitive verb \emph{pi-} 'to say (to tell)' is in annex \ref{anx:11}, table \ref{tab:14} and the negative imperative forms in annex \ref{anx:12}, table \ref{tab:15}):

\begin{example} \label{ex:18}\ \\
	\emph{pi-e-y-u} 'I told you'\\
	\texttt{‑TV.pi+IDO.e6+IND.y4+1.Ø3+DL.u2+DS12A.Ø1}
\end{example}

\begin{example} \label{ex:19}\ \\
	\emph{pi-mu-l-i} 'if you (non-sg) tell me'\\
	\texttt{‑TV.pi+2A.mu23+CND.l4+1.i3+SG.Ø2}
\end{example}

\begin{example} \label{ex:20}\ \\
	\emph{pi-fi-m-u} 'tell him, you both'\\
	\texttt{‑TV.pi+EDO.fi6+IMP.Ø4+2.m3+DL.u2}
\end{example}

\begin{example} \label{ex:21}\ \\
	\emph{pi-ki-fi-l-y-iñ} 'let us (pl) not tell him'\\
	\texttt{‑TV.pi+NEG.ki10+EDO.fi6+CNI.l4+1.y3+PL.iñ2}
\end{example}

\subsubsection{Verb inflectional nominalization} \label{sec:12}

\paragraph{} \label{tp:12} In a \emph{Mapuche} sentence, subordinates are derived from verbs, nominalized by inflectional nominalizers. These suffixes share position in slot 4 with mood markers, therefore, a verb form is either finite or nominalized. Finite forms take mood, person and number. Nominalized forms can not take those suffixes, taking instead one of the inflectional nominalizers.

Besides as subordinates of verbs, nominalized verbs may also act "as subject, direct object, instrumental object or complement noun phrase, indicating an event as such, a participant, an instrument, time, place, reason, purpose or background event" [Smeets, I. 2008: 188] \cite{RefB:21}; as noun modifiers, and as predicates in nominal sentences.

"Some nominalized forms can be used as a finite verb form. The subject of a subordinate is usually indicated by a possessive pronoun, which immediately precedes the subordinate. However, when a subordinate is used as a temporal or causal clause, or as a finite verb form, the subject is indicated by a personal pronoun" [Smeets, I. 2008: 189] \cite{RefB:21}. There are seven inflectional nominalizers:

\begin{itemize} \label{it:06}
	\item[] Agentive verbal noun \emph{-t-}\\
	This suffix may denote an event as such; an instrument or location, and the patient or agent of an event.
	\begin{example} \label{ex:22} [Smeets, I. 2008: 215 (186)] \cite{RefB:21}\\
		\emph{tüfa ñi pi-e-\textbf{t}-ew}\\
		\texttt{-DP.tüfa\_este -SP.ñi\_mi\_su\\-TV.pi\_decir+IDO.e6+AVN.t4+DS3A.ew1}\\
		'this is what he told me' lit: 'this his told me'
	\end{example}
	\item[] Completive subjective verbal noun \emph{-wma-}\\
	This suffix indicates the subject of a completed event.
	\begin{example} \label{ex:23} [Smeets, I. 2008: 400 (24)] \cite{RefB:21}\\
		\emph{füta-nge-\textbf{wma}-rke}\\
		\texttt{‑NN.füta\_marido+VRB.nge36‑IV+CSVN.wma4\\+REP.rke}\\
		'she has been married, some say'
	\end{example}
	\item[] Instrumental verbal noun \emph{-m}\\
	This suffix may indicate an instrument, a location, or an event as such. In combination with \emph{-a-} non-realized action (slot 9), it may indicate purpose. With \emph{-ye-} constant feature (slot 5), it forms a temporal clause.
	\begin{example} \label{ex:24} [Smeets, I. 2008: 206 (137)] \cite{RefB:21} \\
		\emph{iñchiñ ta-yiñ lleg-mu-\textbf{m}}\\
		\texttt{‑NN.-PP.iñchiñ\_nosotros\\-AP.ta\_el-SP.yiñ\_nuestro-s\\-IV.lleg\_crecer+PLPF.mu7+IVN.m4}\\
		'where we (pl) have grown up' lit: 'we the our have grown up place'
	\end{example}
	\item[] Objective verbal noun \emph{-el} $\sim$ \emph{-Ø}\\
	This suffix expresses a passive participle, indicating the patient of the event. It can also be used to indicate an event as such; and rarely it is also used as	an instrumental or locative.
	\begin{example} \label{ex:25} [Smeets, I. 2008: 76 (16)] \cite{RefB:21}\\
		\emph{kuyfi pichi-ka-\textbf{el}}\\
		\texttt{-AV.kuyfi\_antes\\-AJ.pichi\_pequeño+VRB.Ø36+CONT.ka16+OVN.el4}\\
		'long time ago when I was still young' lit: 'before, in the still little'
	\end{example}
	\item[] Plain verbal noun \emph{-n}\\
	This suffix	indicates an event as such, without time mark. It can convert the form into an adjective denoting an attribute or quality of the modified noun. It can also form a noun denoting a person or thing involved in the event referred to by the verb. It is usually translated as an infinitive: \emph{küdaw} 'the work', \emph{küdaw-ün} 'to work'.
	\begin{example} \label{ex:26} [Smeets, I. 2008: 192 (51)] \cite{RefB:21}\\
		\emph{pütrem-tu-\textbf{n} küme-la-y}\\
		\texttt{-NN.pütrem\_tabaco+VRB.tu36+PVN.n4\\-AJ.küme\_bueno+VRB.Ø36+NEG.la10+IND.y4+3.Ø3}\\
		'smoking is not good' lit: 'tobaccoing good not it is'
	\end{example}
	\item[] Subjective verbal noun \emph{-lu} $\sim$ \emph{-Ø}\\
	This suffix denotes the subject of an event. It may also be used as an active participle, and form a temporal or causal clause.
	\begin{example} \label{ex:27} [Smeets, I. 2008: 218 (203)] \cite{RefB:21}\\
		\emph{pichi che kim-nu-\textbf{lu}}\\
		\texttt{-AJ.pichi\_pequeño -NN.che\_persona\\-TV.kim\_saber+NEG.no10+SVN.lu4}\\
		'a child that does not know' lit: 'little person not knower'
	\end{example}
	\item[] Transitive verbal noun \emph{-fiel}\\
	This suffix may be used as an infinitive, passive participle, locative or instrumental.
	\begin{example} \label{ex:28} [Smeets, I. 2008: 237 (16)] \cite{RefB:21}\\
		\emph{iñche müle-y mi pe-a-\textbf{fiel}}\\
		\texttt{-PP.iñche\_yo -IV.müle\_estar+IND.y4+3.Ø3\\-SP.mi\_tuyo -TV.pe\_ver+NRLD.a9+TVN.fiel4}\\
		'I have to see you (sg)' lit: 'I am in your will be seen'
	\end{example}
\end{itemize}

\subsubsection{Verb derivational nominalization} \label{sec:13}

\paragraph{} \label{tp:13} Some non-verbal suffixes can turn a verb into an adjective or a noun; the stem may be formed by a unique root, a verbal compound, a verbalized root, a verbalized compound or a reduplicated root; or even by a complex stem, a root followed by some suffixes, mainly from slots 35, 34 or 33.

\begin{itemize} \label{it:07}
	\item[]\emph{-fal} \texttt{+ADJDO} indicates that the event denoted by the verb can actually be done.
	\begin{example} \label{ex:29} [Smeets, I. 2008: 312 (12)] \cite{RefB:21}\\
		\emph{pepi-l-\textbf{fal}} 'feasible, practicable'\\
		\texttt{-TV.pepi\_poder-hacer+CA.l34+ADJDO.fal}
	\end{example}
	\item[]\emph{-fe} \texttt{+NOMAG} denotes a characteriztic agent.
	\begin{example} \label{ex:30} [Smeets, I. 2008: 311 (1)] \cite{RefB:21}\\
		\emph{kofke-tu-\textbf{fe}} 'bread eater'\\
		\texttt{-NN.kofke\_pan+VRB.tu36+NOMAG.fe}
	\end{example}
	\item[]\emph{-nten} \texttt{+ADJQE} indicates that the event denoted by the verb may be realized quickly and/or easily.
	\begin{example} \label{ex:31} [Smeets, I. 2008: 312 (14)] \cite{RefB:21}\\
		\emph{afü-\textbf{nten}} 'it gets quickly cooked'\\
		\texttt{-IV.afü\_cocinar+ADJQE.nten}
	\end{example}
	\item[]\emph{-we} \texttt{+NOMPI} denotes a characteriztic place or instrument.
	\begin{example} \label{ex:32} [Smeets, I. 2008: 312 (9)] \cite{RefB:21}\\
		\emph{püra-püra-\textbf{we}} 'stairs'\\
		\texttt{-IV.püra\_subir-RVBR+SFR.Ø36-IV+NOMPI.we}
	\end{example}
\end{itemize}

\subsubsection{Non-verbal suffixes} \label{sec:14}

\paragraph{} \label{tp:14} In the previous section there were already presented four suffixes that can turn verbs into adjectives or nouns, these resulting forms may, in turn, be complex verb stems, i.e., a verb converted into an adjective or a noun may be used as a verb stem, it may be "re-verbalized".

There are other suffixes that act upon non-verbal forms; the final form, i.e., "non-verb + suffix", in its turn can also be a complex verb stem (see \ref{sec:16} \nameref{sec:16}, p. \pageref{sec:16}). Some of these suffixes change the class (category) of the form they are attached to, and some others do not.

\begin{itemize} \label{it:08}
	\item[]Class-changing suffixes (\texttt{CC})
	\item[]\emph{-chi} \texttt{+ADJ} changes a noun or nominalized verb into an adjective.
	\begin{example} \label{ex:33} [Smeets, I. 2008: 114 (25)] \cite{RefB:21}\\
		\emph{lef-\textbf{chi} che} 'runner' lit:'running person'\\
		\texttt{-IV.lef\_correr+SVN.Ø4+ADJ.chi\\-NN.che\_persona}
	\end{example}
	\item[]\emph{-tu} \texttt{+ADV} changes a noun or nominalized verb into an adverb.
	\begin{example} \label{ex:34} [Smeets, I. 2008: 114 (b)] \cite{RefB:21}\\
		\emph{amu-n-\textbf{tu}} 'going', 'on my way there'\\
		\texttt{-IV.amu\_ir+PVN.n4+ADV.tu}
	\end{example}
\end{itemize}

\begin{itemize} \label{it:09}
	\item[]Non class-changing suffixes (\texttt{NCC})
	\item[]\emph{-ke} \texttt{+DISTR} is affixed to adjectives, adverbs and numerals. It indicates a whole consisting of several component parts, each of which has the feature expressed by the form it accompanies.
	\begin{example} \label{ex:35} [Smeets, I. 2008: 112 (17)] \cite{RefB:21}\\
		\emph{küla-\textbf{ke}} 'a threesome'\\
		\texttt{-NU.küla\_tres+DISTR.ke}
	\end{example}
	\item[]\emph{-em} \texttt{+EX} is affixed to a noun of which indicates that is dead or no longer in function or existence.
	\begin{example} \label{ex:36} [Smeets, I. 2008: 110 (6)] \cite{RefB:21}\\
		\emph{fey-tüfa ñi küdaw-\textbf{yem}} 'this was my former job'\\
		\texttt{-DP.fey\_que-DP.tüfa\_este -SP.ñi\_mi\_su\\-NN.küdaw\_trabajo+EX.em}
	\end{example}
	\item[]\emph{-ntu} \texttt{+GR} it refers to a group as a whole or a place which is characterized by the presence of many items referred to by the noun.
	\begin{example} \label{ex:37}\ \\
		\emph{küra-\textbf{ntu}} 'scree'\\
		\texttt{-NN.küra\_piedra+GR.ntu}
	\end{example}
	\item[]\emph{-rke} \texttt{+REP} indicates that the situation or thing expressed by the form it accompanies has not been witnessed by the speaker himself. The speaker has been informed by others, he has heard rumours or he has deduced a conclusion. It may express surprise after the sudden realization of something.
	\begin{example} \label{ex:38} [Smeets, I. 2008: 110 (8)] \cite{RefB:21}\\
		\emph{trewa-\textbf{rke}!} 'a dog!', 'what a big dog!', 'it must have been a dog' (at wondering about who ate the meat that disappeared)'\\
		\texttt{-NN.trewa\_perro+REP.rke}
	\end{example}
	\item[]\emph{-we} \texttt{+TEMP} indicates a period subsequent to an orientation moment.
	\begin{example} \label{ex:39}\ \\
		\emph{kechu-\textbf{we} antü} 'in five days'\\
		\texttt{-NU.kechu\_cinco+TEMP.we -NN.antü\_sol\_día}
	\end{example}
	\item[]\emph{-wen} \texttt{+REL} refers to the people relation indicated by the noun it accompanies.
	\begin{example} \label{ex:40}\ \\
		\emph{kompañ-\textbf{wen} iñchiu} 'we are partners' lit: 'we both are partners of one another'\\
		\texttt{-NN.kompañ\_compañero+REL.wen\\-PP.iñchiu\_nosotros-dos}
	\end{example}
\end{itemize}

\subsubsection{Instrumental object suffix -mew} \label{sec:15}

\paragraph{} \label{tp:15} This suffix may never be part of a complex stem but it may be added to nominalized verbs (E\ref{ex:43}), nouns and pronouns. It indicates instrument, place, time, cause and is used in comparative and partitive constructions. It may also refer to the circumstances under which an event takes place. See next examples:

\begin{example} \label{ex:41} [Smeets, I. 2008: 62 (5)] \cite{RefB:21}\\
	\emph{anel-tu-fi-ñ kiñe kuchillo-\textbf{mew}} 'I threatened him with a knife'\\
	\texttt{-TV.anel\_amenazar+TR.tu33+EDO.fi6+IND1SG.n3\\-NU.kiñe\_uno -NN.kuchillu\_cuchillo+INST.mew}
\end{example}

\begin{example} \label{ex:42} [Smeets, I. 2008: 62 (9)] \cite{RefB:21}\\
	\emph{uma-pu-n ta-ñi peñi-\textbf{mu}} 'I stayed at my brother's'\\
	\texttt{-IV.uma\_pernoctar+LOC.pu17+IND1SG.n3\\-AP.ta\_el-SP.ñi\_mi\_su\\ -NN.peñi\_hermano+INST.mew}
\end{example}

\begin{example} \label{ex:43} [Smeets, I. 2008: 62 (7)] \cite{RefB:21}\\
	\emph{are-tu-n-\textbf{mew} monge-li-y} 'he lives on borrowing'\\
	\texttt{-TV.are\_prestar+TR.tu33+PVN.n4+INST.mew\\-IV.monge\_vivir+ST.le28+IND.y4+3.Ø3}
\end{example}

\subsection{Verb stems} \label{sec:16}

\paragraph{} \label{tp:16} We have classified different types of stems depending on the way they are composed. In this respect we do not strictly follow Smeets. Verbs stems are completed by the verbalizing suffix (\texttt{+VRB}) when there is no verb root present, or a stem formative (\texttt{+SFR}) when there is a reduplicated root.

The following list shows the different types of stems from the simplest to the most complex ones:

\begin{itemize} \label{it:10}
	\item[]Simple stems
	\item Verb root
	\begin{example} \label{ex:44} [Smeets, I. 2008: 64 (29)] \cite{RefB:21}\\
		\emph{\textbf{amu}-y-ng-ün} 'they (pl) went'\\
		\texttt{-IV.amu\_ir+IND.y4+3.ng3+PL.ün2}
	\end{example}
	\item Verb compound (verb root + verb root)
	\begin{example} \label{ex:45} [Smeets, I. 2008: 420 (54)] \cite{RefB:21}\\
		\emph{\textbf{amu-mayna}-tu-e-n-ew} 'he made me stumble'\\
		\texttt{-IV.amu\_ir-TV.mayna\_atar-los-pies-CR.TV\\+TR.tu33+IDO.e6+IND1SG.n3+DS3A.ew1}
	\end{example}
	\item Verb compound (verb root + non-verb root / non-verb root + verb root)
	\begin{example} \label{ex:46} [Smeets, I. 2008: 401 (38)] \cite{RefB:21}\\
		\emph{\textbf{ad-kintu}-a-l} 'to have a look'\\
		\texttt{-NN.ad\_forma-TV.kintu\_mirar+NRLD.a9+OVN.el4}
	\end{example}
	\item Non-verb root \texttt{+VRB} (verbalizer suffix)
	\begin{example} \label{ex:47} [Smeets, I. 2008: 68 (56)] \cite{RefB:21}\\
		\emph{küla \textbf{antü-nge}-y} 'it was three days ago'\\
		\texttt{-NU.küla\_tres\\-NN.antü\_sol\_día+VRB.nge36-IV+IND.y4+3.Ø3}
	\end{example}
	\item Reduplicated root \texttt{+SFR} (stem formative suffix)
	\begin{example} \label{ex:48} [Smeets, I. 2008: 112 (22)] \cite{RefB:21}\\
		\emph{\textbf{aku-aku-nge}-y} 'continually arrive (e.g. letters)'\\
		\texttt{-IV.aku\_llegar-RVBR+SFR.nge36-IV+IND.y4+3.Ø3}
	\end{example}
	\item Non-verb compound (non-verb + non-verb) \texttt{+VRB}
	\begin{example} \label{ex:49} [Smeets, I. 2008: 123 (11)] \cite{RefB:21}\\
		\emph{\textbf{trewa-ad-nge}-y} 'he has dog face'\\
		\texttt{-NN.trewa\_perro-NN.ad\_cara+VRB.nge36-IV\\+IND.y4+3.Ø3}
	\end{example}
\end{itemize}

\begin{itemize} \label{it:11}
	\item[]Complex single root stems \medskip
	\item Numeral + non class-changing suffix \texttt{+VRB}
	\begin{example} \label{ex:50} [Smeets, I. 2008: 400 (30)] \cite{RefB:21}\\
		\emph{\textbf{kiñe-ke-l}-fi-y} 'he gave one to each of them'\\
		\texttt{-NU.kiñe\_uno+DISTR.ke+VRB.l36\\+EDO.fi6+IND.y4+3.Ø3}
	\end{example}
	\item Adjective + non class-changing suffix or inflectional nominalizer \texttt{+VRB}
	\begin{example} \label{ex:51} [Smeets, I. 2008: 473 (40)] \cite{RefB:21}\\
		\emph{\textbf{pichi-n-tu}-ki-y} 'it was for little time'\\
		\texttt{-AJ.pichi\_pequeño+PVN.n4+VRB.tu36\\+CF.ke14+IND.y4+3.Ø3}
	\end{example}
	\item Question \texttt{+VRB} + inflectional nominalizer \texttt{+VRB}
	\begin{example} \label{ex:52} [Smeets, I. 2008: 243 (52)] \cite{RefB:21}\\
		\emph{\textbf{chum-nge-n-tu}-y-m-i-?} 'what do you (sg) think (about it)?'\\
		\texttt{-QC.chum\_cómo+VRB.nge36-IV+PVN.n4+VRB.tu36\\+IND.y4+2.m3+SG.i2}
	\end{example}
	\item Adjective + inflectional nominalizer \texttt{+VRB} + derivational nominalizer \texttt{+VRB}
	\begin{example} \label{ex:53} [Smeets, I. 2008: 375 (25)] \cite{RefB:21}\\
		\emph{\textbf{awka-n-tu-fe-nge}-y} 'he is playful'\\
		\texttt{-AJ.awka\_salvaje+PVN.n4+VRB.tu36+NOMAG.fe\\+VRB.nge36-IV+IND.y4+3.Ø3}
	\end{example}
	\item Adverb + non class-changing suffix or inflectional nominalizer + optional class-changing suffix \texttt{+VRB}
	\begin{example} \label{ex:54} [Smeets, I. 2008: 383 (18)] \cite{RefB:21}\\
		\emph{\textbf{alü-n-tu}-y-ng-ün} 'they were more'\\
		\texttt{-AV.alü\_mucho+PVN.n4+VRB.tu36\\+IND.y4+3.ng3+PL.ün2}
	\end{example}
	\item Noun + non class-changing suffix + inflectional nominalizer +  class-changing suffix \texttt{+VRB}
	\begin{example} \label{ex:55} [Smeets, I. 2008: 411 (53)] \cite{RefB:21}\\
		\emph{\textbf{tukuyu-ke-chi}-le-wü-y} 'it looks like (long) fabric'\\
		\texttt{-NN.tukuyu\_tela+DISTR.ke+SVN.Ø4+ADJ.chi\\+VRB.Ø36+ST.le28+REF.w31+IND.y4+3.Ø3}
	\end{example}
	\item Noun + optional transitivizer or factitive + optional reflexive + optional non-realized + class-changing suffix or non class-changing suffix or inflectional nominalizer \texttt{+VRB}
	\begin{example} \label{ex:56} [Smeets, I. 2008: 90 (34)] \cite{RefB:21}\\
		\emph{\textbf{as-ka-w-ün-nge}-y} 'he is capricious'\\
		\texttt{-NN.ad\_costumbre+FAC.ka33+REF.w31+PVN.n4\\+VRB.nge36-IV+IND.y4+3.Ø3}
	\end{example}
	\item Verb + optional causative + optional transitivizer or factitive + optional reflexive + optional stative + optional hither + optional non-realized + inflectional or derivational nominalizer \texttt{+VRB}
	\begin{example} \label{ex:57} [Smeets, I. 2008: 225 (243)] \cite{RefB:21}\\
		\emph{\textbf{llüka-nten-nge}-wma} 'I was someone who easily gets afraid'\\
		\texttt{-IV.llüka\_temer+ADJQE.nten+VRB.nge36-IV\\+CSVN.wma4}
	\end{example}
	\item Reduplicated verb root + causative\\ (verb \texttt{+CA} + verb \texttt{+CA}
	\begin{example} \label{ex:58} [Smeets, I. 2008: 412 (68)] \cite{RefB:21}\\
	\emph{\textbf{ap-üm-ap-üm-ye}-nge-y} 'we have gradually been finished off'\\
	\texttt{-IV.af\_acabar+CA.m34-RVBR+SFR.ye36-TV\\+PASS.nge23+IND.y4+3.Ø3}
	\end{example}
\end{itemize}

In the last example, the form corresponds to two roots and two suffixes, but it is actually a reduplicated stem of one root with a suffix attached, reason to list it as a single root complex stem.

Complex compound stems are the most complex ones, they are not listed here but on the section \nameref{tp:61}, p. \pageref{tp:61}, where we expose them together with the encoding expressions and rules that manage them, these stems are formed by two roots and at least one suffix apart from the verbalizer.

\subsection{Special verbs} \label{sec:17}

\paragraph{} \label{tp:17} Some roots (verbs and non-verbs), due to semantic or grammatical reasons, must co-occur with certain suffixes when forming complete verb forms. There are some exceptions and/or conditions needed for these roots to behave this way. Smeets writes about the conditions, we have found the exceptions.

\subsubsection{Question roots} \label{sec:18}

\paragraph{} \label{tp:18} Interrogative roots may be verbalized, but not all forms take the same verbalizers (see D\ref{def:25}).

\emph{chem-} 'what, which' may be verbalized by suffixes \texttt{-Ø-} and \texttt{-ye-}, see following examples:

\begin{example} \label{ex:59} [Smeets, I. 2008: 434 (86)] \cite{RefB:21}\\
	\emph{chem-lle-a-l-e} 'whatever they would do' \\
	\texttt{-QC.chem\_qué\_cuál+VRB.Ø36+AFF.lle11+NRLD.a9\\+CND.l4+3.e3}
\end{example}

\begin{example} \label{ex:60} [Smeets, I. 2008: 128 (39)] \cite{RefB:21}\\
	\emph{chem-\textbf{ye}-w-üy-m-u} 'how are you both related?' \\
	\texttt{-QC.chem\_qué\_cuál+VRB.ye36+REF.w31\\+IND.y4+2.m3+DL.u2}
\end{example}

\emph{chuchi- $\sim$ tuchi-} 'which' is verbalized by the null suffix \texttt{-Ø-}, see following example:

\begin{example} \label{ex:61} [Smeets, I. 2008: 405 (7)] \cite{RefB:21}\\
	\emph{chuchi-künu-al} 'how they should carry on' \\
	\texttt{-QC.chuchi\_cuál+VRB.Ø36+PFPS.künu32\\+NRLD.a9+OVN.el4}
\end{example}

\emph{chum-} 'how' is verbalized by \texttt{-Ø-} or \texttt{-nge-}, see following examples:

\begin{example} \label{ex:62} [Smeets, I. 2008: 416 (15)] \cite{RefB:21}\\
	\emph{chum-la-e-n-ew} 'he did not do anything to me' \\
	\texttt{-QC.chum\_cómo+VRB.Ø36+NEG.la10+IDO.e6\\+IND1SG.n3+DS3A.ew1}
\end{example}

\begin{example} \label{ex:63} [Smeets, I. 2008: 225 (246)] \cite{RefB:21}\\
	\emph{chum-\textbf{nge}-wma} 'how it was' \\
	\texttt{-QC.chum\_cómo+VRB.nge36-IV+CSVN.wma4}
\end{example}

\emph{tunte- $\sim$ chunte-} may be verbalized by suffixes \texttt{-Ø-}\footnote{Smeets does not specifically mention that \emph{tunte} may be verbalized by \texttt{-Ø-}, but we have deduced it from the following text: "The interrogative \emph{tunten $ \sim $ chunten} is a quantity noun, which contains the plain verbal noun marker \emph{-n} \texttt{+PVN.n4}" [Smeets, I. 2008: 105] \cite{RefB:21}. To be able to be bound to the \texttt{+PVN.n4} suffix, the interrogative pronoun must be verbalized; since there is no realized form between the root \emph{tunte-} and morpheme \emph{-n}, the only possible verbalizer is \texttt{-Ø-}.}, \texttt{-l-} and \texttt{-ntu-}, see following examples:

\begin{example} \label{ex:64} [Smeets, I. 2008: 114 (c)] \cite{RefB:21}\\
    \emph{tunte-n-tu}\footnote{According to Smeets, the form \emph{tuntentu} has two more possible analyses [Smeets, I. 2008: 559 (\emph{tunte})] \cite{RefB:21}: \\ 1) \emph{tunte-ntu-} 'to stay', 'to be for how long' \\ \texttt{-QT.tunte\_cuánto+VRB.ntu36} \smallskip \\ 2) \emph{tunte-n-tu-} 'to take how much' \\ \texttt{-QT.tunte\_cuánto+VRB.Ø36+PVN.n4+VRB.tu36}} 'for how long?' \\
    \texttt{-QT.tunte\_cuánto+VRB.Ø36+PVN.n4+ADV.tu}
\end{example}

\begin{example} \label{ex:65} [Smeets, I. 2008: 128 (33)] \cite{RefB:21}\\
	\emph{tunte-\textbf{l}-e-y-mew} 'how much did he give to you?' \\
	\texttt{-QT.tunte\_cuánto+VRB.l36+IDO.e6\\+IND.y4+3.Ø3+DS3A.mew1}
\end{example}

\begin{example} \label{ex:66} [Smeets, I. 2008: 399 (19)] \cite{RefB:21}\\
	\emph{tunte-\textbf{ntu}-la-y} 'it did not last long' \\
	\texttt{-QT.tunte\_cuánto+VRB.ntu36+NEG.la10\\+IND.y4+3.Ø3}
\end{example}

\subsubsection{Deictic verbs} \label{sec:19}

\paragraph{} \label{tp:19} "Deictic verbs are derived from the roots  \emph{fa-} 'to become like this' and \emph{fe-} 'to become like that'. These roots do not occur without a derivational suffix. A verb which is derived from the root \emph{fa-} denotes a situation which is contextually determined. A verb which is derived from the root \emph{fe-} denotes an instance which is situationally determined" [Smeets, I. 2008: 321] \cite{RefB:21}.

Smeets says that deictic verbs do not occur without a derivational suffix, but there is a case in which \emph{fe-} directly takes a inflectional suffix without any derivational one:

\begin{example} \label{ex:67} [Smeets, I. 2008: 246 (4)] \cite{RefB:21}\\
	\emph{\textbf{fi-y} llemay}\footnote{"The particle \emph{llemay} conveys certainty on the part of the speaker" [Smeets, I. 2008: 334] \cite{RefB:21}. It consists of the affirmative suffix \emph{-lle-} \texttt{+AFF.lle11} and the particle \emph{may} which is used as a rhetoric question or a question expecting an affirmative answer.} 'that is certainly so'\\
	\texttt{-IV.fe\_ser-eso+IND.y4+3.Ø3\\-PT.llemay\_seguro\_ciertamente}
\end{example}

Compare it with the following examples (which follow the rule):

\begin{example} \label{ex:68} [Smeets, I. 2008: 462 (61)] \cite{RefB:21}\\
	\emph{kom \textbf{fe-le}-y} 'they all are that way'\\
	\texttt{-AV.kom\_todo\\-IV.fe\_ser-eso+ST.le28+IND.y4+3.Ø3}
\end{example}

\begin{example} \label{ex:69} [Smeets, I. 2008: 321 (4)] \cite{RefB:21}\\
	\emph{\textbf{fente}}\footnote{\emph{-nte-} is an unproductive derivative suffix that can yield \emph{fe-nte-} 'adv. that much' and \emph{fa-nte-} 'adv. this much'. As it is unproductive, the adverbial form is collected as such in the lexicon.} \emph{-n-üy} 'it is as much/big as...'\\
	\texttt{AV.fente\_tanto+PVN.n4+VRB.Ø36+IND.y4+3.Ø3}
\end{example}

\begin{example} \label{ex:70} [Smeets, I. 2008: 322 (7)] \cite{RefB:21}\\
	\emph{ka \textbf{fe-le-pa-tu}-n} 'I was in the same situation as before'\\
	\texttt{-AJ.ka\_otro\\-IV.fe\_ser-eso+ST.le28+HH.pa17+RE.tu16\\+IND1SG.n3}
\end{example}

\begin{example} \label{ex:71} [Smeets, I. 2008: 321 (1)] \cite{RefB:21}\\
	\emph{\textbf{fa-le}-wma iñche} 'this is how I was'\\
	\texttt{-IV.fa\_ser-esto+ST.le28+CSVN.wma4\\-PP.iñche\_yo}
\end{example}

\begin{example} \label{ex:72} [Smeets, I. 2008: 322 (10)] \cite{RefB:21}\\
	\emph{\textbf{fa-m-nge}-chi küdaw-ke-n} 'I work this way'\\
	\texttt{-IV.fa\_ser-esto+CA.m34+PASS.nge23\\+SVN.Ø4+ADJ.chi\\-IV.küdaw\_trabajar+CF.ke14+IND1SG.n3}
\end{example}

\subsubsection{Defective verbs} \label{sec:20}

\paragraph{} \label{tp:20} Roots of posture (of the body) verbs obligatorily occur together with the perfect persistence marker \texttt{+PFPS.künu32}, the progressive persistence marker \texttt{+PRPS.nie32} or the stative morpheme \texttt{+ST.le28} when they are the only root of a stem, i.e., when they are not in compounds. Otherwise, when these verbs occur as part of compounds they are not compelled to use any of the three suffixes. Verbs identified by Smeets are:

\begin{itemize} \label{it:12}
	\item[]\emph{kopüd-} 'to lie down on one's belly'
	\item[]\emph{kudu-} 'to lie down'
	\item[]\emph{külü-} 'to lean on one's elbow'
	\item[]\emph{llikosh-} 'to sit down on one's heels', 'to squat'
	\item[]\emph{payla-} 'to lie down on one's back'
	\item[]\emph{potri-} 'to lean over'
	\item[]\emph{potrong-} 'to bow forward' (the head)
	\item[]\emph{potrü-} 'to bow forward' (the body)
	\item[]\emph{rekül-} 'to lean'
	\item[]\emph{üñif-} 'to lie down on the floor'
	\item[]\emph{wira-} 'to sit down with spread legs'
	\item[][Smeets, I. 2008: 235] \cite{RefB:21}
\end{itemize}

\begin{example} \label{ex:73} [Smeets, I. 2008: 296 (21)] \cite{RefB:21}\\
	\emph{\textbf{üñif-künu}-a-fi-ñ} 'I will spread it out'\\
	\texttt{-TV.ünif\_extender+PFPS.künu32\\+NRLD.a9+EDO.fi6+IND1SG.n3}
\end{example}

\begin{example} \label{ex:74} [Smeets, I. 2008: 261 (2)] \cite{RefB:21}\\
	\emph{\textbf{kudu-le}-me-we-la-n} 'I am not going to lay down there any more'\\
	\texttt{-IV.kudu\_yacer+ST.le28+TH.me20+PS.we19\\+NEG.la10+IND1SG.n3}
\end{example}

For the combination of these verbs with \texttt{+PRPS.nie32} we have found no examples in Smeets or elsewhere. Smeets provides two meanings for \emph{kopüd-} when taking this suffix, but no example:

\begin{enumerate} \label{it:13}
	\item[]\emph{kopüd-nie-}
	\item to hold someone on his belly [Smeets, I. 2008: 235] \cite{RefB:21}
	\item to hold in a face downward position [Smeets, I. 2008: 519] \cite{RefB:21}
\end{enumerate}

As we are following Smeets' description of \emph{Mapudüngun}, we have implemented what she states about these verbs, but, besides the issue presented about \texttt{+PRPS.nie32}, there are others that do not support her statement about these verbs. Probably she has worked on not published data.

For the verb \emph{kopüd-} 'to lie down on one's belly' there are no examples but the meaning it takes with \texttt{+PFPS.künu32}, \texttt{+PRPS.nie32} and \texttt{+ST.le28}. We have found examples in other texts, but some of them show a different behaviour to what Smeets explains, i.e., they do not present the "obligatory" suffixes.

\begin{example} \label{ex:75} [Febrés, A.]\footnote{All the examples that come from Augusta, F., Febrés, A. and Valdivia, L. has been consulted on-line on the CORLEXIM site \cite{RefB:03}.}\\
	\emph{\textbf{kopu}-n} 'to be face down lying on the floor or head down, or half bent the body.'\\
	\texttt{-IV.kopüd\_yacer-boca-abajo+PVN.n4}
\end{example}

\begin{example} \label{ex:76} [Valdivia, L.] \cite{RefB:03}\\
	\emph{\textbf{kopu}-w-ün} 'to be facing the floor'\\
	\texttt{-IV.kopüd\_yacer-boca-abajo+PS.we19+PVN.n4}
\end{example}

For verbs \emph{kudu-} 'to lie down' and \emph{külü-} 'to lean on one's elbow', Smeets herself presents examples contradicting its obligatory co-occurrence with the treated suffixes. Other texts also contradict her (Augusta, F. \cite{RefB:03}, Febrés \cite{RefB:03}, Zúñiga \cite{RefB:24}, Mösbach \cite{RefB:14}).

\begin{example} \label{ex:77} [Smeets, I. 2008: 349 (17)] \cite{RefB:21}\\
	\emph{\textbf{kudu}-pu-a-el} 'to go to bed'\\
	\texttt{-IV.kudu\_yacer+LOC.pu17+NRLD.a9+OVN.el4}
\end{example}

\begin{example} \label{ex:78} [Smeets, I. 2008: 244 (4)] \cite{RefB:21}\\
	\emph{\textbf{kudu}-nu-l-m-i} 'if you do not go to bed'\\
	\texttt{-IV.kudu\_yacer+NEG.no10+CND.l4+2.m3+SG.i2}
\end{example}

\begin{example} \label{ex:79} [Smeets, I. 2008: 526 (\emph{lüf-})] \cite{RefB:21}\\
	\emph{\textbf{külü}-a-y antü} 'the Sun will lay down'\\
	\texttt{-IV.külü\_apoyar+NRLD.a9+IND.y4+3.Ø3\\-NN.antü\_sol}
\end{example}

For \emph{llikosh-} Smeets gives the meaning it takes with two of the suffixes but no example. The example we show comes from Augusta, F. \cite{RefB:03}.

\begin{itemize} \label{it:14}
	\item[]\emph{llikosh-küle-} 'to squat, to crouch'
	\item[]\emph{llikosh-künu-w-} 'to squat down, to crouch down'
	\item[][Smeets, I. 2008: 528] \cite{RefB:21}
\end{itemize}

\begin{example} \label{ex:80} [Augusta, F.] \cite{RefB:03}\\
	\emph{\textbf{llikod-küle}-n} 'to be snuggled'\\
	\texttt{-IV.llikosh\_acurrucar+ST.le28+PVN.n4}
\end{example}

For \emph{payla-} there are also no examples in Smeets but the definition in combination with two of the three suffixes. Examples coming from other sources contradict Smeets' observations.

\begin{itemize} \label{it:15}
	\item[]\emph{payla-le-} 'to be lying on one's back'
	\item[]\emph{payla-künu-w-} 'to lie down on one's back'
	\item[][Smeets, I. 2008: 543] \cite{RefB:21}
\end{itemize}

\begin{example} \label{ex:81} [Augusta, Febrés \& Valdivia] \cite{RefB:03}. [Mösbach, E. 1936] \cite{RefB:14}\\
	\emph{\textbf{payl'a}-n} 'to lie on one's back'\\
	\texttt{-IV.payla\_yacer-de-espalda+PVN.n4}
\end{example}

\begin{example} \label{ex:82} [Febrés, A.] \cite{RefB:03}\\
	\emph{\textbf{paylla-l}-ün}\footnote{This example seems to support Smeets' statements, if the phonological changes are that stative \emph{-le-} drops its vocalic element and the indicative 1\textsuperscript{st} person adds an epenthetic schwa in presence of the previous consonant. Normally, \emph{-le-} keeps the vocal and the following suffix, beginning in consonant, remains the same.} 'to put or leave something on its own back, or in peace'\\
	\texttt{-IV.payla\_yacer-de-espalda+ST.le28+IND1SG.n3}
\end{example}

For \emph{potri-} 'to lean', Smeets gives two examples and the meaning acquired with one of the three suffixes, and in a compound. We did not find examples of \emph{potri-} in other texts. We believe that \emph{potri-} and \emph{potrü-} 'to bow forward' are two variants of the same verb, even though Smeets defines them differently. The interchangeability between \emph{ü} and \emph{i} is not rare. She provides two different translations for an example with \emph{potrü-}, the second one in alignment with the same example using \emph{potri-} instead. Compare:

\begin{example} \label{ex:83} [Smeets, I. 2008: 549 (\emph{potri-})] \cite{RefB:21}\\
	\emph{\textbf{potri-tripa}-n ti wangku-mu} 'I toppled out of the chair'\\
	\texttt{-IV.potri\_inclinar-IV.tripa\_salir-CR.IV\\+IND1SG.n3\\-AP.ti\_el -NN.wangku\_silla+INST.mew}
\end{example}

\begin{example} \label{ex:84}\ \\
	\emph{\textbf{potrü-tripa}-n ti wangku-mu}
	\item[1.]'I fell backward from the chair' [Smeets, I. 2008: 62 (11)] \cite{RefB:21}
	\item[2.]'I toppled from the chair' [Smeets, I. 2008: 563 (\emph{tripa-})] \cite{RefB:21}\\
	\texttt{-IV.potrü\_inclinar-IV.tripa\_salir-CR.IV\\+IND1SG.n3\\-AP.ti\_el -NN.wangku\_silla+INST.mew}
\end{example}

\begin{itemize} \label{it:16}
	\item[]\emph{potri-le-} 'to be leaning (over)'
	\item[]\emph{potri-tripa-} 'to topple'
	\item[][Smeets, I. 2008: 543] \cite{RefB:21}
\end{itemize}

\begin{example} \label{ex:85} [Smeets, I. 2008: 296 (20)] \cite{RefB:21}\\
	\emph{\textbf{potri-künu}-w-ün} 'I bent forward'\\
	\texttt{-IV.potrü\_inclinar+PFPS.künu32\\+REF.w31+IND1SG.n3}
\end{example}

We did find examples of \emph{potrü-} in other texts, and as in previous cases, they show no concordance with Smeets statements:

\begin{example} \label{ex:86} [Augusta, F.] \cite{RefB:03}\\
	\emph{\textbf{potrü}-w-ün} 'to buck'\\
	\texttt{-IV.potrü\_inclinar+PS.we19+PVN.n4}
\end{example}

For the verb \emph{rekül-} 'to lean' there are some examples in other texts contradicting Smeets.

\begin{example} \label{ex:87} [Augusta, F.] \cite{RefB:03}\\
	\emph{\textbf{rekül}-tu-n} 'to caddle up, to lie down'\\
	\texttt{-IV.rekül\_apoyar+TR.tu33+PVN.n4}\\
	\emph{\textbf{rekül}-tu-we} 'back (of something)'\\
	\texttt{-IV.rekül\_apoyar+TR.tu33+NOMPI.we}
\end{example}

\begin{example} \label{ex:88} [Febrés, A.] \cite{RefB:03}\\
	\emph{\textbf{rekül}-ün} 1.'to cuddle up, to stand or stand on something'
	2.'to lean'\\
	\texttt{-IV.rekül\_apoyar+PVN.n4}
\end{example}

For the verb \emph{üñif-} $\sim$ \emph{ünif-} 'to lie down on the floor' we did not find examples other than the one given by Smeets, see example E\ref{ex:73}, p. \pageref{ex:73}.

For the verb \emph{wira-} 'to sit down with spread legs', Smeets presents no examples but definitions in combination with two of the three suffixes:

\begin{itemize} \label{it:17}
	\item[]\emph{wira-künu-w-} 'to adopt a position with the legs apart'
	\item[]\emph{wira-le-} 'to sit with the legs apart'
	\item[][Smeets, I. 2008: 576] \cite{RefB:21}
\end{itemize}

Examples from other texts:

\begin{example} \label{ex:89} [Augusta, F.] \cite{RefB:03}\\
	\emph{\textbf{wira-le}-n} 'to be with legs open'\\
	\texttt{-IV.wira\_sentar-con-las-piernas-abiertas\\+ST.le28+PVN.n4}
\end{example}

\begin{example} \label{ex:90} [Mösbach, E. 1936] \cite{RefB:14}\\
	\emph{\textbf{wira-l-küle}-chi} 'the open ones'\\
	\texttt{-IV.wira\_sentar-con-las-piernas-abiertas\\+CA.l34+ST.le28+SVN.Ø4+ADJ.chi}
\end{example}

\begin{example} \label{ex:91} [Mösbach, E. 1936] \cite{RefB:14}\\
	\emph{\textbf{wira}}\footnote{It looks like Pascual Koña uses \emph{wira-} with the sense of "two things that spread appart", not only the legs. Mösbach collected Koña's memoires in the book cited in reference \cite{RefB:14}. }\emph{\textbf{-l-künu}-ye-nge-ke-fu-y} 'they remain open'\\
	\texttt{-IV.wira\_sentar-con-las-piernas-abiertas\\+CA.l34+PFPS.künu32+PLR.ye24+PASS.nge23\\+CF.ke14+IPD.fu8+IND.y4+3.Ø3}
\end{example}

\subsubsection{Verbs that need a directional} \label{sec:21}

\paragraph{} \label{tp:21} There is another group of verbs that require a directional to be expressed if they are not part of a compound or take a transitivizer or causative suffix. Directional suffixes are \emph{-me-} thither (slot 20), \emph{-pa-} hither (slot 17) and \emph{-pu-} locative (slot 17). The affected verbs are:

\begin{itemize} \label{it:18}
	\item[]\emph{antü-} 'to spend a day'
	\item[]\emph{fül-} 'to come near'
	\item[]\emph{küyen-} 'to spend a month'
	\item[]\emph{llekü-} 'to approach'
	\item[]\emph{nge-} 'to have been' (it only requires \emph{-me-} thither or \emph{-pa-} hither)
	\item[]\emph{pülle-} 'to come near'
	\item[]\emph{ru-} 'to pass, to go through' (it only requires \emph{-me-} thither or \emph{-pa-} hither)
	\item[]\emph{tripantu-} 'to spend a year'
	\item[]
	\item[]And the following compounds:
	\item[]\emph{kim-kon-} '(know-enter-) to find out, to understand' requires \emph{-pa-} hither
	\item[]\emph{kim-püra-} '(know-go\_up-) to realize' requires \emph{-me-} thither or \emph{-pa-} hither
	\item[]\emph{trem-tripa-} '(grow\_up-go\_out-) to become conscious of while growing up' requires \emph{-pa-} hither
	\item[][Smeets, I. 2008: 325, 326] \cite{RefB:21}
\end{itemize}

For \emph{antü-} 'to spend a day' there are some examples refuting Smeets' observations, i.e., showing the occurrence of the verb without the directionals nor the transitivizer or causative:
\begin{itemize} \label{it:19}
	\item[]Augusta, F. \cite{RefB:03} \emph{antü-n, antü-le-iñ, antü-y, antü-ñma-le-n, antü-ñma-n}
	\item[]Febrés, A. \cite{RefB:03} \emph{antü-n, antü-ku-n}
	\item[]Mösbach, E. \cite{RefB:14} \emph{antü-y}
	\item[]Smeets, I. \cite{RefB:21} \emph{antü-le-y, antü-a-y,  antü-y, antü-le-chi}
	\item[]Zúñiga, F. \cite{RefB:24} \emph{antü-y}
	\item[]Valdivia. L. \cite{RefB:03} \emph{antü-n, antü-n-ku-n}
\end{itemize}

For \emph{fül-} 'to come near', it happens as with \emph{antü-}:
\begin{itemize} \label{it:20}
	\item[]Augusta, F. \cite{RefB:03} \emph{fül-küle-n, fül-ma-n, fül-ün}
	\item[]Mösbach, E. \cite{RefB:14} \emph{fül-a-n, fül-el, fül-la-e-y-ew, fül-küle-le-n, fül-ma-nge-fu-lu, fül-e-y}
	\item[]Smeets, I. \cite{RefB:21} \emph{fül-küle-n}
\end{itemize}

For \emph{küyen-} 'to spend a month', Smeets gives no examples, and we have also found contradicting ones from other authors:
\begin{itemize} \label{it:21}
	\item[]Augusta, F. \cite{RefB:03} \emph{küyen'-ün}
	\item[]Febrés, A. \cite{RefB:03} \emph{küyen-pe-n, küyen-a-y}
\end{itemize}

For \emph{llekü-} 'to approach', Smeets gives one example, and other authors have contrary examples.

\begin{example} \label{ex:92} [Smeets, I. 2008: 503 (\emph{elu-})] \cite{RefB:21}\\
	\emph{\textbf{llekü-pu}-el} 'to come near'\\
	\texttt{-IV.llekü\_acercar+LOC.pu17+OVN.el4}
\end{example}

\begin{itemize} \label{it:22}
	\item[]Augusta, F. \cite{RefB:03} \emph{llekü-n, llekü-le-n, llekü-ñma, llekü-ñma-le-n, llekü-ñma-nie-n}
	\item[]Febrés, A. \cite{RefB:03} \emph{lleku-n, lleku-le-n}
	\item[]Mösbach, E. \cite{RefB:14} \emph{llekü-n, llekü-ñma-nie-lu}
	\item[]Valdivia. L. \cite{RefB:03} \emph{llekü-n, llekü-le-n}
\end{itemize}

For \emph{nge-} 'to have been', Smeets remarks that an exception is when the negation marker \emph{-la-} co-occurs. But we have found that the exception also applies with the negation marker \emph{-no-}. The first two following forms found in Smeets are also realized without directional suffixes:

\begin{itemize} \label{it:23}
	\item[]Exceptions: \emph{nge-n, nge-y}
	\item[]Negation \emph{-la-}: \emph{nge-\textbf{la}-y, nge-\textbf{la}-n, nge-we-\textbf{la}-y, nge-ke-\textbf{la}-fu-y, nge-we-tu-\textbf{la}-y}
	\item[]Negation \emph{-no-}: \emph{nge-\textbf{nu}-n, nge-ke-\textbf{nu}-lu, nge-\textbf{nu}-n-mu}
\end{itemize}

For \emph{pülle-} 'to come near', there are contrary examples even in Smeets:

\begin{itemize} \label{it:24}
	\item[]Augusta, F. \cite{RefB:03} \emph{pülle-le-n, pülle-künu-n, pülle-lu, pülle-le-ye-lu, pülle-le-chi, pülle-le-nu-chi, pülle-nie-gel-chi, pülle-ke-ñma-w-küle-y, pülle-ñma-w-küle-y-u}
	\item[]Mösbach, E. \cite{RefB:14} \emph{pülle-ñma-w-ke-chi}
	\item[]Smeets, I. \cite{RefB:21} \emph{pülle-le-y, pülle-le-lu}
\end{itemize}

For \emph{ru-} 'to pass, to go through' the occurrence are as Smeets describes them, except for two examples found at Febrés, A. \cite{RefB:03} (\emph{ru-n, ru-a-n}), which we have no way yet to confirm as right or wrong.

\begin{example} \label{ex:93} [Augusta, F.] \cite{RefB:03}\\
	\emph{\textbf{ru-l-pa}-nütram-pe-lu} 'interpreter, translator'\\
	\texttt{-IV.ru\_pasar+CA.l34+HH.pa17\\-NN.nütram\_conversación+PX.pe13+SVN.lu4}
\end{example}

For \emph{tripantu-} 'to spend a year' there are also examples from other authors not supporting Smeets' findings:

\begin{itemize} \label{it:25}
	\item[]Augusta, F. \cite{RefB:03} \emph{tripantu-le-n, tripantu-n, tripantu-y,\\ tripantu-chi}
	\item[]Febrés, A. \cite{RefB:03} \emph{tripantu-n, tripantu-a-n, tripantu-y}
	\item[]Mösbach, E. \cite{RefB:14} \emph{tripantu-a-m, tripantu-n, tripantu-el}
	\item[]Valdivia, L. \cite{RefB:03} \emph{tripantu-n}
\end{itemize}

Finally, for compounds \emph{kim-kon-} 'to find out, to understand', \emph{kim-püra-} 'to realize' and \emph{trem-tripa-} 'to become conscious of while growing up', we have found no other examples than Smeets', who also gives a contradictory example: E\ref{ex:94}.

\begin{example} \label{ex:94} [Smeets, I. 2008: 447 (26)] \cite{RefB:21}\\
	\emph{\textbf{kim-kon}-y-iñ} 'we had become aware'\\
	\texttt{-TV.kim\_saber-IV.kon\_entrar-CR.IV\\+IND.y4+1.Ø3+PL.iñ2}
\end{example}

\begin{example} \label{ex:95} [Smeets, I. 2008: 447 (24)] \cite{RefB:21}\\
	\emph{\textbf{kim-kon-pa}-n} 'I understood, I realized'\\
	\texttt{-TV.kim\_saber-IV.kon\_entrar-CR.IV+HH.pa17\\+IND1SG.n3}
\end{example}

\begin{example} \label{ex:96} [Smeets, I. 2008: 381 (1)] \cite{RefB:21}\\
	\emph{\textbf{kim-püra-me}-n} 'I bcame aware, I came to appreciate'\\
	\texttt{-TV.kim\_saber-IV.püra\_subir-CR.IV+TH.me20\\+IND1SG.n3}
\end{example}

\begin{example} \label{ex:97} [Smeets, I. 2008: 446 (11)] \cite{RefB:21}\\
	\emph{\textbf{kim-püra-me-pa}-n} 'I realized'\\
	\texttt{-TV.kim\_saber-IV.püra\_subir-CR.IV+TH.me20\\+HH.pa17+IND1SG.n3}
\end{example}

\begin{example} \label{ex:98} [Smeets, I. 2008: 262 (10)] \cite{RefB:21}\\
	\emph{\textbf{kim-püra-me-pa}-fi-ñ} 'I have come to know him'\\
	\texttt{-TV.kim\_saber-IV.püra\_subir-CR.IV+TH.me20\\+HH.pa17+EDO.fi6+IND1SG.n3}
\end{example}

\begin{example} \label{ex:99} [Smeets, I. 2008: 415 (4)] \cite{RefB:21}\\
	\emph{\textbf{trem-tripa-pa}-y} 'they grew up knowing (about)'\\
	\texttt{-IV.trem\_crecer-IV.tripa\_salir-CR.IV\\+IND.y4+3.Ø3}
\end{example}

\subsection{Morphophonology} \label{sec:22}

\paragraph{} \label{tp:22} As we have mentioned in the introduction, interaction among roots and suffixes creates different contexts in which the form of these elements may be affected. Most common changes correspond to epenthesis and elision, but there are also cases of phoneme alternation, some are obligatory and others, optional. We present all of these changes in section \ref{sec:39} \nameref{sec:39}, p. \pageref{sec:39}, where, at the same time, we explain how these variations have been encoded to be processed by the FST analyser.

\section{The computational approach} \label{sec:23}

\paragraph{} \label{tp:23} In this section, after a basic introduction to computational morphology and Finite State Transducers (FST), we explain how morphophonologic phenomena of \emph{Mapudüngun} have been encoded in order to process \emph{Mapuche} words through FST analysis, and obtain a proper identification of the parts (roots and suffixes) forming these words.

We do a quick review on FOMA implementations. FOMA is the FST compiling program we use to generate the analyser (see section \ref{sec:30}). And finally, in section \ref{sec:31}, we enter into \emph{Mapudüngun} encoding, starting by the alphabet and finishing by the lexicon: roots and suffixes.

\subsection{Computational morphology} \label{sec:24}

\paragraph{} \label{tp:24} Computational morphology is the branch of computational linguistics concerned with word structure\footnote{In this section we follow Gasser [2011: 55] \cite{RefB:08}.}. Two kinds of processing are of interest. One is morphological analysis, by which a surface word form is analysed into a lexical representation, consisting of the word's component morphemes or grammatical features. The other is morphological generation, by which a lexical representation is converted to a surface word form. Consider the \emph{Mapuche} verb \emph{lelien} 'you looked at me'. A basic morphological analysis would simply segment the word into the morphemes that make it up, as seen below:

\begin{example} \label{ex:100}\ \\
	\emph{lelien} → \emph{leli‐e-n}
\end{example}

A more abstract lexical representation output would indicate the lexical and grammatical significance of the morphemes. The word \emph{lelien} could be represented at the lexical level as shown right below:

\begin{example} \label{ex:101}\ \\
	\emph{leli-e-n} 'you looked at me / look at me!'\\
	‑TV.leli\_mirar\footnote{The analyser lexicon is collected with the Spanish translation, that's why all the analyses presented in this article carry the root meaning in Spanish.}+IDO\footnote{Tags meaning are found in annex \ref{anx:01} \nameref{anx:01}.}.e6+IND1SG.n3+DS12A.Ø1
\end{example}

This chain of tags represents the root of the verb \emph{leli-} meaning 'look at', the internal direct object \emph{-e-}, the portmanteau suffix indicative 1\textsuperscript{st} person singular \emph{-n}, and the null dative subject for 2\textsuperscript{nd} or 1\textsuperscript{st} agent persons\footnote{We follow Smeets' descriptions throughout this work, but it is important to know that in this issue there is discrepancy among authors of \emph{Mapudüngun} descriptive grammars. Basically, what Smeets [2008] \cite{RefB:21} identifies as "agent-patient paradigm" is what Zúñiga [2006] \cite{RefB:24} calls "verbal inversion", and Salas [2006] \cite{RefB:19} "person focalization".}.

Three types of information are required to perform morphological analysis: a lexicon, the morphotactics and the (alternation) rules.

A lexicon is composed of roots or stems, which combine with grammatical morphemes to yield surface word forms.

\paragraph{Morphotactics} \label{tp:25} refer to constraints on the order and class of the morphemes that make up a word within a particular category. For example, the morphotactics of \emph{Mapudüngun} verbs specify the following minimal sequence of morphemes: mood (indicative, conditional or imperative, slot 4), subject (slot 3) and number (slot 2).

\paragraph{Alternation rules} \label{tp:26} are responsible for the variation of forms that morphemes take in the presence of other morphemes. For example, the portmanteau suffix of 1\textsuperscript{st} person indicative mood takes two forms, one before vowels, another before consonants, where an epenthetic \emph{-ü} appears.

Together, knowledge of alternation rules, morphotactics, and the forms of roots or stems in the lexicon represent the morphology of a given language.

Morphological analysis may be efficiently handled by finite state transducers (FST). An FST is a network of states and transitions between them, and the analysis of a word is a path through this network. Each of the transitions along the path specifies a correspondence between input characters (or phones) and output characters. The transducer\footnote{A transducer is a device or machine that converts energy from one form into another, e.g., a microphone is a transducer that converts the vibrations captured from the air into analogous electrical impulses. An FST converts a chain of symbols into another chain of symbols.} converts sequences of input characters to sequences of output characters. One very useful property of FSTs is that they may be inverted. This means that the same transducer that implements analysis (surface to lexical representation) for a given rule, it can also implement generation (lexical to surface representation) through simple reversal of the input and output characters. Another useful property is composition: a sequence of FSTs, converting a surface representation into a lexical representation with various intermediate stages, it may be merged into a single FST which behaves the same as the original sequence of FSTs (see \ref{tp:29} \nameref{tp:29}, p. \pageref{tp:29}).

\subsubsection{Finite state method} \label{sec:25}

\paragraph{} \label{tp:27} A finite state transducer (FST) is a piece of software that operates as an enhanced finite state machine (FSM) which in its turn is capable of representing and operating over finite state networks (FSNs)\footnote{In this section we follow the explanation given by Ríos [2015: 18-21] \cite{RefB:17}}.

\subsubsection{Finite state transducers (FSTs)} \label{sec:26}

\paragraph{} \label{tp:28} There is an important distinction between FSMs that are one-sided, and FSTs that have an upper and a lower side, or more generally, an input and an output level. Since an FST has two sides, it can not only decide if a given word is part of its regular language, but it will also return the corresponding output to the given input [Beesley \& Karttunen 2003: 11] \cite{RefB:01}.

An FST accordingly implements a relation between two regular languages: an upper side and a lower side regular languages, and it literally "transduces" strings from one language into the other. In a non-deterministic FST it may produce more than one possible outputs for a given string.

See figure \ref{fig:01} for an example of an FST that contains the relation of two of the following four word forms with the \emph{Mapudüngun} root \emph{miaw-} 'to wander', and their respective morphological analysis\footnote{\emph{miaw-ün} 'I wandered', \emph{miaw-üy-m-i} 'you wandered', \emph{miaw-a-n} 'I will wander', \emph{miaw-a-y-m-i} 'you will wander'.}:

\begin{figure}[H]
	\includegraphics[width=\columnwidth]{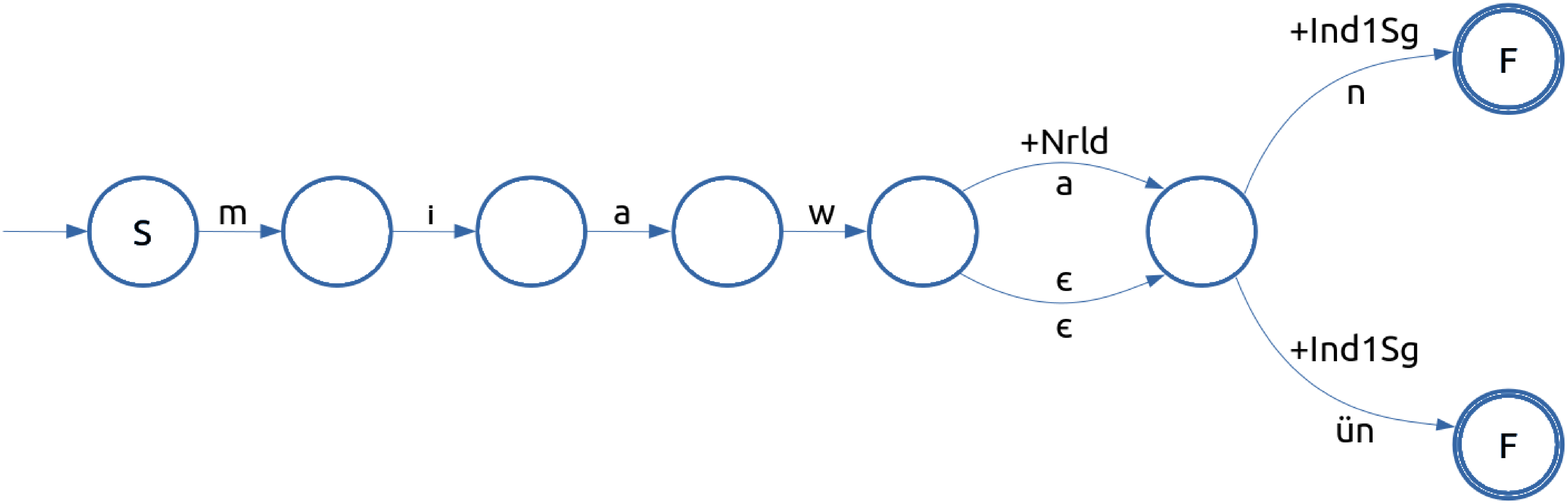}
	\caption{Finite state transducer for \emph{miaw-} with Ind1Sg, present/past or future}
	\label{fig:01}
\end{figure}

\begin{example} \label{ex:102}\ \\
	\emph{miawün} → \texttt{miaw+Ind1Sg3}\footnote{The number at the end of each suffix tag indicates its assigned slot. For a complete explanation on how to read suffixes tags see the note following E\ref{ex:1}, p. \pageref{note:02}.} (in figure \ref{fig:01})\\
		\emph{miawüymi} → \texttt{miaw+Ind4+2p3+Sg2}\\
		\emph{miawan} → \texttt{miaw+Nrld9+Ind1Sg3} (in figure \ref{fig:01})\\
		\emph{miawaymi} → \texttt{miaw+Nrld9+Ind4+2p3+Sg2}
\end{example}

Note that the transducer contains an empty transition \textcyrillic{є:є} which makes the NRLD suffix \emph{-a-} optional. The transducer in figure \ref{fig:01} may be applied in both directions:

\begin{quote} \label{note:03}
	{\small Given \emph{miawüymi} 'you wandered' as input, applied in "upward" direction, it produces:\\ \texttt{‑IV.miaw\_merodear+IND.y4+2.m3+SG.i2} as output. This is the procedure for morphological analysis.
	\\\\
	Given \texttt{‑IV.miaw\_merodear+IND1SG.n3} as input, applied in "downward" direction, it produces \emph{miawün} 'I wandered' as output. This is the procedure for generation.}
\end{quote} 

\begin{figure}[H]
	\includegraphics[width=\columnwidth]{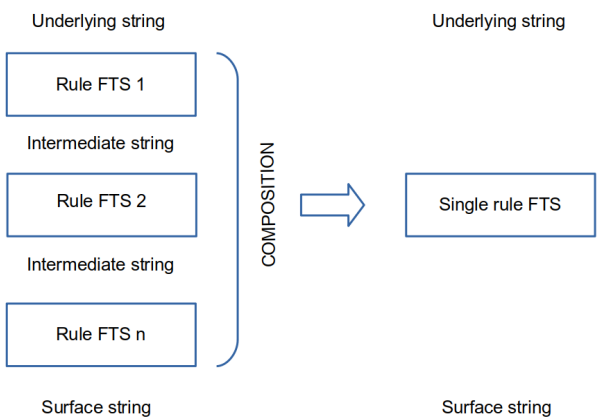}
	\caption{A cascade of rules compiled into finite state transducers may be combined into a single equivalent FST via composition. This mathematical possibility, shown by Johnson, may be performed in practice using a finite state software [Beesley \& Karttunen 2003: 35] \cite{RefB:01}}
	\label{fig:02}
\end{figure}

\paragraph{Composition} \label{tp:29} is a "hard to handle" concept in finite state processing. However, here it suffices to affirm that a cascade of rules compiled into finite state transducers may be combined into a single equivalent FST via composition. See figure \ref{fig:02}.

\subsubsection{Two levels morphology} \label{sec:27}

\begin{figure}[H]
	\includegraphics[width=\columnwidth]{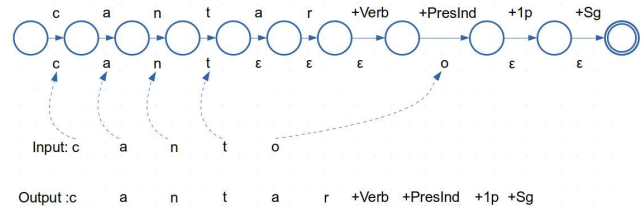}
	\caption{Analysing canto [Beesley \& Karttunen 2003: 13] \cite{RefB:01}}
	\label{fig:03}
\end{figure}

\begin{figure}[H]
	\includegraphics[width=\columnwidth]{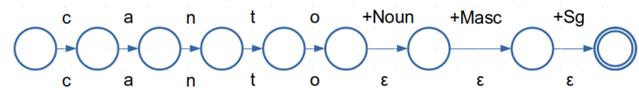}
	\caption{Another path in the Spanish morphological analyser [Beesley \& Karttunen 2003: 13] \cite{RefB:01}}
	\label{fig:04}
\end{figure}

\paragraph{} \label{tp:30} The upper language, also called the abstract level, the lower language also called surface level, and the relation they establish as part of an FST are well explained in Beesley \& Karttunen [2003] \cite{RefB:01}. Related images (figures \ref{fig:03} and \ref{fig:04}) are presented here just to illustrate a general idea about these three concepts: upper and lower languages, and the relation among them.

Table \ref{tab:03} shows that relations contain pairs of strings. For analysis, the lower language is used as input, and the upper language is produced as output [figures \ref{fig:03} and \ref{fig:04}]:

\begin{table}[htb]
	\caption{Article/Determiner/Quantifier distinctions.}
	\label{tab:03}
	\begin{tabular}{|l|l|}
		\hline\noalign{\smallskip}
		Upper: the+Art+Def & Upper: a+Art+Indef\\
		\noalign{\smallskip}\hline\noalign{\smallskip}
		Lower: the & Lower: a\\
		\noalign{\smallskip}\hline
	\end{tabular}
\end{table}

\paragraph{Regular expressions as rules} \label{tp:31}
In the development of this system we mainly used two types of operators "restriction and replacement" (\texttt{=>, ->}), complemented with composition \\(\texttt{.o.}), context (\texttt{||}), and some others operators.

Restriction is one of the fundamental functions in two-level calculus:

\begin{example} \label{ex:103}
	\texttt{[a => c \_ [r|t]];}
\end{example}

E\ref{ex:103} denotes the language of strings that have the property that the string "\textbf{a}" is immediately preceded by the string "\textbf{c}" and immediately followed by the string "\textbf{r}" or "\textbf{t}"; so, the final strings \texttt{"cat"} and \texttt{"car"} satisfy the condition, but strings such as \texttt{"cab"} or
\texttt{"can"} do not.

\begin{example} \label{ex:104}
	\texttt{[y -> \{ie\} || \_ [r|\{st\}]];}
\end{example}
	
E\ref{ex:104} denotes the relation in which the string "\textbf{y}" is transformed into "\textbf{ie}" (here a condition is introduced, in this case is the context represented by the twin pipes (\texttt{||})) when followed by "\textbf{r}" or "\textbf{st}"; so, \texttt{"ugly"} becomes \texttt{"uglier"} or \texttt{"ugliest"}, and \texttt{"pretty"} becomes \texttt{"prettier"} or\\ \texttt{"prettiest"}\footnote{The actual rule to treat this behaviour is much more complex, it is presented in this simplistic way just as an example.}.

Composition is the concatenation of rules, for example, E\ref{ex:104} could be decomposed into two concatenated rules, i.e., they together form a "composition" of rules:

\begin{example} \label{ex:105}
	\texttt{[y -> \{ie\} || \_ r] .o. \newline [y -> \{ie\} || \_ \{st\}];}
\end{example}

Or it could be represented stating two different contexts for the change to occur:

\begin{example} \label{ex:106}
	\texttt{[y -> \{ie\} || \_ r, \_ \{st\}];}
\end{example}

In general, regular expressions may be redundant because some of the operators could easily be defined in terms of others, which means that a string and its restrictions may be expressed in different ways by means of regular expressions.

For more detailed descriptions on regular expressions operators, please consult: FOMA's Regular Expression Reference: \href{https://code.google.com/archive/p/foma/wikis/RegularExpressionReference.wiki}{https://code.google.com/archive/p/foma/wikis/\\RegularExpressionReference.wiki}

Another excellent reference is Beesley \& Karttunen's "Finite State Morphology" book \cite{RefB:01}.

\subsection{The bases of the analyser} \label{sec:28}

\paragraph{} \label{tp:32} The main script of the analyser is the series of regular expressions (regexps) encoding the \emph{Mapudüngun} grammar. \\This is where the different parts forming the \emph{Mapuche} language are declared: roots, suffixes, particles, etc., and the rules for them to interact in the way \emph{Mapudüngun} accepts it.

\paragraph{} \label{tp:33} We follow Smeets description of \emph{Mapudüngun}: "A Grammar of \emph{Mapuche}" \cite{RefB:21} to implement the analyser and generator. We have begun treating only one variant or dialect of this language, which is known as central \emph{Mapudüngun}. The analyser, which is basically the same as the generator, has little broader rules that accept mainly spelling variants, that in some cases come from different dialects, but generally correspond to a certain syncretism (and confusion) generated after the different spelling proposals and the influence of Spanish orthography.

The spelling proposal we follow is AMU, "Alfabeto \emph{Mapuche} Unificado" [Sochil 1986, 1988] \cite{RefB:23}. This "grafemario" is also known as the "academic proposal".

Having half mind in computational technology and the other half in linguistics, finding a description of a language such as the one Smeets does of \emph{Mapudüngun} in her thesis work, leads straight to think in a computational implementation of her grammar. The suffixes organized in slots and the description of interaction rules are the words that reflect what regular expressions can encode. We did further research on other descriptions of \emph{Mapudüngun} while implementing the analyser but only to compare descriptions (Other sources of \emph{Mapudüngun} grammars we have consulted are Fernández-Garay \& Malvestiti \cite{RefB:07}, Lonkon \cite{RefB:12}, Ragileo \cite{RefB:15}, Salas \cite{RefB:19} and Zúñiga \cite{RefB:24}).

\subsection{The analyser} \label{sec:29}

\paragraph{} \label{tp:34} This is a rule based morphological analyser (and generator) applied to the \emph{Mapudüngun} language. It was built with finite state transducers using the algorithms developed by Mans Hulden\footnote{\href{https://en.wikipedia.org/wiki/Mans\_Hulden}{https://en.wikipedia.org/wiki/Mans\_Hulden}} for his FOMA\footnote{\href{https://fomafst.github.io/}{https://fomafst.github.io/}} project, an open source application to compile finite state transducers.

The rules that were generated for the analyser, as well as the tags used in its outputs, are based on the description and study made by Dr. Ineke Smeets of \emph{Mapudüngun}, and published in her book "A Grammar of \emph{Mapuche}" [Smeets, I. 2008] \cite{RefB:21}\footnote{\href{https://www.degruyter.com/view/product/22765}{https://www.degruyter.com/view/product/22765}}.

\subsection{FOMA implementations} \label{sec:30}

\paragraph{} \label{tp:35} The path we follow is Annette Ríos work on Quechua, she have developed various tools, the main one being the finite state morphological analyser and generator [Ríos 2015] \cite{RefB:17}. 

Ríos' development for the analyser and generator was with XFST and other tools released by Xerox [Beesley \& Kartunnen 2003] \cite{RefB:01}. She used FOMA for the spell checker. We decided to use FOMA for everything, mainly because it is open source software, and we wanted to develop a totally free set of linguistic tools for \emph{Mapudüngun}.

FOMA is a very well-known software used by many linguists and computer developers for a wide range of tasks, but mainly for language applications. Searching on the Internet we have looked for the FOMA implementations\footnote{This information was compiled using Google Scholar only. Some publications mention more than one FOMA implementation, some times at different years; to simplify, we have counted the publications about FOMA implementations, and we have left out those that mention FOMA only as a reference. The results vary as the search is repeated.} per year, shown in table \ref{tab:04}:

FOMA has been widely used in Basque [Alegria et al. 2009] related tools, but also in a good amount of American aboriginal languages (Quechua [Rios. 2015]; Arapaho [Kazeminejad, Cowell, \& Hulden. 2017]; Nahuatl [Escobar. 2019]), Turkish [Yıldız, Avar, \& Ercan. 2019], Indonesian [Larasati et al. 2011], Japanese [Sim. 2013], Arabic [Attia, Al-Badrashiny, \& Diab. 2014], Kazakh [Kessikbayeva \& Cicekli. 2014] and others\footnote{There are many more FST implementations for different languages, with other available FST compilers; but as we have developed our own with FOMA, we just list those.}. We list here some of the implementations counted in table \ref{tab:04}.

\begin{itemize} \label{it:26}
	\item[] 2020. \emph{A Finite-State Morphological Analyser for Evenki}. Zueva, A., Kuznetsova, A. \& Tyers, F. (Indiana University).
	\item[]
	\item[] 2019. \emph{Improved Finite-State Morphological Analysis for St. Lawrence Island Yupik Using Paradigm Function Morphology}. Chen, E., Park, H. \& Schwartz, L. (University of Illinois Urbana-Champaign).
	\item[]
	\item[] 2018. \emph{Computational syntactic analysis of Setswana}. Berg, A. (North-West University. Johannesburg).
	\item[]
	\item[] 2017. \emph{Creating lexical resources for polysynthetic languages — the case of Arapaho}. Kazeminejad, A., Cowell, A. \& Hulden, M. (University of Colorado).
	\item[]
	\item[] 2016. \emph{ZeuScansion: A tool for scansion of English poetry}. Agirrezabal, M., Astigarraga. A., Arrieta, B. (University of the Basque Country) \& Hulden, M. (University of Colorado Boulder).
	\item[]
	\item[] 2015. \emph{A Basic Language Technology Toolkit for Quechua}. Ríos, A. (University of Zurich).
	\item[]
	\item[] 2014. \emph{GWU-HASP: Hybrid Arabic Spelling and Punctuation Corrector}. Attia, M., Al-Badrashiny, M. \& Diab, M. (The George Washington University).
	\item[]
	\item[] 2013. \emph{A Morphological Analyzer for Japanese Nouns, Verbs and Adjectives}. Sim, Y. (Carnegie Mellon University).
	\item[]
	\item[] 2012. \emph{Finite State Methods Applied to Hebrew Noun Patterns (Mishqalim)}. Rozenberg, F. (Eberhard Karls Universität. Tübingen, Germany).
	\item[]
	\item[] 2011. \emph{Matxin, an open-source rule-based machine translation system for Basque}. Mayor, A., Alegria, I., Díaz de Ilarraza, A., Labaka, G.,  Lersundi, M. \& Sarasola, K. (University of the Basque Country).
\end{itemize}

\begin{table}[htb]
	\caption{FOMA implementations per year (2011 to 2020)}
	\label{tab:04}
	\begin{tabular}{lllll}
		2020: 12 & 2019: 17 & 2018: 16 & 2017: 14 &	2016: 10 \\
		\noalign{\smallskip}
		2015: 11 & 2014: 13 & 2013: 09 & 2012: 12 & 2011: 13
	\end{tabular}
\end{table}

\subsection{The encoded \emph{Mapuche} alphabet} \label{sec:31}

\paragraph{} \label{tp:36} We present in this section the alternative representation of some graphemes in the analyser, alternative from those of Smeets. Finally, the rules implying variation in the spelling of words, generated either by elision, epenthesis or other similar phenomena.

As it also deals with orthography, this section brings up the subject of \nameref{sec:54} (see \ref{sec:54}, p. \pageref{sec:54}).

The sigma alphabet of the analyser includes every non-epsilon (not empty) symbol that appears in the network, either by itself or as a component of a symbol pair [Beesley \& Karttunen 2003: 57, 58] \cite{RefB:01}. For example, the sigma alphabet of the network compiled from [cat "+Noun":0] consists of the symbols a, c, t and +Noun [Beesley \& Karttunen 2003: 62] \cite{RefB:01}. Then, the sigma alphabet is a list of all the single symbols that occur either on the upper or lower side of the arcs [Beesley \& Karttunen 2003: 99] \cite{RefB:01} (see \ref{sec:26} \nameref{sec:26}, p. \pageref{sec:26}).

The analyser's sigma, among all the symbols it comprises, contains the \emph{Mapuche} alphabet:

\begin{quote} \label{note:04}
	{\small VOW [ a | e | i | o | u | ü ];\\
	SVW [ w | y | g ];\\
	CON [ \{ch\} | d | f | k | l | \{ll\} | m | n | ñ | \{ng\} | p | r | s | \{sh\} | t | \{tr\} ];\\\\
	\{VOW, SVW, CON\}  ${\in}$ $\Sigma $ of the analyser}
\end{quote}

In order to apply morphotactics (see \ref{sec:24} \nameref{sec:24}, p. \pageref{sec:24}) and rules, the alphabet has been divided into three groups:

\subsubsection{Vowels} \label{sec:32}

\paragraph{} \label{tp:37} For this group of characters (see table \ref{tab:02}, p. \pageref{tab:02}) Smeets' proposal was adopted without any variation. They conform the group named \texttt{VOW}.

It is worth to mention though, that "\emph{Mapudüngun} has six vowel phonemes, /\textsci~ë~\textturna~ö~\textupsilon~\textreve/... It should be noted that vowels of \emph{Mapudüngun} have traditionally been treated as the five vowels of Spanish (/i e a o u/), with identical stressed and unstressed allophones, plus a high central unrounded vowel /\textbari/ (commonly known as the 'sixth vowel') having a mid central allophone [\textschwa] in unstressed position" [Sadowsky, S. 2013] \cite{RefB:18}.

The previous paragraph comes to say that the grapheme \emph{ü} represents two sounds: /\textbari/ and  /\textschwa/ which realize in complementary contexts.

\subsubsection{Semivowels:} \label{sec:33}

\paragraph{} \label{tp:38} Smeets identifies them as glides (see table \ref{tab:01}, p. \pageref{tab:01}). She includes the \emph{r}, and counts the glides among the consonants, while in the analyser these are separated from the consonants to form the semi-vowels group, except for the \emph{r}. With this categorization we can treat certain phonological phenomena that involve these semi-vowels, such as elision and epenthesis (see \ref{sec:39}, p. \pageref{sec:39}). We could have called them glides as Smeets does, but following Beesley \& Kartunnen [2003] \cite{RefB:01}, we have called them semi-vowels. Anyway, they do occur as both, semi-vowels and glides\footnote{Glides immediately precede the vowel, semi-vowels immediately follow the vowel, both are less sonorous than the vowel.}.

The sound /{\textgamma}/ usually represented by a \emph{g}, Smeets represents it by a \emph{q} to reflect a difference in sound, which is softer than the one she represents using \emph{g}, for instance, in the loan from Spanish \emph{\textbf{g}ayeta} 'cookie', so written in Smeets' work. She spells the \emph{Mapuche} word for 'seven' as \emph{re\textbf{q}le}, while we spell it \emph{re\textbf{g}le}. This word and all those that Smeets spells with \emph{q} have been traditionally spelled with \emph{g}. This distinction does not affect the meaning of words, then \emph{g} spelling has been adopted by us to represent all the /{\textgamma}/ variants.

For the other two graphemes, \emph{w} and \emph{y}, no changes were made. All three are grouped under the \texttt{SVW} denomination.

\subsubsection{Consonants:} \label{sec:34}

\paragraph{} \label{tp:39} Contrary to Smeets, who states that there are 19 consonants, and because we have counted out the three glides identified in the previous paragraphs as semi-vowels, there are 16 recognized consonants.

Smeets uses \textcrd~ to represent the voiceless\footnote{Not in all dialects this sound is voiceless, prove of that is in the early transcriptions made by Jesuit monks, all of them used \emph{d}, which represents a voiced sound.} interdental fricative /\texttheta/ to make clear the difference with the d from Spanish loans. We use \emph{d} because this distinction does not affect the meaning of words.

For the interdental series \emph{t'}, \emph{n'}, \emph{l'}, we have eliminated the apostrophe to interpret these letters as the alveolar counterpart. Letters with apostrophe appear mainly in the older texts. In more recent texts they sometimes appear, even though not consistently. Some times the writer introduces an interdental in a word, later it does not, not even for the same word. Due to this misleading usage of the interdental variant and the "dying out" of the distinction [Smeets, I. 2008: 31] \cite{RefB:21}, the inclusion of interdental has been avoided for morphological generation, while they are accepted as variants in the analysis.

The elimination of apostrophe broadens the possibilities of analysis. The words and roots to be analysed are collected in the lexicon; if it appears just \emph{newen} 'force, strength' in the lexicon, the variant \emph{n'ewen'} would be taken as an unknown root, while by eliminating the apostrophe before analysis, it allows the transducer to analyse any introduced variant, either \emph{n'ewen}, \emph{n'ewen'}, \emph{newen'} or \emph{newen} as the same root.

Loaned sounds /b/ (bilabial, plosive, voiced), represented by b, and /x/ (velar, fricative, voiceless), represented by j, from Spanish, are not included in our system. They are converted into the corresponding letter of the \emph{Mapudüngun} alphabet by means of the spelling unifier (see \ref{sec:54} below); b into \emph{f} and j into \emph{k}, which are the most usual conversions we have found reflected in some dictionaries: jabón → \emph{kafon}; burro → \emph{furiku}, etc. [Febrés 1882\footnote{Consulted at \href{http://corlexim.cl}{http://corlexim.cl} <2019-07-11>. More examples are found in Febrés dictionary [1882] \cite{RefB:06}.}].

Other possible conversions are directly reflected in the lexicon: vaca → \emph{waka}\footnote{Today it is well-known that in Spanish there was never a difference in pronunciation of words spelled with b or v (see \href{http://lema.rae.es/dpd/srv/search?id=d45ahCOicD6TkHkns8}{http://lema.rae.es/dpd/srv/search?id=d45ahCOicD6TkHkns8} for more information), otherwise, probably the inclusion of vaca into \emph{Mapuche} lexicon should have been as \emph{faka}.}; vehículo → \emph{weikulo}; voto → \emph{woto}; jamón → \emph{kümon}; junio → \emph{kunio}, etc.

Within the consonants group, \texttt{CON}, note the representation for digraphs (\emph{\{ch\}, \{ll\}, \{ng\}, \{sh\}, \{tr\}}) in curly braces. This indicates that these two symbols together form a single representation of a sound. In other words, the concatenation of these two symbols is invariable and univocal.

\subsection{Intermediate representation symbols} \label{sec:35}

\paragraph{} \label{tp:40} These symbols are created to treat different morphophonological changes that occur in the language due to the interaction between suffixes, between roots, and among all of them together. It may be said that they are part of the alphabet, at least in an abstract level, because they are used to represent a certain stage in a change process, from which the surface form (the one we write or read) arises. A few of them are listed here as an example, the complete list of these symbols with their function is found in annex \ref{anx:09} \nameref{anx:09}, p. \pageref{anx:09}:

\begin{example} \label{ex:107}\\
    \texttt{"@G"} is used to treat epenthesis of glottal stops represented by \emph{g} in the spelling, see \ref{sec:59}, E\ref{ex:116}, E\ref{ex:117} and R\ref{R:01}, p. \pageref{ex:116}.
\end{example}

\begin{example} \label{ex:108}
    \texttt{"@Ü"} is used to treat schwa insertion represented by \emph{ü} in the spelling, see E\ref{ex:110}, E\ref{ex:111} and E\ref{ex:112}, p. \pageref{ex:112}.
\end{example}

\begin{example} \label{ex:109}
    \texttt{"@GK"} is used to treat radical alternation in some intransitive verbs which change their last consonant from \emph{g} to \emph{k} if they are in contact with the causative suffix \emph{-üm-}, which transitivizes them, see D\ref{def:19} and D\ref{def:19}, p. \pageref{def:19}.
\end{example}

\subsection{Roots encoding} \label{sec:36}

\paragraph{} \label{tp:41} In the FST, this section headline is "Read in roots", because the system reads the files containing the roots lexicon and incorporate them into the analyser. Lists are separated by part of speech (grammatical categories).

\begin{definition} \label{def:01} \\
    \texttt{define AVROOT @re"roots/avroot.lex"; \\
    define NROOT @re"roots/nroot.lex";}
\end{definition}
	
The expression above introduces the lists of roots for adverb and noun categories, so any listed noun may be found throughout "\texttt{NROOT}", which encodes nouns as shown below:

\begin{definition} \label{def:02} Sample of nouns lexicon:\\
	\texttt{|["‑NN"\{.aylen\_brasa\}]: ["@G"\{aylen\}]\\
		|["‑NN"\{.aywiñ\_sombra\}]: ["@G"\{aywiñ\}]\\
		|["‑NN"\{.chachay\_papá-afectuoso\}]: [\{chacha\}|\\\{chachay\}|\{tatay\}]\\
		|["‑NN"\{.chadi\_sal\}]: \{chadi\}\\
		|["‑NN"\{.chaf\_cáscara\_piel-de-frutas\}]: \{chaf\}\\
		|["‑NN"\{.chafid\_bagazo)\}]: \{chafid\}}
\end{definition}

\begin{quote} \label{note:05}
	{\small In D\ref{def:02} \texttt{"‑NN"} stands for (simple) noun or nominal (root). \texttt{"@G"} is an intermediate language tag to treat glottal stop insertion, see \ref{sec:59}, E\ref{ex:116}, E\ref{ex:117} and R\ref{R:01}, p. \pageref{ex:116}.}
\end{quote}

In the same way D\ref{def:01} shows noun and adverb roots encoding, there is a file per part of speech and other forms (verbs, adjectives, particles, interjections, etc.).

We have divided the lexicon in two major groups: the roots that may be verbalized, and the forms that can not be verbalized. Among the roots, all except for the verb roots may be independent (without suffixes) words.

\begin{enumerate} \label{it:27}
	\item[] Roots (verbalizable)
	\item File \texttt{ajroot.lex}: adjectives
	\item File \texttt{avroot.lex}: adverbs
	\item File \texttt{ivroot.lex}: intransitive verb roots
	\item File \texttt{names.lex}: proper nouns
	\item File \texttt{nroot.lex}: nouns
	\item File \texttt{nuroot.lex}: numerals
	\item File \texttt{onroot.lex}: onomatopoeia
	\item File \texttt{qroot.lex}: question forms
	\item File \texttt{tvroot.lex}: transitive verb roots
\end{enumerate}

\begin{enumerate} \label{it:28}
	\item[] Other lexicon (non-verbalizable)
	\item File \texttt{adverb.lex}: adverbs
	\item File \texttt{anaphpr.lex}: anaphoric pronouns
	\item File \texttt{auxv.lex}: auxiliary verbs
	\item File \texttt{conj.lex}: conjunctions
	\item File \texttt{dempr.lex}: demonstrative pronouns
	\item File \texttt{forexp.lex}: foreign expressions (Spanish loans)
	\item File \texttt{intpr.lex}: interrogative pronouns
	\item File \texttt{itj.lex}: interjections
	\item File \texttt{neg.lex}: negation particles
	\item File \texttt{numbers.lex}: numbers
	\item File \texttt{part.lex}: particles
	\item File \texttt{perspr.lex}: personal pronouns
	\item File \texttt{posspr.lex}: possessive pronouns
	\item File \texttt{prep.lex}: prepositions
	\item File \texttt{punct.lex}: punctuation marks
\end{enumerate}

\nameref{anx:01} is found in annex \ref{anx:01}. There is a list of tags assigned to every part of speech and suffix with the name (in Spanish) identifying them, on:\\ \href{http://www.chandia.net/glosas-del-dungupeyem}{http://www.chandia.net/glosas-del-dungupeyem}

\subsection{Suffixes encoding} \label{sec:37}

\paragraph{} \label{tp:42} As every suffix is assigned to a slot, which in turn is encoded in a file, the addition of such information to the main script is carried out by calling these files under the script section "Read in slots". Each file contains the fillers of the corresponding slot. For example:

\begin{definition} \label{def:03}\ \\
	\texttt{define NEG @re"slots/1-15-Inflectional-Suffixes\\/slot‑10.aff";}
\end{definition}

\begin{figure}[H]
	\includegraphics[width=\columnwidth]{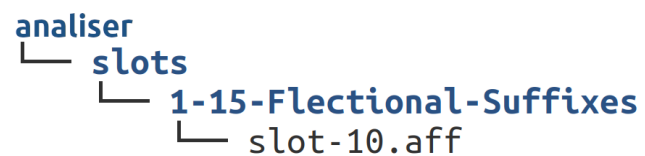}
	\caption{Directory tree structure for the location of a suffix file.}
	\label{fig:05}
\end{figure}

D\ref{def:03} is the definition of an expression named "\texttt{NEG}", which in turn, is composed by the regular expressions stored in the text file\\"\texttt{slot-10.aff}", for which its complete location route is indicated: "\texttt{slots/1-15‑Flexive‑Suffixes/}" [See figure \ref{fig:05}].

This file, deployed in D\ref{def:04} below, contains negation suffixes filling slot 10:

\begin{definition} \label{def:04} \\
	\texttt{["+NEG"{.la10}]: \{la\}\\
		| ["+NEG"\{.ki10\}]: "@NK"\\
		| ["+NEG"\{.no10\}]: [\{no\}|\{nu\}]\\
		| ["+NEG"\{.kino10\}]: [\{kino\}|\{kinu\}];}\footnote{This regexp is written in only one line, but for the sake of better understanding each of its parts is exposed in separated lines.}
\end{definition}

D\ref{def:04} expresses four suffixes filling slot 10. They are separated by the pipe (vertical bar | ) symbol. Every suffix section indicates its upper or abstract level (right of the colon) and lower or surface level (left of the colon). Every abstract form is returned in the analysis process after a cleaning task leaves only the category tag, the suffix form and the slot number: "\texttt{+NEG.no10}". These same forms, before being cleaned up, are used to apply morphotactics in the abstract level at analysis and generation processes. At the surface level we find the forms given by analysis and/or the tags that trigger a process of replacement, like in the case of \texttt{"@NK"} (in D\ref{def:04} above) which further down the script has a rule replacing it for certain form in a defined context.

The complete list of files per slot containing the suffixes is given in annex \ref{anx:02}, p. \pageref{anx:02}.

\section{The \emph{Mapudüngun} FST analyser} \label{sec:38}

\paragraph{} \label{tp:43} In this section we expose and explain how the different aspects of \emph{Mapudüngun} morphology are treated in order to implement our FST analyser and generator. The order of presentation is not necessarily the same as in the encoding script. There are some encoding already presented in previous sections: the \emph{Mapuche} alphabet [section \ref{sec:31}, p. \pageref{sec:31}]; the spelling unifier [section \ref{sec:54}, p. \pageref{sec:54}]; the lexicon's inclusion [section \ref{sec:36}, p. \pageref{sec:36}], and the suffixes' inclusion [section \ref{sec:37}, p. \pageref{sec:37}].

We begin by explaining phonological changes (\ref{sec:39}), including special cases (see \ref{sec:40}, \ref{sec:41}, \ref{sec:42}, p. \pageref{sec:40}). Then we move on to the stems typology and the strategies to manage them. The interaction of suffixes after the stem comes next, this section introduces the verb paradigms and verb nominalization. It also brings up the subject of the mobility of some suffixes and how to deal with it.

Some verb roots present a special behaviour, we treat them at the end of this section.

\subsection{Phonological changes into spelling} \label{sec:39}

\paragraph{} \label{tp:44} The occurrence of roots and suffixes in a verb form generates certain phonological changes when interacting with their neighbours suffixes or roots. In \emph{Mapudüngun} that interaction may be between suffixes, between the root and the consecutive suffix, between roots inside a compound stem, or between all the previous ones inside complex stems.

These changes are encoded in the "lower" or "surface" side of the language (see sections \ref{sec:26} and \ref{sec:27}, p. \pageref{sec:26}), therefore, they are under the section "Lower rules" in the FST script, and they affect the word form.

There are frequent phonological changes in \emph{Mapudüngun}, which are reflected in spelling, except for the epenthesis of the voiced velar fricative  /{\textgamma}/ represented by \emph{g}. Which is optional between the sequences \emph{ii, uu, üü}, and obligatory between the sequences \emph{ae, ea, ee, ai, ia}, where the first vowel of the sequence is the last of a suffix, and the second vowel of the sequence is the first of a consecutive suffix\footnote{In Smeets' texts, this /{\textgamma}/ is never transcribed, but in texts from other authors is sometimes present: \emph{kellu-ke-(\textbf{g})e-n} 'you helped me' \texttt{-NN.kellu\_ayuda+VRB.Ø36+CF.ke14+IDO.e6+IND1SG.n3\\+DS12A.Ø1} [Mösbach, E. \cite{RefB:14}].}.

The epenthesis of schwa (sometimes \emph{i, e, u}) is obligatory between a form (root or suffix) ending in consonant or semi-vowel and the following form beginning in consonant or semi-vowel.

\begin{example} \label{ex:110} [Smeets, I. 2008: 353 (43)] \cite{RefB:21}\\
	\emph{kim-\textbf{u}w-küle-y-iñ} 'we (pl) know each other'\\
	\texttt{‑TV.kim\_saber+REF.w31+ST.le28+IND.y4+1.Ø3\\+PL.iñ2}
\end{example}

\begin{example} \label{ex:111} [Smeets, I. 2008: 68 (56)] \cite{RefB:21}\\
	\emph{puw-\textbf{ü}n} 'to arrive'\\
	\texttt{‑IV.puw\_llegar+PVN.n4}
\end{example}

\begin{example} \label{ex:112} [Smeets, I. 2008: 109 (1)] \cite{RefB:21}\\
	\emph{mamüll-\textbf{e}ntu} 'grove'
	\texttt{‑NN.mamüll\_árbol+GR.ntu}
\end{example}

"A schwa is optionally inserted between a consonant and the suffix sequence \emph{-l-e} \texttt{+CND.l4+3.e3} and between a consonant and the suffix sequence \emph{-y-iñ} \texttt{+IND.y4+1.Ø3+PL.iñ2} and \emph{-y-u} \texttt{+IND.y4+1.Ø3+DL.u2}" [Smeets, I. 2008: 51].\\Sometimes it is \emph{i} instead of schwa.

\begin{example} \label{ex:113} [Smeets, I. 2008: 52] \cite{RefB:21}\\
	\emph{kim-l-e → kim-\textbf{ü}l-e} 'if he knows'\\
	\texttt{‑TV.kim\_saber+CND.l4+3.e3}
\end{example}

\begin{example} \label{ex:114}\ \\
	\emph{lef-y-u → lef-\textbf{ü}y-u} 'we both ran'\\
	\texttt{‑IV.lef\_correr+IND.y4+1.Ø3+DL.u2}
\end{example}

\begin{example} \label{ex:115} [Smeets, I. 2008: 52] \cite{RefB:21}\\
	\emph{lef-y-iñ → lef-\textbf{i}y-iñ} 'we (pl) ran'\\
	\texttt{‑IV.lef\_correr+IND.y4+1.Ø3+PL.iñ2}
\end{example}

Lexical forms must be collected in an intermediate form with the appropriate tags to later apply the rules transforming them into the final surface forms, e.g., the portmanteau suffix for indicative, 1\textsuperscript{st} person, singular may occur as \emph{-ün-, -üñ-, -n-} or \emph{-ñ-}. It is encoded as:

\begin{definition} \label{def:05}
	\texttt{["+IND1SG"\{.n3\}] : ["@Ü"[n|ñ]];}
\end{definition}

The intermediate form is encoding two variants: the tag \texttt{"@Ü"} followed by \emph{n} or \emph{ñ}.

The rules to generate the four forms mentioned above are:

\begin{definition} \label{def:06}
	\texttt{["@Ü" -> ü || [CON|SVW|.\#.] \_ ] \newline .o. ["@Ü" -> 0 || VOW \_ ];}
\end{definition}

These are two rules concatenated by the composition symbol \texttt{.o.} (see figure \ref{fig:02}). They are applicable not only to this case, but wherever the tag \texttt{"@Ü"} is found. The first rule indicates that the tag \texttt{"@Ü"} is replaced by \emph{ü} in the context (twin pipes \texttt{||} indicate the context) where a consonant or semi-vowel precedes it, or when it is at the beginning of the word (\texttt{.\#.} "word boundary character"), the tag position being marked by the underscore "\texttt{\_}". The second rule indicates that the tag is deleted if preceded by a vowel. In combination with the "intermediate representation", the FST compiles the four possible forms for this suffix.

The suffixes affected by these rules are the following ones:

\begin{itemize} \label{it:30}
	\item More implicated object (s29) \emph{-l-} \texttt{["@Ü"l]}
	\item Satisfaction (s25) \emph{-ñmu-} \texttt{["@Ü"\{ñmu\}]}\footnote{Some suffixes have multiple forms, as the case presented in D\ref{def:05} and D\ref{def:06}; or satisfaction \emph{-ñmu-} that it may also be \emph{-ñmo-}. For simplicity, we present here only the most common form of the suffix.}
	\item Interruptive (s18) \emph{-r-} \texttt{["@Ü"r]}
	\item Reportative (s12 \& NCC) \emph{-rke-} \texttt{["@Ü"\{rke\}]}
	\item Conditional (s4) \emph{-l-} \texttt{["@Ü"l]}
	\item Plain verbal noun (s4) \emph{-n-} \texttt{["@Ü"n]}
	\item Indicative 1\textsuperscript{st} singular (s3) \emph{-n-} \texttt{["@Ü"n]}
	\item Adjectiviser quick \& easy (NOM) \emph{-nten-} \texttt{["@Ü"\{nten\}]}
\end{itemize}

Some suffixes may trigger an \emph{-ü-} or an \emph{-u-} (the first two), or only an \emph{-u-} (the last two):

\begin{itemize} \label{it:31}
	\item Pluperfect (s15) \emph{-wye-} \texttt{[["@Ü"|"@U"]\{wye\}]}
	\item Completive subjective verbal noun (s4) \emph{-wma-}\\ \texttt{[["@Ü"|"@U"]\{wma\}]}
	\item Reflexive/reciprocal (s31) \emph{-w-} \texttt{["@U"w]}
	\item 1\textsuperscript{st} person agent (s23) \emph{-w-} \texttt{["@U"w]}
\end{itemize}

Tags are created and assigned arbitrarily. For the rules to work it is necessary to place the corresponding tag in the appropriate position. As the rule is applied to the surface level, the tag is placed at that same level, which is encoded to the right of the colon, e.g.:

\begin{definition} \label{def:07}
	\texttt{["+PLPF"\{.üwye15\}] : ["@Ü"\{wye\}];}
\end{definition}

The encoding above incorporates the pluperfect marker into the system. To the left of the colon is the upper or abstract level, the analysis representation. To the right, it is the lower level, the surface representation, where the tag is added preceding the suffix as the initial character. Once the tag is appropriately replaced, there is another rule that eliminates the unused tags.

Epenthesis of a glottal stop is optional between the ending vowel of a root and the initial vowel of a following root in compounds.

\begin{example} \label{ex:116} [Smeets, I. 2008: 52] \cite{RefB:21}\\
	\emph{dewma-iyal-la-y → dewma-\textbf{g}iyal-la-y} 'he did not prepare food'\\
	\texttt{‑TV.dewma\_hacer-N.iyal}\footnote{\emph{iyal} is a lexicalized form for "food" that may be analysed as \texttt{-TV.i\_comer+NRLD.a9+OVN.el4} 'what will be eaten'.}\texttt{\_comida‑CR.TV+NEG.la10\\+IND.y4+3.Ø3}
\end{example}

To apply the rule described above, we have marked, in the lower level, all the roots beginning in vowel, placing a "@G" before the root; for example:

\begin{example} \label{ex:117}
	\item[]\texttt{["-AJ"\{.allush\_tibio\}]: ["@G"\{allush\}]}
	\item[]\texttt{["-AJ"\{.awka\_rebelde\}]: ["@G"\{awka\}]}
	\item[]\texttt{["-AV"\{.aymüñ\_bastante\}]: ["@G"\{aymüñ\}]}
	\item[]\texttt{["-AV"\{.ina\_cerca\}]: ["@G"\{ina\}]}
	\item[]\texttt{["-IV"\{.echiw\_estornudar\}]: ["@G"\{echiw\}]}
	\item[]\texttt{["-IV"\{.uma\_dormir\}]: ["@G"\{uma\}]}
	\item[]\texttt{["-NN"\{.antü\_sol\}]: ["@G"\{antü\}]}
	\item[]\texttt{["-NN"\{.epew\_cuento\}]: ["@G"\{epew\}]}
	\item[]\texttt{["-TV"\{.ingka\_defender\}]: ["@G"\{ingka\}]}
	\item[]\texttt{["-TV"\{.üngüm\_esperar\}]: ["@G"\{üngüm\}]}
\end{example}

The rules encoding this change are the next ones:

\begin{exercise} \label{R:01} \textbf{Glottal stop insertion (between roots)} \\
    \texttt{define RuGlot ["@G" (->) g || VOW \_ ]\\ .o. ["@G" -> 0];} \end{exercise}

The first rule, to the left of the concatenation (\texttt{.o.}) symbol, encodes the optionality by enclosing the direction operator between parenthesis. The change applies only when the tag is preceded by a vowel, otherwise, the second rule transforms it into 0 (zero = null character), i.e., it is deleted.

A glottal stop is also optionally added between suffixes, but only in the cases where the sequence \emph{ii} is generated. Instead of adding a tag to the suffixes, this change is treated with a more general rule:

\begin{exercise} \label{R:02} \textbf{Glottal stop insertion (between suffixes)} \\
	\texttt{define RuleEliGem}\footnote{This is a wider rule that treats elision, epenthesis and gemination of other phonemes, see R\ref{R:03}.} \texttt{[\{ii\} (->) \{igi\}];} \end{exercise}

Applying this rule we can analyse words like the following one:

\begin{example} \label{ex:118} [Smeets, I. 2008: 51] \cite{RefB:21}\\
	\emph{leli-l-i-iñ → leli-l-i-\textbf{g}iñ} 'if we look'\\
	\texttt{‑TV.leli\_mirar+CND.l4+1.i3+PL.iñ2}
\end{example}

Two equal consonants geminate in careful speech, and became a single sound in colloquial speech. And this fact is sometimes transcribed into written text.

\begin{example} \label{ex:119} [Smeets, I. 2008: 51] \cite{RefB:21}\\
	\emph{ko\textbf{n-n}u-l-i → ko\textbf{n}u-l-i} 'if I do not enter'\\
	\texttt{‑IV.kon\_entrar+NEG.no10+CND.l4+1.i3+SG.Ø2}
\end{example}

To treat gemination we use the same type of rule shown in R\ref{R:02}. The following rule not only encodes the gemination of \emph{n}, but also of \emph{m} and \emph{e}\footnote{In the case of \emph{e} the phonological term is not gemination but lengthening.}:

\begin{exercise} \label{R:03} \textbf{Gemination simplification} \\
	\texttt{define RuleEliGem [\{nn\} (->) n, \\ \{mm\} (->) m, \{ee\} (->) e];}
\end{exercise}

\begin{example} \label{ex:120} [Smeets, I. 2008: 437 (17)] \cite{RefB:21}\\
	\emph{fille\textbf{m-m}ew → fille\textbf{m}ew} 'in every respect'\\
	\texttt{-NN.fillem\_toda-clase-de-cosas+INST.mew}
\end{example}

\begin{example} \label{ex:121} [Smeets, I. 2008: 278 (1)] \cite{RefB:21}\\
	\emph{nie-\textbf{e}-y-u → ni\textbf{e}-y-u} 'I hold you (sg)'\\
	\texttt{-TV.nie\_tener+IDO.e6+IND.y4+1.Ø3+DL.u2\\+DS12A.Ø1}
\end{example}

The non-realized affix \emph{-a-} (s9) separates itself from a preceding \emph{a-} inserting \emph{-y} in between. In this case we have an intermediate representation of the suffix (D\ref{def:08}), and a couple of rules treating the corresponding tag \texttt{"@Y"} in context (R\ref{R:04}):

\begin{definition} \label{def:08} \texttt{["+NRLD"\{.a9\}]: ["@Y"a];} \end{definition}

\begin{exercise} \label{R:04} \textbf{\emph{y} epenthesis} \\
	\texttt{define RuTrEPENTHy ["@Y" (->) y ||[a|.\#.] \_] \\.o. ["@Y" -> 0 || \textbackslash a}\footnote{Term negation (\textbackslash X).	Any single symbol except X. Equivalent to [?~-~X] [Hulden, M. in \href{https://code.google.com/archive/p/foma/wikis/RegularExpressionReference.wiki}{https://code.google.com/archive/p/foma/wikis/\\RegularExpressionReference.wiki}].} \texttt{\_ ];}
\end{exercise}

This concatenated rule says that \texttt{"@Y"} becomes \emph{y} when preceded by an \emph{a} or at the beginning of the word, and it becomes a null character when preceded by any character except \emph{a}.

\begin{example} \label{ex:122} [Smeets, I. 2008: 63 (19)] \cite{RefB:21}\\
	\emph{tripa-a-n → tripa-\textbf{y}a-n} 'I will leave'\\
	\texttt{‑IV.tripa\_salir+NRLD.a9+IND1SG.n3}
\end{example}

Pronouns \emph{engu} 'they (dl)' and \emph{engün} 'they (pl)' are optionally realized as \emph{yengu} and \emph{yengün} respectively, either isolated or forming part of a compound where the previous element ends in vowel. Here also applies definition D\ref{def:08}.

\begin{example} \label{ex:123} [Smeets, I. 2008: 95 (ii)] \cite{RefB:21}\\
	\emph{tüfa-\textbf{y}engu} 'these two'\\
	\texttt{-DP.tüfa\_este-PP.engu\_ellos-dos}
\end{example}

The sequence \emph{ae} is optionally simplified as \emph{a}.

\begin{example} \label{ex:124} [Smeets, I. 2008: 52] \cite{RefB:21}\\
	\emph{leli-l\textbf{a-e}-y-u → leli-l\textbf{a}-y-u} 'I shall not look at you (sg)'\\
	\texttt{‑TV.leli\_mirar+NEG.la10+IDO.e6+IND.y4+1.Ø3\\+DL.u2+DS12A.Ø1}
\end{example}

\begin{example} \label{ex:125} [Smeets, I. 2008: 52] \cite{RefB:21}\\
	\emph{i-me-\textbf{a-e}l → i-me-\textbf{a}-l} 'eat there!' (lit.: 'you will eat there')\\
	\texttt{‑TV.i\_comer+TH.me20+NRLD.a9+OVN.el4}
\end{example}

The sequence \emph{ae} is never simplified when \emph{a} is followed by \emph{-e-n} \texttt{+IDO.e6+IND1SG.n3+DS12A.Ø1} or by \emph{e-n-ew}\\ \texttt{+IDO.e6+IND1SG.n3+DS3A.ew1}.

\begin{example} \label{ex:126} [Smeets, I. 2008: 48] \cite{RefB:21}\\
	\emph{elu-\textbf{a-e}-n} 'you (sg) will give to me'\\ \texttt{‑TV.elu\_dar+NRLD.a9+IDO.e6+IND1SG.n3+DS12A.Ø1}
\end{example}

\begin{example} \label{ex:127} [Smeets, I. 2008: 485 (4)] \cite{RefB:21}\\
	\emph{ayü-l\textbf{a-e}-n-ew} 'she did not love me'\\ \texttt{‑TV.ayü\_amar+NEG.la10+IDO.e6+IND1SG.n3\\+DS3A.ew1}
\end{example}

\subsubsection{Interaction between suffixes of slots 10 to 4}

\paragraph{} \label{tp:45} Most common suffixes in a verb form are those located between slots 10 and 4. There are multiple morphophonological changes depending on the suffixes occurring. The sequence \emph{a-e} mentioned above is only one of them.

All the suffixes triggering morphophonological changes that need special rules and belonging to this series of slots are encoded as follows:

\begin{definition} \label{def:09} Slot 10: Negation\footnote{Some slots of this series encode more suffixes than the ones displayed here, which are the ones having relevance for the rules generated and explained in this section.}\\
	\texttt{["+NEG"\{.la10\}] : \{la\}\\| ["+NEG"\{.ki10\}] : [k"@NK"];}
\end{definition}

\begin{definition} \label{def:10} Slot 9: Non-realized situation\\
	\texttt{["+NRLD"\{.a9\}]: ["@Y"a];}
\end{definition}

\begin{definition} \label{def:11} Slot 8: Impeditive\\
	\texttt{["+IPD"\{.fu8\}] : [f"@IP"];}
\end{definition}

\begin{definition} \label{def:12} Slot 6: Internal and external direct objects\\
	\texttt{["+EDO"\{.fi6\}] : "@ED"\\| ["+IDO"\{.e6\}] : "@ID";}
\end{definition}

\begin{definition} \label{def:13} Slot 4: inflectional nominalizers\\
	\texttt{["+OVN"\{.el4\}] : "@EL";}
\end{definition}

\begin{definition} \label{def:14} Slot 3PTMT: Portmanteau morphs\\
	\texttt{["+IND1SG"\{.n3\}] : ["@Ü"[n|ñ];}
\end{definition}

The following set of rules deals with the suffixes shown above:

\begin{exercise} \label{R:05} \textbf{Negation for imperative forms} (\emph{ki → k}) \\
	\texttt{define RuNegKi ["@NK"(->)[i|e] || \_[e|"@ID"]]\\ .o. ["@NK" -> i]}
\end{exercise}

From slot 10, there are two negation suffixes that can play a role in this set of interconnected rules: one is the negation for indicative \emph{-la-}, the final \emph{a} has to be taken into account when interacting with \texttt{+IDO} suffix \emph{-e-}, slot 6. The other negation suffix is \emph{-ki-} for imperatives. It may drop its final \emph{i}, or replace it by \emph{e} when followed by \emph{e}, which is also the form of the \texttt{+IDO} suffix encoded as \texttt{"@ID"} in the intermediate representation. This is what previous rule R\ref{R:05} manages. "When the negative marker \emph{-ki-} (slot 10) is followed by \emph{e}, the sequence \emph{ie} is optionally replaced by \emph{ee} or contracted to \emph{e}" [Smeets, I. 2008: 52 (8.1.4.3)] \cite{RefB:21}.

\begin{example} \label{ex:128} [Smeets, I. 2008: 52 (8.1.4.3)] \cite{RefB:21}\\
	\emph{sungu-we-k\textbf{i-e}-l-i} $\sim$ \emph{sungu-we-k\textbf{e-e}-l-i} $\sim$ \emph{sungu-we-k-\textbf{e}-l-i}\\ 'don't speak to me any more'\\ \texttt{-NN.düngu\_palabra+VRB.Ø36+PS.we19\\+NEG.ki10+IDO.e6+CNI.l4+1.i3+SG.Ø2+DS12A.Ø1}
\end{example}

Another suffix that has the same conditions of interaction as the negation suffix \emph{-la-}, with \texttt{+IDO} suffix, is \texttt{+NRLD} suffix \emph{-a-}, slot 9. This suffix may also be realized as \emph{-ya-} (see D\ref{def:08} and R\ref{R:04}). We will recall \emph{-la-} \texttt{+NEG} and \emph{-a-} \texttt{+NRLD} further down in R\ref{R:10}.

The occurrence of the suffixes sequence \texttt{+IPD} \emph{-fu-} and \texttt{+EDO} \emph{-fi-} may yield \emph{-fufi-} or \emph{-fwi-} in Smeets' texts, but also \emph{-fui-} in some other texts.

\begin{example} \label{ex:129} [Smeets, I. 2008: 39 (c)] \cite{RefB:21}\\
	\emph{angkad-\textbf{fu-fi}-n} → \emph{angkad-\textbf{fwi}-n} 'the one I had taken on the back of my horse'\\ \texttt{‑TV.angkad\_llevar-en-ancas+IPD.fu8+EDO.fi6\\+IND1SG.n3}
\end{example}

\begin{example} \label{ex:130} [Mösbach, E. 1936: 16] \cite{RefB:14}\\
	\emph{kim-la-\textbf{fu-fi}-ñ} → \emph{kim-la-\textbf{fui}-ñ} 'I didn't know (about) that'\\ \texttt{‑TV.kim\_saber+NEG.la10+IPD.fu8+EDO.fi6\\+IND1SG.n3}
\end{example}

\begin{exercise} \label{R:06} \textbf{Impeditive + EDO}\\
	\texttt{define RuTrIPDEDO 
		\item[] [["@IP" -> [u"@1"|w"@2"] || \_ "@ED"] .o.
		\item[] ["@ED" -> [{fi}|i] || "@1" \_ ] .o.
		\item[] ["@ED" -> i || "@2" \_ ]];}
\end{exercise}

R\ref{R:06} encodes the changes exemplified in E\ref{ex:129} and E\ref{ex:130}. \texttt{RuTrIPDEDO} is composed of three concatenated rules. First sub-rule states that \texttt{"@IP"} is either transformed into \texttt{"u@1"} or \texttt{"w@1"} when followed by \texttt{"@ED"}, which is expressed in the realization context (the underscore marks the position of the treated element): \texttt{|| \_ "@ED"}. It must be taken into account that rules are applied sequentially, so, when the transformation of \texttt{"@IP"} is carried out, this very same tag, which indicates the context for the subsequent change of \texttt{"@ED"}, is lost (it has been transformed in something else). This is why when changing \texttt{"@IP"}, new context marks are given for the subsequent \texttt{"@ED"} change. These new tags (\texttt{"@1"} and \texttt{"@2"}) are only used as context marks to continue processing the forms. In a later step, context tags are cleared out.

Second sub-rule of \texttt{RuTrIPDEDO}, (concatenated by \texttt{.o.}), states that \texttt{"@ED"} is either transformed into \emph{fi} or \emph{i} when preceded by \texttt{"@1"}.

Third sub-rule states that \texttt{"@ED"} is replaced only by \emph{i} when preceded by \texttt{"@2"}. \texttt{"@1"} and \texttt{"@2"} were established as contextual marks by the previous rule. Rules are applied sequentially.

Whenever \texttt{+IPD} (impeditive) is followed by \texttt{+EDO} (external direct object), the context is given, so the rule is applied accepting three combinations for analysis.

The process just described is illustrated in figure \ref{fig:06} below. It is important to be aware that it shows the generation direction because it is easier to explain and understand. Also note that the FSTs are reversible, therefore rules may be applied backwards, i.e., in the analysis direction. The resulting analysis for any of the three possible spellings \emph{-fufi-, -fwi-, -fui-} will yield the analysis \texttt{"+IPD.fu8+EDO.fi6"}.

\begin{figure}[H]
	\includegraphics[width=\columnwidth]{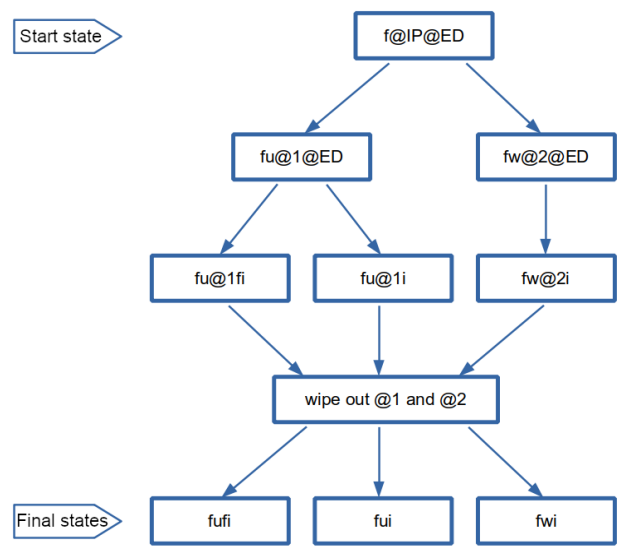}
	\caption{Concatenated rules for generation process: simplified view of rule \texttt{RuTrIPDEDO} that processes the interaction between suffixes impeditive \emph{-fu-} and external direct object \emph{-fi-}.}
	\label{fig:06}
\end{figure}

When \texttt{+IPD} \emph{-fu-} occurs followed by \texttt{+IDO} \emph{-e-} they yield the form \emph{-fue-}, but they may optionally yield the contracted form \emph{-fe-}.

\begin{example} \label{ex:131} [Smeets, I. 2008: 52 (8.1.6)] \cite{RefB:21}\\
	\emph{ellka-l-ke-rke-\textbf{fu-e}-y-ew} → \emph{ellka-l-ke-rke-\textbf{f-e}-y-ew} 'she used to hide it, they say'\\ \texttt{‑TV.ellka\_ocultar+CA.l34+CF.ke14+REP.rke12\\+IPD.fu8+IDO.e6+IND.y4+3.Ø3+DS3A.ew1}
\end{example}

\texttt{+IPD} and \texttt{+OVN} may co-occur in a sequence yielding \mbox{\emph{-fu-el}} or \emph{-f-el}. Smeets gives no examples to this respect, but she writes "The suffix \emph{-fu-} may occur in indicative and conditional forms and in subordinates except those marked with the plain verbal noun suffix \emph{-n} \texttt{+PVN} (s4) or the completive subjective verbal noun suffix \emph{-wma} \texttt{+CSVN} (s4)" [Smeets, I. 2008: 231] \cite{RefB:21}. Then, the sequence \texttt{+IPD} \texttt{+OVN} is feasible, and we do have found examples where \emph{-el} contracts with a previous form ending in \emph{e}, and examples of \emph{-fu-el} from other authors:

\begin{example} \label{ex:132} [Smeets, I. 2008: 245 (10)] \cite{RefB:21}\\
	\emph{ni\textbf{e-e}l} → \emph{ni\textbf{e}l} 'to have had'\\ \texttt{-TV.nie\_tener+OVN.el4}
\end{example}

\begin{example} \label{ex:133} [Smeets, I. 2008: 249 (7)] \cite{RefB:21}\\
	\emph{küdaw-p\textbf{e-e}l} → \emph{küdaw-p\textbf{e}l} 'the own job'\\ \texttt{-IV.küdaw\_trabajar+PX.pe13+OVN.el4}
\end{example}

\begin{example} \label{ex:134} [Smeets, I. 2008: 411 (53)] \cite{RefB:21}\\
	\emph{fende-k\textbf{e-e}l-chi} → \emph{fende-k\textbf{e}l-chi} 'a sold thing'\\ \texttt{-TV.fende\_vender+CF.ke14+OVN.el4+ADJ.chi}
\end{example}

\begin{example} \label{ex:135} [Zúñiga, F. 2006: 144 (54)] \cite{RefB:21}\\
	\emph{nge-we-ke-no-f\textbf{u-e}l} → \emph{nge-we-ke-no-f\textbf{e}l} 'to be no more'\\ \texttt{-IV.nge\_ser\_estar+PS.we19+CF.ke14+NEG.no10\\+IPD.fu8+OVN.el4}
\end{example}

\begin{exercise} \label{R:07} \textbf{Impeditive + Internal Direct Object or \\ Objective Verbal Noun} \\
	\texttt{define RuTrIPDIDOOVN \newline [["@IP" -> [[u"@3"]|"@3"] || \_ ["@ID"|"@EL"]] \\ .o. ["@ID" -> e, "@EL" -> {el} || "@3" \_ ]];}
\end{exercise}

R\ref{R:07} encodes the changes exemplified in E\ref{ex:131} and E\ref{ex:135}. \texttt{RuTrIPDIDOOVN} is composed of two concatenated rules. The first rule states that \texttt{"@IP"} is either transformed into \texttt{"u@3"} or \texttt{"@3"} (the last one being the elision of \emph{u}) when followed by \texttt{"@ID"} or \texttt{"@EL"}. Once that conversion is done, if the sequence is completed by \texttt{"@ID"}, this one is transformed into \emph{e}, yielding two possible intermediate representations, \texttt{fu@3e} or \texttt{f@3e}. If the sequence is completed by \texttt{"@EL"}, this one is transformed into \emph{el}, also yielding two possible intermediate representations, \texttt{fu@3el} or \texttt{f@3el}. Then \texttt{"@3"} is wiped out giving \emph{fue} or \emph{fe} for \texttt{"+IPD+IDO"}, and \emph{fuel} or \emph{fel} for \texttt{"+IPD+OVN"}.

R\ref{R:08} converts \texttt{"@IP"} into \emph{u} in any other context, yielding \emph{fu}, as shown in E\ref{ex:136}:

\begin{exercise} \label{R:08} \textbf{Impeditive} \\
	\texttt{define RuTrIPD ["@IP" -> u];}
\end{exercise}

\begin{example} \label{ex:136} [Smeets, I. 2008: 63 (17)] \cite{RefB:21}\\
	\emph{kutran-\textbf{fu}-n} 'to have been ill'\\ \texttt{-NN.kütran\_enfermedad+VRB.Ø36+IPD.fu8\\+IND1SG.n3}
\end{example}

R\ref{R:09} converts \texttt{"@ED"} into \emph{fi} when is preceded by anything but \texttt{"@IP"}, as shown in E\ref{ex:137}:

\begin{exercise} \label{R:09} \textbf{External direct object} \\	 
	\texttt{define RuTrEDO ["@ED" -> \{fi\} || \textbackslash "@IP" \_ ];}
\end{exercise}

\begin{example} \label{ex:137} [Smeets, I. 2008: 65 (31)] \cite{RefB:21}\\
	\emph{allkü-tu-nie-\textbf{fi}-n} 'I am listening to him'\\ \texttt{-TV.allkü\_oir+TR.tu33+PRPS.nie32+EDO.fi6\\+IND1SG.n3}
\end{example}

\begin{exercise} \label{R:10} \textbf{Internal direct object} \\
	\texttt{define RuTrIDO [["@ID" -> e || [CON|SVW] \_ ] \\ .o. ["@ID" -> e || a \_ "@Ü"n[\{ew\}|.\#.]] \\ .o. ["@ID" (->) e || [VOW|"@NK"] \_ ]];}
\end{exercise}

R\ref{R:10} contemplates other possible contexts of realization for \texttt{+IDO}. It becomes \emph{e} when preceded by any consonant or semi-vowel.

\begin{example} \label{ex:138} [Smeets, I. 2008: 87 (21)] \cite{RefB:21}\\
	\emph{kim-\textbf{e}-y-u} 'I recognized you' lit: 'I knew you'\\ \texttt{-TV.kim\_saber+IDO.e6+IND.y4+1.Ø3+DL.u2\\+DS12A.Ø1}
\end{example}

\texttt{+IDO} also becomes \emph{e} when is preceded by \emph{a} and is followed by the intermediate form \texttt{"@Ü"n}, which corresponds to the portmanteau suffix for indicative, 1\textsuperscript{st} person, singular; which in turn, it either ends the verb form (because it implies the presence of the null suffix \emph{-Ø} \texttt{+DS12A} following it), or it is followed by \emph{-ew} \texttt{+DS3A}.

\begin{example} \label{ex:139} [Smeets, I. 2008: 157 (17)] \cite{RefB:21}\\
	\emph{pe-\textbf{e}-n} 'you saw me' \\ \texttt{-TV.pe\_ver+IDO.e6+IND1SG.n3+DS12A.Ø1}
\end{example}

\begin{example} \label{ex:140} [Smeets, I. 2008: 94 (63)] \cite{RefB:21}\\
	\emph{pe-me-\textbf{e}-n-ew} 'there he saw me' \\ \texttt{-TV.pe\_ver+TH.me20+IDO.e6+IND1SG.n3+DS3A.ew1}
\end{example}

The last context of realization for \texttt{+IDO} says that \texttt{"@ID"} is optionally transformed into \emph{e}, which means that it may be elided, when preceded by a vowel or by the tag for negation in imperatives \texttt{"@NK"} (see R\ref{R:05} and examples E\ref{ex:128} and E\ref{ex:141}).

\begin{example} \label{ex:141} [Smeets, I. 2008: 94 (63)] \cite{RefB:21}\\
	\emph{ina-ni-a-\textbf{Ø}-lu-mu} 'they have been followed' \\ \texttt{-AV.ina\_detrás+VRB.Ø36+PRPS.nie32\\+NRLD.a9+IDO.e6+SVN.lu4+DS3A.mew1}
\end{example}

The last rule of the set treating suffixes between slots 10 and 4, deals with the objective verbal noun suffix \texttt{+OVN} \emph{-el}.

\begin{exercise} \label{R:11} \textbf{Objective Verbal Noun}\\
	\texttt{define RuOVN \newline ["@EL" -> [l|\{el\}] || [a|e|"@ID"] \_ ]\\ .o. ["@EL" -> \{el\}];}
\end{exercise}

R\ref{R:11} specifies that \texttt{"@EL"}, which is how \texttt{+OVN} is encoded, may be converted into \emph{l} or \emph{el} when preceded by \emph{a, e} or \texttt{"@ID"} (e.r. \ref{tp:179}). And in any other context it will be converted into \emph{el}, see examples below:

\begin{example} \label{ex:142} [Smeets, I. 2008: 114 (26)] \cite{RefB:21}\\
	\emph{lang-üm-\textbf{el}-chi ufisha} 'killed sheep' \\ \texttt{-IV.la\_morir+CA.m34+OVN.el4+ADJ.chi\\-NN.ufisha\_oveja}
\end{example}

\begin{example} \label{ex:143} [Smeets, I. 2008: 189 (45)] \cite{RefB:21}\\
	\emph{pi-\textbf{el}-mew} 'of what is said' \\ \texttt{-TV.pi\_decir+OVN.el4+INST.mew}
\end{example}

\begin{example} \label{ex:144} [Smeets, I. 2008: 189 (46)] \cite{RefB:21}\\
	\emph{entu-\textbf{el}} 'what is taken (out)' \\ \texttt{-TV.entu\_sacar+OVN.el4}
\end{example}

\subsubsection{Special case: suffix -nge-} \label{sec:40}

\paragraph{} \label{tp:46} Corresponding to the verbalizer (see \ref{sec:06}) or stem formative (see \ref{sec:07}) located in slot 36, or to the passive suffix, slot 23, the form \emph{-nge-} may alternate with \emph{-ngi-} when followed by the indicative suffix \emph{-y-}, and the verb form corresponds to the 3\textsuperscript{rd} person non-specified for number. All of which is encoded by the next rule:

\begin{exercise} \label{R:12} \textbf{Alternative \emph{nge} form} \\
	\texttt{define RuNGE ["@EY" (->) i || \_ "@Ü"y]\\ .o. ["@EY" -> e];}
\end{exercise}

For this rule to work, all mentioned suffixes were encoded as follows:

\begin{definition} \label{def:15} \\
	\texttt{["+VRB"\{.nge36\}"-IV"] : [\{ng\}"@EY"]; \newline ["+SFR"\{.nge36\}"-IV"] : [\{ng\}"@EY"]; \newline ["+PASS"\{.nge23\}] : [\{ng\}"@EY"];}
\end{definition}

An example of each case is rendered below:

\begin{example} \label{ex:145} [Smeets, I. 2008: 456 (3)] \cite{RefB:21}\\
	\emph{wentru-\textbf{ngi}-y} 'they were men' \\ \texttt{-NN.wentru\_hombre+VRB.nge36-IV+IND.y4+3.Ø3}
\end{example}

\begin{example} \label{ex:146} [Smeets, I. 2008: 305 (2)] \cite{RefB:21}\\
	\emph{weyel-weyel-\textbf{ngi}-y} 'he always swims' \\ \texttt{-IV.weyel\_nadar-RVBR+SFR.nge36-IV+IND.y4+3.Ø3}
\end{example}

\begin{example} \label{ex:147} [Smeets, I. 2008: 445 (3)] \cite{RefB:21}\\
	\emph{elu-\textbf{ngi}-y mapu} 'he was given land' \\ \texttt{-TV.elu\_dar+PASS.nge23+IND.y4+3.Ø3\\ -NN.mapu\_tierra}
\end{example}

\subsubsection{Special case: verb i- 'to eat'} \label{sec:41}

\paragraph{} \label{tp:47} The verb \emph{i-} may be realized as \emph{i-, iy-} or \emph{yi-} depending on the context. To apply the rules (R\ref{R:13}) that regulate this verb form in the different contexts, the verb \emph{i-} has been encoded as follows:

\begin{definition} \label{def:16} \texttt{["-TV"{.i\_comer}]: "@i";} \end{definition}

First sub-rule of R\ref{R:13} avoids \emph{i-} to be recognized and analysed as the final \emph{i} of any word by deleting it.

\begin{exercise} \label{R:13} \textbf{Forms of verb \emph{i-} 'to eat'} \\
	\texttt{define Verbi 
		\item[] ["@i" -> 0 || \_ .\#.] .o.
		\item[]["@i" -> "@G"\{iy\} || \_ [a|e|"@Y"|"@ID"|\\"@EL"]] .o.
		\item[] ["@i" -> \{yi\} || \_ k [i|ü|"@NK"], "@i" \_ ,\\ \_ "@i"] .o.
		\item[] ["@i" (->) \{yi\} || \_ w] .o.
		\item[] ["@i" -> "@G"i]];}
\end{exercise}

Second sub-rule converts \texttt{"@i"} into the intermediate form \texttt{"@G"\{iy\}} which set the verb ready to be part of a compound where \texttt{"@G"} will optionally become \emph{g} if preceded by a vowel (see E\ref{ex:116}, E\ref{ex:117}, p. \pageref{ex:116}; R\ref{R:01} and R\ref{R:02}, p. \pageref{R:02}). This change is carried out when \texttt{"@i"} is followed by \emph{a, e} or the intermediate forms \texttt{"@Y", "@ID", "@EL"}.

\begin{example} \label{ex:148}\ \\
	\emph{dewma-\textbf{iy}-a-l-mew → dewma-\textbf{giy}-a-l-mew} 'while preparing food' \\ \texttt{-TV.dewma\_hacer-TV.i\_comer-CR.TV+NRLD.a9\\+OVN.el4+INST.mew}
\end{example}

\begin{example} \label{ex:149} [Smeets, I. 2008: 204 (125)] \cite{RefB:21}\\
	\emph{\textbf{i}-el → \textbf{iy}-el} 'what had been eaten' \\ \texttt{-TV.i\_comer+OVN.el4}
\end{example}

Third sub-rule in R\ref{R:13} converts \texttt{"@i"} into \emph{yi} in different contexts:

1) when \texttt{"@i"} is followed by \emph{ki, kü} or the intermediate form \texttt{"@NK"} (see R\ref{R:05} and E\ref{ex:128}, p. \pageref{R:05}):

\begin{example} \label{ex:150} [Smeets, I. 2008: 445 (3)] \cite{RefB:21}\\
	\emph{\textbf{i-ki}-fi-l-nge → \textbf{yi-ki}-fi-l-nge} 'you need not eat it' \\ \texttt{-TV.i\_comer+NEG.ki10+EDO.fi6+CNI.l4\\+IMP2SG.nge3}
\end{example}

2) \texttt{"@i"} is converted into \emph{yi} when preceded by itself and followed by itself, this way the rule is managing reduplication of the verb root \emph{i-} 'to eat':

\begin{example} \label{ex:151} [Smeets, I. 2008: 307 (8)] \cite{RefB:21}\\
	\emph{\textbf{i-i}-künu-fi-ñ → \textbf{yi-yi}-künu-fi-ñ} 'I ate it quickly' \\ \texttt{-TV.i\_comer-RVBR+SFR.Ø36-IV+PFPS.künu32\\+EDO.fi6+IND1SG.n3}
\end{example}

In the fourth sub-rule, \texttt{"@i"} is optionally converted into \emph{yi} when followed by \emph{w}:

\begin{example} \label{ex:152} [Smeets, I. 2008: 263 (11)] \cite{RefB:21}\\
	\emph{\textbf{i}-we-me-ke-la-y} 'he no longer goes there to eat (as he used to)' \\ \texttt{-TV.i\_comer+PS.we19+TH.me20+CF.ke14+NEG.la10\\+IND.y4+3.Ø3}
\end{example}

\begin{example} \label{ex:153} [Smeets, I. 2008: 260 (4)] \cite{RefB:21}\\
	\emph{\textbf{i}-we-la-n → \textbf{yi}-we-la-n} 'I eat no more' \\ \texttt{-TV.i\_comer+PS.we19+NEG.la10+IND1SG.n3}
\end{example}

Finally, fifth sub-rule states that \texttt{"@i"} is converted in the intermediate representation \texttt{"@G"i} in any other context. So, if it forms part of a compound being the second member, it can up bring a \emph{g} before itself when the previous element of the compound ends in vowel (see E\ref{ex:148}).

\begin{example} \label{ex:154} [Smeets, I. 2008: 309 (4)] \cite{RefB:21}\\
	\emph{\textbf{i}-püra-fi-ñ} 'I ate it reluctantly' \\ \texttt{-TV.i\_comer+AIML.püda+EDO.fi6+IND1SG.n3}
\end{example}

\begin{example} \label{ex:155} [Smeets, I. 2008: 43 (f)] \cite{RefB:21}\\
	\emph{\textbf{i}-fal-ün} 'I must eat' \\ \texttt{-TV.i\_comer+FORCE.fal25+IND1SG.n3}
\end{example}

\subsubsection{Special case: verb entu- 'to take out'} \label{sec:42}

\paragraph{} \label{tp:48} The verb \emph{entu-} may be realized as \emph{ntu-, entu-} or \emph{nentu-} depending on the context. To apply the rules (R\ref{R:14}) that regulate this verb form in the different contexts, the verb \emph{entu-} has been encoded as follows:

\begin{definition} \label{def:17} \texttt{["-TV"{.entu\_sacar\_quitar}]: ["@VE"\{ntu\}];} \end{definition}

\begin{exercise} \label{R:14} \textbf{Forms of verb \emph{entu-} 'to take out'} \\
	\texttt{define Verbentu
		\item[] ["@VE" -> [["@G"e]|\{ne\}] || [.\#.|a|i] \_ ].o.
		\item[] ["@VE" -> [\{ne\}|0] || ü \_ ] .o. 
		\item[] ["@VE" -> e || [d|f|m] \_ ] .o. 
		\item[] ["@VE" -> \{ne\} || [e|u] \_ ] .o. 
		\item[] ["@VE" -> e ];}
\end{exercise}

There are 5 different contexts shaping the form of this verb. First sub-rule of R\ref{R:14} indicates that \texttt{"@VE"} may be converted into (the intermediate form) \texttt{"@G"e} or \emph{ne} at word beginning, or after \emph{a} or \emph{i}, in both last cases the tag \texttt{"@G"} is optionally converted into \emph{g} (see R\ref{R:01} and E\ref{ex:116}, p. \pageref{R:01}):

\begin{example} \label{ex:156} [Smeets, I. 2008: 448 (32)] \cite{RefB:21}\\
	\emph{\textbf{e}ntu-fi-y-iñ} 'we took him out' \\ \texttt{-TV.entu\_sacar+EDO.fi6+IND.y4+1.Ø3+PL.iñ2}
\end{example}

\begin{example} \label{ex:157} [Smeets, I. 2008: 318 (8)] \cite{RefB:21}\\
	\emph{\textbf{ne}ntu-antü-y} 'they fixed a date' \\ \texttt{-TV.entu\_sacar-NN.antü\_día+IND.y4+3.Ø3}
\end{example}

\begin{example} \label{ex:158} [Smeets, I. 2008: 407 (24)] \cite{RefB:21}\\
	\emph{taym\textbf{a-e}ntu-nge-pa-y} 'they were taken out there' \\ \texttt{-TV.tayma\_eliminar-TV.entu\_sacar-CR.TV\\+PASS.nge23+HH.pa17+IND.y4+3.Ø3}
\end{example}

\begin{example} \label{ex:159} [Smeets, I. 2008: 315] \cite{RefB:21}\\
	\emph{witr\textbf{a-ne}ntu-n} 'I pulled out' \\ \texttt{-TV.witra\_levantar-TV.entu\_sacar-CR.TV\\+IND1SG.n3}
\end{example}

\begin{example} \label{ex:160} [Smeets, I. 2008: 409 (40)] \cite{RefB:21}\\
	\emph{dull\textbf{i-e}ntu-a-y-iñ} 'we will choose him' \\ \texttt{-TV.dulli\_elegir-TV.entu\_sacar-CR.TV\\+NRLD.a9+IND.y4+1.Ø3+PL.iñ}
\end{example}

\begin{example} \label{ex:161} [Smeets, I. 2008: 553 (\emph{rapi-})] \cite{RefB:21}\\
	\emph{rap\textbf{i-ne}ntu-y} 'he threw up' \\ \texttt{-IV.rapi\_vomitar-TV.entu\_sacar-CR.TV\\+IND.y4+3.Ø3}
\end{example}

Second sub-rule of R\ref{R:14} converts \texttt{"@VE"} into \emph{ne} after \emph{ü} or eliminates it, which means that the verb may be realized as \emph{ntu-} when the tag is eliminated, or \emph{nentu-} when the tag is converted:

\begin{example} \label{ex:162} [Smeets, I. 2008: 318 (8)] \cite{RefB:21}\\
	\emph{wem\textbf{ü}-ntu-nge-rume-ye-m} 'they were suddenly expelled\\ without realizing (it)' \\ \texttt{-TV.wemü\_perseguir-TV.entu\_sacar-CR.TV\\+PASS.nge23+SUD.rume21+CF.ye5+IVN.m4}
\end{example}

\begin{example} \label{ex:163} [Smeets, I. 2008: 556 (\emph{rüfü-})] \cite{RefB:21}\\
	\emph{rüf\textbf{ü-ne}ntu-me-ki-y} 'he is busy serving out there' \\ \texttt{-TV.rüfü\_servir-comida-TV.entu\_sacar-CR.TV\\+TH.me20+CF.ke14+IND.y4+3.Ø3}
\end{example}

Third sub-rule converts \texttt{"@VE"} into \emph{e} after \emph{d, f} or \emph{m} yielding the \emph{entu-} form of the verb:

\begin{example} \label{ex:164} [Smeets, I. 2008: 405 (7)] \cite{RefB:21}\\
	\emph{a\textbf{d-e}ntu-a-l} 'how to settle' \\ \texttt{-NN.ad\_forma-TV.entu\_sacar+NRLD.a9+OVN.el4}
\end{example}

\begin{example} \label{ex:165} [Smeets, I. 2008: 201 (102)] \cite{RefB:21}\\
	\emph{ütrü\textbf{f-e}ntu-fi-n} 'I have thrown it away' \\ \texttt{-TV.ütrüf\_tirar-TV.entu\_sacar-CR.TV\\+EDO.fi6+IND1SG.n3}
\end{example}

\begin{example} \label{ex:166} [Smeets, I. 2008: 486 (16)] \cite{RefB:21}\\
	\emph{ki\textbf{m-e}ntu-a-n} 'I shall declare' \\ \texttt{-TV.kim\_saber-TV.entu\_sacar-CR.TV\\+NRLD.a9+IND1SG.n3}
\end{example}

Fourth sub-rule converts \texttt{"@VE"} into \emph{ne} when preceded by \emph{e} or \emph{u} yielding the \emph{nentu-} form of the verb:

\begin{example} \label{ex:167} [Smeets, I. 2008: 88 (23)] \cite{RefB:21}\\
	\emph{weñ\textbf{e-ne}ntu-nge-r-pu-y} 'it would eventually be robbed' \\ \texttt{-TV.weñe\_robar-TV.entu\_sacar-CR.TV\\+PASS.nge23+ITR.r18+LOC.pu17+IND.y4+3.Ø3}
\end{example}

\begin{example} \label{ex:168} [Smeets, I. 2008: 556 (\emph{rüfü-})] \cite{RefB:21}\\
	\emph{utr\textbf{u-ne}ntu-y} 'she spilled it out' \\ \texttt{-TV.utru\_derramar-TV.entu\_sacar-CR.TV\\+IDO.e6+IND.y4+3.Ø3+DS12A.Ø1}
\end{example}

The last sub-rule converts \texttt{"@VE"} into \emph{e} in any other context not considered in the R\ref{R:14} set of rules, yielding the \emph{entu-} form of the verb.

\subsubsection{Special case: radical consonant alternation before causative -üm-} \label{sec:43}

\paragraph{} \label{tp:49} \emph{Mapuche} verb roots which have an intransitive meaning may be transitivized by adding causatives suffixes \emph{-el-, -ül-, -üm-}, slot 34, the factitive \emph{‑ka-} or transitivizer \emph{-tu-} suffixes, slot 33. Few roots undergo a change through this process, actually, Smeets says that it is an "unproductive relic phenomena" [Smeets 2008: 53] \cite{RefB:21}. She gives the following exhaustive list:

\begin{itemize} \label{it:32}
	\item[] \emph{af-} 'to come to an end' → \emph{a\textbf{p}-üm-} 'to finish'
	\item[] \emph{lef-} 'to run' → \emph{le\textbf{p}-üm-} 'to make run (animals)'
	\item[] \emph{traf-} 'to fit in/on' → \emph{tra\textbf{p}-üm-} 'to cause to fit in/on'
	\item[] \emph{lleg-} 'to come up (plants)' → \emph{lle\textbf{k}-üm-} 'to plant' (tr.),\\ but \emph{lleg-üm-} 'to make come up'
	\item[] \emph{nag-} 'to go down' → \emph{na\textbf{k}-üm-} 'to carry down',\\ but \emph{nag-üm-} 'to take down'
	\item[] \emph{la-} 'to die' → \emph{la\textbf{ng}-üm-} 'to kill'
\end{itemize}

We have also found some other cases:

\begin{itemize} \label{it:33}
	\item[] \emph{trof-} 'to explode, crack' (itr.) → \emph{tro\textbf{p}-üm-} 'to crack' (tr.)
	\item[] \emph{nel-} 'to get loose' →  \emph{nel(\textbf{k})-üm-} 'to let loose, to set free'
	\item[] \emph{lüf-} 'to burn' (itr.) → \emph{lü\textbf{p}-üm-} 'to burn' (tr.), 'to set fire'
\end{itemize}

As the goal of the FST is analysis, the system was set for the maximum analysis possible. So, instead of introducing both forms (intransitive and transitive), only the intransitive verb form was introduced in the lexicon, together with the creation of a rule to handle the radical change.

\begin{definition} \label{def:18} Encoding of forms with radical change: \texttt{
		\item[] ["‑IV"{.af\_acabar}]: ["@G"a"@FP"];
		\item[] ["‑IV"{.la\_morir}]: [{la}"@NG"];
		\item[] ["‑IV"{.lef\_correr}]: [{le}"@FP"];
		\item[] ["‑IV"{.lleg\_crecer}]: [{lle}"@GK"];
		\item[] ["-IV"{.lüf\_quemar}]: [{lü}"@FP"];
		\item[] ["‑IV"{.nag\_bajar}]: [{na}"@GK"];
		\item[] ["-IV"{.nel\_soltar}]: [{nel}"@GK"];
		\item[] ["‑IV"{.traf\_encajar}]: [{tra}"@FP"];
		\item[] ["-IV"{.trof\_romper}]: [{tro}"@FP"];}
\end{definition}

The forms on the list have a tag on the right side. \texttt{"@FP"} when the root has to end in \emph{f} for the intransitive meaning and in \emph{p} for the transitive sense. \texttt{"@NG"} appears when the root does not change anything regarding intransitiveness, and add \emph{ng} when transitive. \texttt{"@GK"}, intransitive ending in \emph{g}, transitive ending either in \emph{g} or \emph{k}. The later are the ones with a "but" on the "list of forms with radical change" (\ref{def:18}).

\begin{definition} \label{def:19} Causative \emph{-üm-} encoding: \\ 
	\texttt{["+CA"{.m34}]: ["@ÜC"m];}
\end{definition}

\begin{exercise} \label{R:15} \textbf{Radical consonant alternation before \emph{-üm-}} \\
    \texttt{define RuCAlt01
	\item[] [["@NG" -> \{ng\}"@4", "@FP" -> p"@5", \\"@GK" -> [[k|g]"@6"] || \_ "@ÜC"] .o. 
	\item[] ["@ÜC" -> ü || ["@4"|"@5"|"@6"|CON|SVW] \_ ]\\ .o.
	\item[] ["@4"|"@5"|"@6" -> 0]]; \bigskip
	\item[] define RuCAlt02 
	\item[] [["@NG" -> 0, "@FP" -> f, "@GK" -> [g|0]] \\.o. 
	\item[] ["@ÜC" -> 0 || ["@NG"|VOW] \_ ] .o.
	\item[] ["@ÜC" -> ü || ["@FP"|"@GK"|CON|SVW] \_ ] .o.
	\item[] ["@NG"|"@FP"|"@GK"|"@ÜC" -> 0]];}
\end{exercise}

The above set of rules is similar to the one defined by \texttt{RuTrIPDEDO} (R\ref{R:06} p. \pageref{R:06}), in the sense that it follows the same logic. Basically, when any of the tags \texttt{"@NG"}, \texttt{"@FP"} or \texttt{"@GK"} enters in contact with \texttt{"@ÜC"}, the transitivizing option is activated implying a new context tag \texttt{"@4"}, \texttt{"@5"}, \texttt{"@6"} to allow the subsequent change of \texttt{"@ÜC"} into \emph{ü}. After these two steps, context tags (\texttt{"@4"}, etc.) are wiped out.

Rule \texttt{RuCAlt02} operates on the intransitive change, i.e., it either eliminates the tag or transforms it into the intransitive form. The following analyses show that processes described above are successfully carried out:

\begin{example} \label{ex:169} [Smeets, I. 2008: 192 (52)] \cite{RefB:21}\\
	\emph{a\textbf{f}-a-y} 'it will stop' \\
	\texttt{‑IV.af\_acabar+NRLD.a9+IND.y4+3.Ø3}
\end{example}

\begin{example} \label{ex:170} [Smeets, I. 2008: 313 (15)] \cite{RefB:21}\\
	\emph{a\textbf{p}-üm-fal-iy} 'it can be finished' \\
	\texttt{‑IV.af\_acabar+CA.üm34+ADJDO.fal}\footnote{Smeets labels \emph{-fal} as a nominalizer putting it under the category of derivative nominalizers as a broad term for non-verbal suffixes (see chap. 28.1 of 'A Grammar of \emph{Mapuche}'), but we have tagged it as adjectivizer because \emph{-fal} indicates that the action denoted by the verb is applicable to the subject of the phrase (e.g., edible) [Smeets 2008: 312] \cite{RefB:21}.}\texttt{+VRB.Ø36\\+IND.y4+3.Ø3}
\end{example}

\begin{example} \label{ex:171} [Smeets, I. 2008: 34] \cite{RefB:21}\\
	\emph{le\textbf{f}-iy} 'he ran' \\
	\texttt{‑IV.lef\_correr+IND.y4+3.Ø3}
\end{example}

\begin{example} \label{ex:172} [Smeets, I. 2008: 265 (6)] \cite{RefB:21}\\
	\emph{le\textbf{p}-üm-kantu-nge-y} 'they made it run' (they made a mare run for exercise) \\
	\texttt{‑IV.lef\_correr+CA.üm34+PLAY.kantu22\\+PASS.nge23+IND.y4+3.Ø3}
\end{example}

\begin{example} \label{ex:173} [Smeets, I. 2008: 304 (23)] \cite{RefB:21}\\
	\emph{tra\textbf{f}-me-n} 'I went to meet' (somebody) \\
	\texttt{‑IV.traf\_encajar+TH.me20+IND1SG.n3}
\end{example}

\begin{example} \label{ex:174} [Smeets, I. 2008: 560 (\emph{traf-})] \cite{RefB:21}\\
	\emph{tra\textbf{p}-üm-a-fi-n} 'I will gather' (it) \\
	\texttt{‑IV.traf\_encajar+CA.üm34+NRLD.a9\\+EDO.fi6+IND1SG.n3}
\end{example}

\begin{example} \label{ex:175} [Smeets, I. 2008: 206 (137)] \cite{RefB:21}\\
	\emph{lle\textbf{g}-mu-m} 'where it had grown up' \\
	\texttt{‑IV.lleg\_crecer+PLPF.mu7+IVN.m4}
\end{example}

\begin{example} \label{ex:176} [Smeets, I. 2008: 528 (\emph{lleg-})] \cite{RefB:21}\\
	\emph{lle\textbf{k}-üm-fi-ñ} 'I grew it' \\
	\texttt{‑IV.lleg\_crecer+CA.üm34+EDO.fi6+IND1SG.n3}
\end{example}

\begin{example} \label{ex:177} [Zúñiga, F. 2006: 306 (parir)] \cite{RefB:24}\\
	\emph{lle\textbf{g}-üm-ün} 'I grew' \\
	\texttt{‑IV.lleg\_crecer+CA.üm34+IND1SG.n3}
\end{example}

\begin{example} \label{ex:178} [Smeets, I. 2008: 49] \cite{RefB:21}\\
	\emph{na\textbf{k}-üm-fi-y-u} 'we brought him down' \\
	\texttt{‑IV.nag\_bajar+CA.üm34+EDO.fi6+IND.y4\\+1.Ø3+DL.u2}
\end{example}

\begin{example} \label{ex:179} [Smeets, I. 2008: 137 (37)] \cite{RefB:21}\\
	\emph{na\textbf{g}-ün} 'it went down / the going down' \\
	\texttt{‑IV.nag\_bajar+PVN.n4}
\end{example}

\begin{example} \label{ex:180} [Smeets, I. 2008: 243 (1)] \cite{RefB:21}\\
	\emph{la-le-la-y} 'she is not dead' \\
	\texttt{‑IV.la\_morir+ST.küle28+NEG.la10+IND.y4+3.Ø3}
\end{example}

\begin{example} \label{ex:181} [Smeets, I. 2008: 243 (2)] \cite{RefB:21}\\
	\emph{lan\textbf{g}-üm-ki-fi-l-nge} 'don't kill it' \\
	\texttt{‑IV.la\_morir+CA.üm34+NEG.ki10+EDO.fi6+CNI.l\\+IMP2SG.nge3}
\end{example}

\begin{example} \label{ex:182} [Guevara 1913: 77] \cite{RefB:09}\\
	\emph{tro\textbf{f}-lu} 'the exploding one' \\
	\texttt{‑IV.trof\_explotar+SVN.lu4}
\end{example}

\begin{example} \label{ex:183} [Augusta, F. (\emph{tropümün})] \cite{RefB:03}\\
	\emph{tro\textbf{p}-üm-ün} 'snap, shoot' \\
	\texttt{‑IV.trof\_explotar+CA.üm34+PVN.n4}
\end{example}

\begin{example} \label{ex:184} [Augusta, F. (\emph{nel-})] \cite{RefB:03}\\
	\emph{nel-ün kawellu} 'loose horse' \\
	\texttt{-IV.nel\_soltar+IND1SG.n3\\-NN.kawellu\_caballo}
\end{example}

\begin{example} \label{ex:185} [Smeets, I. 2008: 441 (60)] \cite{RefB:21}\\
	\emph{nel\textbf{k}-üm-nge-nu-a-l} 'not to get fired' \\
	\texttt{‑IV.nel\_soltar+CA.üm34+PASS.nge23\\+NEG.no10+NRLD.a9+OVN.el4}
\end{example}

\begin{example} \label{ex:186} [Smeets, I. 2008: 526 (\emph{lüf-})] \cite{RefB:21}\\
	\emph{lü\textbf{f}-a-y} 'it will burn' \\
	\texttt{‑IV.lüf\_quemar+NRLD.a9+IND.y4+3.Ø3}
\end{example}

\begin{example} \label{ex:187} [Augusta, F. (encender)] \cite{RefB:03}\\
	\emph{lü\textbf{p}-üm-ün} 'to set fire to' \\
	\texttt{‑IV.lüf\_quemar+CA.üm34+PVN.n4}
\end{example}

\subsection{Morphotactics: constructing the verb form} \label{sec:44}

\paragraph{} \label{tp:50} As it was explained in \nameref{tp:25}, p. \pageref{tp:25}, morphotactics is the set of constraints that regulates the co-occurrence of morphemes. Once the lexicon and suffixes that interact in the verb form are declared (see \ref{sec:36} \nameref{sec:36}, p. \pageref{sec:36} and \ref{sec:37} \nameref{sec:37}, p. \pageref{sec:37}.), it is necessary to regulate their interaction.

We have introduced the \emph{Mapuche} verb form in section \ref{sec:04}, p \pageref{sec:04}. In a concise way, the verb is a stem followed by a series of suffixes that complete the verb form.

\subsubsection{Stems codification} \label{sec:45}

\paragraph{} \label{tp:51} Section \ref{sec:16} \nameref{sec:16}, p. \pageref{sec:16} exposes different stem configurations. Most simple stem type is formed by a single verbal root. Verb suffixes may be added immediately to this type of stem. See example E\ref{ex:44}, p. \pageref{ex:44}.

Other type of stem that accepts verb suffixes immediately is the simple (implying no suffixes) compound where one of the members is a verbal root, the other member may be another verbal root, an adjectival, adverbial, nominal, numeral or a question root. See examples E\ref{ex:45}, E\ref{ex:46}, p. \pageref{ex:45} and following table.

\begin{table}[htb]
	\caption{Simple stem forms}
	\label{tab:05}
	\begin{tabular}{|r|c|}
		\hline\noalign{\smallskip}
		Stem & Suffixes\\
		\noalign{\smallskip}\hline\noalign{\smallskip}
		Verbal root & +Suffixes\\
		\noalign{\smallskip}\hline\noalign{\smallskip}
		Verbal root + Verbal root & +Suffixes\\
		\noalign{\smallskip}\hline\noalign{\smallskip}
		Verbal root + Non-verbal root & +Suffixes\\
		\noalign{\smallskip}\hline\noalign{\smallskip}
		~~~~~~~~~~ Non-verbal root + Verbal root & +Suffixes\\
		\noalign{\smallskip}\hline
	\end{tabular}
\end{table}

\begin{definition} \label{def:20} Simple stems encoding\\ 
	\texttt{define CMPVBVAL \newline [CAjVbVSTEM|CAvVbSTEM|CNnVbSTEM|CQtVbSTEM|\\CVbAjSTEM|CVbAvSTEM|CVbNnSTEM|CVbVbSTEM];}
\end{definition}

\begin{itemize} \label{it:34}
	\item[] \texttt{CAjVbVSTEM} Complex adjective+verb compound stem
	\item[] \texttt{CAvVbSTEM} Complex adverb+verb compound stem
	\item[] \texttt{CNnVbSTEM} Complex noun+verb compound stem
	\item[] \texttt{CQtVbSTEM} Complex question+verb compound stem
	\item[] \texttt{CVbAjSTEM} Complex verb+adjective compound stem
	\item[] \texttt{CVbAvSTEM} Complex verb+adverb compound stem
	\item[] \texttt{CVbNnSTEM} Complex verb+noun compound stem
	\item[] \texttt{CVbVbSTEM} Complex verb+verb compound stem
\end{itemize}

Definition D\ref{def:20} above, defined as \texttt{CMPVBVAL}, digests, among others in the FST script, the possible \emph{Mapuche} simple stems. \texttt{CMPVBVAL} stands for "verbal compounds with their corresponding valence" (see \ref{tp:54} below).

\paragraph{Compounds encoding.} \label{tp:52} In D\ref{def:21}, \texttt{formCNnVbROOT} encodes the form of a "noun + verb" compound. The whole form is enclosed in brackets and the tag \texttt{"-NVCR"} is attached to it. Then, sub-rule \texttt{CNnVbSTEM} applies neutralization of tags (see \ref{tp:53} below) and verb valence (see \ref{tp:54} below) to the compound.

\begin{definition} \label{def:21}\ Noun + verb compound \\ 
	\texttt{define formCNnVbROOT \newline [[NROOT [IVROOT|TVROOT]]"-NVCR"];\\
	define CNnVbSTEM [RuIVCNnVb .o. RuTVCNnVb .o.\\ RuCNnVb01 .o. [neutCNnVb .o. formCNnVbROOT]];}
\end{definition}

\paragraph{Neutralization of tags.} \label{tp:53} Compound stems, complex stems (see \nameref{tp:59}, p. \pageref{tp:59}) and complex compound stems (see \nameref{tp:61}, p. \pageref{tp:61}), are qualified as "complex" because they incorporate suffixes into the stem, and have their own rules of interaction. For this reason, PoS and suffixes tags are converted into different tags while applying the inner compound rules (R\ref{R:16}). We call this process "neutralization" because it makes general rules not affect stems. This change is reverted before the analysis output, so the user does not have to interpret a wider set of tags. Neutralization is applied first, and then compound rules are applied to the resulting form, therefore the rules are generated taking into account the converted tags (see \texttt{CNnVbSTEM} in D\ref{def:21}). R\ref{R:16} is an example of how neutralization is applied:

\begin{exercise} \label{R:16} \textbf{Neutralization of PoS and suffixes tags} (sample)\\
	\texttt{define NeutAj ["-aj0" <- "-AJ"];\\
	define NeutNn ["-nn0" <- "-NN"];\\
	define NeutIv ["-iv0" <- "-IV"];\\
	define NeutTv ["-tv0" <- "-TV"];\\
	define NeutAdjdo ["+adjdo0" <- "+ADJDO"];\\
	define NeutCa [{"+ca0"} <- {"+CA"}];\\
	define NeutDistr ["+distr0" <- "+DISTR"];\\
	define NeutHh ["+hh0" <- "+HH"];\\
	define NeutNomag ["+nomag0" <- "+NOMAG"];\\
	define NeutPvn ["+pvn0" <- "+PVN"];\\
	define NeutRef ["+ref0" <- "+REF"];\\
	define NeutTh ["+th0" <- "+TH"];\\
	define NeutTr ["+tr0" <- "+TR"];}
\end{exercise}

\paragraph{Valence in compounds.} \label{tp:54} When one of the roots in a compound is a verb and the other is not, the resulting compound gets the valence from the verb root. When both members of a compound are verb roots, the valence is derived from the second. This needs to be encoded because transitive verbs take suffixes that intransitive ones do not.

\begin{exercise} \label{R:17} \textbf{Valence in verbal compounds} (sample) \\
	\texttt{define RuIVCNnVb ["-CR.IV" <- "-NVCR" ||\\
	"-nn0" \$["-iv0"]}\footnote{This notation is equivalent to  \texttt{?* "-iv0" ?*}: \texttt{"-iv0"} surrounded by none or any amount of elements to the right and to the left.}\texttt{ \_ ];\\
	define RuTVCNnVb ["-CR.TV" <- "-NVCR" ||\\
	"-nn0" \$["-tv0"] \_ ];\\
	define RuIVCVbVb ["-CR.IV" <- "-VCR" || \newline
	["-tv0"|"-iv0"] \$["-iv0"] \_ ];\\
	define RuTVCVbVb ["-CR.TV" <- "-VCR" || \newline
	["-tv0"|"-iv0"] \$["-tv0"] \_ ];}
\end{exercise}

R\ref{R:17} has two examples of valence application, one for "noun + verb" compounds and another for "verb + verb" compounds. Each of them have a rule for transitive and another for intransitive valences. The tag \texttt{"-NVCR"}, that was added to the compound in rule \texttt{formCNnVbROOT} (D\ref{def:21}), is transformed into \texttt{"-CR.IV"} when preceded by the sequence of neutralized tags \texttt{"-nn0" ?* "-iv0"} (\texttt{?*} indicates zero or more elements in between). This process establishes the intransitive valence for this compound. Transitive process is analogous.

When the compound is made up of two adjectives, two nouns or two verbs, we need to process the compound in a way to not accept equal roots, in which case it would not be a compound but a reduplicated root.

\begin{definition} \label{def:22}\ Verb + verb compound \\ 
	\texttt{define formCVbVbROOT [[\%< [TVROOT|IVROOT] \%\#\\ \%< [IVROOT|TVROOT] \%\#]"-VCR"];\\
	define neutCVbVb [NeutIv .o. NeutTv];\\
	define preCVbVbROOT [\_eq(formCVbVbROOT,\\ \%< , \%\#)];\\
	define CVbVbSTEM [RuIVCVbVb .o. RuTVCVbVb .o. \newline [neutCVbVb .o. formCVbVbROOT - preCVbVbROOT]];}
\end{definition}

As in the case of the "noun + verb" compound (D\ref{def:21}), in D\ref{def:22} the first rule \texttt{formCVbVbROOT} defines the elements and their order in the compound, but it also adds some marks to the roots. Both roots are marked with \texttt{<} on the left and \texttt{\#} on the right: "\texttt{\%< ROOT \%\#}". The \texttt{\%} (percentage) escapes the symbols to read them literally. Then, the whole form is enclosed in brackets and the tag \texttt{"-VCR"} is attached to it. The next rule \texttt{neutCVbVb}, defines the tags to be neutralized (see \nameref{tp:53}, p. \pageref{tp:53}).

Rule \texttt{preCVbVbROOT} filters from the output side of\\ \texttt{formCVbVbROOT} all those strings where some sub-strings occurring between the delimiters \texttt{<} and \texttt{\#} are different. This rule\footnote{"\texttt{\_eq(X,L,R)}\\ Filters from the output side of \texttt{X} all those strings where some sub-strings occurring between the delimiters \texttt{L} and \texttt{R} are different. Example:\\ Consider the language \texttt{\%< a* b \%> \%< a b* \%>}, which contains an infinite number of strings:\\ \texttt{<b><a> <b><ab> <ab><a> <ab><ab> <ab><abbb> ...}\\ However, only one of the strings in this language has identical sub-strings between all instances of \texttt{<} and \texttt{>}, namely \texttt{<ab><ab>}. Hence, the language containing the single string\\ \texttt{<ab><ab>}\\ is produced by the regular expression:\\ \texttt{\_eq(\%< a* b \%> \%< a b* \%> , \%< , \%>) ;}\\ This operation is mostly used to model reduplication in natural language lexicons. Usually, the bare words to be reduplicated are marked with delimiters, say \texttt{<} and \texttt{>}, after which one can produce the reduplicated forms. For example:\\ \texttt{define Lexicon \{cat\}|\{dog\}|\{horse\};\\ define RLexicon \%< Lexicon \%> (\%- \%< \textbackslash [\%<|\%>]+ \%>);\\ regex \_eq(RLexicon, \%<, \%>) .o. \%<|\%> -> 0 ;}\\ and now we get:\\ \texttt{foma[1]: lower-words cat cat-cat dog dog-dog horse horse-horse.}" \newline [Hulden, M. in \href{https://code.google.com/archive/p/foma/wikis/RegularExpressionReference.wiki}{https://code.google.com/archive/p/foma/wikis/\\RegularExpressionReference.wiki}].} is meant to treat reduplicated roots, but we have modified it a little, so we can apply it to the compounds in order to not analyse reduplication as composition. Actually, what we do is subtract from the form (defined by \texttt{formCVbVbROOT}) the result of the calculus made at \texttt{preCVbVbROOT}, obtaining only those forms where both members are different.

Finally, \texttt{CVbVbSTEM} holds the result of applying neutralization, valence definition (see \nameref{tp:54}, p. \pageref{tp:54}) and the subtraction explained in the previous paragraph.

All type of stems will be later collected under the rule \texttt{VERBSTEM} (D\ref{def:28}, p \pageref{def:28}), where \texttt{CMPVBVAL} (see table \ref{tab:05} and D\ref{def:20}) is one of them.

\paragraph{Stems formed with a verbalizer suffix.} \label{tp:55} One more degree of complexity is given by the necessity of some single roots or compounds of adding a verbalizing suffix in slot 36 (see section \ref{sec:06} \nameref{sec:06}, p. \pageref{sec:06}) to be used as verb stems. Single roots that need this kind of suffix are adjectives, adverbs, nouns, numerals, onomatopoeia, proper nouns and question forms. Reduplicated roots of any category also need these suffixes, which are called "stem formative" in this case (see section \ref{sec:07} \nameref{sec:07}, p. \pageref{sec:07}). Compounds where none of the two roots forming them is a verb, also need a verbalizer in slot 36.

Table \ref{tab:06} summarizes what have been explained in the previous paragraph.

\begin{table}[htb]
	\caption{Simple stem forms: 1\textsuperscript{o} complexity}
	\label{tab:06}
	\begin{tabular}{|r|c|c|}
		\hline\noalign{\smallskip}
		Stem & \makecell{Verbalizers\\Slot 36} & Suffixes\\
		\noalign{\smallskip}\hline\noalign{\smallskip}
		Non-verbal root & +VRB & +Suffixes\\
		\noalign{\smallskip}\hline\noalign{\smallskip}
		~~~~~~~~~~ Non-verbal compound & +VRB & +Suffixes\\
		\noalign{\smallskip}\hline\noalign{\smallskip}
		Reduplicated root & +SFR & +Suffixes\\
		\noalign{\smallskip}\hline
	\end{tabular}
\end{table}

\paragraph{Single non-verbal roots.} \label{tp:56} They need a verbalizer to become verbal stems; see them collected in definition D\ref{def:24}, encoded as\\ \texttt{SPNVBROOT}. Verbalizers, slot 36, are encoded in D\ref{def:23} below.

\begin{definition} \label{def:23} Verbalizers (slot 36)\\ 
	\texttt{["+VRB"\{.Ø36\}] : 0\\
		| ["+VRB"\{.nge36\}"-IV"] : [\{ng\}"@EY"]\\
		| ["+VRB"\{.tu36\}] : \{tu\}\\
		| ["+VRB"\{.ntu36\}] : ["@N"\{tu\}]\\
		| ["+VRB"\{.l36\}] : l\\
		| ["+VRB"\{.ye36\}] : \{ye\};}
\end{definition}

\begin{definition} \label{def:24}\ Single non-verbal roots + verbalizer\\ 
	\texttt{define SPNVBROOT [AJROOT|AVROOT|NROOT|NUROOT|\\PROPN|QROOT] SVRB;}
\end{definition}

\texttt{SPNVBROOT} states that any of the single roots it collects must be followed by a verbalizer (collected under \texttt{SVRB}) in order to occur with verbal suffixes. R\ref{R:18} exposes the rules that regulate the suffixation of verbalizers by category (see section \ref{sec:06}, p \pageref{sec:06}):

\begin{exercise} \label{R:18} \textbf{Non-verb roots forming verb stems} \\ 
	\texttt{define RuAj [["-AJ"|"-CAJ"] =>\\ \_ ?* [\{.Ø36\}|\{.l36\}|\{.nge36\}|\{.ntu36\}]]; \medskip \\
		define RuAv [["-AV"|"-CAV"] =>\\ \_ ?* [\{.Ø36\}|\{.l36\}|\{.nge36\}|\{.ntu36\}]]; \medskip \\
		define RuNn [["-NN"|"-PN"|"-CNN"|"-CPN"] =>\\ \_ ?* [\{.Ø36\}|\{.nge36\}|\{.tu36\}|\{.ye36\}]]; \medskip \\
		define RuNu ["-NU" =>\\ \_ ?* [\{.Ø36\}|\{.l36\}|\{.nge36\}]]; \medskip \\
		define RuQc [["@Q1" => \_ ?* [\{.Ø36\}|\{.ye36\}]]\\ .o. ["@Q2" => \_ ?* \{.Ø36\}]\\ .o. ["@Q3" => \_ ?* [\{.Ø36\}|\{.nge36\}]]]; \medskip \\
		define RuQt ["-QT" => \_ ?* [\{.l36\}|\{.ntu36\}]]}
\end{exercise}

Rule \texttt{RuAj} in R\ref{R:18} allows adjectives, compounds made of two adjectives and complex adjective stems (see ref{tp:59} \nameref{tp:59}, p. \pageref{tp:59}) to be completed as verbal stems by suffixes \texttt{-Ø-, -l-, -nge-} or \texttt{-ntu-}, slot 36 (see section \ref{sec:06} \nameref{sec:06}, p. \pageref{sec:06}).

Rule \texttt{RuAv} allows adverbs and complex adverb stems (see \ref{tp:59}, p. \pageref{tp:59}) to be completed as verbal stems by suffixes \texttt{-Ø-, -l-, -nge-} or \texttt{-ntu-}, slot 36.

Rule \texttt{RuNn} allows nouns, proper nouns, nominal compounds and complex noun stems (\ref{tp:59}, p. \pageref{tp:59}) to be completed as verbal stems by suffixes \texttt{-Ø-, -nge-, -tu-} or \texttt{-ye-}, slot 36.

Rule \texttt{RuNu} allows numerals to form verbal stems with suffixes \texttt{-Ø-, -l-} or \texttt{-nge-}, slot 36.

Rules \texttt{RuQc} and \texttt{RuQt} regulate verbalizing suffixes for question roots, there are only four question roots and they have diverse behaviour, so they have been encoded distinctively, as shown in D\ref{def:25}:

\begin{definition} \label{def:25}\ Question roots\\ 
	\texttt{["-QC""@Q1"\{.chem\_qué\_cuál\}]:\{chem\}\\
		|["-QC""@Q2"\{.chuchi\_cuál\}]:[\{chuchi\}|\{tuchi\}]\\
		|["-QC""@Q3"\{.chum\_cómo\}]:\{chum\}\\
		|["-QT"\{.tunte\_cuánto\}]:\{tunte\};}
\end{definition}

Question root tagged \texttt{"@Q1"} is verbalized by suffixes \texttt{-Ø-} and \texttt{-ye-} (see E\ref{ex:59} and E\ref{ex:59}, p. E\pageref{ex:59}).

Question root tagged \texttt{"@Q2"} is verbalized by suffix \texttt{-Ø-} (see E\ref{ex:61}, p. E\pageref{ex:61}).

Question roots tagged \texttt{"@Q3"} are verbalized by suffixes \texttt{-Ø-} and \texttt{-nge-} (see E\ref{ex:62} and E\ref{ex:63}, p. E\pageref{ex:62}).

Question root identified by \texttt{-QT} is verbalized by suffixes \texttt{-Ø-}, \texttt{-l-} and \texttt{-ntu-} (see E\ref{ex:64}, E\ref{ex:65} and E\ref{ex:66}, p. E\pageref{ex:64}).

All restrictions encoded in R\ref{R:18} are applied to\\ \texttt{SPNVBROOT} (D\ref{def:24}) by means of a new rule, \texttt{SPNVBSTEM} (D\ref{def:26}) displayed below, which in turn is collected by \texttt{VERBSTEM} (see D\ref{def:28}, p \pageref{def:28}).

\begin{definition} \label{def:26}\ Verbalizable single non-verbal roots\\ 
	\texttt{define SPNVBSTEM [RuAj .o. RuAv .o. RuNn .o.\\ RuNu .o. RuQc .o. RuQt .o. SPNVBROOT];}
\end{definition}

\paragraph{Non-verbal compounds.} \label{tp:57} One type is made up by two adjectives, which is recognized as an adjectival compound; another types are "adjective + noun", or two nouns, both recognized as nominal compounds. Another compound, not registered by Smeets, but present in other authors' texts, is "numeral + noun", also recognized as nominal compound.

As single non-verbal roots, these compounds may be verbalized by a suffix of slot 36. The same suffixes that verbalize single adjectives, verbalize also adjective compounds. The same suffixes that verbalize single noun roots, verbalize nominal compounds. These are collected in their own rule: \texttt{CPNVBROOT} in  D\ref{def:27}. Then, \texttt{CPNVBSTEM} applies verbalization restrictions:

\begin{definition} \label{def:27}\ Non-verbal simple compounds + verbalizer\\ 
	\texttt{define CPNVBROOT [CAjAjROOT|CAjNnROOT|\\CNnNnROOT|CNuNnROOT] SVRB;\\
	define CPNVBSTEM [CLEANu .o. RuAj .o. RuNn\\ .o. CPNVBROOT .o. CLEANd];}
\end{definition}

Forms resulting from \texttt{CPNVBSTEM} are also collected in \texttt{VERBSTEM} (see D\ref{def:28}, p \pageref{def:28}).

\paragraph{Reduplicated root stems.} \label{tp:58} As shown in table \ref{tab:06} (p \pageref{tab:06}), reduplication, even verbal one, needs what Smeets calls a stem formative (slot 36) to further attach verb suffixes. We explain the case of nominal reduplication encoding, which is analogous to the other two types, verbal and onomatopoeic.

\begin{exercise} \label{R:19} \textbf{Nominal root reduplication} \\ 
	\texttt{define NROOTNT [NeutNn .o. NROOT];\\
	define NROOTx2 [\%< NROOTNT \%>"-RNNR"];\\
	define InsNRoot [[..] -> \%< NROOTNT \%> ||\\ \%> \$[\_] "-RNNR"];\\
	define APPLYNN [NROOTx2 .o. InsNRoot];\\
	define REDNNROOT [0 <- \%<|\%> .o.\\ \_eq(APPLYNN, \%<,\%>) .o. \%<|\%>|"-RNNR" -> 0];}
\end{exercise}

First rule in R\ref{R:19} neutralizes the nominal tag (see \ref{tp:54}, p. \pageref{tp:54}). Second rule marks the reduplicated element and adds a tag to the entire structure. In \texttt{InsNRoot, [..]} (Epsilon modifier\footnote{Epsilon modifier [..]\\The LHS of a rule may be wrapped in the epsilon modifier, in which case any epsilons on the LHS get a special interpretation, where only one empty string is assumed to exist between each symbol in the input string. For example, the rule:\\ \texttt{[.a*.] -> x} will produce a transducer that maps the input string \texttt{a} unambiguously to \texttt{xxx}.\\Also, \texttt{[..]} will simply produce a rule that inserts one instance of the RHS whenever the context is matched:\\ \texttt{[..] -> x} will map \texttt{aaa} to \texttt{xaxaxax}. \newline [Hulden, M. in \href{https://code.google.com/archive/p/foma/wikis/RegularExpressionReference.wiki}{https://code.google.com/archive/p/foma/wikis/\\RegularExpressionReference.wiki}].}) produces a rule that inserts one instance of \texttt{< NROOTNT >} in between \texttt{>} and \texttt{"-RNNR"}, which is the right side of the form defined in the previous rule \texttt{NROOTx2}. Rule \texttt{APPLYNN} combines and applies previous configurations. Finally, \texttt{REDNNROOT} cleans \texttt{<} and \texttt{>} from the grammatical representation, filter the form out of the previous rule, and clean any tag from the lexical side, to end up in a clean analysis (see examples E\ref{ex:32}, p. \pageref{ex:32}; E\ref{ex:48}, p. \pageref{ex:48}; E\ref{ex:146}, p. \pageref{ex:146} and E\ref{ex:151}, p. \pageref{ex:151}):

\begin{exercise} \label{R:20} \textbf{Reduplicated roots stem formation} \\ 
	\texttt{define REDSTEMS [[REDONROOT|REDVBROOT|\\REDNNROOT] SSFR];\\
		define RuRdOnSt ["-RONR" => \_ ?* \{.Ø36\}];\\
		define RuRdVbSt ["-RVBR" => \_ ?* [\{.Ø36\}|\\ \{.nge36\}|\{.tu36\}|\{.ye36\}]];\\
		define RuRdNnSt ["-RNNR" => \_ ?* [\{.nge36\}|\\ \{.tu36\}]];\\
		define RDROOTSTEM [RuRdOnSt .o. RuRdVbSt .o.\\ RuRdNnSt .o. REDSTEMS];}
\end{exercise}

R\ref{R:20} assigns the appropriate \texttt{+SFR} to each type of reduplicated root to convert them into verbal stems. Reduplicated noun, onomatopoeia and verb roots forming stems are collected in \texttt{RDROOTSTEM}, which in turn will be part of the \texttt{VERBSTEM} definition (see D\ref{def:28}, p \pageref{def:28}).

\paragraph{Complex single root stems.} \label{tp:59} As we have explained before (\nameref{tp:53}, p. \pageref{tp:53}), stems made up by a single root, a compound or a reduplicated root that incorporates at least one suffix (rarely more than three) into the structure are considered "complex stems".

Complex single root stems (one root plus one or more suffixes forming a verb stem, see \ref{it:11}, p. \pageref{it:11}) that we encode are adjectival, adverbial, nominal, numeral, questions and nominalized verbs (for the later see sections \ref{sec:12} \nameref{sec:12}, p. \pageref{sec:12}; and \ref{sec:13} \nameref{sec:13}, p. \pageref{sec:13}).

We explain here the complex nominalized verb stem. The other ones follow the same procedure with the appropriate rules for their category; they were listed in item \ref{it:11}, p. \pageref{it:11} as "Complex single root stems".

\begin{exercise} \label{R:21} \textbf{Complex nominalized verb stem: 1\textsuperscript{st} step} \\ 
	\texttt{define formCXVBROOT [[IVROOT|TVROOT] (CA)\\ (TRFAC) (REF) (ST) (HH) (NRLD) \newline [FLECNOM|NMZ] SVRB];}
\end{exercise}

In R\ref{R:21} we have a composition of 9 rules, some of them including two or three sub-rules. As in the treatment of compounds, we first define the form and order of elements in the stem. In this case \texttt{formCXVBROOT} states that the stem begins with a transitive or intransitive verb root. Then there is a series of suffixes that are optional, which means that they can co-occur (rarely more than three) in any combination, respecting the order. These suffixes are causative \emph{-l-} or \emph{-m-}, slot 34; factitive \emph{-ka-} or transitivizer \emph{-tu-}, slot 33; reflexive/reciprocal \emph{-w-}, slot 31; stative \emph{-le-}, slot 28; hither \emph{-pa-}, slot 17; and non-realized situation \emph{-a-}, slot 9. Then come the obligatory nominalizers, those may be inflectional (see \ref{sec:12}, p. \pageref{sec:12}) or derivational (see \ref{sec:13}, p. \pageref{sec:13}). A verbalizer, slot 36, completes the stem.

\begin{exercise} \label{R:22} \textbf{Complex nominalized verb stem: 2\textsuperscript{nd} step} \\
	\texttt{define neutCXVb [NeutAdjdo .o. NeutAdjqe .o.\\ NeutCa .o. NeutFac .o. NeutHh .o. NeutNrld .o. NeutNomag .o. NeutPvn .o. NeutRef .o. NeutSt\\ .o. NeutSvn .o. NeutTr .o. NeutIv .o. NeutTv]; \medskip
	define CXVBROOT [neutCXVb .o. formCXVBROOT]; \smallskip \\
	define RuCxV01 [$\sim$\$[["-iv0"|"-tv0"] ?* ["+OVN"\\|"+IVN"|"+TVN"|"+AVN"|\{.Ø4\}|"+CSVN"|"+NOMPI"\\|"+NOM"]]];}
\end{exercise}

Tag neutralization of all the members in the stem, and the application to the stem form comes in the second and third rules. Rule \texttt{RuCxV01} specifies which of the nominalizing suffixes do not form part of this stem. Those are not neutralized because while forbidding them, they need no further interaction rules.

\begin{exercise} \label{R:23} \textbf{Complex nominalized verb stem: 3\textsuperscript{rd} step} \\
	\texttt{define RuCxV02 [["+pvn0"|"+nomag0"|"+adjqe0"]\\ => \_ ?* [\{.Ø36\}|\{.nge36\}]]; \smallskip \\
	define RuCxV03 [[\{.lu4\}|"+adjdo0"]\\ => \_ ?* \{.Ø36\}]; \smallskip \\
	define RuCxV04 [["+ca0" => \_ ?* "+adjdo0"]\\ .o. ["+ref0" => \_ \$["+pvn0"] \{.nge36\}] .o. \newline ["+tr0" => \_ \$["+nomag0"|"+pvn0"] \{.nge36\}] \\ .o. ["+st0" => \_ ?* "+svn0"]]; \smallskip \\
	define RuCxV05 [$\sim$\$["+ca0" ?* ["+tr0"|"+fac0"|\\"+ref0"|"+st0"|"+nrld0"|\{.Ø4\}]]] .o.	\newline [$\sim$\$[["+tr0"|"+ref0"] ?* ["+st0"|"+nrld0"|\\ \{.Ø4\}]]];}
\end{exercise}

Rules \texttt{RuCxV02, RuCxV03, RuCxV04, RuCxV05} regulate the interaction of all possible suffixes in the stem, including the verbalizers.

\begin{exercise} \label{R:24} \textbf{Complex nominalized verb stem: 4\textsuperscript{th} step} \\
	\texttt{define CXVBSTEM [RuCxV01 .o. RuCxV02 .o.\\ RuCxV03 .o. RuCxV04 .o. RuCxV05 .o. RuCCXVbSt\\ .o. RuPr50 .o. CXVBROOT];}
\end{exercise}

Final rule \texttt{CXVBSTEM} compiles all together producing the final possible forms for this type of stem. All complex single root stems are also collected in\\ \texttt{VERBSTEM} (see D\ref{def:28}).

\paragraph{Complex reduplicated root stem.} \label{tp:60} As it was explained with example E\ref{ex:58}, p. \pageref{ex:58}, this stem is listed as a single root complex stem because it is "one" root and "one" stem that reduplicate, i.e., the whole stem reduplicates. This form was not encoded as a compound nor as a single root, but in the section that deals with reduplicated roots. The difference between the rule for this case and the one presented in \ref{R:19} for nominal reduplication, is that the root is encoded together with the suffix, and that the identifying tag suits with the category of the stem, see below:

\begin{exercise} \label{R:25} \textbf{Verbal root reduplication}\\ 
    \texttt{define VBROOTNT [[IVROOT|TVROOT] (CA)];\\
	define VBROOTx2 [\%< [IVROOT|TVROOT] (CA)\\ \%>"-RVBR"];\\
	define RuVbCA [$\sim$\$[["-IV"|"-TV"] ?* \{.l34\}]];\\
	define InsVBRoot [[..] -> \%< VBROOTNT \%> ||\\ \%> \$[\_] "-RVBR"];\\
	define APPLYVB [RuVbCA .o. VBROOTx2 .o.\\ InsVBRoot];\\
	define REDVBROOT [0 <- \%<|\%> .o. \_eq(APPLYVB,\\ \%<,\%>) .o. \%<|\%>|"-RVBR" -> 0];}
\end{exercise} 

The differences we have mentioned in the paragraph above are found in the line starting with \texttt{"define VBROOTx2"}, where there is an optional suffix \texttt{CA} (causative), and the corresponding tag for the reduplicated verb root \texttt{"-RVBR"}. The causative suffix \emph{-üm-} is the only one found in a reduplicated stem, at least in Smeets' texts.

\paragraph{Complex compound stems.} \label{tp:61} Basically, this type of stem is formed in the same way as the "complex single root stem" (p. \pageref{tp:10}), but implicating two roots. Complex compound stems (see p. \pageref{it:11}) that we encode are:

\begin{itemize} \label{it:35}
	\item adjective (+ suffixes)\footnote{Parenthesis express optionality.} + noun (+ suffixes): see E\ref{ex:192};
	\item adjective (+ suffixes) +verb +nominalizer, see E\ref{ex:193};
	\item adverb (+suffxes) + verb, see E\ref{ex:188};
	\item noun (+suffxes) + noun (+suffxes), see E\ref{ex:194};
	\item noun (+suffxes) + verb, see E\ref{ex:189};
	\item verb (+suffxes) + noun, see E\ref{ex:191};
	\item verb (+suffxes) + verb, see E\ref{ex:190};
\end{itemize}

All complex compound stems are collected together with simple compounds in \texttt{CMPVBVAL}, see D\ref{def:20}. And as for the previous types of stems, the later ones are also collected in rule \texttt{VERBSTEM} (D\ref{def:28}), which is summarized in table \ref{tab:07}:

\begin{definition} \label{def:28} Verb stems\\ 
\texttt{define VERBSTEM [IVROOT|TVROOT|CXVBSTEM|\\CXNNSTEM|CXNNSTEM2|CXAJSTEM|CXAJSTEM2|\\CXAVSTEM|CXNUSTEM|CXQUSTEM|RDROOTSTEM|\\CMPVBVAL|CPNVBSTEM|SPNVBSTEM];}
\end{definition}

\begin{itemize} \label{it:36}
	\item \texttt{IVROOT}: Intransitive verb root
	\item \texttt{TVROOT}: Transitive verb root
	\item \texttt{CXVBSTEM}: Complex verb root stem (R\ref{R:21})
	\item \texttt{CXNNSTEM}: Complex noun root stem
	\item \texttt{CXNNSTEM2}: Complex noun root stem (form 2)
	\item \texttt{CXAJSTEM}: Complex adjective root stem
	\item \texttt{CXAJSTEM2}: Complex adjective root stem (form 2)
	\item \texttt{CXAVSTEM}: Complex adverb root stem
	\item \texttt{CXNUSTEM}: Complex numeral root stem
	\item \texttt{CXQUSTEM}: Complex question root stem
	\item \texttt{RDROOTSTEM}: Reduplicated root stems (R\ref{R:20})
	\item \texttt{CMPVBVAL}: Verbal compound stem with valence (D\ref{def:20})
	\item \texttt{CPNVBSTEM}: Verbalized non-verbal compounds (D\ref{def:27})
	\item \texttt{SPNVBSTEM}: Verbalized single non-verbal roots (D\ref{def:26})
\end{itemize}

\begin{table}[htb]
	\caption{\emph{Mapudüngun} stems}
	\label{tab:07}
	\begin{tabular}{|r|c|c|}
		\hline\noalign{\smallskip}
		Stem & \makecell{Verbalizers\\Slot 36} & Suffixes\\
		\noalign{\smallskip}\hline\noalign{\smallskip}
		Verbal root & & +Suffixes\\
		\noalign{\smallskip}\hline\noalign{\smallskip}
		Verbal root + Verbal root & & +Suffixes\\
		\noalign{\smallskip}\hline\noalign{\smallskip}
		Verbal root + Non-verbal root & & +Suffixes\\
		\noalign{\smallskip}\hline\noalign{\smallskip}
		Non-verbal root + Verbal root & & +Suffixes\\
		\noalign{\smallskip}\hline\noalign{\smallskip}
		Non-verbal root & +VRB & +Suffixes\\
		\noalign{\smallskip}\hline\noalign{\smallskip}
		Non-verbal compound & +VRB & +Suffixes\\
		\noalign{\smallskip}\hline\noalign{\smallskip}
		Reduplicated root & +SFR & +Suffixes\\
		\noalign{\smallskip}\hline\noalign{\smallskip}
		Root + suffixes & +VRB & + Suffixes\\
		\noalign{\smallskip}\hline\noalign{\smallskip}
		Root + suffixes + Root & +VRB & + Suffixes\\
		\noalign{\smallskip}\hline\noalign{\smallskip}
		(Root + suffix) reduplicated & +SFR & + Suffixes\\
		\noalign{\smallskip}\hline\noalign{\smallskip}
		Root + suffixes + Root + suffixes & +VRB & + Suffixes\\
		\noalign{\smallskip}\hline
	\end{tabular}
\end{table}

Different conformations of stems where identified in section \ref{sec:16} \nameref{sec:16}, p. \pageref{sec:16}; in this point, we expose the rules that regulate the interaction among the elements introduced above, roots (section \ref{sec:36}) and suffixes (section \ref{sec:37}), which take part of the different types of stems.

\paragraph{Complex compound stems} \label{tp:62} "Adverb + optional causative + optional transitivizer + verb" \\
Rule:\footnote{Rules are presented here in a simple way, just to show the elements involved, but actually, rules are much more complex in the system because they have to deal with the generation of the compounds, the addition of tags to carry out the processes, and the elimination of these tags once used. See this example of one of the simplest rules in the FST script, which does not have to deal with the addition of a verbalizer because there is a verb root implied:\\
\texttt{\#\#\# Question / Verb\\
define ensCQtVbROOT \newline [[\%< QROOT \%\# \%< [IVROOT2|TVROOT2]]"-QVCR"];\\
define neutCQtVb \newline [NeutIv .o. NeutQc .o. NeutQt .o. NeutTv];\\
define preCQtVbROOT [\_eq(ensCQtVbROOT, \%< , \%\#)];\\
define CQtVbSTEM [RuIVCQtVb .o. RuTVCQtVb .o. RuCCXVbSt .o. [neutCQtVb .o. formCQtVbROOT]];\\
define CMPVBVAL [CLEANu .o. CQtVbSTEM .o. CLEANd];}} \texttt{AVROOT (CA) (TR) [IVROOT2|TVROOT2];}

\begin{example} \label{ex:188} [Smeets, I. 2008: 387 (26)] \cite{RefB:21}\\
	\emph{ñi \textbf{pülle-tu-pe}-lu} 'he came close to see'\\	\texttt{-SP.ñi\_mi\_su\\-AV.pülle\_cerca+TR.tu33-TV.pe\_ver+SVN.lu4}\\
\end{example}
"Noun + optional transitivizer or factitive + verb"\\
Rule: \texttt{NROOT (TRFAC) [IVROOT2|TVROOT2];}

\begin{example} \label{ex:189} [Smeets, I. 2008: 358 (5)] \cite{RefB:21}\\
	\emph{\textbf{trari-ntuku}-künu-nge-ke-fu-y} 'they were caught and left tied up'\\
	\texttt{-NN.trari\_amarra-TV.tuku\_poner+PFPS.künu32\\+PASS.nge23+CF.ke14+IPD.fu8+IND.y4+3.Ø3}\\
\end{example}
"Verb + optional experiencer + optional causative + optional transitivizer or factitive + optional reflexive + optional hither or locative + verb"\\
Rule: \texttt{[TVROOT|IVROOT] (EXPOO) (CA) (TRFAC)\\ (REF) (HHLOC) [IVROOT2|TVROOT2];}

\begin{example} \label{ex:190} [Smeets, I. 2008: 408 (28)] \cite{RefB:21}\\
	\emph{ñi \textbf{ru-pa-aku}-lu} 'he has gone by'\\
	\texttt{-SP.ñi\_mi\_su\\-IV.ru\_pasar+HH.pa17-IV.aku\_llegar-CR.IV\\+SVN.lu4}\\
	\end{example}
"Verb + optional causative + optional transitivizer or factititve + optional reflexive + optional hither + noun"\\
Rule: \texttt{[TVROOT|IVROOT] (CA) (TRFAC) (REF)\\ (HH) NROOT2;}

\begin{example} \label{ex:191} [Smeets, I. 2008: 456 (8)] \cite{RefB:21}\\
	\emph{\textbf{kim-el-tu-che}-ke-fu-y} 'he used to teach people'\\
	\texttt{-TV.kim\_saber+CA.l34+TR.tu33-NN.che\_persona\\+CF.ke14+IPD.fu8+IND.y4+3.Ø3}\\
\end{example}
"Adjective + transitivizer or factitive + noun + optional derivational nominalizer \texttt{+VRB}"\\
Rule: \texttt{AJROOT (TRFAC) NROOT2 (NMZ) SVRB;}

\begin{example} \label{ex:192} [Smeets, I. 2008: 90 (36)] \cite{RefB:21}\\
	\emph{\textbf{wisa-ka-sungu}-n, ta eymi} 'what a dirty talker you [are]!'\\
	\texttt{-AJ.weda\_malo+FAC.ka33-NN.düngu\_palabra\\+VRB.Ø36+PVN.n4\\-AP.ta\_el\\-PP.eymi\_tu}\\
\end{example}
"Adjective + optional transitivizer or factitive + verb + derivational nominalizer \texttt{+VRB}"\\
Rule: \texttt{AJROOT (TRFAC) [IVROOT2|TVROOT2] NMZ SVRB;}

\begin{example} \label{ex:193} \\
	\emph{\textbf{küme-ka-puru-fe-nge}-y} 'he is (always) a good dancer'\\
	\texttt{-AJ.küme\_bueno+FAC.ka33-IV.puru\_bailar\\+NOMAG.fe+VRB.nge36-IV+IND.y4+3.Ø3}\\
\end{example}
"Noun + optional transitivizer or factitive or non class-change suffixes + noun + optional non class-change suffixes or derivational nominalizers \texttt{+VRB}"\\
Rule: \texttt{NROOT (TRFAC|NCC) NROOT2 (NCC|NMZ) SVRB;}

\begin{example} \label{ex:194} [Smeets, I. 2008: 459 (36)] \cite{RefB:21}\\
	\emph{ta-yiñ pu \textbf{peñi-wen-lamngen-wen-nge}-n} 'we are all related as brothers and sisters' lit: 'this is our brothers relation sisters relation'\\
	\texttt{-AP.ta\_este-SP.yiñ\_nuestro-COLL.pu\\-NN.peñi\_hermano+REL.wen-NN.lamngen\_hermana\\+REL.wen+VRB.nge36-IV+PVN.n4}
\end{example}

\subsection{Morphotactics of verb suffixes} \label{sec:46}

\paragraph{} \label{tp:63} In section "\ref{sec:05} \nameref{sec:05}", p. \pageref{sec:05}, we have explained that suffixes belonging to the same slot are mutually exclusive. There are about eighty verbal suffixes spread in thirty-six slots. Some suffixes exclude others for grammatical or semantic reasons, for example, once a verb has taken an inflectional nominalizer, slot 4, it can not take suffixes of mood (slot 4), person (slot 3) and number (slot 2).

To start treating suffixes co-occurrence, we first established the suffix sequence with all the possible variants generated by suffix mobility (see \ref{sec:10}, p. \pageref{sec:10} and \ref{sec:50}, p. \pageref{sec:50}), see next rule:

\begin{definition} \label{def:29}\ Verb suffixes\\ 
	\texttt{define VERBSUFFIX [(REF) (EXPOO) (PASS) (REF)\\ (TR) (CA) (REF) (TRFAC) (FORCE) (BEN) (FORCE)\\ (PRPSPFPS) (REF) (HH) (CIRCINT) (PLAYSIM)\\ (MIO) (STPR) (BEN) (OS) (IMMSUD) (PLR) (IO)\\ (PASS) (FORCESAT) (PLR) (FORCE) (TH)\\ (PASS1A2A) (PLAYSIM) (IMMSUD) (TH) (PS) (ITR)\\ (HHLOC) (TH) (PS) (REF) (RE) (RECONT) (PLPF15) (CF14) (PX) (REP) (RE) (AFF) (NEG) (NRLD)\\ (IPD) (PLPF07) (EIDO) (CF05) \newline [[[(MOOD) [PERSON|PTMT] (NUMBER)] (DS)]| \newline [[FLECNOM|NMZ] (DS) (NCC) (CC) (INST)]]];}
\end{definition}

The names or tags appearing in D\ref{def:29} encode the suffixes assigned to each slot (see \ref{sec:37} \nameref{sec:37}, p. \pageref{sec:37}).

The first thing that may call the attention is repetition of some tags in different positions, e.g. \texttt{PASS, REF, FORCE, IMMSUD}, etc. This is to deal with suffix mobility (see \ref{sec:10}, p. \pageref{sec:10} and \ref{sec:50}, p. \pageref{sec:50}).

Also note that almost all suffixes are marked as optional, they are between parenthesis, except for \texttt{PERSON, PTMT, FLECNOM} and \texttt{NMZ}. The \emph{Mapuche} verb is either finite (\texttt{PERSON} and \texttt{PTMT}) or nominalized (\texttt{FLECNOM} or \texttt{NMZ}). These are the obligatory suffixes for those forms.

\paragraph{Methodology.} \label{tp:64} To encode suffixes occurrence in the \emph{Mapuche} verb form, we started incorporating the minimal verb form, i.e., an intransitive verb root plus suffixes expressing mood, person and number (see annex \ref{anx:10} "\nameref{anx:10}", p. \pageref{anx:10}), which are obligatory in a finite verb form. We continued adding the transitive verb related suffixes. So, we first established a set of rules dealing with the minimal forms for both, intransitive and transitive verbs (see annex \ref{anx:11} "\nameref{anx:11}", p. \pageref{anx:11}).

\subsubsection{Verb paradigms} \label{sec:47}

\paragraph{} \label{tp:65} In D\ref{def:29} above, the last two lines reflect the two forms a verb may take. Penultimate line corresponds to finite forms; in slot 4 is mood, in slot 3 is person or the portmanteau morphs\footnote{"Portmanteau morphs which include a subject marker are assigned subject position (slot 3)" [Smeets 2008: 152] \cite{RefB:21}. We have also seen that assigning portmanteau morphs in this position allows the conditional marker, obligatory in negative imperative forms, appears in its natural position, slot 4 for mood.} (see \texttt{slot-03PTMT.aff} in annex \ref{anx:03}, p. \pageref{it:53}); in slot 2 is number, and dative subject (used in transitive forms) is in slot 1.

It was also necessary to incorporate suffixes assigned to slots 23 and 6, as they complete the transitive verb paradigm (see section \ref{sec:11} \nameref{sec:11}, p. \pageref{sec:11}), and negation suffixes in slot 10, even though they are not strictly obligatory and occur in transitive and intransitive forms, they complement with mood suffixes and have a particular incidence in the case of imperative negative forms, (see annex \ref{anx:12} "\nameref{anx:12}", p. \pageref{anx:12}).

\begin{table}[htb]
	\caption{Intransitive and transitive suffixes per slot}
	\label{tab:08}
	\begin{tabular}{|c|c|c|c|c|c|c|c|}
		\hline\noalign{\smallskip}
		Slot & 23 & 10 & 6 & 4 & 3 & 2 & 1 \\
		\noalign{\smallskip}\hline\noalign{\smallskip}
		Itr & - & neg. & - & mood & \makecell{pers.\\ptmt} & num. & - \\
		\noalign{\smallskip}\hline\noalign{\smallskip}
		Tr & agent & neg. & obj. & mood & \makecell{pers.\\ptmt} & num. & \makecell{dative\\subj.} \\
		\noalign{\smallskip}\hline
	\end{tabular}
\end{table}

Table \ref{tab:08} shows suffixes per slot\footnote{"It is remarkable that the subject-object paradigm is completed with suffixes which occupy a position in between derivational suffixes, away from the inflectional block at the end of a verb form. The suffixes \emph{-mu-} \texttt{+2A} and \emph{-w-} \texttt{+1A} share their position, slot 23, with the passive marker \emph{-nge-} [Smeets, I. 2008: 161] \cite{RefB:21}.} implied in transitive and intransitive \emph{Mapuche} verbs. Not all suffixes in table \ref{tab:08} co-occur in a transitive form, for instance, agent markers (slot 23) do not co-occur with direct objects (slot 6) or dative subjects (slot 1).

To regulate the verbal paradigms, thirty-three rules were necessary, some of them containing sub-rules, and some including the interaction with inflectional nominalizers, slot 4. No reference to mood, person or number may be made when a verb takes one of the nominalizers, but nominalized verbs may include agents (E\ref{ex:195}, E\ref{ex:196}) or objects with the corresponding dative subject (E\ref{ex:195}, E\ref{ex:198}):

\begin{example} \label{ex:195} [Smeets, I. 2008: 269 (11)] \cite{RefB:21}\\
	\emph{mütrüm-\textbf{uw}-lu} 'his calling to' \\
	\texttt{-TV.mütrüm\_llamar+1A.w23+SVN.lu4}
\end{example}

\begin{example} \label{ex:196} [Smeets, I. 2008: 269 (14)] \cite{RefB:21}\\
	\emph{fey-pi-\textbf{mu}-a-fiel} 'what you will tell me' \\
	\texttt{-TV.feypi\_decir+2A.mu23+NRLD.a9+TVN.fiel4}
\end{example}

\begin{example} \label{ex:197} [Smeets, I. 2008: 394 (38)] \cite{RefB:21}\\
	\emph{chem-pi-\textbf{e}-t-\textbf{ew}} 'what they where told by' \\
	\texttt{-QC.chem\_qué-TV.pi\_decir-CR.TV\\+IDO.e6+AVN.t4+DS3A.ew1}
\end{example}

\begin{example} \label{ex:198} [Smeets, I. 2008: 485 (5)] \cite{RefB:21}\\
	\emph{pe-\textbf{fi}-lu iñche} 'at my seeing her' \\
	\texttt{-TV.pe\_ver+EDO.fi6+SVN.lu4\\-PP.iñche\_yo}
\end{example}

\begin{exercise} \label{R:26} \textbf{Dependency rule 1} \\
    \texttt{define RuDp01 \newline [["+DS3A"|"+DS12A"] => "+IDO" ?* \_ ];}
\end{exercise}

R\ref{R:26} is what we call a "dependency" rule, it says that for suffixes \texttt{+DS3A} and \texttt{+DS12A} to occur it must previously occur the suffix \texttt{+IDO}, i.e., \texttt{+DS3A} and \texttt{+DS12A} depend on \texttt{+IDO} occurrence.

\begin{exercise} \label{R:27} \textbf{Prohibition rule 10} \\
    \texttt{define RuPr10 [$\sim$\$["+CND" ?* [["+1"\{.Ø3\}]\\ | ["+3"[\{.Ø3\}|\{.ng3\}]]]]];}
\end{exercise}

R\ref{R:27} is a prohibition rule. The combination of symbols \texttt{$\sim$\$}\footnote{\texttt{$\sim$X} calculates the complement of \texttt{X}, i.e., finds all the elements in the group that are not part of \texttt{X}, or that are not \texttt{X}. \texttt{\$X} denotes the language that contains a sub-string drawn from the language \texttt{X} [Hulden, M. in \href{https://code.google.com/archive/p/foma/wikis/RegularExpressionReference.wiki}{https://code.google.com/archive/p/foma/wikis/\\RegularExpressionReference.wiki}].} may be read as "it can not be the case that", and the rest of this regexp is read as "the conditional is followed by a 1\textsuperscript{st} person suffix in its null form \emph{-Ø-}, or the 3\textsuperscript{st} person suffix in its forms null or \emph{-ng-}.

\begin{exercise} \label{R:28} \textbf{Obligation rule 9} \\
    \texttt{define RuOb09 [[["+NEG"[\{.ki10\}|\{.kino10\}]]\\ => \_ ?* "+CNI"];}
\end{exercise}

R\ref{R:28} is an obligation rule which regulates the obligatory occurrence of the conditional marker when there is a negation in the imperative form (see e.r. \ref{tp:135} and annex \ref{anx:12} \nameref{anx:12}, p. \pageref{anx:12}).

\subsubsection{Nominalized verbs} \label{sec:48}

\paragraph{} \label{tp:66} Last line of D\ref{def:29} (p. \pageref{def:29}), reflects the form of a nominalized verb, either by inflectional (see section \ref{sec:12} \nameref{sec:12}, p. \pageref{sec:12}) or derivational (see section \ref{sec:13} \nameref{sec:13}, p. \pageref{sec:13}) nominalizers. In both cases a nominalized verb may be followed by a dative subject (see E\ref{ex:197}), a non class-changing suffix (see \texttt{NCC.aff} in annex \ref{anx:06}, p. \pageref{anx:06}), a class-changing suffix (see \texttt{CC.aff} in annex \ref{anx:06}, p. \pageref{anx:06}), or the instrumental suffix (see section \ref{sec:15} \nameref{sec:15}, p. \pageref{sec:15}).

To regulate verb nominalization twelve rules were added. Note that these rules regulate co-occurrence among the suffixes of "\texttt{[FLECNOM|NMZ] (DS) (NCC) (CC) (INST)}",\\ and some times with suffixes from other slots; but in general, there are other rules to deal with co-occurrence of these suffixes, or the ones belonging to the transitive and intransitive paradigms, and the derivational ones.

\begin{exercise} \label{R:29} \textbf{nominalization prohibition for completive subjective verbal noun} \\
    \texttt{define RuPr19 [$\sim$\$["+CSVN" ?* ["+DS3A"|\\"+DS12A"|"+INST"|"+ADJ"]]];}
\end{exercise}

R\ref{R:29} forbids dative subject suffixes (slot 1), instrumental, or adjectivizer (class-changing suffix), to appear when the verb has been nominalized by the "completive subjective verbal noun" (slot 4).

\begin{exercise} \label{R:30} \textbf{Obligation for agentive verbal noun} \\
    \texttt{define RuOb12 ["+AVN" =>\\ "+IDO" \$[\_] ["+DS3A"{.ew1}]];}
\end{exercise}

R\ref{R:30} forces the "agentive verbal noun" (slot 4) to occur together with the "internal direct object" (slot 6) and the "dative subject for 3\textsuperscript{rd} person agent" (slot 1) in its form \emph{-ew} (see E\ref{ex:22}, p. \pageref{ex:22} and E\ref{ex:197}, p. \pageref{ex:197}).

\begin{exercise} \label{R:31} \textbf{Only plain verbal noun may be adverbialized} \\
    \texttt{define RuDp05 ["+ADV" => "+PVN" ?* \_ ];}
\end{exercise}

R\ref{R:31} states that the class-changing suffix \emph{-tu} may only adverbialize a verb nominalized by the "plain verbal noun" \emph{-n-} (see E\ref{ex:34}, p. \pageref{ex:34}). In other words the adverbializer depends on the "plain verbal noun" to occur with a verb.

Examples of inflectionally nominalized verbs may be found through e.r. \ref{tp:178}, e.r. \ref{tp:179}, e.r. \ref{tp:180}, e.r. \ref{tp:181}, e.r. \ref{tp:182}, e.r. \ref{tp:183}, e.r. \ref{tp:184} and e.r. \ref{tp:185}.

Examples of derivationally nominalized verbs are: E\ref{ex:29}, E\ref{ex:30}, E\ref{ex:31}, E\ref{ex:32}, E\ref{ex:87}, and the following ones:

\begin{example} \label{ex:199} [Smeets, I. 2008: 314 (\emph{-Ø})] \cite{RefB:21}\\
	\emph{anü-m-ka} 'planting' \\
	\texttt{-IV.anü\_sentar+CA.m34+FAC.ka33+NOM.Ø}
\end{example}

\begin{example} \label{ex:200} [Smeets, I. 2008: 314 (\emph{-Ø})] \cite{RefB:21}\\
	\emph{ül-kantu} 'song' \\
	\texttt{-NN.ül\_canto+VRB.Ø36+PLAY.kantu22+NOM.Ø}
\end{example}

\begin{example} \label{ex:201} [Smeets, I. 2008: 314 (\emph{-Ø})] \cite{RefB:21}\\
	\emph{yall-tuku} 'illegitimate child' \\
	\texttt{-NN.yall\_hijo-de-un-hombre\\-TV.tuku\_poner-CR.TV+NOM.Ø}
\end{example}

\begin{example} \label{ex:202} [Smeets, I. 2008: 314 (\emph{-Ø})] \cite{RefB:21}\\
	\emph{ru-pa} 'time' \\
	\texttt{-IV.ru\_pasar+HH.pa17+NOM.Ø}
\end{example}

\begin{example} \label{ex:203} [Smeets, I. 2008: 312 (8)] \cite{RefB:21}\\
	\emph{angkü-m-tu-\textbf{we}} 'poison', 'device to dry things' \\
	\texttt{-IV.angkü\_secar+CA.m34+TR.tu33+NOMPI.we}
\end{example}

\subsubsection{Occurrence of suffixes between slots 5 and 35} \label{sec:49}

\paragraph{} \label{tp:67} There are thirty more rules to regulate the occurrence of suffixes that are not obligatory in the minimal transitive or intransitive forms. Most of the rules come from descriptions of the suffixes made by Smeets, for example the rule for the reflexive reciprocal \emph{-w-}: "The suffix \emph{-w-}  does not combine with a suffix in slot 23, 6 or 1. The reflexive morpheme \emph{-w-} may occur with intransitive verbs, i.e., with verbs which do not take a suffix in slot 6" [Smeets, I. 2008: 291] \cite{RefB:21}; R\ref{R:32} reflects the previous description:

\begin{exercise} \label{R:32} \textbf{Reflexive do not occur in transitive forms} \\
    \texttt{define RuPr48 [$\sim$\$["+REF" ?* ["+REF"|"+PASS"\\|"+1A"|"+2A"|"+IDO"]]];}
\end{exercise}

In the description of (non) combinations of \texttt{+REF}, Smeets also mentions suffixes of slot 1. These are not collected in R\ref{R:32} because there is a previous dependency rule (R\ref{R:26}, p. \pageref{R:26}) stating that dative subjects (slot 1) need the \texttt{+IDO} suffix (slot 6) to occur; as this one is forbidden to occur with \texttt{+REF} the condition does not fulfil for the \texttt{+DS} (slot 1) to occur.

\begin{exercise} \label{R:33} \textbf{More involved object obligatory contexts} \\
    \texttt{define RuOb16 ["+MIO" => "+CIRC" ?* \_ ,\\ \_ ?* ["+PASS"|"+EDO"|"+TVN"]];}
\end{exercise}

The rule presented in R\ref{R:33} derives from what we have found in Smeets' examples, there are no explicit combination rules for the more involved object suffix labelled \texttt{+MIO}. It is important to rule this suffix due to its form \emph{-l-} after vowel, \emph{-ül-} after consonant or semi-vowel, sometimes \emph{-el-} after \emph{r}. These forms coincide with other suffixes forms like stative's or causative's ones, which are proximate in their occurrence position, therefore, they may be erroneously identified.

There are fourteen examples given by Smeets, where \texttt{+MIO} (slot 29) is present. In six of them is preceded by \texttt{+CIRC} (slot 30), circular (erratic) movement suffix \emph{-iaw-}. There are another six where it co-occurs with \texttt{+EDO} (slot 6), external direct object suffix \emph{-fi-}. One where it co-occurs with \texttt{+PASS} (slot 23), passive \emph{-nge-}. And one more where it co-occurs with \texttt{+TVN} (slot 4), transitive verbal noun \emph{-fiel-}.

\texttt{+MIO} "indicates a more direct, intense or complete involvement of the patient in the event" [Smeets, I. 2008: 287] \cite{RefB:21}; \texttt{+CIRC} "denotes an ongoing event which involves movement in no particular direction" [Smeets, I. 2008: 288] \cite{RefB:21}.

It is not clear for us the semantic or grammatical relation between \texttt{+MIO} and \texttt{+CIRC}, but when these two suffixes co-occur, nor \texttt{+PASS} nor \texttt{+EDO} occur. On the other hand, the other three suffixes in the rule have a grammatical relation with \texttt{+MIO}. For \texttt{+EDO}, the external direct object, \texttt{+MIO} gives a further degree of prominence to the object, see following example:

\begin{example} \label{ex:204} [Smeets, I. 2008: 288 (6)] \cite{RefB:21}\\
	\emph{koyla-tu-künu-\textbf{l-fi}-ñ} 'I lied to him' \\
	\texttt{-NN.koyla\_mentira+VRB.tu36+PFPS.künu32\\+MIO.l29+EDO.fi6+IND1SG.n3}
\end{example}

The same happens with the objects denoted in a passive \texttt{+PASS} construction (E\ref{ex:205}), or in a transitive \texttt{+TVN} clause (E\ref{ex:206}). In the case of a verb nominalized by the transitive verbal noun suffix, \texttt{+CIRC} and \texttt{+TVN} may co-occur (E\ref{ex:207}), rule  R\ref{R:33} does not prevent it.

\begin{example} \label{ex:205} [Smeets, I. 2008: 397 (62)] \cite{RefB:21}\\
	\emph{yiñ ngünen-ka-\textbf{l-nge}-we-no-a-m} 'we are no longer deceived' \\
	\texttt{-SP.yiñ\_nuestro\\-NN.ngünen\_engaño+VRB.Ø36+FAC.ka33+MIO.l29\\+PASS.nge23+PS.we19+NEG.no10+NRLD.a9+IVN.m4}
\end{example}

\begin{example} \label{ex:206} [Smeets, I. 2008: 288 (5)] \cite{RefB:21}\\
	\emph{eymi mi wirar-\textbf{ül}-meke-ke-\textbf{fiel}-mew iñche} 'you are always shouting at me' \\
	\texttt{-PP.eymi\_tu -SP.mi\_tu\_tuyo\\-IV.wirar\_gritar+MIO.l29+PR.meke28+CF.ke14\\+TVN.fiel4+INST.mew}
\end{example}

\begin{example} \label{ex:207} [Smeets, I. 2008: 398 (5)] \cite{RefB:21}\\
	\emph{ñi küdaw-\textbf{kiaw-ül}-el-\textbf{fiel} pu ülmen} 'he worked around for the rich people' \\
	\texttt{-SP.ñi\_mi\_su\\-IV.küdaw\_trabajar+CIRC.iaw30+MIO.l29\\+BEN.el27+TVN.fiel4 \\-COLL.pu -NN.ülmen\_adinerado}
\end{example}

\begin{exercise} \label{R:34} \textbf{Circular movement context restrictions} \\
    \texttt{define RuPr41 [$\sim$\$[["+CIRC"|"+INT"|"+ST"|"+PR"] ?* ["+ST"|"+PR"]]];}
\end{exercise}

\subsubsection{Treating suffix mobility} \label{sec:50}

\paragraph{} \label{tp:68} As it was explained in section \ref{sec:10}, p. \pageref{sec:10}, some suffixes may occur in different positions, this is call "mobility". To deal with it, we have declared the slot containing the suffix in all the positions it may appear along the suffixes chain (see D\ref{def:29}, p. \pageref{def:29}). When a slot holds more than one suffix and only one of them is mobile, we have created a new file containing only the mobile suffix. For example, slot 17, encoded in file \texttt{slot-17.aff}, holds hither \emph{-pa-} and locative \emph{-pu-} and only \texttt{+HH} is mobile. We have created the file \texttt{slot-17M.aff} that holds the hither only, and is declared as \texttt{HH} in the script. In  D\ref{def:29}, \texttt{HH} is found once alone and once together with \texttt{LOC}, as \texttt{HHLOC}, these are the two positions where it may occur; the same stands for the other mobile suffixes.

In R\ref{R:34} above, also in R\ref{R:32}, p. \pageref{R:32}, there are suffixes repeated on both sides of the expression, left and right of the \texttt{?*} symbols. These are prohibition rules, interpreted as "it can not be the case that", as we have explained for R\ref{R:27}, p. \pageref{R:27}. So, "it can not be the case that \texttt{+ST} co-occurs with \texttt{+ST}, or \texttt{+PR} with \texttt{+PR}", in R\ref{R:34}. And "it can not be the case that \texttt{+REF} co-occurs with \texttt{+REF}", in R\ref{R:32}, p. \pageref{R:32}. All of this is to avoid that the same suffix may be recognized twice in a verb form, due to being declared in two different positions. These rules are also used to exclude the co-occurrence of these suffixes with different ones for grammatical or semantic reasons, e.g., R\ref{R:34} states that "it can not be the case that \texttt{+CIRC} co-occurs with \texttt{+ST}, one expresses the opposite idea of the other: circular movement / stative; or \texttt{+INT} (intensifier) with \texttt{+PR} (progressive)" .

\subsubsection{Over-generation} \label{sec:51}

\paragraph{} \label{tp:69} One of the problems derived from encoding homograph suffixes, or suffixes that do not phonetically realize (null suffixes \emph{Ø}) is that first ones may be mistakenly recognized, and the seconds could be virtually recognized at any position. To solve these issues we have generated some rules that do not have to do with the \emph{Mapudüngun} morphotactics, but help in preventing wrong recognition. These rules have derived from the observation of analysing results (see \nameref{tp:91}, p. \pageref{tp:91}).

\begin{exercise} \label{R:35} \textbf{Forbidden null morpheme sequence occurrence} \\
    \texttt{define RuPr46 [$\sim$\$["+NOM" ?* [["+OVN"{.Ø4}]|\newline["+SVN"\{.Ø4\}]|"+IMP"|["+1"\{.Ø3\}]|["+3"\{.Ø3\}]|\newline["+SG"\{.Ø2\}]|"+DS12A"]]];}
\end{exercise}

Rule R\ref{R:35} forbids the nominalizer which has a null form \emph{-Ø-} to be followed by other suffixes that also have a null realization: \texttt{+OVN, +SVN, +IMP, +1, +3, +SG, +DS12A}

\paragraph{Complete forms.} \label{tp:70} There is another set of eight rules that acts upon the entire verb forms when the stems are verbalized roots, complex stems, compounds or complex compounds. Some suffixes in the stem condition the entire form.

\begin{exercise} \label{R:36} \textbf{nominalization of verbalized noun} \\
    \texttt{define RuPr51 [$\sim$\$[["-nn0"|"-pn0"] \$["+VRB"] \\ \$["+OVN"|"+IVN"|"+TVN"|"+AVN"|"+SVN"\\|"+CSVN"] "+ADV"]];}
\end{exercise}

The prohibition rule R\ref{R:36} states that a noun or proper noun once verbalized may be nominalized by any of the inflectional nominalizers but not adverbialized. Note that in this rule coexist neutralized tags belonging to the stem with non-neutralized ones that belong to suffixe chain of the verb.

\subsubsection{Special roots} \label{sec:52}

\paragraph{} \label{tp:71} In this section we explain how roots of \nameref{sec:19}, p. \pageref{sec:19}, and \nameref{sec:20}, p. \pageref{sec:20}, are encoded. Also, some other special cases are explained:

\begin{itemize} \label{it:37}
	\item[] List of special roots {\fontsize{8pt}{11pt}\selectfont
		\item[]\texttt{["-AV""@SC01"\{.fül\_cerca\}]: \{fül\}}
		\item[]\texttt{["-AV""@SC01"\{.pülle\_cerca\}]: \{pülle\}}
		\item[]\texttt{["-IV""@SC01"\{.llekü\_acercar\}]: \{llekü\}}
		\item[]\texttt{["-IV""@SC02"\{.chekod\_encuclillar\}]: \{chekod\}}
		\item[]\texttt{["-IV""@SC02"\{.kopüd\_yacer-boca-abajo\}]: \{kopüd\}}
		\item[]\texttt{["-IV""@SC02"\{.kudu\_yacer\}]: \{kudu\}}
		\item[]\texttt{["-IV""@SC02"\{.külü\_apoyar\}]: \{külü\}}
		\item[]\texttt{["-IV""@SC02"\{.llikosh\_encuclillar\}]: \{llikosh\}}
		\item[]\texttt{["-IV""@SC02"\{.payla\_yacer-de-espalda\}]: \{payla\}}
		\item[]\texttt{["-IV""@SC02"\{.potrong\_inclinar-la-cabeza\}]:\{potrong\}}
		\item[]\texttt{["-IV""@SC02"\{.potrü\_inclinar\}]: [\{potrü\}|\{potri\}]}
		\item[]\texttt{["-IV""@SC02"\{.rekül\_apoyar\}]: \{rekül\}}
		\item[]\texttt{["-TV""@SC02"\{.ünif\_extender\}]: ["@G"[\{ünif\}|\{üñif\}]]}
		\item[]\texttt{["-IV""@SC02"\{.wira\_sentar-en-ancas\}]: \{wira\}}
		\item[]\texttt{["-IV""@SC03"\{.trem\_crecer\}]: \{trem\}}
		\item[]\texttt{["-TV""@SC03"\{.kim\_saber\}]: \{kim\}}
		\item[]\texttt{["-IV""@SC04"\{.kon\_entrar\}]: \{kon\}}
		\item[]\texttt{["-IV""@SC04"\{.tripa\_salir\}]: [\{tripa\}|\{chipa\}]}
		\item[]\texttt{["-IV""@SC05"\{.püra\_subir\}]: [\{püra\}|\{ñpüra\}]}
		\item[]\texttt{["-IV""@SC06"\{.müle\_estar\_vivir\}]: [\{müle\}|\{müli\}]}
		\item[]\texttt{["-TV""@SC06"\{.meke\_ocupar\}]: [\{mek\}"@EI"]}
		\item[]\texttt{["-TV""@SC06"\{.nie\_tener\}]: [\{nee\}|\{ne\}|[\{ni\}"@E0"]]}
		\item[]\texttt{["-IV""@SC07"\{.miaw\_merodear\}]: \{miaw\}}
		\item[]\texttt{["-IV""@SC08"\{.nge\_ser\_estar\}]: \{nge\}}
		\item[]\texttt{["-IV""@SC09"\{.pepi\_ser-capaz-de\}]: \{pepi\}}
		\item[]\texttt{["-IV""@SC10"\{.ru\_pasar\}]: \{ru\}}
		\item[]\texttt{["-IV""@FA"\{.fa\_ser-esto\}]: \{fa\}}
		\item[]\texttt{["-IV""@FE"\{.fe\_ser-eso\}]: \{fe\}}
		\item[]\texttt{["-AJ""@CF"\{.küme\_bueno\}]: \{küme\}}
		\item[]\texttt{["-AJ""@CF"\{.weda\_malo\}]: [\{weda\}|\{wesha\}]}
		\item[]\texttt{["-IV""@CF"\{.aye\_reír\}]: ["@G"\{aye\}]}
		\item[]\texttt{["-IV""@CF"\{.lladkü\_entristecer\_enojar\}]: \{lladkü\}}
		\item[]\texttt{["-IV""@CF"\{.llüka\_asustar\_temer\}]: \{llüka\}}
		\item[]\texttt{["-IV""@CF"\{.welu\_intercambiar\}]: \{welu\}}
		\item[]\texttt{["-NN""@CF"\{.lofo\_lobo\_salvaje\}]: \{lofo\}}
		\item[]\texttt{["-TV""@CF"\{.yewe\_avergonzar\_respetar\}]: \{yewe\}}
		}
\end{itemize}

\begin{exercise} \label{R:37} \textbf{Realization contexts for 1\textsuperscript{st} type of defective\\ verbs} \\
    \texttt{define RuVSC01 ["@SC01" =>\\ \_ \$["+CA"|"+TR"] ["+HH"|"+TH"|"+LOC"],\\ \_ ?* ["+CA"|"+ca0"|"+TR"|"+tr0"|"+ST"],\\ \_ ?* ["+HH"|"+TH"|"+LOC"]];}
\end{exercise}

\paragraph{Defective roots,} \label{tp:72} marked \texttt{@SC01} have three obligatory contexts of realization. In the first possible context they have to be followed by causatives \emph{-l-}, \emph{-m-} (slot 34) or transitivizer \emph{-tu-} (slot 33), which in turn have to co-occur with hither \emph{-pa-}, locative \emph{-pu-} (slot 17), or thither \emph{-me-} (slot 20).

In the second possible context for roots tagged \texttt{@SC01}, they have to be followed by causatives \emph{-l-}, \emph{-m-} (slot 34), transitivizer \emph{-tu-} (slot 33) or stative \emph{-le-} (slot 28). Note that in this second context is found the neutralized tag for causative \texttt{+ca0} and transitivizer \texttt{+tr0}, this is because the rule also applies when these roots form a complex stem or a complex compound stem:

\begin{example} \label{ex:208} [Smeets, I. 2008: 316 (5)] \cite{RefB:21}\\
	\emph{fül-\textbf{üm}-tuku-fi-n} 'I put it closer to' \\
	\texttt{-AV.fül\_cerca+CA.m34-TV.tuku\_poner-CR.TV\\+EDO.fi6+IND1SG.n3}
\end{example}

In the third possible context, these roots must co-occur with hither \emph{-pa-}, locative \emph{-pu-} (slot 17), or thither \emph{-me-} (slot 20):

\begin{example} \label{ex:209} [Smeets, I. 2008: 419 (46)] \cite{RefB:21}\\
	\emph{pülle-\textbf{pu}-el} 'going to a near place' \\
	\texttt{-AV.pülle\_cerca+VRB.Ø36+LOC.pu17+OVN.el4}\\
\end{example}

\begin{exercise} \label{R:38} \textbf{Realization contexts for 2\textsuperscript{nd} type of defective\\ verbs} \\
    \texttt{define RuVSC02 ["@SC02" =>\\ \_ ?* ["-CR.IV"|"-CR.TV"],\\ \_ ?* ["+ST"|"+PRPS"|"+PFPS"|"+LOC"|\{.no10\}|\\"+NRLD"]] .o.\\ $\sim$\$["@SC02" ?* ["-CR.IV"|"-CR.TV"]\\ ?* ["+ST"|"+PRPS"|"+PFPS"]];}
\end{exercise}

Roots tagged \texttt{@SC02} have different contexts of realization too, but they also have forbidden contexts, this is why R\ref{R:38} is a concatenation of two rules. The first rule treats \texttt{@SC02} roots in compounds, when this roots are before tags \texttt{-CR.IV, -CR.TV} is because they form part of a compound, \texttt{-CR} stands for "compound root":

\begin{example} \label{ex:210} [Smeets, I. 2008: 522 (\emph{külü-})] \cite{RefB:21}\\
	\emph{\textbf{külü-ru}-pa-n antü} 'when the sun is going down' \\
	\texttt{-IV.külü\_apoyar-IV.ru\_pasar-CR.IV+HH.pa17\\+PVN.n4 -NN.antü\_sol}
\end{example}

The second context, obligatorily places \texttt{@SC02} roots together with stative \emph{-le-} (slot 28), progressive persistent \emph{-nie-} or perfect persistent \emph{-künu-} (slot 32), locative \emph{-pu-} (slot 17), negation for conditional \emph{-no-} (slot 10) or non-realized situation \emph{-a-} (slot 9):

\begin{example} \label{ex:211} [Smeets, I. 2008: 281] \cite{RefB:21}\\
	\emph{\textbf{rekül-künu}-w-üy} 'then, he leaned over' \\
	\texttt{-IV.rekül\_apoyar+PFPS.künu32+REF.w31\\+IND.y4+3.Ø3}
\end{example}

The concatenated prohibition rule states that \texttt{@SC02} roots can not be part of a compound and form a verb taking the suffixes stative \emph{-le-} (slot 28), progressive persistent \emph{-nie-} or perfect persistent \emph{-künu-} (slot 32).

\begin{exercise} \label{R:39} \textbf{Realization contexts for 1\textsuperscript{st} type of compounded defective verbs} \\
    \texttt{define RuVSC03 ["@SC03" ?* "@SC05"] =>\\ \_ ?* ["+TH"|"+HH"];}
\end{exercise}

R\ref{R:39} obliges compounds made with \texttt{@SC03} and \texttt{@SC05} roots to occur together with markers for thither \emph{-me-} (slot 20) or hither \emph{-pa-} (slot 17):

\begin{example} \label{ex:212} [Smeets, I. 2008: 262 (9)] \cite{RefB:21}\\
	\emph{\textbf{kim-püra-me-pa}-n} 'there I realized' \\
	\texttt{-TV.kim\_saber-IV.püra\_subir-CR.IV\\+TH.me20+HH.pa17}\footnote{In this example both suffixes appear, but only one is obligatory.}\texttt{+IND1SG.n3}
\end{example}

\begin{example} \label{ex:213} [Smeets, I. 2008: 381 (1)] \cite{RefB:21}\\
	\emph{\textbf{kim-püra-me}-n} 'I came to appreciate' \\
	\texttt{-TV.kim\_saber-IV.püra\_subir-CR.IV+TH.me20\\+IND1SG.n3}
\end{example}

\begin{exercise} \label{R:40} \textbf{Realization contexts for 3\textsuperscript{rd} type of defective\\ verbs} \\
    \texttt{define RuVSC04\\ $\sim$\$["@SC06" ?* ["+ST"|"+PR"|"+PRPS"|"+PFPS"]];}
\end{exercise}

Prohibition rule R\ref{R:40} forbids roots tagged \texttt{@SC06} to take suffixes stative \emph{-le-} or progressive \emph{-meke-} (slot 28), progressive persistent \emph{-nie-} or perfect persistent \emph{-künu-} (slot 32).

\begin{exercise} \label{R:41} \textbf{Realization contexts for 4\textsuperscript{th} type of defective\\ verbs} \\
    \texttt{define RuVSC05 $\sim$\$["@SC07" ?* ["+CIRC"|"+ST"|\\"+PR"|"+PRPS"|"+PFPS"]];}
\end{exercise}

Prohibition rule R\ref{R:41} forbids roots tagged \texttt{@SC07} to take suffixes of circular movement \emph{-iaw-} (slot30), stative \emph{-le-} or progressive \emph{-meke-} (slot 28), progressive persistent \emph{-nie-} or perfect persistent \emph{-künu-} (slot 32).

\begin{exercise} \label{R:42} \textbf{Realization contexts for verb \emph{nge-} 'to be'} \\
    \texttt{define RuVSC06 ["@SC08" => \_ ?* ["+CA"|"+TR"],\\ \_ ?* ["+HH"|"+TH"],\\ \_ ?* "+NEG"] .o.\\ $\sim$\$["@SC08"?*["+HH"|"+TH"]?*["+NEG"\{.la10\}]];}
\end{exercise}

R\ref{R:42} is another rule made by composition, the first sub-rule defines the contexts where the verb root \emph{nge-} 'to be / to have' must happen. The first context for this root, tagged \texttt{@SC08}, demands it to be followed either by \texttt{+CA} causative (slot 34) \emph{-l-} or \emph{-m-}, or by \texttt{+TR} transitivizer (slot 33) \emph{-tu-}:

\begin{example} \label{ex:214} [Smeets, I. 2008: 126 (28)] \cite{RefB:21}\\
	\emph{\textbf{nge-l-me}-fi-ñ} 'I have taken them' \\
	\texttt{-IV.nge\_ser+CA.l34+TH.me20}\footnote{The obligatory suffixes of this and next rule co-occur in this verb, which is another possible context.}\texttt{+EDO.fi6+IND1SG.n3}
\end{example}

The second possible context for \emph{nge-} makes it occur with hither \emph{-pa-} (slot 17) or thither \emph{-me-} (slot 20):

\begin{example} \label{ex:215} [Smeets, I. 2008: 231 (1)] \cite{RefB:21}\\
	\emph{\textbf{nge-me}-fu-n} 'I was there' \\
	\texttt{-IV.nge\_ser+TH.me20+IPD.fu8+IND1SG.n3}
\end{example}

\begin{example} \label{ex:216} [Smeets, I. 2008: 534 (\emph{mungel})] \cite{RefB:21}\\
	\emph{\textbf{nge-pa}-yaw-ki-y-m-i} 'You were hanging around here' \\
	\texttt{-IV.nge\_ser+HH.pa17+CIRC.iaw30+CF.ke14\\+IND.y4+2.m3+SG.i2}
\end{example}

The last possible context for \emph{nge-} says that it must be followed by a negation marker, slot 10. This rule may be read as an exception to the previous rules: \emph{nge-} must be followed by \emph{-l-}, \emph{-m-}, \emph{-tu-}, \emph{-pa-} or \emph{-me-}, except when it takes a negation suffix, making previous suffixes optional:

\begin{example} \label{ex:217} [Smeets, I. 2008: 407 (17)] \cite{RefB:21}\\
	\emph{\textbf{nge}-ke-\textbf{la}-fu-y} 'they were not' \\
	\texttt{-IV.nge\_ser+CF.ke14+NEG.la10+IPD.fu8\\+IND.y4+3.Ø3}
\end{example}

\begin{example} \label{ex:218} [Smeets, I. 2008: 194 (64)] \cite{RefB:21}\\
	\emph{\textbf{nge-nu}-n} 'there was not' \\
	\texttt{-IV.nge\_ser+NEG.no10+PVN.n4}
\end{example}

Finally, the prohibition concatenated rule for \emph{nge-} 'to be' prevents its co-occurrence with \texttt{+HH}, \texttt{+TH} and the negation for indicative \emph{-la-}.

\begin{exercise} \label{R:69} \textbf{Realization contexts for verb \emph{pepi-} 'to be able to'} \\
    \texttt{define RuVSC07 ["@SC09" =>\\ \_ ?* [["+CA"\{.ül34\}]|"+FAC"]];}
\end{exercise}

The root \emph{pepi-} 'to be able to' in a single root stem must always be followed by \emph{-l-} causative form or by factitive \emph{-ka-}:

\begin{example} \label{ex:219} [Smeets, I. 2008: 402 (45)] \cite{RefB:21}\\
	\emph{\textbf{pepi-ka}-w-ün} 'the setting of preparations' \\
	\texttt{-TV.pepi\_poder-hacer+FAC.ka33+REF.w31+PVN.n4}
\end{example}

\begin{example} \label{ex:220} [Smeets, I. 2008: 545 (\emph{pepi})] \cite{RefB:21}\\
	\emph{\textbf{pepi-l}-fal-la-y} 'it can not be done' \\
	\texttt{-TV.pepi\_poder-hacer+CA.l34+FORCE.fal25\\+NEG.la10+IND.y4+3.Ø3}
\end{example}

\begin{exercise} \label{R:44} \textbf{Realization contexts for verb \emph{ru-} 'to go through'} \\
    \texttt{define RuVSC08 ["@SC10" =>\\ \_ ?* ["+HH"|"+hh0"|"+TH"|"th0"]];}
\end{exercise}

Root \emph{ru-} 'to go through' does not occur without direction markers \emph{-me-} (thither slot 20) or \emph{-pa-} (hither slot 17), even in complex stems or compounds.

\begin{example} \label{ex:221} [Smeets, I. 2008: 247 (3)] \cite{RefB:21}\\
	\emph{amu-\textbf{ru-me}-y mawün-mew} 'he went through the rain' \\
	\texttt{-IV.amu\_ir-IV.ru\_pasar-CR.IV+TH.me20\\+IND.y4+3.Ø3\\-NN.mawün\_lluvia+INST.mew}
\end{example}

\begin{example} \label{ex:222} [Smeets, I. 2008: 515 (\emph{kata-})] \cite{RefB:21}\\
	\emph{kata-\textbf{ru}-l-\textbf{me}-y} 'it pierced through' \\
	\texttt{-TV.kata\_perforar-IV.ru\_pasar-CR.IV\\+CA.l34+TH.me20+IND.y4+3.Ø3}
\end{example}

\begin{example} \label{ex:223} [Smeets, I. 2008: 555 (\emph{ru-})] \cite{RefB:21}\\
	\emph{külü-\textbf{ru-pa}-n antü} 'after noon' \\
	\texttt{-IV.külü\_apoyar-IV.ru\_pasar-CR.IV\\+HH.pa17+PVN.n4}
\end{example}

\begin{example} \label{ex:224} [Smeets, I. 2008: 462 (63)] \cite{RefB:21}\\
	\emph{\textbf{ru}-l-\textbf{pa}-antü-le-y-iñ} 'we spent the day' \\
	\texttt{-IV.ru\_pasar+CA.l34+HH.pa17-NN.antü\_día-CR.IV\\+ST.le28+IND.y4+1.Ø3+PL.iñ2}
\end{example}

\paragraph{Deictic roots.} \label{tp:73} See section \ref{sec:19}, p. \pageref{sec:19}. Root \emph{fa-} is tagged \texttt{@FA} and root \emph{fe-} is tagged \texttt{@FE}, this allows us to apply rules that verbs formed from these roots require.

\begin{exercise} \label{R:45} \textbf{Realization contexts for deictic verbs} \\
    \texttt{define RuVSC09 ["@FA" => \_ ?* ["+CA"|"+ST"]]\\ .o. ["@FE" => \_ ?* ["+CA"|"+ST"],\\ \_ \$["+IND"] "+3"];}
\end{exercise}

R\ref{R:45} states that verbs containing \emph{fa-} 'to be like this' or \emph{fe-} 'to be like that' must also contain causative suffixes \mbox{\emph{-l-}}, \emph{-m-}, or stative suffix \emph{-le-}. However, verbs derived from \emph{fe-} do not obligatory fit this rule when they end in \emph{-y} which corresponds to indicative mood, 3\textsuperscript{rd} person (see examples E\ref{ex:67} to E\ref{ex:72}, p. \pageref{ex:67}).

\paragraph{Verbs with causative and factitive.} \label{tp:74} Finally, there is a number of verbs, which roots are tagged \texttt{@CF}, that "do not take the causative suffix \emph{-l-} \texttt{+CA} (slot 34) without simultaneously taking the factitive morpheme \emph{-ka-} \texttt{+FAC} (slot 33)" [Smeets, I. 2008: 301] \cite{RefB:21}:

\begin{exercise} \label{R:46} \textbf{Realization contexts for deictic verbs occurring with causative} \\
    \texttt{define RuVSC10 [["@CF" ?*[["+CA"|"+ca0"]{.l34}]]\\ => \_ ?* ["+FAC"|"+fac0"]];}
\end{exercise}

R\ref{R:46} do not force roots marked \texttt{@CF} to be followed by causative suffix, instead, it states that the sequence "\mbox{\texttt{@CF}\emph{-l-}}" must co-occur with factitive \emph{-ka-}. All of this may be read as "if a \texttt{@CF} root is followed by \texttt{+CA} it must also follow \texttt{+FAC}. This rule also covers complex stems an complex compounds by means of "neutralized" tags (see \nameref{tp:53}, p. \pageref{tp:53}):

\begin{example} \label{ex:225} [Smeets, I. 2008: 66 (42)] \cite{RefB:21}\\
	\emph{\textbf{küme}-y} 'it is good' \\
	\texttt{-AJ.küme\_bueno+VRB.Ø36+IND.y4+3.Ø3}
\end{example}

\begin{example} \label{ex:226} [Smeets, I. 2008: 255 (3)] \cite{RefB:21}\\
	\emph{\textbf{küme-l-ka}-le-tu-n} 'I am well' \\
	\texttt{-AJ.küme\_bueno+VRB.Ø36+CA.l34+FAC.ka33\\+ST.le28+RE.tu16+IND1SG.n3}
\end{example}

\begin{example} \label{ex:227} [Smeets, I. 2008: 349 (17)] \cite{RefB:21}\\
	\emph{\textbf{llüka}-le-n} 'I am afraid' \\
	\texttt{-IV.llüka\_temer+ST.le28+IND1SG.n3}
\end{example}

\begin{example} \label{ex:228} [Smeets, I. 2008: 375 (25)] \cite{RefB:21}\\
	\emph{\textbf{llüka-l-ka}-che-ke-y} 'he frightens people' \\
	\texttt{-IV.llüka\_asustar+CA.l34+FAC.ka33\\-NN.che\_persona-CR.IV+CF.ke14+IND.y4+3.Ø3}
\end{example}

\begin{example} \label{ex:229} [Smeets, I. 2008: 572 (\emph{welu\textsuperscript{2}})] \cite{RefB:21}\\
	\emph{ti lifru \textbf{welu}-y} 'the book was exchanged' \\
	\texttt{-AP.ti\_el -NN.lifru\_libro\\-IV.welu\_intercambiar+IND.y4+3.Ø3}
\end{example}

\begin{example} \label{ex:230} [Smeets, I. 2008: 572 (\emph{welu\textsuperscript{2}})] \cite{RefB:21}\\
	\emph{\textbf{welu-l-ka}-ñma-fi-ñ} 'I
exchanged it' \\
	\texttt{-IV.welu\_intercambiar+CA.l34+FAC.ka33\\+IO.ñma26+EDO.fi6+IND1SG.n3}
\end{example}

\section{Beyond "A grammar of \emph{Mapuche}"} \label{sec:53}

\paragraph{} \label{tp:75} "A grammar of \emph{Mapuche}" [Smeets, I. 2008] \cite{RefB:21}, our development base, describes the central \emph{Mapuche} dialect. We have added into the analyser words that are not in Smeets' work. Compounds that she have not found throughout her study but some other authors mention. Also some minor dialectal variations.

\subsection{The spelling unifier} \label{sec:54}

\paragraph{} \label{tp:76} There is a significant variation in \emph{Mapudüngun} spelling, mainly due to the existence of different spelling proposals, together with the strong influence of Spanish orthography. Some texts may present a mixture of these orthographic proposals along with Castilianized orthography. This is something to sort out before analysing a text, either with rule based analysers or statistical ones, because the divergence in input means more rules for the first ones and poor results for the second ones.

The task of the unifier is to replace characters; from a series of possible inputs, a single output is returned. This process is called "unification". Initially, the idea was to change different graphemic proposals for the \emph{Mapuche} language into one single spelling. But the strong influence of Spanish orthography in written \emph{Mapudüngun} was noticed along the way. Therefore, the final implementation reflects mainly this fact, including anyway a couple of rules related to some of the graphemic proposals. This process is embedded in the analyser.

\emph{Mapudüngun} vowels should not have accentuation marks, if there is any, it is transformed into its non accentuated version:

\begin{quote} \label{note:06}
	{\small á → \emph{a}, é → \emph{e}, etc.}
\end{quote}

The morphological analysis is performed on words in lowercase letters. The unifier section makes the main FST interpret every uppercase letter as a lowercase one, which does not mean it gives a lowercase output, as it is demonstrated below:

\begin{quote} \label{note:07}
	{\small \emph{Kasinta} → ‑PN.Kasinta\_Jacinta}
\end{quote}

Other changes are:
\begin{quote} \label{note:08}
	{\small b → \emph{f}\\
	ca → \emph{ka}, co → \emph{ko}, cu → \emph{ku}\\
	ce → \emph{se}, ci → \emph{si}\\
	gue → \emph{ge}, gui → \emph{gi}\\
	hua → \emph{wa}, hue → \emph{we}, hui → \emph{wi}, huo → \emph{wo}, huu → \emph{wu}, when \emph{h} is not preceded by \emph{c}\\
	ha → \emph{a}, he → \emph{e}, hi → \emph{i}, ho → \emph{o}, hu → \emph{u} when \emph{h} is not preceded by \emph{c} (\emph{che})\\
	j → \emph{k}, qu → \emph{k}\\
	v → \emph{ü} when v is between consonants or semi-vowels, v → \emph{f} otherwise\\
	q → \emph{g} when q is not followed by u\\
	tx → \emph{tr}, x → \emph{tr}\\
	z → \emph{d}}
\end{quote}

From all these changes, only the last three and the first part of the fourth backwards (v into \emph{ü}) do not have to do with Spanish but with some graphemic proposal for \emph{Mapudüngun}.

The \emph{Mapuche} alphabet we encode has twenty-five graphemes, five of which are digraphs. There are six vowels, three semi-vowels and seventeen consonants:

\begin{quote} \label{note:09}
	{\small \emph{a, ch, d, e, f, g, i, k, l, ll, m, n, ng, ñ, o, p, r, s, sh, t, tr, u, ü, w, y}}
\end{quote}

There are additional graphemes that we accept for analysis, which belong to other graphemic proposals for \emph{Mapudüngun} or to Spanish:

\begin{quote} \label{note:10}
	{\small b, c, h, j, l', n', q, t', tx, x, v, z}
\end{quote}

\subsection{Lexicon} \label{sec:55}

\paragraph{} \label{tp:77} We have augmented our lexicon mostly from Augusta's dictionaries \cite{RefB:03}. But not only new words have been introduced, also many variants to the already collected words from Smeets, do not forget that \emph{Mapudüngun} spelling is not fixed yet. But what most variants generates are the differences in pronunciation of some sounds, for instance, final \emph{-n} is usually interchanged by final \emph{-ñ}; \emph{ü} in any positions is commonly interchanged with \emph{u, i} and sometimes \emph{e}, or the other way around; \emph{tr} with \emph{ch}, \emph{d} with \emph{s}, etc., see table \ref{tab:09}\footnote{Spaces with a dash in Smeets' column of table \ref{tab:09} mean the introduction of a new term not found in her work.} for some examples:

\begin{table}[htb]
	\caption{Spelling variants in lexicon}
	\label{tab:09}
	\begin{tabular}{|c|c|c|c|}
		\hline\noalign{\smallskip}
		PoS & Smeets & Variant & Meaning \\
		\noalign{\smallskip}\hline\noalign{\smallskip}
		\texttt{-AJ} & \emph{arke\textbf{n}} & \emph{arke\textbf{ñ}} & evaporated\\
		\noalign{\smallskip}\hline\noalign{\smallskip}
		\texttt{-AJ} & \emph{kol\textbf{ü}$\sim$koll\textbf{ü}} & \emph{kol\textbf{i}$\sim$koll\textbf{i}} & \makecell{brown, reddish brown,\\beige}\\
		\noalign{\smallskip}\hline\noalign{\smallskip}
		\texttt{-AJ} & - & \emph{liuke} & clean, clear, pure (water)\\
		\noalign{\smallskip}\hline\noalign{\smallskip}
		\texttt{-AV} & \emph{ki\textbf{s}u$\sim$ki\textbf{sh}u} & \emph{ki\textbf{d}u} & alone, self, own\\
		\noalign{\smallskip}\hline\noalign{\smallskip}
		\texttt{-AV} & - & \emph{kashill} & near\\
		\noalign{\smallskip}\hline\noalign{\smallskip}
		\texttt{-AV} & \emph{küt\textbf{u}} & \emph{küt\textbf{o}} & even, also\\
		\noalign{\smallskip}\hline\noalign{\smallskip}
		\texttt{-IV} & - & \emph{yawa} & make noise\\
		\noalign{\smallskip}\hline\noalign{\smallskip}
		\texttt{-IV} & \emph{w\textbf{i}tra} & \emph{w\textbf{ü}tra} & get up\\
		\noalign{\smallskip}\hline\noalign{\smallskip}
		\texttt{-IV} & \emph{cheko\textbf{d}$\sim$lliko\textbf{sh}} & \emph{lliko\textbf{d}} & \makecell{to squat,\\to sit down on one's heels}\\
		\noalign{\smallskip}\hline\noalign{\smallskip}
		\texttt{-NN} & \emph{achaw\textbf{all}} & \makecell{\emph{achaw}$\sim$\\ \emph{achaw\textbf{üll}}} & chicken\\
		\noalign{\smallskip}\hline\noalign{\smallskip}		
		\texttt{-NN} & \emph{chaf\textbf{o}} & \emph{chaf\textbf{a}} & cough, catarrh, cold \\
		\noalign{\smallskip}\hline\noalign{\smallskip}
		\texttt{-NN} & - & \makecell{\emph{da\textbf{g}llu}$\sim$\\ \emph{da\textbf{w}llu}} & river shrimp\\
		\noalign{\smallskip}\hline\noalign{\smallskip}
		\texttt{-TV} & \emph{in\textbf{g}ka} & \emph{inka} & to defend \\
		\noalign{\smallskip}\hline\noalign{\smallskip}
		\texttt{-TV} & - & \makecell{\emph{kedi\textbf{n}}$\sim$\\ \emph{kedi\textbf{ñ}}} & to shear \\
		\noalign{\smallskip}\hline\noalign{\smallskip}
		\texttt{-TV} & \emph{\textbf{ü}trüf} & \emph{\textbf{i}trüf} & to throw \\
		\noalign{\smallskip}\hline
	\end{tabular}
\end{table}

\subsection{\emph{Williche} verb forms} \label{sec:56}

\paragraph{} \label{tp:78} "In \emph{Williche}\footnote{\texttt{-NN.willi\_sur-NN.che\_persona} 'Southern people'}, the southernmost dialect of \emph{Mapudüngun}, transitive verbs expressing the 1 → 2 relationship (with a total number of participants greater than two) is indicated by the combination of \emph{-e-} and a 2\textsuperscript{nd} person subject marker in slot 3 (Mösbach 1962: 80, and Augusta 1903: 84–86 (cited by Salas 1979a: 307)), e.g. \emph{pe-e-y-m-i} 'I saw you (sg)', \emph{pe-e-y-m-u} 'I saw you (dl)'; \emph{pe-e-y-m-ün} 'I saw you (pl)'" [Smeets, I. 2008: 160] \cite{RefB:21}. This very same relations are included in a single form of central \emph{Mapudüngun} when participants are more than two \texttt{+1A.w23+IND.y4+1.Ø3+PL.iñ2}, which may be disambiguated by means of personal or possessive pronouns. But Smeets also gives as example \emph{pe-e-y-m-i} 'I saw you (sg)' which are actually two participants, central \emph{Mapudüngun} and \emph{Williche} also differ in this form. Note that \emph{Williche} forms are ended by the null morpheme of dative subject \texttt{+DS12A} demanded by the \texttt{+IDO} marker \emph{-e-}, see the following examples to compare:

\begin{example} \label{ex:231}\ central \emph{Mapudungün} 1s → 2s [Smeets, I. 2008: 157 (20)] \cite{RefB:21}\\
	\emph{pe-\textbf{e}-y-\textbf{u}} 'I see you (sg)' \\
	\texttt{-TV.pe\_ver+IDO.e6+IND.y4+1.Ø3+DL.u2+DS12A.Ø1}
\end{example}

\begin{example} \label{ex:232}\ \emph{Williche} 1s → 2s [Smeets, I. 2008: 160] \cite{RefB:21}\\
	\emph{pe-\textbf{e}-y-\textbf{m-i}} 'I see you (sg)' \\
	\texttt{-TV.pe\_ver+IDO.e6+IND.y4+2.m3+SG.i2+DS12A.Ø1}
\end{example}

\begin{example} \label{ex:233}\ central \emph{Mapudungün} 1 → 2 (more than two participants) [Smeets, I. 2008: 572 ()] \cite{RefB:21}\\
	\emph{kellu-\textbf{w}-y-\textbf{iñ}} 'I helped you (d/p), we (d/p) helped you (s/d/p)' \\
	\texttt{-NN.kellu\_ayuda+VRB.Ø36\\+1A.w23+IND.y4+1.Ø3+PL.iñ2}
\end{example}

\begin{example} \label{ex:234}\ \emph{Williche} 1s → 2d [Smeets, I. 2008: 160] \cite{RefB:21}\\
	\emph{pe-\textbf{e}-y-\textbf{m-u}} 'I see you (dl)' \\
	\texttt{-TV.pe\_ver+IDO.e6+IND.y4+2.m3+DL.u2+DS12A.Ø1}
\end{example}

\begin{example} \label{ex:235}\ \emph{Williche} 1s → 2p [Smeets, I. 2008: 160] \cite{RefB:21}\\
	\emph{pe-\textbf{e}-y-\textbf{m-ün}} 'I see you (pl)' \\
	\texttt{-TV.pe\_ver+IDO.e6+IND.y4+2.m3+PL.ün2+DS12A.Ø1}
\end{example}

Rules generated for central \emph{Mapudüngun} correctly analyse the \emph{Williche} form \emph{pe-e-y-m-u} 'I saw you (dl)'. For the other two forms we had to change two rules in the system, compare:

\begin{exercise} \label{R:47} \textbf{Central \emph{Mapudüngun} imperative/plural form} \\
    \texttt{define RuPr06 $\sim$\$[["+IMP1SG"|"+PL"] ?* \newline [["+DS3A"\{.ew1\}]|"+DS12A"]]}
\end{exercise}

\begin{exercise} \label{R:48} \textbf{\emph{Williche} imperative/plural form} \\
    \texttt{define RuPr06 $\sim$\$[["+IMP1SG"|"+PL"] ?* \newline ["+DS3A"{.ew1}]] .o. $\sim$\$["+IMP1SG" ?* "+DS12A"];}
\end{exercise}

In central \emph{Mapudüngun}, the plural \texttt{+PL} \emph{-ün-} can not be followed by the dative subject for 1\textsuperscript{st} or 2\textsuperscript{nd} person agent \emph{-Ø-}. On the contrary, it is necessary in the \emph{Williche} dialect, and correctly analysed as shown in E\ref{ex:235}.

\begin{exercise} \label{R:49} \textbf{Central \emph{Mapudüngun} dative subject occurrence} \\
\texttt{define RuPr12 $\sim$\$[["+EDO"|"+PVN"|"+TVN"\\|["+SG"\{.i2\}]] ?* ["+DS3A"|"+DS12A"]]}
\end{exercise}

\begin{exercise} \label{R:50} \textbf{\emph{Williche} dative subject occurrence} \\
    \texttt{define RuPr12 $\sim$\$[["+EDO"|"+PVN"|"+TVN"\\|["+SG"\{.i2\}]] ?* "+DS3A"] .o. \\ $\sim$\$[["+EDO"|"+PVN"|"+TVN"] ?* "+DS12A"]}
\end{exercise}

As in the previous case, in central \emph{Mapudüngun}, the singular \texttt{+SG} \emph{-i-} can not be followed by the dative subject for 1\textsuperscript{st} or 2\textsuperscript{nd} person agent \emph{-Ø}. Which is also necessary in the \emph{Williche} dialect, and correctly analysed as shown in E\ref{ex:232}.

\subsection{Following Zúñiga} \label{sec:57}

\paragraph{} \label{tp:79} Even though Zúñiga seems to base "\emph{Mapudüngun}. El habla \emph{mapuche}" [Zúñiga, F. 2006] \cite{RefB:24} in central \emph{Mapudüngun}, texts included in his work present some variations respect Smeets. We have included these divergences in the analyser.

\subsubsection{Different indicative form} \label{sec:58}

\paragraph{} \label{tp:80} "The mark for indicative mood is \emph{-i-}. It appears as a vowel if the root ends in a consonant, as a semi-vowel \emph{-y-} if the root ends in a vowel other than \emph{i-}, and it does not appear if the root ends in \emph{i-}"\footnote{Translation is ours, the original is in Spanish} [Zúñiga, F. 2006: 105] \cite{RefB:24}. Instead of the "root ending", it should be said "the preceding sound", because this one may actually belong to the root, but also to a previous suffix. We have found that it realizes as \emph{-i-} after semi-vowel too (E\ref{ex:237}); and is a null suffix when found either preceded or followed by \emph{i} (E\ref{ex:239}). In Smeets work, the indicative is either \emph{-y-} after vowel or semi-vowel, or \emph{-üy-} after consonant; there is a \emph{-iy-} variant for the latter.

To treat the variant presented by Zúñiga, we have added an \texttt{@IZ} tag to the already existent encoding of \texttt{+IND} suffix, D\ref{def:30}. And added a new rule to deal with contexts of conversion of the intermediate representation \texttt{@IZ} (R\ref{R:51}):

\begin{definition} \label{def:30} \texttt{["+IND"\{.y4\}]: [[["@ÜI"|"@Ü"]y]|"@IZ"]} \end{definition}

\begin{exercise} \label{R:51} \textbf{Zúñiga's indicative form} \\
    \texttt{define RuIndZu ["@IZ" -> 0 || i \_ , \_ i]\\ .o. ["@IZ" -> i];}
\end{exercise}

D\ref{def:30} encodes indicative mood suffix, we have introduced the \texttt{@IZ} tag to treat Zúñiga's variant. Preceding this new tag is the suffix as Smeets presents it (E\ref{ex:238}). \texttt{@Ü} is for \mbox{"\texttt{CON}-\emph{üy-}"}, "\texttt{VOW|SVW}-\emph{y-}"; \texttt{@ÜI} is for "\texttt{CON}-\emph{iy-}".

R\ref{R:51} defines, by context, the form \texttt{@IZ} should take, transforming it always into \emph{i}, except when occurs either before (E\ref{ex:239}) or after (E\ref{ex:240}) \emph{i}, in which case it is transformed into \emph{0}, acting as a null suffix. Otherwise, the tag remains until the end of the process, when it is cleared out (E\ref{ex:236}, E\ref{ex:237} and E\ref{ex:238}).

\begin{example} \label{ex:236} [Zúñiga, F. 2006: 105 (Cuadro III-3a / 2\textsuperscript{a})] \cite{RefB:24}\\
	\emph{kon-\textbf{i}-m-i} 'you enter' \\
	\texttt{-IV.kon\_entrar+IND.y4+2.m3+SG.i2}
\end{example}

\begin{example} \label{ex:237} [Zúñiga, F. 2006: 283 (\emph{pewma})] \cite{RefB:24}\\
	\emph{chum-yaw-\textbf{i-m}-i} 'what are you doing around?' \\
	\texttt{-QC.chum\_cómo+VRB.Ø36+CIRC.iaw30\\+IND.y4+2.m3+SG.i2}
\end{example}

\begin{example} \label{ex:238} [Zúñiga, F. 2006: 105 (Cuadro III-3b / 2\textsuperscript{a})] \cite{RefB:24}\\
	\emph{tripa-\textbf{y}-m-i} 'you leave' \\
	\texttt{-IV.tripa\_salir+IND.y4+2.m3+SG.i2}
\end{example}

\begin{example} \label{ex:239} [Zúñiga, F. 2006: 227 (84)] \cite{RefB:24}\\
	\emph{amu-a-iñ} 'let us go' \\
	\texttt{-IV.amu\_ir+NRLD.a9+IND.Ø4+1.Ø3+PL.iñ2}
\end{example}

\begin{example} \label{ex:240} [Zúñiga, F. 2006: 105 (Cuadro III-3c / 2\textsuperscript{a})] \cite{RefB:24}\\
	\emph{pi-m-i} 'you say' \\
	\texttt{-TV.pi\_decir+IND.Ø4+2.m3+SG.i2}
\end{example}

\subsubsection{Glottal stop before \texttt{+IDO}} \label{sec:59}

\paragraph{} \label{tp:81} When the internal direct object \emph{-e-} is preceded by \emph{a-}, there is an optional epenthesis of a glottal stop in between, reflected as \emph{-g-} in spelling. As there is already a rule that treat glottal stop epenthesis in compounds where the second root starts in vowel. We just added the appropriate tag \texttt{@G} to the \texttt{+IDO} suffix, D\ref{def:31} (see E\ref{ex:116}, E\ref{ex:117}, E\ref{ex:118}, R\ref{R:01} and R\ref{R:02}, p. \pageref{R:01}):

\begin{definition} \label{def:31} Encoding of the \texttt{+IDO} suffix\\ 
	\texttt{["+IDO"\{.e6\}]: ["@G""@ID"]}
\end{definition}

\begin{example} \label{ex:241} [Zúñiga, F. 2006: 274 (56)] \cite{RefB:24}\\
	\emph{kulli-\textbf{a-g-e}-y-u} 'I will pay you both' \\
	\texttt{-TV.kulli\_pagar+NRLD.a9+IDO.e6\\+IND.y4+1.Ø3+DL.u2+DS12A.Ø1}
\end{example}

\begin{example} \label{ex:242} [Zúñiga, F. 2006: 279 (109)] \cite{RefB:24}\\
	\emph{elu-tu\textbf{a-g-e}-n} 'give it back to me' \\
	\texttt{-TV.elu\_dar+RE.tu16+NRLD.a9+IDO.e6\\+IND1SG.n3+DS12A.Ø1}
\end{example}

\begin{example} \label{ex:243} [Zúñiga, F. 2006: 130 (note 10)] \cite{RefB:24}\\
	\emph{elu-\textbf{la-g-e}-n} 'you did not give me' \\
	\texttt{-TV.elu\_dar+NEG.la10+IDO.e6+IND1SG.n3+DS12A.Ø1}
\end{example}

\subsubsection{Nominal compounds} \label{sec:60}

\paragraph{} \label{tp:82} We have only explained the \emph{Mapuche} verb throughout this article. We have not introduced the nominal forms as such, but as one of the possible verbal stems. Nominal forms are much more simpler than the verbal ones. We do not want to make this article too extensive adding encoding details that are well explained with the verb form, it suffices to say that there is nominal compounding in \emph{Mapudüngun}.

Zúñiga defines \emph{püle} 'by, towards' as a post-position [Zú-ñiga, F. 2006: 195] \cite{RefB:24}. Smeets defines it as a post-position too, 'side'; but also as a noun [Smeets, I. 2008: 69 (10.4)] \cite{RefB:24}, which is how we have incorporated \emph{püle} 'side, direction' into our lexicon.

Even classifying \emph{püle} as a post-position, in many occasions this word forms compounds in Zúñiga's texts, which are not recognized by the Smeets derived rules of the analyser.

\begin{exercise} \label{R:52} \textbf{Nominal compound (\emph{püle})} \\ 
	\texttt{define formCXNN\newline [[\%<[DEMPR|IVROOT|INTPR]\%\# \%<NROOT\%\#]"-CNN"];\\
	define preCXNN\newline [\_eq(formCXNN, \%< , \%\#)];\\
	define RuleCXNN1\newline ["-NN" => ["@TYA"|"@FA"|"@T"] ?* \_ ];\\
	define CXNN\newline [CLEANu .o. RuleCXNN1 .o. \\ formCXNN - RuleCXNN1 .o. CLEANNVFd];}
\end{exercise}

The mechanism for compounding was already explained at \nameref{tp:52}, p. \pageref{tp:52}, so here we add that in R\ref{R:52} the form of these nominal compounds is defined as having a first member that may be a demonstrative pronoun, an intransitive verb or an interrogation pronoun. \texttt{RuleCXNN1} specifies which forms, out of these categories, are actually accepted to form the compound with the noun as a second member. The specific forms has been tagged to filter them out, \texttt{@TYA, @FA}\footnote{The verb root tagged \texttt{@FA} is also identified as a deictic verb with the same tag (see \nameref{tp:73}, p. \pageref{tp:73}).} and \texttt{@T} respectively; actually, only one member of each category is tagged. In the examples we show compounds that are not recognized following Smeets, compounds that take \emph{püle} as a second member (In Smeets, \emph{püle} may be the second noun in a nominal compound):

\begin{example} \label{ex:244} [Zúñiga, F. 2006: 275 (65)] \cite{RefB:24}\\
	\emph{fa-püle} 'around here' \\
	\texttt{-IV.fa\_ser-esto-NN.püle\_lugar}
\end{example}

\begin{example} \label{ex:245} [Zúñiga, F. 2006: 191 (Cuadro III-17)] \cite{RefB:24}\\
	\emph{kañ}\footnote{This is an epenthetic \emph{ñ}.}\emph{-püle} 'somewhere else' \\
	\texttt{-AJ.ka\_otro-NN.püle\_lugar}\footnote{There is a colloquial expression in Chile, 'salta pal lao' which means something like 'I don't believe you', 'are you kidding', 'you better not...', depending on the situation. \emph{Payllafilu} says that this expression is equivalent to \emph{kañpüle} in \emph{Mapudüngun}.}
\end{example}

\begin{example} \label{ex:246} [Zúñiga, F. 2006: 273 (49)] \cite{RefB:24}\\
	\emph{tie-püle} 'over there' \\
	\texttt{-DP.tüye\_aquel-de-allá-NN.püle\_lugar}
\end{example}

\begin{example} \label{ex:247} [Zúñiga, F. 2006: 182 (95.d)] \cite{RefB:24}\\
	\emph{tuchi-püle} 'wherever' \\
	\texttt{-IP.chuchi\_qué\_cuál-NN.püle\_lugar}
\end{example}

\begin{example} \label{ex:248} [Zúñiga, F. 2006: 275 (67)] \cite{RefB:24}\\
	\emph{üye-püle} 'over there' \\
	\texttt{-AV.üye\_allí\_allá-NN.püle\_lugar}
\end{example}

\paragraph{Augusta's nominal compound.} \label{tp:83} Another nominal compound integrated into the analyser was found in Augusta, F. \cite{RefB:03}. It has a numeral as first member and a noun as second member:

\begin{example} \label{ex:249} [Augusta, F. (\emph{epuange})] \cite{RefB:24}\\
	\emph{epu-ange} 'two faces\footnote{Two faces, (be of) two faces, a designation of a \emph{wekufü} which owns the sea or lake and is also called \emph{Millalongko} 'golden head' or \emph{Kawekufü} 'water daemon'. \emph{Kutranelenew Epuange} '\emph{Epuange} has made me sick'. Epithet that in some places \emph{Mapuche} give as first name to a god, e.g. \emph{Epuange ngünechen} 'two faces father regulator', either because they represent two sexes, or because with this expression they allude to the benign and serene heaven and to the unfavourable heaven; or to the severity and benignity that the supreme being can demonstrate to men. In addition, the idea is applied to both God and Mayorwekufu 'major daemon'. V. Augusta (1910, p. 227) [Augusta, F. \emph{epuange}] \cite{RefB:03}}' \\
	\texttt{-NU.epu\_dos-NN.ange\_cara}
\end{example}

\begin{example} \label{ex:250} [Zúñiga, F. 2006: 319 (\emph{kiñepüle})] \cite{RefB:24}\\
	\emph{kiñe-püle} 'by/towards this side/place' \\
	\texttt{-NU.kiñe\_uno-NN.püle\_lugar}
\end{example}

\subsubsection{Instrumental and ad-position \emph{mew}} \label{sec:61}

\paragraph{} \label{tp:84} Smeets classifies \emph{-mew} as an instrumental suffix that follows nouns, deverbal nouns and pronouns (see Smeets, I. 2008: 61 - 67, "10.1 The instrumental \emph{-mew} $\sim$ \emph{-mu}" \cite{RefB:21}).

Zúñiga defines \emph{mew} as an ad-position that is realized separated from the noun or deverbal noun that follows, but together with pronouns or adverbs (see Zúñiga, F. 2006: 194 - 197. "4.1 Las adposiciones y los sustantivos relacionales" \cite{RefB:24}).

In order to recognize \emph{mew} as an independent form, but still as the instrumental suffix, it was declared as such in the non-verbal section of the script:
\begin{example} \label{ex:251} [Zúñiga, F. 2006: 195 (105.a)] \cite{RefB:24}\\
	\emph{müle-ka-n ruka \textbf{mew}} 'I am still at home' \\
	\texttt{-IV.müle\_estar+CONT.ka16+IND1SG.n3\\ -NN.ruka\_casa +INST.mew}
\end{example}

\begin{example} \label{ex:252} [Zúñiga, F. 2006: 195 (105.b)] \cite{RefB:24}\\
	\emph{amu-tu-n waria \textbf{mew}} 'I went back to the city' \\
	\texttt{-IV.amu\_ir+RE.tu16+IND1SG.n3\\ -NN.wariya\_ciudad +INST.mew}
\end{example}

\begin{example} \label{ex:253} [Zúñiga, F. 2006: 195 (105.c)] \cite{RefB:24}\\
	\emph{waria \textbf{mew} küpa-n} 'I came from the city' \\
	\texttt{-NN.wariya\_ciudad +INST.mew\\ -IV.küpa\_venir+IND1SG.n3}
\end{example}

\begin{example} \label{ex:254} [Zúñiga, F. 2006: 201 (2.b)] \cite{RefB:24}\\
	\emph{fey-\textbf{mew} kintu‐ke‐y‐ng‐ün meli mamüll} 'then they looked for four trees' \\
	\texttt{-AV.fey\_entonces+INST.mew\\ -TV.kintu\_buscar+CF.ke14+IND.y4+3.ng3+PL.ün2\\ -NU.meli\_cuatro -NN.mamüll\_árbol}
\end{example}

\subsection{\emph{Wüño} as auxiliary} \label{sec:62}

\paragraph{} \label{tp:85} According to Smeets, in \emph{Mapudüngun} there are five auxiliary verbs. These are elements separated from the main verb. They are verbal stems without inflection, which immediately precede the main verb, without any other element in between [Smeets, I. 2008: 175 (25.4)] \cite{RefB:21}. These are:

\begin{itemize} \label{it:38}
	\item[] \emph{kalli} 'enabling'
	\item[] \emph{kim} 'knowing how to'
	\item[] \emph{küpa} 'wishing'
	\item[] \emph{pepi} 'being able'
	\item[] \emph{shinge} 'moving up/along'.
\end{itemize}

Lonkon calls these elements, "modal prefixes". She identifies four of them: \emph{kalli, kim-, küpa- and pepi-}, giving them the same values as Smeets. But Lonkon considers them prefixes, therefore, attached to the verb they are moulding, except for the permissive/enabling \emph{kalli} [Lonkon, E. 2011: 249] \cite{RefB:12}.

Zúñiga identifies two of these modal elements: \emph{kim-} and \emph{pepi-}. He says that they form part of "complex verb stems", which means that both verb roots, the modal and the moulded one, form a compound. However, he says, they can also be expressed as separated elements, then he calls them "pre-verbal particles" [Zúñiga, F. 2006: 136] \cite{RefB:24}.

Zúñiga also displays a list of verb roots that form complex verb stems, and, in fact, he devotes a paragraph to explain that they may be formed by radical concatenation, and by nominal incorporation. In the list is found the verb root \emph{wüño-} 're-', 'return, come back'. In the list are also \emph{kalli-} and \emph{küpa-}, but not \emph{shinge-} [Zúñiga, F. 2006: 179] \cite{RefB:24}.

Zúñiga explains that these forms are frequently found as pre-verbal particles, i.e., separated from the main verb, which is reflected in spelling. He treats them as radical concatenation, though.

Salas calls them "modals", and he identifies \emph{kim-} 'know', \emph{küpa-} 'wish' and \emph{pepi-} 'be able'. He says that they function as prefixes of single and complex stems [Salas, A. 2006: 192] \cite{RefB:19}.

Augusta defines \emph{wüño-} as a suffix equivalent to the prefix morpheme 're-', which mainly expresses the idea of redoing the action indicated by the following verb. We have realized that \emph{wüño} also expresses the idea of retrospective and/or backwards action [Augusta, F. \emph{wüño}] \cite{RefB:03}.

Among the examples given by Augusta, there are some that show \emph{wüño} separated from the main verb, and others, forming a compound with the moulded verb.

Many \emph{Mapudüngun} native speakers perceive \emph{wüño} separated from the verb it is moulding, and they reflect it as Zúñiga shows in its examples.

An affix does not realize isolated in \emph{Mapudüngun}, it must be attached to a verb or another root (adjective, noun, etc.). It can neither work as a root accepting suffixes to be attached. \emph{wüño}, on the other hand, also works as a verb root that may be inflected by attaching suffixes to it. Therefore, \emph{wüño} meets the auxiliary definition given by Smeets.

\emph{wüño} $\sim$ \emph{wiño} either forms a compound, originates an inflected verb, or is a separated element. If some native speakers perceive it as a separate element, it may indicate that it is a modal element, thus, following  Smeets, this verb would fulfil the "auxiliary" function as she defines it. Or the "pre-verbal particle" function, as Zúñiga calls it.

Anyway, not all \emph{Mapudüngun} native speakers spell \emph{wüño} $\sim$ \emph{wiño} separated from the main verb. Many of them use it in the verb as a compound, i.e., as a "modal prefix", as Lonkon calls it.

We have added \emph{wüño} as auxiliary into the lexicon list through the following entry:

\begin{definition} \label{def:32}\ \\
		\texttt{["-XV"\{.wüño\_re-\_volver-a\}]: [\{wüño\}|\{wiño\}]}
\end{definition}

\begin{example} \label{ex:255} [Zúñiga, F. 2006: 148 (63)] \cite{RefB:24}\\
	\emph{\textbf{wüño}-witra-me-tu-a-fiel } 'to go there to recover them' \\
	\texttt{-IV.wüño\_volver-IV.wütra\_levantar-CR.IV\\+TH.me20+RE.tu16+NRLD.a9+TVN.fiel4}
\end{example}

\begin{example} \label{ex:256} [Augusta, F. \emph{contestar}] \cite{RefB:03}\\
	\emph{\textbf{wüño} fey-pin} 'to answer' \\
	\texttt{-XV.wüño\_re-\_volver-a -TV.feypi\_decir+PVN.n4}
\end{example}

\begin{example} \label{ex:257} [Lonkon, E. 2017] \cite{RefB:13}\\
	\emph{\textbf{wiño} wütra-m-püra-m-nge-tu-a-fu-y} 'it revitalized' \\
	\texttt{-XV.wüño\_re-\_volver-a\\ -IV.wütra\_levantar+CA.m34-IV.püra\_subir-CR.IV\\+CA.m34+PASS.nge23+RE.tu16+NRLD.a9\\+IPD.fu8+IND.y4+3.Ø3}
\end{example}

\subsection{Proposing \emph{-ñma} as adverbializer} \label{sec:63}

\paragraph{} \label{tp:86} Smeets lists \emph{-ñma} as an unproductive suffix [Smeets, I. 2008: 116] \cite{RefB:21}, giving a series of examples with this suffix:

\begin{enumerate} \label{it:39}
	\item \emph{fücha-ñma} 'very long' (\emph{fücha} 'long')
	\item \emph{we-ñma} 'very new' (\emph{we} 'new')
	\item \emph{wesha-ñma} 'very bad' (\emph{wesha} 'bad')
	\item \emph{rume-ñma} 'extremely' (\emph{rume} 'very')
	\item \emph{welu-ñma} 'wrong, reversely' (\emph{welu} 'but, wrong, reversely')
	\item \emph{alü-ñma} 'for a long time' (\emph{alü} 'much'), cf. \emph{alü-ñma-mew} 'much later, a long
time after that'
	\item \emph{fentre-ñma $\sim$ fentre-yma} 'for a long time' (\emph{fentre} 'much')
	\item \emph{epu-ñma} 'with the two of us' (\emph{epu} 'two')
	\item \emph{ka-ruka-ñma} 'neighbour' (\emph{ka-ruka} 'neighbour')
\end{enumerate}

Except for the examples 5, 8 and 9, it is quite evident the part of the meaning contributed by \emph{-ñma}, 'very'. In other texts we have found:

\begin{enumerate} \label{it:40}
	\item \emph{weda-ñma} 'evil, too bad' (\emph{weda} 'bad') [Zúñiga, F. 2006: 270 (15)] \cite{RefB:24}
	\item \emph{pichi-ñma} 'just now, recently' (\emph{pichi} 'little') [Zúñiga, F. 2006: 271 (30)] \cite{RefB:24}
	\item \emph{fücha-ñma} 'very big' (\emph{fücha} 'big, old') [Augusta, F.\\ (\emph{füchañma})] \cite{RefB:03}
	\item \emph{llekü-ñma} 'very close' (\emph{llekü} 'close, near') [Augusta, F. (\emph{llekü-ñma})] \cite{RefB:03}
\end{enumerate}

We have not gone through an exhaustive research of this suffix, that is why it is just a proposal. We think it is quite clear that \emph{-ñma} adds the 'very' part of the meaning, but we have observed that it is applied only to adjectives and certain adverbs with this sense, and only as part of the nominal compounding. There is only one case, mentioned by Smeets, where it seems to be part of a verbal stem, but there are no examples of it; so in the cases of E\ref{ex:258} it could be either the indirect object marker, slot 26, or the experience suffix, slot 35, which share the form \emph{-ñma}:

\begin{example} \label{ex:258} [Smeets, I. 2008: 574 (\emph{wesha})] \cite{RefB:21}\\
	\emph{wesha-\textbf{ñma}-nge-} (Vi) to be a bad person; \\
	\emph{wesha-\textbf{ñma}-w-} (Vi) to break down, to fall apart, to become a bad person; \\
	\emph{wesha-\textbf{ñma}-w-küle-} (Vi) to be broken/in pieces, to feel awful;
\end{example}

\begin{definition} \label{def:33} \texttt{["+ADV"\{.ñma\}]: ["@ÜÑ"\{ma\}];} \end{definition}

\begin{exercise} \label{R:53} \textbf{Prevent adverbializer to appear as verbal suffix} \\
    \texttt{define RuPr51 $\sim$\$[\$["+ADV"\{.ñma\}]]];} \end{exercise}

\begin{exercise} \label{R:54} \textbf{Apply adverbializer only to adjectives and adverbs} \\
    \texttt{define NvsOpRu01 [["+ADV"\{.ñma\}] => \newline ["-AJ"|"-AV"] ?* \_ ];} \end{exercise}

D\ref{def:33} encodes \emph{-ñma} as adverbializer. R\ref{R:53} prevents it to appear along the verb sequence of suffixes. R\ref{R:54} restricts it to co-occur only with adjectives and adverbs. The other forms, nouns, numerals, that occur with this suffix are collected as lexicalized forms in the lexicon:

\begin{definition} \label{def:34} \\
	\texttt{["-AV"{.epuñma\_con-nosotros-dos}]:["@G"\{epuñma\}]}
\end{definition}

\section{Analyser dimensions} \label{sec:64}

\paragraph{} \label{tp:87} This section exposes data referent to the amount of each type of element interacting in the system: lexicon, suffixes, rules, states, etc.

\begin{figure}[H]
	\includegraphics[width=15.5cm,height=19.5cm]{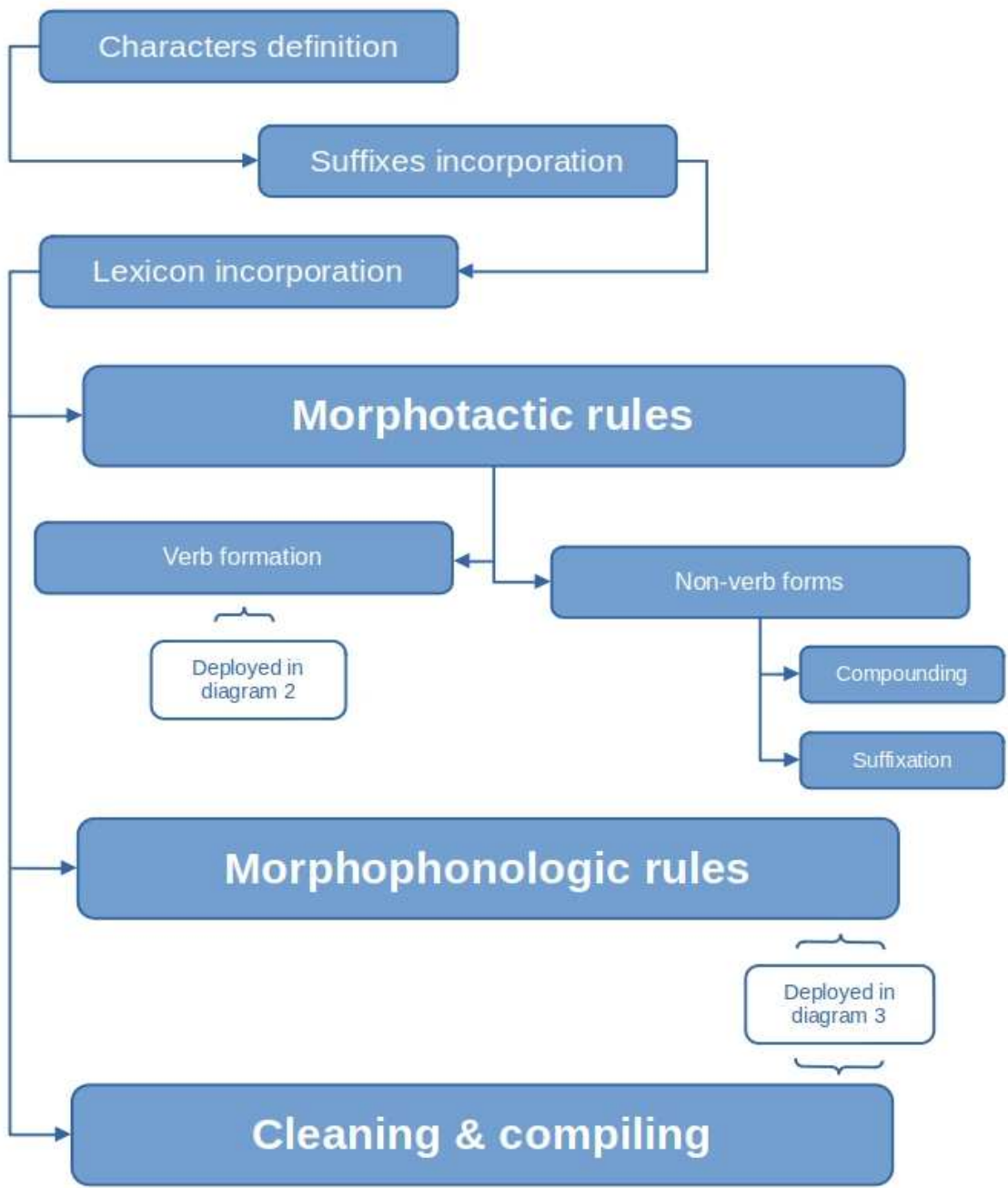}
	\caption{FST Analyser flux diagram 1: general view.}
	\label{fig:09}
\end{figure}

\begin{figure}[H]
	\includegraphics[width=\textwidth]{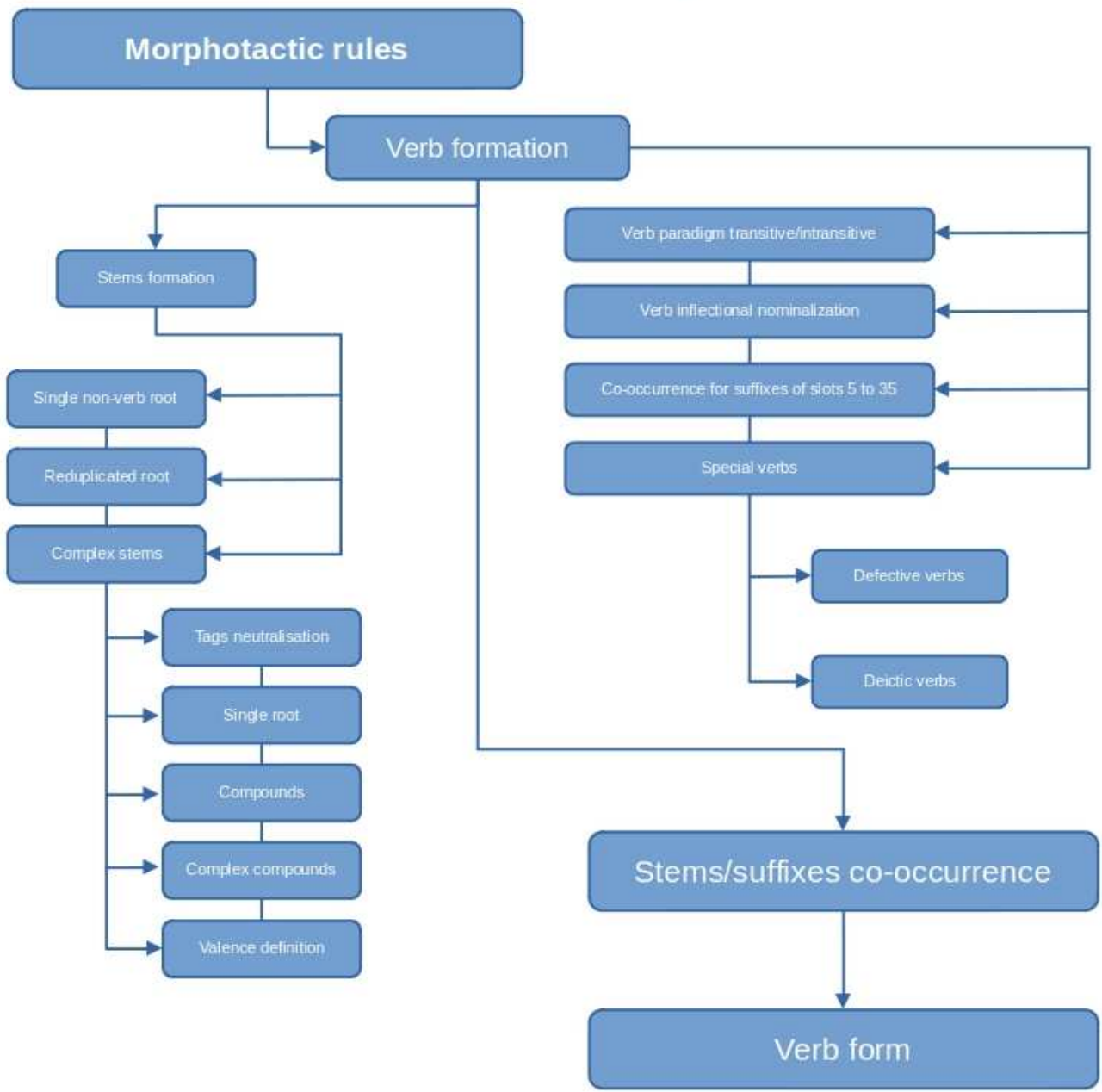}
	\caption{FST Analyser flux diagram 2: verb formation view.}
	\label{fig:10}
\end{figure}

\begin{figure}[H]
	\includegraphics[width=14cm,height=24cm]{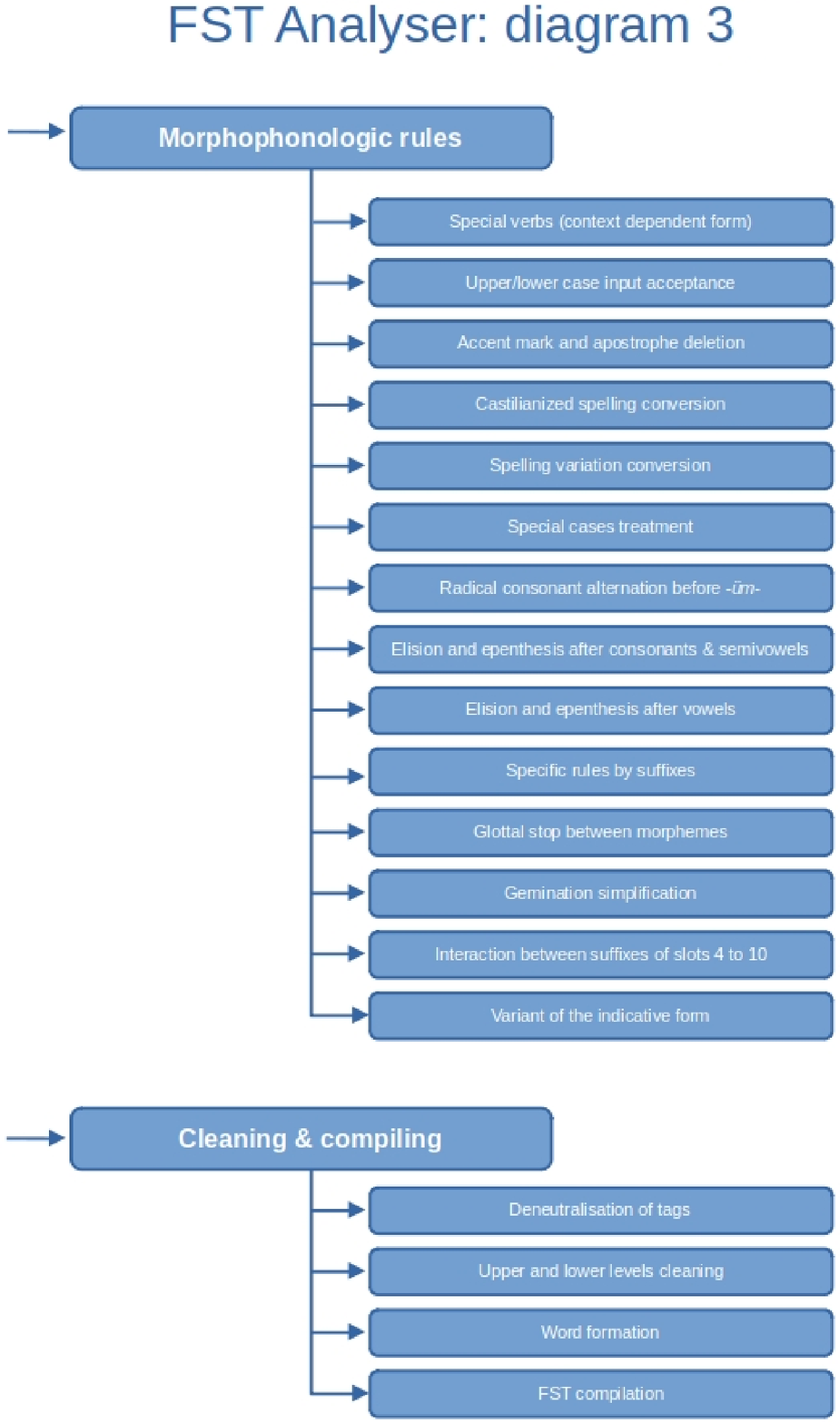}
	\caption{FST Analyser flux diagram 3: morphophonologic rules, cleaning \& compiling view.}
	\label{fig:11}
\end{figure}

There is a flow chart (deployed in figures \ref{fig:09}, \ref{fig:10} and \ref{fig:11}) showing the interconnection of all the analyser elements along the process in annex \ref{anx:13}, p. \pageref{anx:13}.

\begin{itemize} \label{it:41}
    \item[] \textbf{Roots (verbalizable lexicon): 2,096}
    \item[] Adjectives: 128
    \item[] Adverbs: 24
    \item[] Intransitive verbs: 257
    \item[] Proper nouns: 68
    \item[] Nouns: 1,325
    \item[] Numerals: 14
    \item[] Onomatopoeia: 12
    \item[] Questions: 4
    \item[] Transitive verbs: 264
\end{itemize}

\begin{itemize} \label{it:42}
    \item[] \textbf{Non verbalizable lexicon: 266}
    \item[] Adverbs: 88
    \item[] Anaphoric pronouns: 5
    \item[] Auxiliaries: 8
    \item[] Conjunctions: 9
    \item[] Demonstrative pronouns: 6
    \item[] Foreign expressions: 7
    \item[] Interrogative pronouns: 8
    \item[] Interjections: 27
    \item[] Negations: 1
    \item[] Numbers: 10\footnote{\texttt{["-NBR"]: [[\%0|1|2|3|4|5|6|7|8|9]+];} This regex declares the unities and the + sign encodes any combination formed by one to infinite unities, all of which would be tagged \texttt{"-NBR"}.}
    \item[] Particles: 19
    \item[] Personal pronouns: 9
    \item[] Possessive pronouns: 6
    \item[] Prepositions: 5
    \item[] Punctuation marks: 58
\end{itemize}

\begin{itemize} \label{it:43}
    \item[] \textbf{Suffixes: 116}
\begin{itemize}
    \item[] \textbf{Verb suffixes: 101}
    \item[] Inflectionals: 56
    \item[] Mobile derivationals: 20
    \item[] Fix\footnote{In this category not all derivational are fix, but most of them; as in the previous category not all suffixes are mobile, but many of them.} derivationals: 24
    \item[] Non-slot assigned: 1
\end{itemize}
\begin{itemize} \label{it:44}
    \item[] \textbf{Non-verb\footnote{These suffixes may actually be added to nominalized verbs, but never to finite verb forms, i.e., verbs that have mood, person and number.} suffixes: 15}
    \item[] Class-changing: 3
    \item[] Instrumental: 1
    \item[] Non class-changing: 6
    \item[] Nominalizers: 5
\end{itemize}
\end{itemize}

\begin{itemize} \label{it:45}
    \item[] \textbf{Rules: 472}
    \item[] Regexs (files\footnote{These are files containing regular expressions encoding the lexicon and suffixes, which a are separated from the main script.}) location: 76
    \item[] Character definitions\footnote{These are the lists of consonants, vowels and semi-vowels.}: 3
    \item[] Phonological: 44
    \item[] Morphological: 345
    \item[] Cleaning\footnote{These rules clear symbols used as marks when processing morphological and phonological changes.}: 4
\end{itemize}

\begin{itemize} \label{it:46}
    \item[] \textbf{Compilation values}
    \item[] Size: 200.3 MB.
    \item[] States: 2,858,426
    \item[] Arcs: 13,128,696
    \item[] FST type: cyclic
\end{itemize}

\section{Evaluating the analyser} \label{sec:65}

\subsection{Corpora in use} \label{sec:66}

\subsubsection{Gold standard} \label{sec:67}

\paragraph{} \label{tp:88} We have collected a corpus made of sentences coming from "A grammar of \emph{Mapuche}" [Smeets, I. 2008] \cite{RefB:21}, which is our "Gold standard" corpus. Words in the corpus were analysed and disambiguated by Smeets.

The Gold Standard corpus includes all the sentences from chapters 10 to 18, and 21 [Smeets, I. 2008: 61-116, 121-128] \cite{RefB:21}. These chapters deal with nouns, adjectives, adverbs, numerals, demonstratives and anaphoric pronouns, personal pronouns, possessive pronouns, interrogative pronouns, suffixation and verbalization. The corpus also contains all the seventeen texts of "Part VIII - Texts" [Smeets, I. 2008: 369-487] \cite{RefB:21}, which is obviously the most abundant source of \emph{Mapuche} writings in the Gold standard. Texts titles are:

\begin{itemize} \label{it:47}
	\item[] Text 1. Demons
	\item[] Text 2. Work
	\item[] Text 3. Youth
	\item[] Text 4. Missionary
	\item[] Text 5. The war
	\item[] Text 6. An old man
	\item[] Text 7. Olden times
	\item[] Text 8. Conversation about demons
	\item[] Text 9. Conversation about youth
	\item[] Text 10. Conversation about work on big farms
	\item[] Text 11. Conversation about land disappropriation
	\item[] Text 12. Our reservation
	\item[] Text 13. My father
	\item[] Text 14. Brick
	\item[] Text 15. Song 1
	\item[] Text 16. Song 2
	\item[] Text 17. Song 3
\end{itemize}

\subsubsection{Control corpus} \label{sec:68}

\paragraph{} \label{tp:89} Out of the Gold standard we have extracted a control corpus of 240 sentences containing a total of 1,671 words, which correspond to 650 forms.

\begin{example} \label{ex:259} \\
	\emph{ñi trewa, ñi ñarki ka ñi kawell} 'my dog, my cat and my horse'
\end{example}

In example E\ref{ex:259} there are seven words, but 5 forms which are \emph{ka, kawell, ñarki, ñi} and \emph{trewa}. The 3 \emph{ñi} words count as 1 form.

The control corpus is used to check results obtained from the analyser. A correct analysis for every word of this corpus must appear in the output.

\subsubsection{Comparison corpus} \label{sec:69}

\paragraph{} \label{tp:90} This corpus is collected from Zúñiga's texts [Zúñiga, F. 2006: 270 (15)] \cite{RefB:24}. It is made of 170 sentences that contain 1,256 words which correspond to 511 forms (see previous section \ref{sec:68} and example E\ref{ex:259}). The texts are extracted from "\emph{Mapudüngun}. El habla \emph{Mapuche}", chapter V. Textos en \emph{mapudüngun} [Zúñiga, F. 2006: 266 - 288] \cite{RefB:24}. Text titles are:

\begin{enumerate} \label{it:48}
	\item \emph{Feychi ngürü afngünengelu} 'That crafty fox'
	\item \emph{Mawün} 'Rain'
	\item \emph{Pewma} 'A dream'
	\item \emph{Ngillañ mawün} 'Asking for rain'
	\item \emph{We tripantu} 'New year'
	\item Pausa\_Historia 'Pause\_History'
	\item Abuela\_Voz 'Granny\_Voice'
\end{enumerate}

\subsection{Establishing the ambiguity parameter} \label{sec:70}

\paragraph{Ambiguity.} \label{tp:91} In any language some words present ambiguity. These word forms, known as homographs, correspond to different meanings, which are disambiguated by context; for example, the English word \textbf{drop} means 'a little amount of liquid' in "a \textbf{drop} of coffee stained my letter". And it means 'to fall' or 'let fall'  in "do not \textbf{drop} papers on the floor". The only way to know which meaning \textbf{drop} is referring to, i.e., the only way to "disambiguate" it, is by putting it into context.

In \emph{Mapudüngun} there are quite few ambiguous words (homophones which once written become homographs) that appear very often in texts, like \emph{ka}, \emph{fey} and most of the words ending in \emph{n} which is the shared form between the "1\textsuperscript{st} person, singular" suffix and the "plain verbal noun" suffix, examples of them are below (see \ref{sec:51} \nameref{sec:51}, p. \pageref{sec:51}):

\begin{itemize} \label{it:49}
	\item[]\emph{ka} \texttt{→ -AJ.ka\_otro}
	\item[]\emph{ka} \texttt{ → -CJ.ka\_y}
	\item[]\emph{ka} \texttt{ → -PT.ga\_ciertamente\_indignación\_cinismo}
	\item[]
	\item[]\emph{fey} \texttt{ → -AV.fey\_entonces}
	\item[]\emph{fey} \texttt{ → -DP.fey\_que\_aquel\_ese}
	\item[]\emph{fey} \texttt{ → -IV.fe\_ser-eso+IND.y4+3.Ø3}
	\item[]\emph{fey} \texttt{ → -PP.fey\_él\_ella\_ellos}
	\item[]
	\item[]\emph{küdawün} \texttt{ → -IV.küdaw\_trabajar+IND1SG.n3}
	\item[]\emph{küdawün} \texttt{ → -IV.küdaw\_trabajar+PVN.n4}
\end{itemize}

\paragraph{Mapudüngun ambiguity calculus.} \label{tp:92} If we add up all the analysis of three previous words: \texttt{9}, and divide it by the number of words we are taking into account: \texttt{3}, we obtain the "average \emph{Mapudüngun} ambiguity" (\textbf{ama}) of these words: \texttt{3}. We use this result as a reference parameter when comparing analysis results.

As we have explained before, the development of the analyser arises from Smeets' description of \emph{Mapudüngun} morphology [Smeets, I. 2008] \cite{RefB:21}. We have first developed an analyser that strictly fits with Smeets work, we have generated an FST from it, we call it "SmeetsAnalyser". Then, we have expanded the analyser in lexicon and rules that fit another variants of \emph{Mapudüngun}, provisionally, we call it \emph{Düngupeyem}\footnote{\emph{düngu} 'word, speech, language'; \emph{pe} 'proximity suffix, it indicates physic and or temporal proximity with the action expressed by the verb; \emph{ye} 'constant feature suffix'; \emph{m} 'it indicates instrument or location'. All together it is something like 'instrument always used to do things with language'. We intended to say 'language tool'. That is one of the reasons to be a provisional name. 'Tool' in \emph{Mapudüngun} is \emph{küdaw-ka-we}, so another possibility is to call our analyser \emph{Düngu-ka-we}. Another reason to be a provisional name is that we have found no native \emph{Mapuche} yet, who validates the name.}.

To calculate the average \emph{Mapudüngun} ambiguity, and establish it as a reference parameter, we have analysed the control corpus with the "SmeetsAnalyser".

\begin{itemize} \label{it:50}
	\item[] Control corpus analysed with "SmeetsAnalyser"
	\item word-forms = 650
	\item produced analyses = 2,232
	\item unknown words = 2
	\item[]
	\item[] \fbox{$\mathbf{2,232 \div (650 - 2) = 3.59 ~ ama}$}
\end{itemize}

Following this calculus we obtain \textbf{3.59 ama}, this means that every word-form has an average of 3.59 possible analyses, this is our reference parameter.

\paragraph{Increased ambiguity calculus.} \label{tp:93} The incorporations we have made in the system, those mentioned in section \ref{sec:53}, [p. \pageref{sec:53}], increase the ambiguity in analysis results because they imply more ways of analysis for every form. Increasing the lexicon also produces this effect.

Now we calculate the "average increased ambiguity" (\textbf{aia}) to be able to compare the results and verify if the system is still reliable.

\begin{itemize} \label{it:51}
	\item[] Control corpus analysed with \emph{Düngupeyem}
	\item word-forms = 650
	\item produced analyses = 2,477
	\item unknown words = 2
	\item[]
	\item[] \fbox{$\mathbf{2,477 \div (650 - 2) = 3.82 ~ aia}$}
\end{itemize}

As expected, ambiguity rises, but not too much, less than 0.24 points per word-form, which indicates that the reliability of the analyser is quite good.

\subsubsection{Comparison corpus results} \label{sec:71}

\paragraph{} \label{tp:94} To check if the additions (see \ref{sec:53}, p. \pageref{sec:53}) we have made to the FST really worth it, we analyse the comparison corpus.

\begin{itemize} \label{it:52}
	\item[] Comparison corpus analysed with "SmeetsAnalyser"
	\item word-forms = 511
	\item produced analyses = 1,368
	\item unknown words = 120
	\item[]
	\item[] \fbox{$\mathbf{1,368 \div (511 - 120) = 3.49 ~ aa}$}
\end{itemize}

\begin{itemize} \label{it:53}
	\item[] Comparison corpus analysed with \emph{Düngupeyem}
	\item word-forms = 511
	\item produced analyses = 1,828
	\item unknown words = 10
	\item[]
	\item[] \fbox{$\mathbf{1,828 \div (511 -10) = 3.64 ~ aa}$}
\end{itemize}

At first glance, these results confirm a good performance of \emph{Düngupeyem}, the difference in the average ambiguity (\textbf{aa}) even diminishes from 0.24 to 0.15. But there is a factor that we did not considerate in the calculus with the control corpus, it would have made no difference in results. There were too few unrecognized words (only 2), and the same amount for both analysers.

Analysing the comparison corpus, we obtain 120 unrecognized words from "SmeetsAnalyser", and only 10 from \emph{Düngupeyem}. To measure the impact of this factor in the ambiguity index, we have used a "root guesser\footnote{The root guesser is a tool derived from the analyser, which have no lexicon of roots, but the possible root structures in terms of consonants, vowels and semi-vowels, e.g., CVC, CVSV, CVCVCVC are valid \emph{Mapuche} root structures. The lexicon of suffixes is included, also their combination rules. This machine first check the possible root structures and then the possible suffixes combinations. This FST is not described in this article because that would have made it too extensive.}" to count the possible analyses the unknown words generate. Then, we add the resulting possible analyses to the known analysis and recalculate the ambiguity index.

\begin{itemize} \label{it:54}
	\item[] 120 unrecognized words from "SmeetsAnalyser"
	\item words producing no possible analyses = 49
	\item words producing possible analyses = 71
	\item total number of possible analyses = 636
	\item[]
	\item[] \fbox{$\mathbf{(1,368 + 636) \div (511 - 49) = 4.33 ~ aa}$}
\end{itemize}

\begin{itemize} \label{it:55}
	\item[] 10 unrecognized words from \emph{Düngupeyem}
	\item words producing no possible analyses = 3
	\item words producing possible analyses = 7
	\item total number of possible analyses = 56
	\item[]
	\item[] \fbox{$\mathbf{(1,828 + 56) \div (511 - 3) = 3.70 ~ aa}$}
\end{itemize}

These calculi confirm that the analyser is reliable, even more after adding the considerations for other dialects of \emph{Mapudüngun} and more lexicon. The average analyses (\textbf{aa}) have raised, but too little as to consider it a problem, only 0.06 between the analyses that does not take into account the possible analyses for unknown roots (3.64), and the one that does take them into account (3.70).

\subsection{Comparing against other analysers} \label{sec:72}

\paragraph{} \label{tp:95} For this purpose we have used the Gold standard corpus (see \ref{sec:67}, p. \pageref{sec:67}) to train two different analysers, and we have analysed the comparison corpus (see \ref{sec:69}, p. \pageref{sec:69}) with these tools and our analyser to compare results.

\subsubsection{Trainable systems to compare against} \label{sec:73}

\paragraph{} \label{tp:96} A trainable system receives a disambiguated corpus to learn from, i.e., it stores in its memory (or database) the correct forms that have been introduced in the training process. Then, using different algorithms, it compares an input text to the stored data and tags input accordingly. The trainable systems we have compared our analyser against are RFTagger\footnote{We wanted to use another tagger, one built with neuronal networks that we have trained last year. But we could not make it work now, it seems that Python modules have change too much, and we only get errors with it.} [Schmid \& Laws 2008] \cite{RefB:20} and Morfette [Chrupala et al. 2008] \cite{RefB:04}.

\paragraph{RFTagger.} \label{tp:97} This is a tool for the annotation of text with fine-grained part of speech tags\footnote{This is the developers own definition found on the RFTagger site: \href{https://www.cis.uni-muenchen.de/~schmid/tools/RFTagger/}{https://www.cis.uni-muenchen.de/$\sim$schmid/tools/RFTagger/} <22/08/2020>.}. It is a Hidden-Markov-Model tagging method that is particularly suitable for PoS tag sets with numerous detailed tags [Schmid \& Laws 2008: 1] \cite{RefB:20}. An HMM part of speech tagger calculates the most likely sequence of PoS tags for a given word sequence. The difference with our analyser is that RFTagger identifies a complete form and tags it accordingly, while our morphological analyser breaks down the input form and gives a tag for each part that constructs the full form.

\paragraph{Morfette.} \label{tp:98} This is a tool for supervised learning of inflectional morphology. Given a corpus of sentences annotated with lemmas and morphological labels, and optionally a lexicon, Morfette learns how to morphologically analyse new sentences\footnote{This is the developers own definition found on the Morfette site: \href{https://hack-age.haskell.org/package/morfette}{https://hack-age.haskell.org/package/morfette} <22/08/2020>}. This is a data-driven modular probabilistic system that learns to perform morphological tagging from morphologically annotated corpora. The system is composed of two learning modules that use a maximum entropy classifier to predict morphological tags. The third module dynamically combines the predictions of the Maximum-Entropy models and generates a probability distribution over the sequences of tag-lemma pairs [Chrupala et al. 2008: 1] \cite{RefB:04}.

\paragraph{Training RFTagger and Morfette.} \label{tp:99} The text of the Gold standard corpus has to be tokenized, one token per line. For each token, a tag is entered after a tab space. For verbs, the tag shows 38 slots for attributes. 36 verbal slots and two additional ones for nominal, temporal, instrumental or nominalizing suffixes; each unavailable attribute\footnote{Grammatical categories are collected with attributes, which are the aspects that detail the category, for example, the word "plastic" has the category Noun; a fine-grained tag for it would be NSM, where N stands for "noun", S for  "singular" and M for "masculine", which are the attributes.}, which fits into a slot, is indicated by a 0 (zero), as shown in the following example:

\begin{example} \label{ex:260}\ \emph{amu-ke-fu-y} 'he used to go'\\
	\texttt{amukefuy IV.0.0.0.0.0.0.0.0.0.0.0.0.0.0.0.0.0.\\0.0.0.0.0.+CF@ke14.0.0.0.0.0.+IPD@fu8.0.0.0.\\+IND@üy4.+3@Ø3.0.0.0.0}
\end{example}

For non-verbal forms, the tag holds four positions for attributes, the first one occupied by the category of the form and the last three by possible nominal suffixes, as shown in the following example:

\begin{example} \label{ex:261}\ \emph{peñi-mu} 'at my brother's'\\
	\texttt{peñimu NN.0.0.+INST@mew}
\end{example}

Morfette requires a bit more information. There is an additional column between the form and the tag to enter the lemma, where the root was added instead. Unlike RFTagger, attributes of the tag are not preceded by dots, compare E\ref{ex:260} with the following example:

\begin{example} \label{ex:262}\ \emph{amu-ke-fu-y} 'he used to go'\\
	 \texttt{amukefuy amu IV0000000000000000000000+CF@ke14\\00000+IPD@fu8000+IND@üy4+3@Ø30000}
\end{example}

Both machines were trained in their standard way, that is, with their default settings, except that in the case of Morfette we chose the vector representations of words (see E\ref{ex:260}, E\ref{ex:261}, E\ref{ex:262}). The vector representation for tags uses a \texttt{0} (zero) for each possible attribute, where the realized ones replace them in the appropriate position. It is the only configuration that RFTagger accepts.

\subsubsection{Results from the three analysers} \label{sec:74}

\paragraph{} \label{tp:100} Results from the two comparison tools are not ambiguous, they output only one tag per form, which gives us a binary value, either right or wrong tag assignation. They always set a tag for every form, so there are no unknown forms neither. We plan to add a disambiguator to the system so we can analyse words in context and, instead of outputting all possible analyses per word, only output the most suitable one.

The training corpus, the "Gold standard", is made up of 1,220 sentences, which in turn are made up of 9,209 tokens of which 1,998 are punctuation marks; when subtracting the latter, the total of unambiguously annotated word-forms is 7,211. This material is actually insufficient\footnote{In data driven tools the more training data receive the tool the better it performs its tasks. Text applied tools use to be trained with thousands of sentences. For the tools we have used (RFTagger and Morfette) we believe that about 15 thousand sentences would give good quality results; in Smeets' work there are about 1,800 morphologically tagged sentences.} for an accurate training that can obtain acceptable results. Note that this is one of the main reasons for developing a morphological analyser based on rules, as there are not enough \emph{Mapuche} annotated corpora available to carry out the data-driven type of approach with computational tools.

\begin{table}[htb]
	\caption{Analysing results for the comparison corpus (see \ref{sec:69})}
	\label{tab:10}
	\begin{tabular}{|c|c|c|c|}
		\hline\noalign{\smallskip}
		Data type & RFTagger & Morfette & \emph{Düngupeyem} \\
		\noalign{\smallskip}\hline\noalign{\smallskip}
		Tokens & 1,471 - 100\% & 1,471 - 100\% & 1,471 - 100\% \\
		\noalign{\smallskip}\hline\noalign{\smallskip}
		Right & 723 - 49.1\% & 786 - 53.4\% & 1,460 - 99,2\% \\
		\noalign{\smallskip}\hline\noalign{\smallskip}
		Wrong & 749 - 50.8\% & 685 - 46.5\% & 0 \\
		\noalign{\smallskip}\hline\noalign{\smallskip}
		Unknown & - & - & 11 - 0.75\%\\
		\noalign{\smallskip}\hline\noalign{\smallskip}
		Analyses & 1,471 & 1,471 & 4,533 \\
		\noalign{\smallskip}\hline
	\end{tabular}
\end{table}

\paragraph{RFTagger results.} \label{tp:101} This tool can not be blamed for this low performance (nor Morfette either). The insufficient amount of training material is the cause of the poor results. RFTagger was not developed for agglutinative languages explicitly, or the complexities of \emph{Mapudüngun}. Also keep in mind that agglutinative languages have potentially an infinite number of words, as English, for example, has potentially an infinite number of phrases. Therefore, it is more difficult for the system to learn from such a variety of forms how to tag accurately.

\paragraph{Morfette results.} \label{tp:102} This tool can handle the analysis process a little better than RFTagger. More words are correctly analysed, and less words are incorrectly analysed.

Throughout the results validation we have noticed that Morfette breaks down the forms to apply its inferred assumptions on analysed suffixes. But both machines fail though, in assuming almost all the words starting by a capital letter as a proper noun.

\paragraph{Düngupeyem results.} \label{tp:103} To make the results comparable, we have generated a mathematical formula that approach the two types of data and give us a way to explain the numbers.

The average ambiguity (\textbf{aa}) of the comparison corpus (see \ref{it:53}, p. \pageref{it:53}) is \textbf{3.64 aa}. On subtracting the number of analyses corresponding to that percentage, of \emph{Düngupeyem}'s correct analyses (1,460) we relate this amount to those of Morfette and RFTagger right analyses. Then, we recalculate the percentage obtaining a closer degree of correspondence among the three machines results:\\

\minibox[frame]{ \medskip $\mathbf{3,64(aa)~\%~1,460~=~53.29~(ambiguous~words)}$ \\
	\medskip $\mathbf{1,460~-~53.29~=~1,406.71}$ \\
	\medskip $\mathbf{1,471~=~100\%~(analysed~words)}$ \\
	$\mathbf{1,406.71~=~95.62\%~(correctly~analysed~words)}$} \bigskip

After these results, it is again clear the system's reliability, and the analyses' high degree of accuracy. We see that regardless counting not recognized words as if they were wrong analyses, failed analyses reach only 0.75\%, while success overcomes 95\%. The other two tools are around 50\% in right and wrong analyses.

\subsection{Comparing against a Quechua analyser} \label{sec:75}

\paragraph{} \label{tp:104} Our work follows the path that Annette Ríos\footnote{\href{https://www.cl.uzh.ch/de/people/team/compling/arios.html}{https://www.cl.uzh.ch/de/people/team/compling/arios.html}} has drawn for Quechua. This step is not an exception. Although there
are notable differences, \emph{Düngupeyem}'s evaluation is largely based on the same process for Quechua FST tools carried out by Ríos [Ríos, A. 2015: 36-39] \cite{RefB:17}.

The Quechua system is made up of 5 cascading transducers [Ríos, A. 2015: 22] \cite{RefB:17}, and some of these FSTs use a disambiguator for specific parts of the forms. For example, to unravel whether a root is nominal or verbal; or on another transducer to reveal if a suffix is of one or another type, etc. So the final Quechua form has been disambiguated throughout the analysis process.

Training material also differs. Quechua system used about 3,000 disambiguated sentences. \emph{Mapuche} training material (see \ref{sec:70}, p. \pageref{sec:70} and \ref{sec:73}, p. \pageref{sec:73}) consists of less than the half of those sentences, 1,220 to be exact. In both cases, however, this is too little material to train data-driven systems well enough.

Comparison corpus are similar. Ours contains 170 sentences with 2,207 tokens (1,475 words + 732 punctuation marks) (see \ref{sec:69}, p. \pageref{sec:69}), Quechua has 322 sentences containing 2,142 tokens [Ríos, A. 2015: 36] \cite{RefB:17}. 

Although the procedure was also performed differently, the final counting (the addition of results) may serve as comparison data. In the Quechua case, both texts of the comparison corpus were tagged separately, to later compare them with each other. In the \emph{Mapuche} case, by contrast, texts were tagged at once.

\subsubsection{Quechua results} \label{sec:76}

\paragraph{} \label{tp:105} Quechua results are transferred and presented in the style used in this article, different from their original presentation, for comparison reasons\footnote{Data are taken from table 2.13: Evaluation: Disambiguated Texts [Ríos, A. 2015: 38] \cite{RefB:17}.}.

\begin{table}[htb]
	\caption{Analysis results. RFTagger / Morfette / Quechua system}
	\label{tab:11}
	\begin{tabular}{|c|c|c|c|}
		\hline\noalign{\smallskip}
		Data type & RFTagger & Morfette & Quechua S. \\
		\noalign{\smallskip}\hline\noalign{\smallskip}
		Tokens & 2,142 - 100\% & 2,142 - 100\% & 2,091 - 97.6\% \\
		\noalign{\smallskip}\hline\noalign{\smallskip}
		Right & 1,459 - 68.1\% & 1,505 - 70.2\% & 2,041 - 95.2\% \\
		\noalign{\smallskip}\hline\noalign{\smallskip}
		Wrong & 683 - 31.8\% & 637 - 29.7\% & 50 - 2.33\% \\
		\noalign{\smallskip}\hline
	\end{tabular}
\end{table}

\begin{table}[htb]
	\caption{Results RFTagger / Morfette - Quechua / \emph{Mapudüngun}}
	\label{tab:12}
	\begin{tabular}{|c|c|c|c|c|}
		\hline\noalign{\smallskip}
		 & \multicolumn{2}{c|}{RFTagger} & \multicolumn{2}{c|}{Morfette} \\
		\noalign{\smallskip}\hline\noalign{\smallskip}
		 & Right & Wrong & Right & Wrong \\
		\noalign{\smallskip}\hline\noalign{\smallskip}
		Quechua & 68.11\% & 31.88\% & 70.26\% & 29.73\% \\
		\noalign{\smallskip}\hline\noalign{\smallskip}
		\emph{Mapudüngun} & 49.12\% & 50.88\% & 53.43\% & 46.57\% \\
		\noalign{\smallskip}\hline\noalign{\smallskip}
		Difference & 18.99\% & 19\% & 16.83\% & 16.84\% \\
		\noalign{\smallskip}\hline\noalign{\smallskip}
		Average difference & \multicolumn{2}{c|}{18.99\%} & \multicolumn{2}{c|}{16,83\%} \\
		\noalign{\smallskip}\hline
	\end{tabular}
\end{table}

\paragraph{RFTagger results.} \label{tp:106} table \ref{tab:11} shows that increasing the amount of training data (1,220 to 3,000 sentences), results improve consequently for both, RFTagger and Morfette.

RFTagger worked well with 68\% of the Quechua forms, which is 19\% more than with \emph{Mapuche} ones. The percentage of wrongly analysed forms is also better, being 19\% lower. Thus, performance of RFTagger with Quechua was 19\% better than with \emph{Mapudüngun} (see table \ref{tab:12}). However, be aware of the differences between the two evaluation processes (see \ref{sec:75}, p. \pageref{sec:75}).

\paragraph{Morfette results.} \label{tp:107} This tool also performs better for Quechua (compare results in table \ref{tab:12}). As with \emph{Mapuche} texts, Morfette performs better than RFTagger in the Quechua texts tagging.

Morfette results with Quechua texts were 16.8\% better than with \emph{Mapuche} texts (see table \ref{tab:12}). It produced 16.8\% more of correct analysis, and the same percentage less of incorrect analyses. The difference is smaller, as it is in the \emph{Mapuche} texts analysis.

Recall that the percentage of correctly analysed words of our analyser was 95.62\% (see \nameref{tp:103}, p. \pageref{tp:103}). table \ref{tab:11} shows, in the case of the Quechua system, 95.2\% of correctness. Despite the differences between both systems (see \ref{sec:75}, p. \pageref{sec:75}) results are similar. So, if Ríos considers her machine performs successfully based on her analyses results, we may go along with the same consideration for our system.

\subsection{Not analysed!} \label{sec:77}

\paragraph{} \label{tp:108} In analyses of the control corpus (\ref{it:51}, p. \pageref{it:51}), made up of Smeets' sentences [Smeets, I. 2008] \cite{RefB:21}, there were two unknown words. We analyse these cases below.

\subsubsection{Smeets' not analysed words.} \label{sec:78}

\paragraph{eluwün-antü \emph{'funeral day'}} \label{tp:109} [Smeets, I. 2008: 402 (74)] \cite{RefB:21}. Presented this way, this word is a nominal compound. But we have not collected \emph{eluwün} 'funeral' as a noun because it is actually a nominalized verb:

\begin{example} \label{ex:263} [Smeets, I. 2008: 404 (7)] \cite{RefB:21} \\
	\emph{el-uw-ün} 'funeral', lit: 'the leaving behind/going' \\
	\texttt{-TV.el\_dejar-atrás\_partir+REF.w31+PVN.n4}
\end{example}

\begin{example} \label{ex:264} [Smeets, I. 2008: 66 (48)] \cite{RefB:21} \\
	\emph{antü} 'day' \\
	\texttt{-NN.antü\_sol\_día\_tiempo}
\end{example}

This is another type of compound that we have not encoded in our system. We think it is not worth to include it because, up to this moment, it has only appear two or three times. Adding variants of whatever form increases the ambiguity, which negatively affects the performance of the analyser.

\paragraph{mapuche-domo \emph{'Mapuche woman'}} \label{tp:110} [Smeets, I. 2008: 399 (18)] \cite{RefB:21}. This is another nominal compound that Smeets presents as made up two nouns. But these are actually three noun roots:

\begin{example} \label{ex:265} [Smeets, I. 2008: 117] \cite{RefB:21} \\
	\emph{mapu-che} 'land person' \\
	\texttt{-NN.mapu\_tierra-NN.che\_persona}
\end{example}

\begin{example} \label{ex:266} [Smeets, I. 2008: 76 (20)] \cite{RefB:21} \\
	\emph{domo} 'woman' \\
	\texttt{-NN.domo\_mujer}
\end{example}

A compound of three roots is not contemplated in the system either, for the same reasons of the previous compound. But, a solution to successfully analyse these type of compounds may be to cascade another FST capable of analysing three roots compounds after the main FST is unable to do it, this way we do not increase ambiguity in the main analyser.

\subsubsection{Zúñiga's not analysed words.} \label{sec:79}

\paragraph{} \label{tp:111} In the analyses of the comparison corpus (\ref{it:53}, p. \pageref{it:53}), made up of Zúñiga's sentences [Zúñiga, F. 2006: 266 - 288] \cite{RefB:24}, there were ten unknown words (\emph{puru} 'to dance' is repeated, so, actually nine). In this section we try to find out why these words could not successfully go through the analyser.

\paragraph{are-tu \emph{'borrowed'}} \label{tp:112} [Zúñiga, F. 2006: 280 (122)] \cite{RefB:24}. In Zú-ñiga and Augusta, F. \cite{RefB:03}, this root is an adjective. From the perspective of Spanish or English it should be the participle form of the verb. Smeets and Augusta collect it as verb, \emph{are-} 'to lend', and when followed by the suffix \emph{-tu-}, \emph{are-tu-} 'to borrow'.

"An \emph{-n} form occurs as an adjective denoting an attribute or quality of the modified noun" [Smeets, I. 2008: 190] \cite{RefB:21}, among other functions of the plain verbal noun. This lead us to think that \emph{aretu} may be analysed as \emph{are-tu-Ø}, where the plain verbal noun is elided, or realized as a null suffix.

A solution, then, to correctly analyse this form, would be to add the null form of the plain verbal noun to the list of suffixes. The problem is that this action would enormously increase the ambiguity, because every nonverbal root allowed to take the verbalizer \emph{Ø}, would be analysed as \texttt{root} and \texttt{root+VRB.Ø36+PVN.Ø4}.

\paragraph{ina-lef-nepe-n \emph{'I startled wake up'}} \label{tp:113} [Zúñiga, F. 2006: 283\\ (\emph{pewma})] \cite{RefB:24}. From the meaning Zúñiga gives to this word, we think that it is a short form for:

\begin{example} \label{ex:267}\ \emph{ina-lef-ün nepe-n} lit: 'I woke up on the run' \\
	\texttt{-AV.ina\_a-través-IV.lef\_correr-CR.IV\\+PASS.nge23+PX.pe13+PVN.n4\\-IV.nepe\_despertar+IND1SG.n3 }
\end{example}

\paragraph{ngen-ko \emph{'god of the waters'}} \label{tp:114} [Zúñiga, F. 2006: 285 (\emph{we tripantü})] \cite{RefB:24}. It is the same case as \emph{eluwün-antü} 'funeral day' (see \nameref{tp:109}, p. \pageref{tp:109}). In Smeets is found as in the following example:

\begin{example} \label{ex:268} [Smeets, I. 2008: 138 (41)] \cite{RefB:21} \\
	\emph{nge-n ko} lit: 'owner/master of the water' \\
	\texttt{-TV.nge\_tener+PVN.n4\\-NN.ko\_agua}
\end{example}

\paragraph{puru \emph{'to dance'}} \label{tp:115} [Zúñiga, F. 2006: 287 (Pausa\_Historia)] \cite{RefB:24}. We think it is the same case as \emph{are-tu}, even though, this is a single root, and is not collected as 'danced', not by Zúñiga nor by Augusta (see \nameref{tp:112}, p. \pageref{tp:112}). So, it is probably:

\begin{example} \label{ex:269}\ \emph{puru-Ø} 'to dance' \\
	\texttt{-IV.puru\_bailar+PVN.Ø4}
\end{example}

\paragraph{purunenutuy} \label{tp:116} [Zúñiga, F. 2006: 287 (Pausa\_Historia)] \cite{RefB:24}. We are not certain how Zúñiga translates it. The form \emph{enu} inside this word is unknown for us, but we guess it may be: 

\begin{example} \label{ex:270}\ \emph{puru-nentu-tu-y} 'get oneself dancing\footnote{Maybe from the expression in Spanish 'sacar a bailar' 'ask someone to dance with'.}'\\ \texttt{-IV.puru\_bailar-TV.entu\_sacar-CR.TV\\+RE.tu16+IND.y4+3.Ø3}
\end{example}

\paragraph{taku-tu-mu-tu-y \emph{'she sheltered herself'}} \label{tp:117} [Zúñiga, F. 2006: 279 (108)] \cite{RefB:24}. If the suffix \emph{-mu-} of the form corresponds to the 2\textsuperscript{nd} person agent, slot 23, then this form is not possible following Smeets description. "The subject (slot 3) of a verb which takes the morpheme \emph{-mu-} (slot 23) indicates 1\textsuperscript{st} person. The participant which is deleted from the situation indicated by a \emph{-mu-} form must be a 2\textsuperscript{nd} person. It cannot be 1\textsuperscript{st} person because the subject marker indicates 1\textsuperscript{st} person. The participant which is deleted from the situation cannot be 3\textsuperscript{rd} person (for then one would have used the passive marker \mbox{\emph{-nge-}}), nor can it be included in the subject referent (for then one would have used the reflexive marker \emph{-w-}) ... The suffix \emph{-mu-} is used when the total number of participants is greater than two. The number marker (slot 2) co-refers to the subject marker and may indicate singular, dual or plural" [Smeets, I. 2008: 268] \cite{RefB:21}.

Following Smeets, and fitting Zúñiga's translation, the verb should have been:

\begin{example} \label{ex:271}\ \emph{taku-tu-nge-tu-y} lit: 'she was sheltered by herself'\\ \texttt{-TV.taku\_cubrir+TR.tu33+PASS.nge23+RE.tu16\\+IND.y4+3.Ø3}
\end{example}

Another possibility is that \emph{-mu-} corresponds to an alternative form for another suffix (that is not in our system) like \emph{-me-}, thither, slot 20 or \emph{-m-}, causative, slot 34.

Finally, if we add the suffix \emph{-u} (dual) at the end of the form, we do obtain an analysis, because this form implies the 1\textsuperscript{st} person in its null form \emph{-Ø-}, but we distance from the meaning given by Zúñiga:

\begin{example} \label{ex:272}\ \emph{taku-tu-mu-tu-y-u} 'you sheltered us both'\\ \texttt{-TV.taku\_cubrir+TR.tu33+2A.mu23+RE.tu16\\+IND.y4+1.Ø3+DL.u2}
\end{example}

\paragraph{uf-kün-tuku-pa-y \emph{'they camped in memory'}} \label{tp:118} [Zúñiga, F. 2006: 288 (Abuela\_Voz)] \cite{RefB:24}. First, note that this word is at a poem. Then, following Zúñiga's translation, roots and suffixes interpretations we do for this word are as follows:

\begin{example} \label{ex:273}\ \emph{uf-kün-tuku-pa-y} lit: 'they tight up and put memory/knowledge there'\\ \texttt{-TV.uf\_apretar\_afirmar-TV.kim\_saber\_recordar\\-TV.tuku\_poner+HH.pa17+IND.y4+3.Ø3}
\end{example}

The first reason for this analysis not being produced by our machine is the three roots stem, as we have explained in \nameref{tp:110} (p. \pageref{tp:110}). Then, we are guessing the stem to be composed as \emph{uf-kün-tuku-}.

Smeets collects \emph{üfi-} 'to become tight, to tighten' [Smeets, I. 2008: 567] \cite{RefB:21}. Augusta, F. \cite{RefB:03}, \emph{uf-ün} 'tighten the straws with bands (when roofing)', \emph{uf-tüku-n} 'tighten with tools'. But also \emph{üfü-n} 'tighten with something to tie'. So \emph{uf-} and \emph{üfi-} are, very likely, variants of each other.

In Smeets' dictionary are \emph{kim-tuku-} 'to have known/un-derstood for some time' and \emph{kim-tu-} 'to remember' [Smeets, I. 2008: 559] \cite{RefB:21}. Febrés' dictionary presents \emph{kün-tüku-l-ün} 'make someone else to remember'; \emph{kün-tüku-n} 'to remember'; \emph{kün-tüku-pe-m} 'the memory' Febrés, A. \cite{RefB:03}. It is quite probable that \emph{kim-} and \emph{kün-} are also variants of each other.

\emph{tuku-} and \emph{tüku-} are undoubtedly variants of each other, and it can not be \emph{tü-ku-} because there is no attested \emph{-ku-} suffix.

\paragraph{wima-kütu-ye-nge-y \emph{'she was whipped'}} \label{tp:119} [Zúñiga, F. 2006: 281 (123)] \cite{RefB:24}. A more literal translation would be 'she was taken to be whipped all along' if the parts forming this verb are the ones we suggest:

\emph{wima-} 'dipstick, thin stick'.

\emph{-kütu-} it seems to correspond to a suffix that adds the sense of 'all along'.

Augusta defines it as 'suffix and post-position: (Variant used in \emph{Pangi}). From (temporarily). \emph{kuyfi \textbf{kütu}} 'from a long time'. Conjunction: and even, until' Augusta, F. \cite{RefB:03}.

Valdivia recognizes in it a locative sense, 'from and to; \emph{fa \textbf{kütu} tüye \textbf{kütu}} 'from here to there'' Valdivia, L. \cite{RefB:03}.

Smeets only recognizes it as an adverb, \emph{küto $ \sim $ kütu} 'even, also'. There is a root, though, that Smeets collects as \emph{weñangkü-} 'to get sad' [Smeets, I. 2008: 573] \cite{RefB:21}. Augusta collects it as \emph{weñang-} 'to have pain, annoyance, desire', but also \emph{weñang-kü-n} 'get sad'. So, maybe \emph{-kütu-} is formed by two suffixes \emph{-kü-tu-}, being \emph{-kü-} this suffix of the 'all along' sense, and \emph{-tu-} the transitivizer suffix. This is not totally rare in \emph{Mapudüngun}, see the case of \emph{llemay} in the note to E\ref{ex:67} (p. \pageref{ex:67}).

Concluding, the verb is made up of a complex stem, which is formed by a 'noun root + 1 or 2 suffixes + verb root', possibly:

\begin{example} \label{ex:274}\ \emph{wima-kü-tu-ye-nge-y} \\ \texttt{-NN.wima\_vara+THR.kü35+TR.tu33\\-TV.ye\_traer\_llevar+PASS.nge23+IND.y4+3.Ø3}
\end{example}

\begin{example} \label{ex:275}\ \emph{wima-kütu-ye-nge-y} \\ \texttt{-NN.wima\_vara+THR.kütu35\\-TV.ye\_traer\_llevar+PASS.nge23+IND.y4+3.Ø3}
\end{example}

We have tagged \texttt{THR} the new suffix, from 'through', that comes from the idea of 'all along'. An we have assigned it to slot 35, which seems to be a suitable position for this suffix, of course all of this is tentative.

\paragraph{witra-n-püra-may-a-n \emph{'That raise it up'}} \label{tp:120} [Zúñiga, F. 2006: 284 (\emph{Ngillañmawün})] \cite{RefB:24}. The division we propose for this verb would give a meaning like 'I rose it up in assent' or ' I rose up my assenting X'. This way, the stem is complex and formed by three roots, a not allowed analysis in our system. The analysis would be:

\begin{example} \label{ex:276}\ \emph{witra-n-püra-may-a-n} \\ \texttt{-TV.witra\_levantar+PVN.n4-IV.püra\_subir\\-TV.may\_asentir-CR.TV+NRLD.a9+IND1SG.n3}
\end{example}

\section{Public user interface} \label{sec:80}

\paragraph{} \label{tp:121} In this section we briefly present the exploitation interface we have developed for open access to our analyser.

The URL to access it is:\\ \href{http://www.chandia.net/dungupeyem}{http://www.chandia.net/dungupeyem}

\begin{figure}[H]
	\includegraphics[width=\columnwidth]{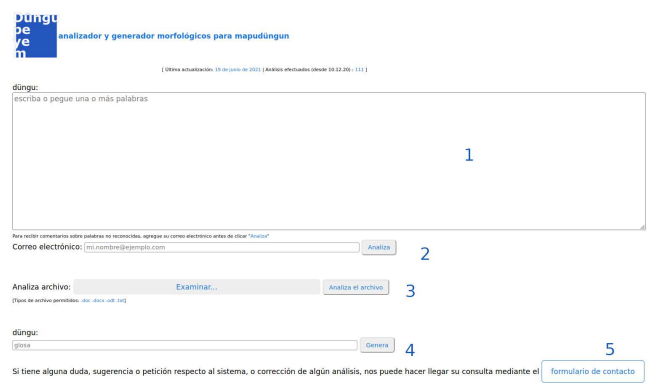}
	\caption{Analyser public web interface.}
	\label{fig:07}
\end{figure}

\begin{enumerate} \label{it:56}
	\item[] Numbers in figure \ref{fig:07} mark:
	\item Text box to paste or type \emph{Mapuche} words to be analysed.
	\item An e-mail field to add the user address in order to receive comments about unknown words. And the "analyse" button to submit the text.
	\item A field that allows to upload a .doc, .docx, .odt or .txt file to be analysed.
	\item A text field to input analyse glosses in order to generate \emph{Mapuche} words.
	\item A link to a contact form in case the user needs some feedback from us.
\end{enumerate}

\begin{figure}[H]
	\includegraphics[width=\columnwidth]{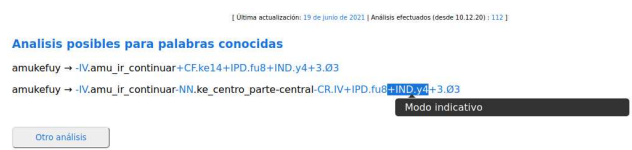}
	\caption{Analysis results on screen.}
	\label{fig:08}
\end{figure}

Figure \ref{fig:08} shows the analysis results for the word \emph{amu-ke-fu-y} 'he used to go', where all tags are in blue and show the tag name when hovered.

When the user uploads a file to be analysed, the system gives a link back to download a text file containing all the analyses.

The user can review information about the tags on a complementary page called "Glosas del \emph{Düngupeyem}". There is also a search block available where tags and suffixes may be queried.

Whenever a user submits an unknown word to the system, or uploads a file to analyse, we receive an e-mail with this information to check if there were some issues, and improve the system if necessary.

\section{Conclusions} \label{sec:81}

\paragraph{} \label{tp:122} The FST analyser has been developed and it has proved to be a reliable tool, of course it is improvable, and it can extend its use to other tools, something that we have already started to do, but not explained in this article, which is devoted to the analyser implementation, specifically in the \emph{Mapuche} verb treatment.

Throughout this article, the process and results have been described in detail to explain the quality, scale and precision of the system. 
We continue working on tools derived from the analyser, which is the basic system for current and future work on automatic processing of \emph{Mapudüngun}.

\begin{acknowledgements}
We deeply and sincerely thank Iñaki Alegria who have guided us along the confection of this article. He has spent countless hours of revision to improve and present this work in a clear, concise and understandable manner.
\end{acknowledgements}

\section{Annexes}

\subsection{Tags meaning} \label{anx:01}
\begin{itemize} \label{it:57}
	\item[]Parts of Speech (and other elements)
	\item[] \texttt{-AJ}: Adjective
	\item[] \texttt{-AV}: Adverb
	\item[] \texttt{-AP}: Anaphoric pronoun
	\item[] \texttt{-CJ}: Conjunction
	\item[] \texttt{-COLL}: Collectivizer
	\item[] \texttt{-CR.IV}: Intransitive verb compound
	\item[] \texttt{-CR.TV}: Transitive verb compound
	\item[] \texttt{-DP}: Demonstrative pronoun
	\item[] \texttt{-FE}: Foreign expression
	\item[] \texttt{-IJ}: Interjection
	\item[] \texttt{-IP}: Interrogative pronoun
	\item[] \texttt{-IV}: Intransitive verb
	\item[] \texttt{-LOC}: Locative
	\item[] \texttt{-NN}: Noun
	\item[] \texttt{-NBR}: Digit (number)
	\item[] \texttt{-NG}: Negation particle
	\item[] \texttt{-NU}: Numeral
	\item[] \texttt{-ON}: Onomatopoeia
	\item[] \texttt{-PCT}: Punctuation mark
	\item[] \texttt{-PN}: Proper name
	\item[] \texttt{-PP}: Personal pronoun
	\item[] \texttt{-PPN}: Possible proper noun
	\item[] \texttt{-PR}: Preposition
	\item[] \texttt{-PT}: Particle
	\item[] \texttt{-QC}: Interrogative \emph{chum}
	\item[] \texttt{-QT}: Interrogative \emph{tunte}
	\item[] \texttt{-RNNR}: Reduplicated noun
	\item[] \texttt{-RONR}: Reduplicated onomatopoeia
	\item[] \texttt{-RVBR}: Reduplicated verb
	\item[] \texttt{-SP}: Possessive pronoun
	\item[] \texttt{-TV}: Transitive verb
	\item[] \texttt{-UNK}: Unknown root or word
	\item[] \texttt{-XV}: Auxiliary
\end{itemize}

\begin{itemize} \label{it:58}
	\item[]Suffixes
	\item[] \texttt{+1}: 1\textsuperscript{st} person
	\item[] \texttt{+1A}: 1\textsuperscript{st} person agent
	\item[] \texttt{+2}: 2\textsuperscript{nd} person
	\item[] \texttt{+2A}: 2\textsuperscript{nd} person agent
	\item[] \texttt{+3}: 3\textsuperscript{rd} person
	\item[] \texttt{+ADJ}: Adjectiviser
	\item[] \texttt{+ADJDO}: Adjectiviser doable
	\item[] \texttt{+ADJQE}: Adjectiviser quick and easy
	\item[] \texttt{+ADV}: Adverbializer
	\item[] \texttt{+AFF}: Affirmative
	\item[] \texttt{+AIML}: Aimless/involuntary
	\item[] \texttt{+AVN}: Agentive verbal noun
	\item[] \texttt{+BEN}: Benefactive
	\item[] \texttt{+CA}: Causative
	\item[] \texttt{+CF}: Constant feature
	\item[] \texttt{+CIRC}: Circular movement
	\item[] \texttt{+CND}: Conditional
	\item[] \texttt{+CNI}: Conditional in imperative
	\item[] \texttt{+CONT}: Continuative
	\item[] \texttt{+CSVN}: Completive subjective verbal noun
	\item[] \texttt{+DL}: Dual
	\item[] \texttt{+DS12A}: Dative subject, 1\textsuperscript{st} or 2\textsuperscript{nd} person agent
	\item[] \texttt{+DS3A}: Dative subject, 3\textsuperscript{rd} person agent
	\item[] \texttt{+DISTR}: Distributive
	\item[] \texttt{+EDO}: External direct object
	\item[] \texttt{+EX}: Discontinuative (ex)
	\item[] \texttt{+EXP}: Experience
	\item[] \texttt{+FAC}: Factitive
	\item[] \texttt{+FORCE}: Force
	\item[] \texttt{+GR}: Group(alizer)
	\item[] \texttt{+HH}: Hither
	\item[] \texttt{+IDO}: Internal direct object
	\item[] \texttt{+IMM}: Immediate
	\item[] \texttt{+IMP}: Imperative
	\item[] \texttt{+IMP1SG}: Imperative, 1\textsuperscript{st} person, singular
	\item[] \texttt{+IMP2SG}: Imperative, 2\textsuperscript{nd} person, singular
	\item[] \texttt{+IMP3}: Imperative, 3\textsuperscript{rd} person
	\item[] \texttt{+IND}: Indicative
	\item[] \texttt{+IND1SG}: Indicative, 1\textsuperscript{st} person, singular
	\item[] \texttt{+INST}: Instrumental object
	\item[] \texttt{+INT}: Intensive
	\item[] \texttt{+IO}: Indirect object
	\item[] \texttt{+IPD}: Impeditive
	\item[] \texttt{+ITR}: Interruptive
	\item[] \texttt{+IVN}: Instrumental verbal noun
	\item[] \texttt{+LOC}: Locative
	\item[] \texttt{+MIO}: More involved object
	\item[] \texttt{+NEG}: Negation
	\item[] \texttt{+NOM}: nominalizer
	\item[] \texttt{+NOMAG}: nominalizer agent
	\item[] \texttt{+NOMPI}: nominalizer place or instrument
	\item[] \texttt{+NRLD}: Non-realized
	\item[] \texttt{+OO}: Oblique object
	\item[] \texttt{+OVN}: Objective verbal noun
	\item[] \texttt{+PASS}: Passive
	\item[] \texttt{+PFPS}: Perfect persistent
	\item[] \texttt{+PL}: Plural
	\item[] \texttt{+PLAY}: Play
	\item[] \texttt{+PLPF}: Pluperfect
	\item[] \texttt{+PLR}: Pluralizer
	\item[] \texttt{+PRPS}: Progressive persistent
	\item[] \texttt{+PR}: Progressive
	\item[] \texttt{+PS}: Persistence
	\item[] \texttt{+PVN}: Plain verbal noun
	\item[] \texttt{+PX}: Proximity
	\item[] \texttt{+RE}: Iterative/restorative
	\item[] \texttt{+REF}: Reflexive/reciprocal
	\item[] \texttt{+REL}: Relative
	\item[] \texttt{+REP}: Reportative
	\item[] \texttt{+SAT}: Satisfaction
	\item[] \texttt{+SFR}: Stem formative
	\item[] \texttt{+SG}: Singular
	\item[] \texttt{+SIM}: Simulative
	\item[] \texttt{+ST}: Stative
	\item[] \texttt{+SUD}: Sudden
	\item[] \texttt{+SVN}: Subjective verbal noun
	\item[] \texttt{+TEMP}: Temporal
	\item[] \texttt{+TH}: Thither
	\item[] \texttt{+TR}: Transitivizer
	\item[] \texttt{+TVN}: Transitive verbal noun
	\item[] \texttt{+VRB}: Verbalizer
\end{itemize}

\subsection{Suffixes by slot} \label{anx:02}

\subsubsection{Slots 1 to 15: inflectional suffixes} \label{anx:03}

\begin{itemize} \label{it:59}
	\item \texttt{slot-01.aff}: Dative Subject
	\item[] \emph{-Ø}: 1\textsuperscript{st} or 2\textsuperscript{nd} person agent "\texttt{+DS12A}"
	\item[] \emph{-ew $\sim$ -mew $\sim$ -mu}: 3\textsuperscript{rd} person agent "\texttt{+DS3A}"
	\item \texttt{slot-02.aff}: Number
	\item[] \emph{-Ø}: Singular (1\textsuperscript{st} \& 2\textsuperscript{nd} indicative) "\texttt{+SG}"
	\item[] \emph{-i}: Singular (2\textsuperscript{nd} indicative \& conditional) "\texttt{+SG}"
	\item[] \emph{-u}: Dual "\texttt{+DL}"
	\item[] \emph{-iñ}: Plural (1\textsuperscript{st} indicative \& conditional) "\texttt{+PL}"
	\item[] \emph{-ün}: Plural (2\textsuperscript{nd} all moods \& 3\textsuperscript{rd} indicative) "\texttt{+PL}"
	\item \texttt{slot-03.aff}: Person
	\item[] \emph{-Ø}: 1\textsuperscript{st} non-singular "\texttt{+1}" \& 3\textsuperscript{rd} "\texttt{+3}" persons
	\item[] \emph{-i}: 1\textsuperscript{st} person "\texttt{+1}"
	\item[] \emph{-m}: 2\textsuperscript{nd} person "\texttt{+2}"
	\item[] \emph{-e}: 3\textsuperscript{rd} person "\texttt{+3}"
	\item[] \emph{-ng}: 3\textsuperscript{rd} person non-singular "\texttt{+3}"
	\item[] \emph{-y}: 1\textsuperscript{st} "\texttt{+1}" \& 3\textsuperscript{rd} "\texttt{+3}" persons agent
	\item \texttt{slot-03PTMT.aff}: Portmanteau [mood, person, number]
	\item[] \emph{-n}: Indicative, 1\textsuperscript{st} person, singular "\texttt{+IND1SG}"
	\item[] \emph{-chi}: Imperative, 1\textsuperscript{st} person, singular "\texttt{+IMP1SG}"
	\item[] \emph{-nge}: Imperative, 2\textsuperscript{nd} person, singular "\texttt{+IMP2SG}"
	\item[] \emph{-pe}: Imperative, 3\textsuperscript{rd} person "\texttt{+IMP3}"
	\item \texttt{slot-04.aff}: Mood
	\item[] \emph{-Ø}: Imperative "\texttt{+IMP}"
	\item[] \emph{-l}: Conditional "\texttt{+CND}" \& "\texttt{+CNI}"
	\item[] \emph{-y}: Indicative "\texttt{+IND}"
	\item \texttt{slot-04NF.aff}: inflectional nominalizers
	\item[] \emph{-Ø}: Objective "\texttt{+OVN}" \& subjective "\texttt{+SVN}" verbal noun "\texttt{+OVN}"
	\item[] \emph{-el}: Objective verbal noun "\texttt{+OVN}"
	\item[] \emph{-fiel}: Transitive verbal noun "\texttt{+TVN}"
	\item[] \emph{-lu}: Subjective verbal noun "\texttt{+SVN}"
	\item[] \emph{-m}: Instrumental verbal noun "\texttt{+IVN}"
	\item[] \emph{-n}: Plain verbal noun "\texttt{+PVN}"
	\item[] \emph{-t}: Agentive verbal noun "\texttt{+AVN}"
	\item[] \emph{-wma}: Completive subjective verbal noun "\texttt{+CSVN}"
	\item \texttt{slot-05.aff}: Constant feature
	\item[] \emph{-ye}: "\texttt{+CF}"
	\item \texttt{slot-06.aff}: Internal \& External direct objects
	\item[] \emph{-e}: Internal direct object "\texttt{+IDO}"
	\item[] \emph{-fi}: External direct object "\texttt{+EDO}"
	\item \texttt{slot-07.aff}: Pluperfect
	\item[] \emph{-mu}: "\texttt{+PLPF}"
	\item \texttt{slot-08.aff}: Impeditive
	\item[] \emph{-fu}: "\texttt{+IPD}"
	\item \texttt{slot-09.aff}: Non-realized situation
	\item[] \emph{-a}: "\texttt{+NRLD}"
	\item \texttt{slot-10.aff}: Negation
	\item[] \emph{-ki}: Imperative "\texttt{+NEG}"
	\item[] \emph{-la}: Indicative "\texttt{+NEG}"
	\item[] \emph{-no}: Conditional "\texttt{+NEG}"
	\item \texttt{slot-11.aff}: Affirmative
	\item[] \emph{-lle}: "\texttt{+AFF}"
	\item \texttt{slot-12.aff}: Reportative
	\item[] \emph{-rke}: "\texttt{+REP}"
	\item \texttt{slot-13.aff}: Proximity
	\item[] \emph{-pe}: "\texttt{+PX}"
	\item \texttt{slot-14.aff}: Constant feature
	\item[] \emph{-ke}: "\texttt{+CF}"
	\item \texttt{slot-15.aff}: Pluperfect
	\item[] \emph{-wye}: "\texttt{+PLPF}"
\end{itemize}

\subsubsection{Slots 16 to 27: mobile derivational suffixes} \label{anx:04}

\begin{itemize} \label{it:60}
	\item \texttt{slot-16.aff}: Repetitive/restorative \& continuative
	\item[] \emph{-ka}: Continuative "\texttt{+CONT}"
	\item[] \emph{-tu}: Repetitive/restorative "\texttt{+RE}"
	\item \texttt{slot-16M.aff}: Repetitive/restorative mobile
	\item[] \emph{-tu}: "\texttt{+RE}"
	\item \texttt{slot-17.aff}: Hither \& locative
	\item[] \emph{-pa}: Hither "\texttt{+HH}"
	\item[] \emph{-pu}: Locative "\texttt{+LOC}"
	\item \texttt{slot-17M.aff}: Hither mobile
	\item[] \emph{-pa} "\texttt{+HH}"
	\item \texttt{slot-18.aff}: Interruptive
	\item[] \emph{-r}: One interruption "\texttt{+ITR}"
	\item[] \emph{-yeku}: Repetead interruptions "\texttt{+ITR}"
	\item \texttt{slot-19.aff}: Persistence
	\item[] \emph{-we} "\texttt{+PS}"
	\item \texttt{slot-20.aff}: Thither
	\item[] \emph{-me}: "\texttt{+TH}"
	\item \texttt{slot-21.aff}: Immediate \& sudden
	\item[] \emph{-fem}: Immediate "\texttt{+IMM}"
	\item[] \emph{-rume}: Sudden "\texttt{+SUD}"
	\item \texttt{slot-22.aff}: Play \& simulation
	\item[] \emph{-faluw}: Simulation "\texttt{+SIM}"
	\item[] \emph{-kantu}: Play "\texttt{+PLAY}"
	\item \texttt{slot-23.aff}: Passive, 1\textsuperscript{st} \& 2\textsuperscript{nd} persons agent
	\item[] \emph{-w}: 1\textsuperscript{st} person agent "\texttt{+1A}"
	\item[] \emph{-mu}: 2\textsuperscript{nd} person agent "\texttt{+2A}"
	\item[] \emph{-nge}: Passive "\texttt{+PASS}"
	\item \texttt{slot-23M.aff}: Passive mobile
	\item[] \emph{-nge}: "\texttt{+PASS}"
	\item \texttt{slot-24.aff}: Pluralizer
	\item[] \emph{-ye}: "\texttt{+PL}"
	\item \texttt{slot-25.aff}: Force \& satisfaction
	\item[] \emph{-fal}: Force "\texttt{+FORCE}"
	\item[] \emph{-ñmu}: Satisfaction "\texttt{+SAT}"
	\item \texttt{slot-25M.aff}: Force mobile
	\item[] \emph{-fal} "\texttt{+FORCE}"
	\item \texttt{slot-26.aff}: Indirect object
	\item[] \emph{-ñma} "\texttt{+IO}"
	\item \texttt{slot-27.aff}: Beneficiary
	\item[] \emph{-el} "\texttt{+BEN}"
\end{itemize}

\subsubsection{Slots 28 to 36: fixed derivational suffixes} \label{anx:05}

\begin{itemize} \label{it:61}
	\item \texttt{slot-28.aff}: Stative \& progressive
	\item[] \emph{-le}: Stative "\texttt{+ST}"
	\item[] \emph{-meke}: Progressive "\texttt{+PR}"
	\item \texttt{slot-28M.aff}: Stative mobile
	\item[] \emph{-le}: "\texttt{+ST}"
	\item \texttt{slot-29.aff}: More involved object
	\item[] \emph{-l}: "\texttt{+MIO}"
	\item \texttt{slot-30.aff}: Circular movement \& intensive
	\item[] \emph{-iaw}: Circular movement "\texttt{+CIRC}"
	\item[] \emph{-tie}: Intensive "\texttt{+INT}"
	\item \texttt{slot-31.aff}: Reflexive/reciprocal
	\item[] \emph{-w}: "\texttt{+REF}"
	\item \texttt{slot-32.aff}: Progressive persistent \& perfect persistent
	\item[] \emph{-künu}: Perfect persistent "\texttt{+PFPS}"
	\item[] \emph{-nie}: Progressive persistent "\texttt{+PRPS}"
	\item \texttt{slot-33.aff}: Trasitivizer \& factitive
	\item[] \emph{-ka}: Factitive "\texttt{+FAC}"
	\item[] \emph{-tu}: Transitivizer "\texttt{+TR}"
	\item \texttt{slot-33M.aff}: Transitivizer mobile
	\item[] \emph{-tu}: "\texttt{+TR}"
	\item \texttt{slot-34.aff}: Causatives
	\item[] \emph{-l}: "\texttt{+CA}"
	\item[] \emph{-m}: "\texttt{+CA}"
	\item \texttt{slot-35.aff}: Experience \& oblique object
	\item[] \emph{-ma}: Experience "\texttt{+EXP}"
	\item[] \emph{-ye}: Oblique object "\texttt{+OO}"
	\item \texttt{slot-36S.aff}: Stem formative
	\item[] \emph{-Ø}: "\texttt{+SFR}"
	\item[] \emph{-nge}: "\texttt{+SFR}"
	\item[] \emph{-tu}: "\texttt{+SFR}"
	\item[] \emph{-ye}: "\texttt{+SFR}"
	\item \texttt{slot-36V.aff}: Verbalizers
	\item[] \emph{-Ø}: "\texttt{+VRB}"
	\item[] \emph{-l}: "\texttt{+VRB}"
	\item[] \emph{-nge}: "\texttt{+VRB}"
	\item[] \emph{-ntu}: "\texttt{+VRB}"
	\item[] \emph{-tu}: "\texttt{+VRB}"
	\item[] \emph{-ye}: "\texttt{+VRB}"
\end{itemize}

\subsubsection{Nominal suffixes} \label{anx:06}

\begin{itemize} \label{it:62}
	\item \texttt{CC.aff}: Class-changing
	\item[] \emph{-chi}: Adjectiviser "\texttt{+ADJ}"
	\item[] \emph{-tu}: Adverbializer "\texttt{+ADV}"
	\item[] \emph{-ñma}: Adverbializer "\texttt{+ADV}"
	\item \texttt{INST.aff}: Instrumental
	\item[] \emph{-mew $\sim$ -mu}: "\texttt{+INST}"
	\item \texttt{NCC.aff}: Non class-changing
	\item[] \emph{-em}: Ex (discontinuative) "\texttt{+EX}"
	\item[] \emph{-ke}: Distributive "\texttt{+DISTR}"
	\item[] \emph{-ntu}: Group "\texttt{+GR}"
	\item[] \emph{-rke}: Reportative "\texttt{+REP}"
	\item[] \emph{-we}: Temporal "\texttt{+TEMP}"
	\item[] \emph{-wen}: Relative "\texttt{+REL}"
	\item \texttt{NOM.aff}: nominalizers
	\item[] \emph{-Ø}: nominalizer "\texttt{+NOM}"
	\item[] \emph{-fal}: Doable "\texttt{+ADJDO}"
	\item[] \emph{-fe}: Agentive "\texttt{+NOMAG}"
	\item[] \emph{-nten}: Quick \& easy "\texttt{+ADJQE}"
	\item[] \emph{-we}: Place or instrument "\texttt{+NOMPI}"
\end{itemize}

\subsubsection{Other suffixes} \label{anx:07}

\begin{itemize} \label{it:63}
	\item \texttt{OS.aff}: Aimless/involuntary
	\item[] \emph{-püda}:  "\texttt{+AIML}"
\end{itemize}

\subsubsection{Examples by suffixes} \label{anx:08}

\paragraph{\emph{Verbalizer} -Ø- \emph{(slot 36):}} \label{tp:123} E\ref{ex:25}, E\ref{ex:26}, E\ref{ex:55}, E\ref{ex:59}, E\ref{ex:61}, E\ref{ex:62}, E\ref{ex:64}, E\ref{ex:69}, E\ref{ex:128}, E\ref{ex:136}, E\ref{ex:141}, E\ref{ex:170}, E\ref{ex:192}, E\ref{ex:200}, E\ref{ex:205}, E\ref{ex:209}, E\ref{ex:225}, E\ref{ex:226}, E\ref{ex:233}, E\ref{ex:237}.

\paragraph{\emph{Verbalizer} -l- \emph{(slot 36):}} \label{tp:124} E\ref{ex:50}, E\ref{ex:65}.

\paragraph{\emph{Verbalizer} -nge- \emph{(slot 36):}} \label{tp:125} E\ref{ex:2}, E\ref{ex:5}, E\ref{ex:23}, E\ref{ex:47}, E\ref{ex:49}, E\ref{ex:52}, E\ref{ex:53}, E\ref{ex:56}, E\ref{ex:57}, E\ref{ex:63}, E\ref{ex:145}, E\ref{ex:193}, E\ref{ex:194}.

\paragraph{\emph{Verbalizer} -ntu- \emph{(slot 36):}} \label{tp:126} E\ref{ex:66}.

\paragraph{\emph{Verbalizer} -tu- \emph{(slot 36):}} \label{tp:127} E\ref{ex:26}, E\ref{ex:30}, E\ref{ex:51}, E\ref{ex:52}, E\ref{ex:53}, E\ref{ex:54}, E\ref{ex:204}.

\paragraph{\emph{Verbalizer} -ye- \emph{(slot 36):}} \label{tp:128} E\ref{ex:60}.

\paragraph{\emph{Stem formative} -Ø- \emph{(slot 36):}} \label{tp:129} E\ref{ex:32}, E\ref{ex:151}.

\paragraph{\emph{Stem formative} -nge- \emph{(slot 36):}} \label{tp:130} E\ref{ex:48}, E\ref{ex:146}.

\paragraph{\emph{Stem formative} -tu- \emph{(slot 36):}} \label{tp:131} E\ref{ex:3}.

\paragraph{\emph{Stem formative} -ye- \emph{(slot 36):}} \label{tp:132} E\ref{ex:58}.

\paragraph{\emph{Causative} -l- \emph{(slot 34):}} \label{tp:133} E\ref{ex:29}, E\ref{ex:90}, E\ref{ex:91}, E\ref{ex:93}, E\ref{ex:131}, E\ref{ex:191}, E\ref{ex:214}, E\ref{ex:220}, E\ref{ex:222}, E\ref{ex:224}, E\ref{ex:226}, E\ref{ex:228}, E\ref{ex:230}.

\paragraph{\emph{Causative} -m- \emph{(slot 34):}} \label{tp:134} E\ref{ex:58}, E\ref{ex:72}, E\ref{ex:142}, E\ref{ex:199}, E\ref{ex:203}, E\ref{ex:208}, E\ref{ex:257}.

\paragraph{\emph{Factitive} -ka- \emph{(slot 33):}} \label{tp:135} E\ref{ex:56}, E\ref{ex:192}, E\ref{ex:193}, E\ref{ex:199}, E\ref{ex:205}, E\ref{ex:219}, E\ref{ex:226}, E\ref{ex:228}, E\ref{ex:230}.

\paragraph{\emph{Trasitivizer} -tu- \emph{(slot 33):}} \label{tp:136} E\ref{ex:41}, E\ref{ex:43}, E\ref{ex:45}, E\ref{ex:87}, E\ref{ex:137}, E\ref{ex:188}, E\ref{ex:191}, E\ref{ex:203}, E\ref{ex:271}, E\ref{ex:272}, E\ref{ex:274}.

\paragraph{\emph{Perfect persistent} -künu- \emph{(slot 32):}} \label{tp:137} E\ref{ex:61}, E\ref{ex:73}, E\ref{ex:85}, E\ref{ex:91}, E\ref{ex:151}, E\ref{ex:189}, E\ref{ex:204}, E\ref{ex:211}.

\paragraph{\emph{Progressive persistent} -nie- \emph{(slot 32):}} \label{tp:138} E\ref{ex:1}, E\ref{ex:121}, E\ref{ex:132}, E\ref{ex:137}, E\ref{ex:141}.

\paragraph{\emph{Reflexive/reciprocal} -w- \emph{(slot 31):}} \label{tp:139} E\ref{ex:55}, E\ref{ex:56}, E\ref{ex:60}, E\ref{ex:85}, E\ref{ex:110}, E\ref{ex:211}, E\ref{ex:219}, E\ref{ex:263}.

\paragraph{\emph{Circular movement} -iaw- \emph{(slot 30):}} \label{tp:140} E\ref{ex:207}, E\ref{ex:216}, E\ref{ex:237}.

\paragraph{\emph{More involved object} -l- \emph{(slot 29):}} \label{tp:141} E\ref{ex:204}, E\ref{ex:205}, E\ref{ex:206}, E\ref{ex:207}.

\paragraph{\emph{Stative} -le- \emph{(slot 28):}} \label{tp:142} E\ref{ex:7}, E\ref{ex:43}, E\ref{ex:55}, E\ref{ex:68}, E\ref{ex:70}, E\ref{ex:71}, E\ref{ex:74}, E\ref{ex:80}, E\ref{ex:82}, E\ref{ex:89}, E\ref{ex:90}, E\ref{ex:110}, E\ref{ex:224}, E\ref{ex:226}, E\ref{ex:227}.

\paragraph{\emph{Beneficiary} -el- \emph{(slot 27):}} \label{tp:143} E\ref{ex:8}, E\ref{ex:207}

\paragraph{\emph{Indirect object} -ñma- \emph{(slot 26):}} \label{tp:144} E\ref{ex:1}, E\ref{ex:230}.

\paragraph{\emph{Force} -fal- \emph{(slot 25):}} \label{tp:145} E\ref{ex:155}, E\ref{ex:220}.

\paragraph{\emph{Passive} -nge- \emph{(slot 23):}} \label{tp:146} E\ref{ex:6}, E\ref{ex:58}, E\ref{ex:72}, E\ref{ex:91}, E\ref{ex:147}, E\ref{ex:158}, E\ref{ex:162}, E\ref{ex:167}, E\ref{ex:172}, E\ref{ex:185}, E\ref{ex:189}, E\ref{ex:205}, E\ref{ex:257}, E\ref{ex:267}, E\ref{ex:271}, E\ref{ex:274}, E\ref{ex:275}.

\paragraph{\emph{2\textsuperscript{nd} person agent} -mu- \emph{(slot 23):}} \label{tp:147} E\ref{ex:8}, E\ref{ex:19}, E\ref{ex:196}, E\ref{ex:272}.

\paragraph{\emph{1\textsuperscript{st} person agent} -w- \emph{(slot 23):}} \label{tp:148} E\ref{ex:195}, E\ref{ex:233}.

\paragraph{\emph{Simulation} -faluw- \emph{(slot 22):}} \label{tp:149} E\ref{ex:7}.

\paragraph{\emph{Play} -kantu- \emph{(slot 22):}} \label{tp:150} E\ref{ex:172}, E\ref{ex:200}.

\paragraph{\emph{Immediate} -fem- \emph{(slot 21):}} \label{tp:151} E\ref{ex:6}.

\paragraph{\emph{Sudden} -rume- \emph{(slot 21):}} \label{tp:152} E\ref{ex:162}.

\paragraph{\emph{Thither} -me- \emph{(slot 20):}} \label{tp:153} E\ref{ex:8}, E\ref{ex:9}, E\ref{ex:10}, E\ref{ex:74}, E\ref{ex:96}, E\ref{ex:97}, E\ref{ex:98}, E\ref{ex:125}, E\ref{ex:140}, E\ref{ex:152}, E\ref{ex:163}, E\ref{ex:173}, E\ref{ex:212}, E\ref{ex:213}, E\ref{ex:214}, E\ref{ex:215}, E\ref{ex:221}, E\ref{ex:222}, E\ref{ex:255}.

\paragraph{\emph{Persistence} -we- \emph{(slot 19):}} \label{tp:154} E\ref{ex:9}, E\ref{ex:74}, E\ref{ex:76}, E\ref{ex:86}, E\ref{ex:128}, E\ref{ex:135}, E\ref{ex:152}, E\ref{ex:153}, E\ref{ex:205}.

\paragraph{\emph{Interruptive (one)} -r- \emph{(slot 18):}} \label{tp:155} E\ref{ex:1}, E\ref{ex:167}.

\paragraph{\emph{Interruptive (repeated)} -yeku- \emph{(slot 18):}} \label{tp:156} E\ref{ex:10}.

\paragraph{\emph{Hither} -pa- \emph{(slot 17):}} \label{tp:157} E\ref{ex:70}, E\ref{ex:93}, E\ref{ex:95}, E\ref{ex:97}, E\ref{ex:98}, E\ref{ex:158}, E\ref{ex:190}, E\ref{ex:202}, E\ref{ex:210}, E\ref{ex:212}, E\ref{ex:216}, E\ref{ex:223}, E\ref{ex:224}, E\ref{ex:273}.

\paragraph{\emph{Locative} -pu- \emph{(slot 17):}} \label{tp:158} E\ref{ex:1}, E\ref{ex:42}, E\ref{ex:77}, E\ref{ex:92}, E\ref{ex:167}, E\ref{ex:209}.

\paragraph{\emph{Continuative} -ka- \emph{(slot 16):}} \label{tp:159} E\ref{ex:25}, E\ref{ex:251}.

\paragraph{\emph{Repetitive/restorative} -tu- \emph{(slot 16):}} \label{tp:160} E\ref{ex:1}, E\ref{ex:10}, E\ref{ex:70}, E\ref{ex:226}, E\ref{ex:242}, E\ref{ex:252}, E\ref{ex:255}, E\ref{ex:257}, E\ref{ex:270}, E\ref{ex:271}, E\ref{ex:272}.

\paragraph{\emph{Pluperfect} -wye- \emph{(slot 15):}} \label{tp:161} E\ref{ex:4}, E\ref{ex:5}.

\paragraph{\emph{Constant feature} -ke- \emph{(slot 14):}} \label{tp:162} E\ref{ex:9}, E\ref{ex:51}, E\ref{ex:72}, E\ref{ex:91}, E\ref{ex:131}, E\ref{ex:134}, E\ref{ex:135}, E\ref{ex:152}, E\ref{ex:163}, E\ref{ex:189}, E\ref{ex:191}, E\ref{ex:206}, E\ref{ex:216}, E\ref{ex:217}, E\ref{ex:228}, E\ref{ex:254}.

\paragraph{\emph{Proximity} -pe- \emph{(slot 13):}} \label{tp:163} E\ref{ex:93}, E\ref{ex:133}, E\ref{ex:267}.

\paragraph{\emph{Reportative} -rke- \emph{(slot 12):}} \label{tp:164} E\ref{ex:5}, E\ref{ex:131}.

\paragraph{\emph{Affirmative} -lle- \emph{(slot 11):}} \label{tp:165} E\ref{ex:59}.

\paragraph{\emph{Negation (imperative)} -ki- \emph{(slot 10):}} \label{tp:166} E\ref{ex:15}, E\ref{ex:17}, E\ref{ex:21}, E\ref{ex:128}, E\ref{ex:150}, E\ref{ex:181}.

\paragraph{\emph{Negation (indicative)} -la- \emph{(slot 10):}} \label{tp:167} E\ref{ex:9}, E\ref{ex:12}, E\ref{ex:26}, E\ref{ex:62}, E\ref{ex:66}, E\ref{ex:74}, E\ref{ex:116}, E\ref{ex:124}, E\ref{ex:127}, E\ref{ex:130}, E\ref{ex:152}, E\ref{ex:153}, E\ref{ex:180}, E\ref{ex:217}, E\ref{ex:220}, E\ref{ex:243}, E\ref{ex:277}.

\paragraph{\emph{Negation (conditional)} -no- \emph{(slot 10):}} \label{tp:168} E\ref{ex:13}, E\ref{ex:27}, E\ref{ex:78}, E\ref{ex:119}, E\ref{ex:135}, E\ref{ex:185}, E\ref{ex:205}, E\ref{ex:218}.

\paragraph{\emph{Non-realized situation} -a- \emph{(slot 9):}} \label{tp:169} E\ref{ex:28}, E\ref{ex:46}, E\ref{ex:59}, E\ref{ex:61}, E\ref{ex:73}, E\ref{ex:77}, E\ref{ex:79}, E\ref{ex:122}, E\ref{ex:125}, E\ref{ex:126}, E\ref{ex:141}, E\ref{ex:148}, E\ref{ex:160}, E\ref{ex:164}, E\ref{ex:166}, E\ref{ex:169}, E\ref{ex:174}, E\ref{ex:185}, E\ref{ex:186}, E\ref{ex:196}, E\ref{ex:205}, E\ref{ex:239}, E\ref{ex:241}, E\ref{ex:242}, E\ref{ex:255}, E\ref{ex:257}, E\ref{ex:276}, E\ref{ex:277}.

\paragraph{\emph{Impeditive} -fu- \emph{(slot 8):}} \label{tp:170} E\ref{ex:3}, E\ref{ex:91}, E\ref{ex:129}, E\ref{ex:130}, E\ref{ex:131}, E\ref{ex:135}, E\ref{ex:136}, E\ref{ex:189}, E\ref{ex:191}, E\ref{ex:215}, E\ref{ex:217}, E\ref{ex:257}.

\paragraph{\emph{Pluperfect} -mu- \emph{(slot 7):}} \label{tp:171} E\ref{ex:24}, E\ref{ex:175}.

\paragraph{\emph{Internal direct object} -e- \emph{(slot 6):}} \label{tp:172} E\ref{ex:1}, E\ref{ex:18}, E\ref{ex:22}, E\ref{ex:45}, E\ref{ex:62}, E\ref{ex:65}, E\ref{ex:101}, E\ref{ex:121}, E\ref{ex:124}, E\ref{ex:126}, E\ref{ex:127}, E\ref{ex:128}, E\ref{ex:131}, E\ref{ex:138}, E\ref{ex:139}, E\ref{ex:140}, E\ref{ex:141}, E\ref{ex:168}, E\ref{ex:197}, E\ref{ex:231}, E\ref{ex:232}, E\ref{ex:234}, E\ref{ex:235}, E\ref{ex:241}, E\ref{ex:242}, E\ref{ex:243}.

\paragraph{\emph{External direct object} -fi- \emph{(slot 6):}} \label{tp:173} E\ref{ex:20}, E\ref{ex:21}, E\ref{ex:41}, E\ref{ex:50}, E\ref{ex:73}, E\ref{ex:98}, E\ref{ex:129}, E\ref{ex:130}, E\ref{ex:137}, E\ref{ex:150}, E\ref{ex:151}, E\ref{ex:154}, E\ref{ex:156}, E\ref{ex:165}, E\ref{ex:174}, E\ref{ex:176}, E\ref{ex:178}, E\ref{ex:181}, E\ref{ex:198}, E\ref{ex:204}, E\ref{ex:208}, E\ref{ex:214}, E\ref{ex:230}.

\paragraph{\emph{Constant feature} -ye- \emph{(slot 5):}} \label{tp:174} E\ref{ex:162}.

\paragraph{\emph{Imperative} -Ø- \emph{(slot 4):}} \label{tp:1xx} E\ref{ex:14}, E\ref{ex:20}.

\paragraph{\emph{Conditional} -l- \emph{(slot 4):}} \label{tp:175} E\ref{ex:13}, E\ref{ex:19}, E\ref{ex:59}, E\ref{ex:77}, E\ref{ex:112}, E\ref{ex:113}, E\ref{ex:118}, E\ref{ex:119}.

\paragraph{\emph{Conditional (in imperatives)} -l- \emph{(slot 4):}} \label{tp:176} E\ref{ex:17}, E\ref{ex:21}, E\ref{ex:128}, E\ref{ex:150}.

\paragraph{\emph{Indicative} -y- \emph{(slot 4):}} \label{tp:177} E\ref{ex:1}, E\ref{ex:4}, E\ref{ex:5}, E\ref{ex:6}, E\ref{ex:8}, E\ref{ex:9}, E\ref{ex:10}, E\ref{ex:11}, E\ref{ex:12}, E\ref{ex:16}, E\ref{ex:18}, E\ref{ex:26}, E\ref{ex:28}, E\ref{ex:43}, E\ref{ex:44}, E\ref{ex:47}, E\ref{ex:48}, E\ref{ex:49}, E\ref{ex:50}, E\ref{ex:51}, E\ref{ex:52}, E\ref{ex:53}, E\ref{ex:54}, E\ref{ex:55}, E\ref{ex:56}, E\ref{ex:58}, E\ref{ex:60}, E\ref{ex:65}, E\ref{ex:66}, E\ref{ex:67}, E\ref{ex:68}, E\ref{ex:69}, E\ref{ex:79}, E\ref{ex:91}, E\ref{ex:94}, E\ref{ex:99}, E\ref{ex:110}, E\ref{ex:114}, E\ref{ex:115}, E\ref{ex:116}, E\ref{ex:121}, E\ref{ex:124}, E\ref{ex:131}, E\ref{ex:138}, E\ref{ex:145}, E\ref{ex:146}, E\ref{ex:147}, E\ref{ex:152}, E\ref{ex:156}, E\ref{ex:157}, E\ref{ex:158}, E\ref{ex:160}, E\ref{ex:161}, E\ref{ex:163}, E\ref{ex:167}, E\ref{ex:168}, E\ref{ex:169}, E\ref{ex:170}, E\ref{ex:171}, E\ref{ex:172}, E\ref{ex:178}, E\ref{ex:180}, E\ref{ex:186}, E\ref{ex:189}, E\ref{ex:191}, E\ref{ex:193}, E\ref{ex:211}, E\ref{ex:216}, E\ref{ex:217}, E\ref{ex:220}, E\ref{ex:221}, E\ref{ex:222}, E\ref{ex:224}, E\ref{ex:225}, E\ref{ex:228}, E\ref{ex:229}, E\ref{ex:231}, E\ref{ex:232}, E\ref{ex:233}, E\ref{ex:234}, E\ref{ex:235}, E\ref{ex:236}, E\ref{ex:237}, E\ref{ex:238}, E\ref{ex:241}, E\ref{ex:254}, E\ref{ex:257}, E\ref{ex:270}, E\ref{ex:271}, E\ref{ex:272}, E\ref{ex:273}, E\ref{ex:274}, E\ref{ex:275}, E\ref{ex:277}.

\paragraph{\emph{Subjective verbal noun} -Ø- \emph{(slot 4):}} \label{tp:178} E\ref{ex:33}, E\ref{ex:55}, E\ref{ex:72}, E\ref{ex:90}.

\paragraph{\emph{Objective verbal noun} -el- \emph{(slot 4):}} \label{tp:179} E\ref{ex:25}, E\ref{ex:46}, E\ref{ex:61}, E\ref{ex:77}, E\ref{ex:92}, E\ref{ex:125}, E\ref{ex:132}, E\ref{ex:133}, E\ref{ex:134}, E\ref{ex:135}, E\ref{ex:142}, E\ref{ex:143}, E\ref{ex:144}, E\ref{ex:148}, E\ref{ex:149}, E\ref{ex:164}, E\ref{ex:185}, E\ref{ex:209}.

\paragraph{\emph{Transitive verbal noun} -fiel- \emph{(slot 4):}} \label{tp:180} E\ref{ex:28}, E\ref{ex:196}, E\ref{ex:206}, E\ref{ex:207}, E\ref{ex:255}.

\paragraph{\emph{Subjective verbal noun} -lu- \emph{(slot 4):}} \label{tp:181} E\ref{ex:27}, E\ref{ex:93}, E\ref{ex:141}, E\ref{ex:182}, E\ref{ex:188}, E\ref{ex:190}, E\ref{ex:195}, E\ref{ex:198}.

\paragraph{\emph{Instrumental verbal noun} -m- \emph{(slot 4):}} \label{tp:182} E\ref{ex:24}, E\ref{ex:162}, E\ref{ex:175}, E\ref{ex:205}.

\paragraph{\emph{Plain verbal noun} -n- \emph{(slot 4):}} \label{tp:183} E\ref{ex:26}, E\ref{ex:34}, E\ref{ex:43}, E\ref{ex:51}, E\ref{ex:52}, E\ref{ex:53}, E\ref{ex:54}, E\ref{ex:56}, E\ref{ex:64}, E\ref{ex:69}, E\ref{ex:75}, E\ref{ex:76}, E\ref{ex:80}, E\ref{ex:81}, E\ref{ex:86}, E\ref{ex:87}, E\ref{ex:88}, E\ref{ex:89}, E\ref{ex:111}, E\ref{ex:179}, E\ref{ex:183}, E\ref{ex:187}, E\ref{ex:192}, E\ref{ex:194}, E\ref{ex:210}, E\ref{ex:218}, E\ref{ex:219}, E\ref{ex:223}, E\ref{ex:256}, E\ref{ex:263}, E\ref{ex:267}, E\ref{ex:268}, E\ref{ex:276}.

\paragraph{\emph{Agentive verbal noun} -t- \emph{(slot 4):}} \label{tp:184} E\ref{ex:22}, E\ref{ex:197}.

\paragraph{\emph{Completive subjective verbal noun} -wma- \emph{(slot 4):}} \label{tp:185}  E\ref{ex:23}, E\ref{ex:57}, E\ref{ex:63}, E\ref{ex:71}.

\paragraph{\emph{1\textsuperscript{st} person non-singular} -Ø- \emph{(slot 3):}} \label{tp:186} E\ref{ex:1}, E\ref{ex:8}, E\ref{ex:16}, E\ref{ex:18}, E\ref{ex:94}, E\ref{ex:110}, E\ref{ex:114}, E\ref{ex:115}, E\ref{ex:121}, E\ref{ex:124}, E\ref{ex:138}, E\ref{ex:156}, E\ref{ex:160}, E\ref{ex:178}, E\ref{ex:224}, E\ref{ex:231}, E\ref{ex:233}, E\ref{ex:239}, E\ref{ex:241}, E\ref{ex:272}.

\paragraph{\emph{3\textsuperscript{rd} person non-singular} -Ø- \emph{(slot 3):}} \label{tp:187} E\ref{ex:4}, E\ref{ex:5}, E\ref{ex:6}, E\ref{ex:9}, E\ref{ex:26}, E\ref{ex:28}, E\ref{ex:43}, E\ref{ex:47}, E\ref{ex:48}, E\ref{ex:49}, E\ref{ex:50}, E\ref{ex:51}, E\ref{ex:53}, E\ref{ex:55}, E\ref{ex:56}, E\ref{ex:58}, E\ref{ex:65}, E\ref{ex:66}, E\ref{ex:67}, E\ref{ex:68}, E\ref{ex:69}, E\ref{ex:79}, E\ref{ex:91}, E\ref{ex:99}, E\ref{ex:116}, E\ref{ex:131}, E\ref{ex:145}, E\ref{ex:146}, E\ref{ex:147}, E\ref{ex:152}, E\ref{ex:157}, E\ref{ex:158}, E\ref{ex:161}, E\ref{ex:163}, E\ref{ex:167}, E\ref{ex:168}, E\ref{ex:169}, E\ref{ex:170}, E\ref{ex:171}, E\ref{ex:172}, E\ref{ex:180}, E\ref{ex:186}, E\ref{ex:189}, E\ref{ex:191}, E\ref{ex:193}, E\ref{ex:211}, E\ref{ex:217}, E\ref{ex:220}, E\ref{ex:221}, E\ref{ex:222}, E\ref{ex:225}, E\ref{ex:228}, E\ref{ex:229}, E\ref{ex:257}, E\ref{ex:270}, E\ref{ex:271}, E\ref{ex:273}, E\ref{ex:274}, E\ref{ex:275}.

\paragraph{\emph{Imperative 1\textsuperscript{st} person singular} -chi- \emph{(slot 3):}} \label{tp:188} E\ref{ex:15}.

\paragraph{\emph{1\textsuperscript{st} person} -i- \emph{(slot 3):}} \label{tp:189} E\ref{ex:19}, E\ref{ex:118}, E\ref{ex:119}, E\ref{ex:128}.

\paragraph{\emph{3\textsuperscript{rd} person} -e- \emph{(slot 3):}} \label{tp:190} E\ref{ex:59}, E\ref{ex:113}.

\paragraph{\emph{Indicative 1\textsuperscript{st} person singular} -n- \emph{(slot 3):}} \label{tp:191} E\ref{ex:2}, E\ref{ex:3}, E\ref{ex:7}, E\ref{ex:41}, E\ref{ex:42}, E\ref{ex:45}, E\ref{ex:62}, E\ref{ex:70}, E\ref{ex:72}, E\ref{ex:73}, E\ref{ex:74}, E\ref{ex:82}, E\ref{ex:83}, E\ref{ex:84}, E\ref{ex:85}, E\ref{ex:95}, E\ref{ex:96}, E\ref{ex:97}, E\ref{ex:98}, E\ref{ex:101}, E\ref{ex:102}, E\ref{ex:122}, E\ref{ex:126}, E\ref{ex:127}, E\ref{ex:129}, E\ref{ex:130}, E\ref{ex:136}, E\ref{ex:137}, E\ref{ex:139}, E\ref{ex:140}, E\ref{ex:151}, E\ref{ex:153}, E\ref{ex:154}, E\ref{ex:155}, E\ref{ex:159}, E\ref{ex:165}, E\ref{ex:166}, E\ref{ex:173}, E\ref{ex:174}, E\ref{ex:176}, E\ref{ex:177}, E\ref{ex:184}, E\ref{ex:204}, E\ref{ex:208}, E\ref{ex:212}, E\ref{ex:213}, E\ref{ex:214}, E\ref{ex:215}, E\ref{ex:226}, E\ref{ex:227}, E\ref{ex:230}, E\ref{ex:242}, E\ref{ex:243}, E\ref{ex:251}, E\ref{ex:252}, E\ref{ex:253}, E\ref{ex:267}, E\ref{ex:276}.

\paragraph{\emph{3\textsuperscript{rd} person non-singular} -ng- \emph{(slot 3):}} \label{tp:192} E\ref{ex:10}, E\ref{ex:44}, E\ref{ex:54}, E\ref{ex:254}.

\paragraph{\emph{Imperative 2\textsuperscript{nd} person singular} -nge- \emph{(slot 3):}} \label{tp:193} E\ref{ex:150}, E\ref{ex:181}.

\paragraph{\emph{2\textsuperscript{nd} person} -m- \emph{(slot 3):}} \label{tp:194} E\ref{ex:11}, E\ref{ex:12}, E\ref{ex:13}, E\ref{ex:14}, E\ref{ex:20}, E\ref{ex:52}, E\ref{ex:60}, E\ref{ex:78}, E\ref{ex:216}, E\ref{ex:232}, E\ref{ex:234}, E\ref{ex:235}, E\ref{ex:236}, E\ref{ex:237}, E\ref{ex:238}, E\ref{ex:240}, E\ref{ex:277}.

\paragraph{\emph{1\textsuperscript{st} persons agent} -y- \emph{(slot 3):}} \label{tp:195} E\ref{ex:17}, E\ref{ex:21}.

\paragraph{\emph{Singular} -Ø- \emph{(slot 2):}} \label{tp:196} E\ref{ex:19}, E\ref{ex:119}, E\ref{ex:128}.

\paragraph{\emph{Singular} -i- \emph{(slot 2):}} \label{tp:197} E\ref{ex:11}, E\ref{ex:52}, E\ref{ex:78}, E\ref{ex:216}, E\ref{ex:232}, E\ref{ex:236}, E\ref{ex:237}, E\ref{ex:238}, E\ref{ex:240}, E\ref{ex:277}.

\paragraph{\emph{Plural} -iñ- \emph{(slot 2):}} \label{tp:198} E\ref{ex:1}, E\ref{ex:8}, E\ref{ex:21}, E\ref{ex:94}, E\ref{ex:110}, E\ref{ex:115}, E\ref{ex:118}, E\ref{ex:156}, E\ref{ex:224}, E\ref{ex:233}, E\ref{ex:239}.

\paragraph{\emph{Dual} -u- \emph{(slot 2):}} \label{tp:199} E\ref{ex:12}, E\ref{ex:14}, E\ref{ex:16}, E\ref{ex:17}, E\ref{ex:18}, E\ref{ex:20}, E\ref{ex:60}, E\ref{ex:112}, E\ref{ex:114}, E\ref{ex:121}, E\ref{ex:124}, E\ref{ex:138}, E\ref{ex:178}, E\ref{ex:231}, E\ref{ex:234}, E\ref{ex:241}, E\ref{ex:272}.

\paragraph{\emph{Plural} -ün- \emph{(slot 2):}} \label{tp:200} E\ref{ex:10}, E\ref{ex:13}, E\ref{ex:44}, E\ref{ex:54}, E\ref{ex:235}, E\ref{ex:254}.

\paragraph{\emph{Dative subject 1\textsuperscript{st} or 2\textsuperscript{nd} persons agent} -Ø- \emph{(slot 1):}} \label{tp:201} E\ref{ex:18}, E\ref{ex:101}, E\ref{ex:121}, E\ref{ex:124}, E\ref{ex:126}, E\ref{ex:128}, E\ref{ex:138}, E\ref{ex:139}, E\ref{ex:168}, E\ref{ex:231}, E\ref{ex:232}, E\ref{ex:234}, E\ref{ex:235}, E\ref{ex:241}, E\ref{ex:242}, E\ref{ex:243}.

\paragraph{\emph{Dative subject 3\textsuperscript{rd} person agent} -mew- \emph{(slot 1):}} \label{tp:202} E\ref{ex:1}, E\ref{ex:22}, E\ref{ex:45}, E\ref{ex:62}, E\ref{ex:65}, E\ref{ex:127}, E\ref{ex:131}, E\ref{ex:140}, E\ref{ex:141}, E\ref{ex:197}.

\paragraph{\emph{nominalizer} -Ø-:} \label{tp:203} E\ref{ex:199}, E\ref{ex:200}, E\ref{ex:201}, E\ref{ex:202}.

\paragraph{\emph{Adjecivizer} -chi-:} \label{tp:204} E\ref{ex:33}, E\ref{ex:55}, E\ref{ex:72}, E\ref{ex:90}, E\ref{ex:134}, E\ref{ex:142}.

\paragraph{\emph{Ex (discontinuative)} -em-:} \label{tp:205} E\ref{ex:36}.

\paragraph{\emph{Doable} -fal-:} \label{tp:206}  E\ref{ex:29}, E\ref{ex:170}.

\paragraph{\emph{Agentive} -fe-:} \label{tp:207} E\ref{ex:30}, E\ref{ex:53}, E\ref{ex:193}.

\paragraph{\emph{Distributive} -ke-:} \label{tp:208} E\ref{ex:35}, E\ref{ex:50}, E\ref{ex:55}.

\paragraph{\emph{instrumental} -mew-:} \label{tp:209} E\ref{ex:41}, E\ref{ex:42}, E\ref{ex:43}, E\ref{ex:83}, E\ref{ex:84}, E\ref{ex:120}, E\ref{ex:143}, E\ref{ex:148}, E\ref{ex:206}, E\ref{ex:221}, E\ref{ex:251}, E\ref{ex:252}, E\ref{ex:253}, E\ref{ex:254}.

\paragraph{\emph{Quick \& easy} -nten-:} \label{tp:210} E\ref{ex:31}, E\ref{ex:57}.

\paragraph{\emph{Group} -ntu-:} \label{tp:211} E\ref{ex:37}, E\ref{ex:112}.

\paragraph{\emph{Aimless/involuntary} -püda-:} \label{tp:212} E\ref{ex:154}.

\paragraph{\emph{Reportative} -rke-:} \label{tp:213} E\ref{ex:23}, E\ref{ex:38}.

\paragraph{\emph{Adverbializer} -tu-:} \label{tp:214} E\ref{ex:34}, E\ref{ex:64}.

\paragraph{\emph{Place or instrument} -we-:} \label{tp:215} E\ref{ex:32}, E\ref{ex:87}, E\ref{ex:203}.

\paragraph{\emph{Temporal} -we-:} \label{tp:216} E\ref{ex:39}.

\paragraph{\emph{Relative} -wen-:} \label{tp:217} E\ref{ex:40}, E\ref{ex:194}. \\\\

\subsubsection{Intermediate language symbols} \label{anx:09}

\paragraph{\emph{\texttt{"@1", "@2", "@3"}}}. Provisional contextual marks to treat changes derived from the interaction between impeditive \emph{-fu-} (slot 8) and internal direct object \emph{-e-}, external direct object \emph{-fi-} (slot 6) or objective verbal noun \emph{-el-} (slot 4); see R\ref{R:06}, R\ref{R:07} and figure \ref{fig:06}.

\paragraph{\emph{\texttt{"@4", "@5", "@6"}}}. Provisional contextual marks to treat radical consonant alternation of some intransitive verbs when interacting with causative \emph{-üm-}, which transitivizes them. \texttt{"@4"} operates in the insertion of \emph{ng}; \texttt{"@5"} operates in the interchange between \emph{f} and \emph{p}; \texttt{"@6"} operates in the interchange between \emph{g} and \emph{k}; see \ref{sec:43} and R\ref{R:15}.

\paragraph{\emph{\texttt{"@E0"}}} treats the elision of the final \emph{e} of progressive persistent \emph{-nie-} (slot 32) when followed by \emph{a}; see \ref{it:37}.

\paragraph{\emph{\texttt{"@ED"}}} stands for the external direct object \emph{-fi-} (slot 6) while processing its interaction with impeditive \emph{-fu-} (slot 8) or objective verbal noun \emph{-el-} (slot 4); see D\ref{def:12}, R\ref{R:06}, R\ref{R:09}, figure \ref{fig:06} and 1\textsuperscript{st} item on this list. 

\paragraph{\emph{\texttt{"@EI"}}} treats the change of \emph{e} into \emph{i} of constant feature \emph{-ke-} (slot 14) when followed by \emph{ü} or \emph{y}; see \ref{it:37}.

\paragraph{\emph{\texttt{"@EL"}}} stands for objective verbal noun \emph{-el-} (slot 4) while processing its interaction with impeditive \emph{-fu-} (slot 8), internal direct object \emph{-e-} or external direct object \emph{-fi-} (slot 6), depending on that, it could take the form \emph{-el-} or \emph{-l-}; see D\ref{def:13}, R\ref{R:07}, R\ref{R:11}, R\ref{R:13}, figure \ref{fig:06} and 1\textsuperscript{st} item on this list.

\paragraph{\emph{\texttt{"@EY"}}} treats the optional realization as \emph{-ngi-} of \emph{-nge-} forms; see \ref{sec:40}, R\ref{R:12}, D\ref{def:15}, D\ref{def:23}.

\paragraph{\emph{\texttt{"@FP"}}} regulates the alternation of the radical consonant \emph{f} into \emph{p} of intransitive verbs when transitivized by causative \emph{-üm-} (slot 34); see \ref{sec:43}, D\ref{def:18}, R\ref{R:15} and 2\textsuperscript{nd} item on this list.

\paragraph{\emph{\texttt{"@G"}}} treats the optional epenthesis of a glottal stop between vowels belonging to different morphemes, usually roots in compounds; see \ref{sec:59}, D\ref{def:02}, E\ref{ex:117}, R\ref{R:01}, R\ref{R:13}, R\ref{R:14}, D\ref{def:18}, \ref{it:37}, D\ref{def:31}, D\ref{def:34}.

\paragraph{\emph{\texttt{"@GK"}}} regulates the alternation of the radical consonant \emph{g} into \emph{k} of intransitive verbs when transitivized by causative \emph{-üm-} (slot 34); see \ref{sec:43}, D\ref{def:18}, R\ref{R:15}.

\paragraph{\emph{\texttt{"@i"}}} is used to regulate different realizations of the verb \emph{i-} 'to eat', which depends on the morphophonological context where it appears; see \ref{sec:41}, D\ref{def:16}, R\ref{R:13}.

\paragraph{\emph{\texttt{"@ID"}}} stands for the internal direct object \emph{-e-} (slot 6) while processing its interaction with negation for imperative \emph{-ki-} (slot 10), impeditive \emph{-fu-} (slot 8) or objective verbal noun \emph{-el-} (slot 4); see D\ref{def:12}, R\ref{R:05}, R\ref{R:07}, R\ref{R:10}, R\ref{R:11}, R\ref{R:13}, D\ref{def:31}, figure \ref{fig:06} and 1\textsuperscript{st} item on this list.

\paragraph{\emph{\texttt{"@IP"}}} stands for impeditive \emph{-fu-} (slot 8) while processing its interaction with internal direct object \emph{-e-}, external direct object \emph{-fi-} (slot 6) or objective verbal noun \emph{-el-} (slot 4); see D\ref{def:11}, R\ref{R:06}, R\ref{R:07}, R\ref{R:08}, R\ref{R:09}, figure \ref{fig:06} and 1\textsuperscript{st} item on this list.

\paragraph{\emph{\texttt{"@IZ"}}} treats the indicative (slot 4) alternative realization \emph{-i-}, see \ref{sec:58}, D\ref{def:30}, R\ref{R:51}.

\paragraph{\emph{\texttt{"@K"}}} treats the insertion of \emph{kü} in stative \emph{-le-} (slot 27) giving \emph{-küle-}, mandatory when preceded by semi-vowel or any consonant but \emph{m}, optional when preceded by \emph{m} or \emph{u}.

\paragraph{\emph{\texttt{"@KG"}}} enables the optional initial \emph{g} of perfect persistent \emph{-künu-} (slot 32) when preceded by \emph{d, f} or \emph{m}.

\paragraph{\emph{\texttt{"@KI"}}} treats the form alternation of circular movement \emph{-kiaw-} (slot 30) after semi-vowel or any consonant but \emph{m}; \emph{-yaw-} after \emph{m} or any vowel.

\paragraph{\emph{\texttt{"@KÜ"}}} (capital \emph{u} with diaereses) treats the insertion of \emph{kü} at the beginning of intensive marker \emph{-tie-} (slot 30) when preceded by \emph{g, i, n, r} or \emph{u}.

\paragraph{\emph{\texttt{"@L"}}} manages the alternation between \emph{-e-} and \emph{-le-} that may be added to causative \emph{-l-} (slot 34) and benefactive \emph{-el-} (slot 27). Both may take \emph{l, el, lel} forms depending on previous phoneme.

\paragraph{\emph{\texttt{"@M"}}}. In \emph{Mapuche} speech, interjections are important because they mark the intention in discourse. \emph{hmm} is widely used but its transcription may bare one or more \emph{m}. We use the next rule to recognize the interjection spite being written with different amounts of \emph{m}: \begin{exercise} \label{R:55} \textbf{\emph{hm} interjection recognition} \\
    \texttt{define RuM ["@M" -> m*];} \end{exercise}

\paragraph{\emph{\texttt{"@N"}}} regulates the epenthesis of \emph{n} at the beginning of a form which is obligatory when the form is preceded by \emph{a, e, i, o}, and optional when preceded by \emph{u, ü} or at form boundary; see D\ref{def:23}.

\paragraph{\emph{\texttt{"@NG"}}} regulates the radical addition of \emph{ng} in some intransitive verbs when transitivized by causative \emph{-üm-} (slot 34); see \ref{sec:43}, D\ref{def:18}, R\ref{R:15}.

\paragraph{\emph{\texttt{"@NK"}}} is used to treat the negation for imperative \emph{-ki-}, it optionally takes the forms \emph{-ke-} or \emph{-k-} when followed by \emph{e}; see D\ref{def:04}, D\ref{def:09}, R\ref{R:05}, R\ref{R:10}, R\ref{R:13}.

\paragraph{\emph{\texttt{"@Ñ"}}} regulates the insertion of \emph{ñ} for experience marker \emph{-ma-} (slot 35) when preceded by vowels.

\paragraph{\emph{\texttt{"@ÑF"}}} treats the optional \emph{ñ} insertion in some forms ending in vowel as word ending or when followed by consonant.

\paragraph{\emph{\texttt{"@U"}}} indicates the insertion of a \emph{u} between a form beginning in \emph{w} and a previous consonant or semi-vowel; see \ref{it:31}.

\paragraph{\emph{\texttt{"@Ü"}}} (capital \emph{u} with diaereses) is used to treat schwa insertion represented by \emph{ü} in the spelling, this is the most common epenthesis in \emph{Mapudüngun}; see D\ref{def:05}, D\ref{def:06}, \ref{it:30}, \ref{it:31}, D\ref{def:07}, D\ref{def:14}, R\ref{R:10}, R\ref{R:12}, D\ref{def:30}.

\paragraph{\emph{\texttt{"@ÜC"}}} (capital \emph{u} with diaereses) is used to treat schwa insertion in causative suffixes \emph{-l-} and \emph{-m-} (slot 34), but also the radical consonant alternation of some intransitive verbs which take this suffix, see \ref{sec:43}, D\ref{def:19}, R\ref{R:15}.

\paragraph{\emph{\texttt{"@ÜI"}}} (capital \emph{u} with diaereses) is used to treat the optional alternative form of indicative \emph{-iy-} (slot 4) after \emph{f, l, m}; see D\ref{def:30}.

\paragraph{\emph{\texttt{"@ÜÑ"}}} (capital \emph{u} with diaereses) regulates the form of indirect object (slot 26) and adverbializer \emph{-ñma-}. They realize as \emph{-ma-} when preceded by \emph{f, l, sh}. They realize as \emph{-ñma-} when preceded by \emph{e, i, o, u}. They realize as \emph{-ñma-} or \emph{-yma-} when preceded by \emph{a, ü}. And they realize as \emph{-üñma-} when preceded by semi-vowel or \emph{ch, d, k, ll, m, n, ñ, ng, p, r, s, t, tr}; see D\ref{def:33}.

\paragraph{\emph{\texttt{"@VE"}}} treats the form of the verb \emph{entu-} 'to take out', this is explained in \ref{sec:42}, p. \pageref{sec:42}, see D\ref{def:17}, R\ref{R:14}.

\paragraph{\emph{\texttt{"@Y"}}} regulates the insertion of a \emph{y} between a form beginning in \emph{a} or \emph{e} and a previous \emph{a}, \emph{e} or boundary; see D\ref{def:08}, R\ref{R:04}, D\ref{def:10}, R\ref{R:13}.

\paragraph{\emph{\texttt{"@YY"}}} it also regulates the insertion of a \emph{y}, but in different contexts. Obligatory after vowel or form boundary, optional after semi-vowel, in both cases the following form begins in \emph{e}.

\onecolumn

\subsection{Conjugation of the intransitive verb \emph{küpa-} 'to come'} \label{anx:10}

\paragraph{} \label{tp:218} The root \emph{küpa-} is returned as \texttt{‑IV.küpa\_venir} by the analyser. To make the conjugation reading simpler, only the \emph{Mapuche} root is presented in the table. Imperative forms preceded by an * are identical to those of indicative mood, and may be used adhortatively.

\begin{table}[htb]
	\caption{Conjugation of the intransitive verb \emph{küpa-} 'to come'}
	\label{tab:13}
	\resizebox{\textwidth}{!}{\begin{tabular}{|c|c|c|c|}
		\hline\noalign{\smallskip}
		Person & Indicative & Conditional & Imperative\\
		\noalign{\smallskip}\hline\noalign{\smallskip}
		1s & \shortstack{\emph{küpa-n} \\ \texttt{‑IV.küpa+IND1SG.n3}} & \shortstack{\emph{küpa-l-i} \\ \texttt{‑IV.küpa+CND.l4+1.i3+SG.Ø2}} & \shortstack{\emph{küpa-chi} \\ \texttt{‑IV.küpa+IMP1SG.chi3}}\\
		\noalign{\smallskip}\hline\noalign{\smallskip}
		1d & \shortstack{\emph{küpa-y-u} \\ \texttt{‑IV.küpa+IND.y4+1.Ø3+DL.u2}} & \shortstack{\emph{küpa-l-i-u} \\ \texttt{‑IV.küpa+CND.l4+1.i3+DL.u2}} & \shortstack{*\emph{küpa-y-u} \\ \texttt{‑IV.küpa+IND.y4+1.Ø3+DL.u2}}\\
		\noalign{\smallskip}\hline\noalign{\smallskip}
		1p & \shortstack{\emph{küpa-y-iñ} \\ \texttt{‑IV.küpa+IND.y4+1.Ø3+PL.iñ2}} & \shortstack{\emph{küpa-l-i-iñ} \\ \texttt{‑IV.küpa+CND.l4+1.i3+PL.iñ2}} & \shortstack{*\emph{küpa-y-iñ} \\ \texttt{‑IV.küpa+IND.y4+1.Ø3+PL.iñ2}}\\
		\noalign{\smallskip}\hline\noalign{\smallskip}
		2s & \shortstack{\emph{küpa-y-m-i} \\ \texttt{‑IV.küpa+IND.y4+2.m3+SG.i2}} & \shortstack{\emph{küpa-l-m-i} \\ \texttt{‑IV.küpa+CND.l4+2.m3+SG.i2}} & \shortstack{\emph{küpa-nge} \\ \texttt{‑IV.küpa+IMP2SG.nge3}}\\
		\noalign{\smallskip}\hline\noalign{\smallskip}
		2d & \shortstack{\emph{küpa-y-m-u} \\ \texttt{‑IV.küpa+IND.y4+2.m3+DL.u2}} & \shortstack{\emph{küpa-l-m-u} \\ \texttt{‑IV.küpa+CND.l4+2.m3+DL.u2}}  & \shortstack{\emph{küpa-m-u} \\ \texttt{‑IV.küpa+IMP.Ø4+2.m3+DL.u2}}\\
		\noalign{\smallskip}\hline\noalign{\smallskip}
		2p & \shortstack{\emph{küpa-y-m-ün} \\ \texttt{‑IV.küpa+IND.y4+2.m3+PL.ün2}} & \shortstack{\emph{küpa-l-m-ün} \\ \texttt{‑IV.küpa+CND.l4+2.m3+PL.ün2}} & \shortstack{\emph{küpa-m-ün} \\ \texttt{‑IV.küpa+IMP.Ø4+2.m3+PL.ün2}}\\
		\noalign{\smallskip}\hline\noalign{\smallskip}
		3 & \shortstack{\emph{küpa-y} \\ \texttt{‑IV.küpa+IND.y4+3.Ø3}} & \shortstack{\emph{küpa-l-e} \\ \texttt{‑IV.küpa+CND.l4+3.e3}} & \shortstack{\emph{küpa-pe} \\ \texttt{‑IV.küpa+IMP3.pe3}}\\
		\noalign{\smallskip}\hline\noalign{\smallskip}
		3d & \shortstack{\emph{küpa-y-ng-u} \\ \texttt{‑IV.küpa+IND.y4+3.ng3+DL.u2}} & & \\
		\noalign{\smallskip}\hline\noalign{\smallskip}
		3p & \shortstack{\emph{küpa-y-ng-ün} \\ \texttt{‑IV.küpa+IND.y4+3.ng3+PL.ün2}} & & \\
		\noalign{\smallskip}\hline
	\end{tabular}}
\end{table}

\newpage
\subsection{Conjugation of the transitive verb \emph{pi-} 'to say (to tell)'} \label{anx:11}

\paragraph{} \label{tp:219} \texttt{‑TV.pi\_decir}. Only the \emph{Mapuche} root is shown in the table. Imperative forms with * are identical to the indicative counterpart, those marked with ** may be used adhortatively.

\begin{table}[htb]
	\caption{Conjugation of the transitive verb \emph{pi-} 'to say (to tell)'}
	\label{tab:14}
	\resizebox{\textwidth}{!}{\begin{tabular}{|c|c|c|c|}
		\hline\noalign{\smallskip}
		Persons & Indicative & Conditional & Imperative\\
		\noalign{\smallskip}\hline\noalign{\smallskip}
		1s → 2s & 
		\shortstack{\emph{pi-e-y-u} \\ \texttt{‑TV.pi+IDO.e6+IND.y4+1.Ø3+DL.u2+DS12A.Ø1}} & 
		\shortstack{\emph{pi-e-l-y-u} \\ \texttt{‑TV.pi+IDO.e6+CND.l4+1.y1+DL.u2+DS12A.Ø1}} &\\
		\noalign{\smallskip}\hline\noalign{\smallskip}
		\shortstack{1 → 2 \\ (more than 2 \\ participants)} & 
		\shortstack{\emph{pi-w-y-iñ} \\ \texttt{‑TV.pi+1A.w23+IND.y4+1.Ø3+PL.iñ2}} & 
		\shortstack{\emph{pi-w-l-i-iñ} \\ \texttt{‑TV.pi+1A.w23+CND.l4+1.i3+PL.iñ2}} &\\
		\noalign{\smallskip}\hline\noalign{\smallskip}
		2s → 1s & 
		\shortstack{\emph{pi-e-n} \\ \texttt{‑TV.pi+IDO.e6+IND1SG.n3+DS12A.Ø1}} & 
		\shortstack{\emph{pi-e-l-i} \\ \texttt{‑TV.pi+IDO.e6+CND.l4+1.i3+SG.Ø2+DS12A.Ø1}} & 
		\shortstack{*\emph{pi-e-n} \\ \texttt{‑TV.pi+IDO.e6+IND1SG.n3+DS12A.Ø1}}\\
		\noalign{\smallskip}\hline\noalign{\smallskip}
		2d/p → 1s & 
		\shortstack{\emph{pi-mu-n} \\ \texttt{‑TV.pi+2A.mu23+IND1SG.n3}} & 
		\shortstack{\emph{pi-mu-l-i} \\ \texttt{‑TV.pi+2A.mu23+CND.l4+1.i3+SG.Ø2}} & 
		\shortstack{\emph{pi-mu-chi} \\ \texttt{‑TV.pi+2A.mu23+IMP1SG.chi3}}\\
		\noalign{\smallskip}\hline\noalign{\smallskip}
		2s/d/p → 1d & 
		\shortstack{\emph{pi-mu-y-u} \\ \texttt{‑TV.pi+2A.mu23+IND.y4+1.Ø3+DL.u2}} & 
		\shortstack{\emph{pi-mu-l-i-u} \\ \texttt{‑TV.pi+2A.mu23+CND.l4+1.i3+DL.u2}} & 
		\shortstack{*\emph{pi-mu-y-u} \\ \texttt{‑TV.pi+2A.mu23+IND.y4+1.Ø3+DL.u2}}\\
		\noalign{\smallskip}\hline\noalign{\smallskip}
		2s/d/p → 1p & 
		\shortstack{\emph{pi-mu-y-iñ} \\ \texttt{‑TV.pi+2A.mu23+IND.y4+1.Ø3+PL.iñ2}} & 
		\shortstack{\emph{pi-mu-l-i-iñ} \\ \texttt{‑TV.pi+2A.mu23+CND.l4+1.i3+PL.iñ2}} & 
		\shortstack{**\emph{pi-mu-y-iñ} \\ \texttt{‑TV.pi+2A.mu23+IND.y4+1.Ø3+PL.iñ2}}\\
		\noalign{\smallskip}\hline\noalign{\smallskip}
		1s → 3 & 
		\shortstack{\emph{pi-fi-n} \\ \texttt{‑TV.pi+EDO.fi6+IND1SG.n3}} & 
		\shortstack{\emph{pi-fi-l-i} \\ \texttt{‑TV.pi+EDO.fi6+CND.l4+1.i3+SG.Ø2}} & 
		\shortstack{\emph{pi-fi-chi} \\ \texttt{‑TV.pi+EDO.fi6+IMP1SG.chi3}}\\
		\noalign{\smallskip}\hline\noalign{\smallskip}
		1d → 3 & 
		\shortstack{\emph{pi-fi-y-u} \\ \texttt{‑TV.pi+EDO.fi6+IND.y4+1.Ø3+DL.u2}} & 
		\shortstack{\emph{pi-fi-l-i-u} \\ \texttt{‑TV.pi+EDO.fi6+CND.l4+1.i3+DL.u2}} & 
		\shortstack{**\emph{pi-fi-y-u} \\ \texttt{‑TV.pi+EDO.fi6+IND.y4+1.Ø3+DL.u2}}\\
		\noalign{\smallskip}\hline\noalign{\smallskip}
		1p → 3 & 
		\shortstack{\emph{pi-fi-y-iñ} \\ \texttt{‑TV.pi+EDO.fi6+IND.y4+1.Ø3+PL.iñ2}} & 
		\shortstack{\emph{pi-fi-l-i-iñ} \\ \texttt{‑TV.pi+EDO.fi6+CND.l4+1.i3+PL.iñ2}} & 
		\shortstack{*\emph{pi-fi-y-iñ} \\ \texttt{‑TV.pi+EDO.fi6+IND.y4+1.Ø3+PL.iñ2}}\\
		\noalign{\smallskip}\hline\noalign{\smallskip}
		2s → 3 & 
		\shortstack{\emph{pi-fi-y-m-i} \\ \texttt{‑TV.pi+EDO.fi6+IND.y4+2.m3+SG.i2}} & 
		\shortstack{\emph{pi-fi-l-m-i} \\ \texttt{‑TV.pi+EDO.fi6+CND.l4+2.m3+SG.i2}} & 
		\shortstack{\emph{pi-fi-nge} \\ \texttt{‑TV.pi+EDO.fi6+IMP2SG.nge3}}\\
		\noalign{\smallskip}\hline\noalign{\smallskip}
		2d → 3 & 
		\shortstack{\emph{pi-fi-y-m-u} \\ \texttt{‑TV.pi+EDO.fi6+IND.y4+2.m3+DL.u2}} & 
		\shortstack{\emph{pi-fi-l-m-u} \\ \texttt{‑TV.pi+EDO.fi6+CND.l4+2.m3+DL.u2}} & 
		\shortstack{\emph{pi-fi-m-u} \\ \texttt{‑TV.pi+EDO.fi6+IMP.Ø4+2.m3+DL.u2}}\\
		\noalign{\smallskip}\hline\noalign{\smallskip}
		2p → 3 & 
		\shortstack{\emph{pi-fi-y-m-ün} \\ \texttt{‑TV.pi+EDO.fi6+IND.y4+2.m3+PL.ün2}} & 
		\shortstack{\emph{pi-fi-l-m-ün} \\ \texttt{‑TV.pi+EDO.fi6+CND.l4+2.m3+PL.ün2}} & 
		\shortstack{\emph{pi-fi-m-ün} \\ \texttt{‑TV.pi+EDO.fi6+IMP.Ø4+2.m3+PL.ün2}}\\
		\noalign{\smallskip}\hline\noalign{\smallskip}
		3 → 3 & 
		\shortstack{\emph{pi-fi-y} \\ \texttt{‑TV.pi+EDO.fi6+IND.y4+3.Ø3}} & 
		\shortstack{\emph{pi-fi-l-e} \\ \texttt{‑TV.pi+EDO.fi6+CND.l4+3.e3}} & 
		\shortstack{\emph{pi-fi-pe} \\ \texttt{‑TV.pi+EDO.fi6+IMP3.pe3}}\\
		\noalign{\smallskip}\hline\noalign{\smallskip}
		3d → 3 & 
		\shortstack{\emph{pi-fi-y-ng-u} \\ \texttt{‑TV.pi+EDO.fi6+IND.y4+3.ng3+DL.u2}} &  &\\
		\noalign{\smallskip}\hline\noalign{\smallskip}
		3p → 3 & 
		\shortstack{\emph{pi-fi-y-ng-ün} \\ \texttt{‑TV.pi+EDO.fi6+IND.y4+3.ng3+PL.ün2}} &  &\\
		\noalign{\smallskip}\hline\noalign{\smallskip}
		3 → 1s & 
		\shortstack{\emph{pi-e-n-ew} \\ \texttt{‑TV.pi+IDO.e6+IND1SG.n3+DS3A.ew1}} & 
		\shortstack{\emph{pi-e-l-i-mew} \\ \texttt{‑TV.pi+IDO.e6+CND.l4+1.i3+SG.Ø2+DS3A.mew1}} & 
		\shortstack{\emph{pi-e-chi-mew} \\ \texttt{‑TV.pi+IDO.e6+IMP1SG.chi3+DS3A.mew1}}\\
		\noalign{\smallskip}\hline\noalign{\smallskip}
		3 → 1d & 
		\shortstack{\emph{pi-e-y-u-mew} \\ \texttt{‑TV.pi+IDO.e6+IND.y4+3.Ø3+DL.u2+DS3A.mew1}} & 
		\shortstack{\emph{pi-e-l-y-u-mew} \\ \texttt{‑TV.pi+IDO.e6+CND.l4+1.y3+DL.u2+DS3A.mew1}} &\\
		\noalign{\smallskip}\hline\noalign{\smallskip}
		3 → 1p & 
		\shortstack{\emph{pi-e-y-iñ-mew} \\ \texttt{‑TV.pi+IDO.e6+IND.y4+1.Ø3+PL.iñ2+DS3A.mew1}} & \shortstack{\emph{pi-e-l-y-iñ-mew} \\ \texttt{‑TV.pi+IDO.e6+CND.l4+1.y3+PL.iñ2+DS3A.mew1}} &\\
		\noalign{\smallskip}\hline\noalign{\smallskip}
		3 → 2s & 
		\shortstack{\emph{pi-e-y-mew} \\ \texttt{‑TV.pi+IDO.e6+IND.y4+2.m3+SG.Ø2+DS3A.ew1}} & 
		\shortstack{\emph{pi-e-l-mew} \\ \texttt{‑TV.pi+IDO.e6+CND.l4+2.m3+SG.Ø2+DS3A.ew1}} &\\
		\noalign{\smallskip}\hline\noalign{\smallskip}
		3 → 2d & 
		\shortstack{\emph{pi-e-y-m-u-mew} \\ \texttt{‑TV.pi+IDO.e6+IND.y4+2.m3+DL.u2+DS3A.mew1}} & 
		\shortstack{\emph{pi-e-l-m-u-mew} \\ \texttt{‑TV.pi+IDO.e6+CND.l4+2.m3+DL.u2+DS3A.mew1}} &\\
		\noalign{\smallskip}\hline\noalign{\smallskip}
		3 → 2p & 
		\shortstack{\emph{pi-e-y-m-ün-mew} \\ \texttt{‑TV.pi+IDO.e6+IND.y4+2.m3+PL.ün2+DS3A.mew1}} & 
		\shortstack{\emph{pi-e-l-m-ün-mew} \\ \texttt{‑TV.pi+IDO.e6+CND.l4+2.m3+PL.ün2+DS3A.mew1}} &\\ 
		\noalign{\smallskip}\hline\noalign{\smallskip}
		3 → 3 & 
		\shortstack{\emph{pi-e-y-ew} \\ \texttt{‑TV.pi+IDO.e6+IND.y4+3.Ø3+DS3A.ew1}} & 
		\shortstack{\emph{pi-e-l-y-ew} \\ \texttt{‑TV.pi+IDO.e6+CND.l4+3.y3+DS3A.ew1}} &\\
		\noalign{\smallskip}\hline
	\end{tabular}}
\end{table}

\newpage
\subsection{Negative imperative forms of the transitive verb \emph{pi-} 'to say (to tell)'} \label{anx:12}

\paragraph{} \label{tp:220} Negative imperative forms are infrequent. Note that a negative command can also be expressed by a negative indicative form which is marked with the suffix \emph{-a-} \texttt{+NRLD.a9} for non-realized situation, as shown below.

\begin{example} \label{ex:277} [Smeets, I. 2008: 173 (82)] \cite{RefB:21}\\
	\emph{amu-la-ya-y-m-i} 'you must not go' (lit.: 'you will not go')\\
	\texttt{‑IV.amu\_ir+NEG.la10+NRLD.a9+IND.y4+2.m3+SG.i2}
\end{example}

\begin{table}[htb]
	\caption{Negative imperative forms of the transitive verb \emph{pi-} 'to say (to tell)'}
	\label{tab:15}
	\resizebox{\textwidth}{!}{\begin{tabular}{|c|c|c|}
		\hline\noalign{\smallskip}
		Persons & Negative imperative & Translation\\
		\noalign{\smallskip}\hline\noalign{\smallskip}
		2s → 1s & \shortstack{\emph{pi-ki-e-l-i} \\ \texttt{‑TV.pi+NEG.ki10+IDO.e6+CNI.l4+1.i3+SG.Ø2+DS12A.Ø1}} & Don't (you/s) tell me\\
		\noalign{\smallskip}\hline\noalign{\smallskip}
		2d/p → 1s & \shortstack{\emph{pi-mu-ki-l-chi} \\ \texttt{‑TV.pi+2A.mu23+NEG.ki10+CNI.l4+IMP1SG.chi3}} & Don't (you/d/p) tell me\\
		\noalign{\smallskip}\hline\noalign{\smallskip}
		2s/d/p → 1d & \shortstack{\emph{pi-mu-ki-l-y-u} \\ \texttt{‑TV.pi+2A.mu23+NEG.ki10+CNI.l4+1.y3+DL.u2}} & Don't (you/s/d/p) tell us/d\\
		\noalign{\smallskip}\hline\noalign{\smallskip}
		2s/d/p → 1p & \shortstack{\emph{pi-mu-ki-l-y-iñ} \\ \texttt{‑TV.pi+2A.mu23+NEG.ki10+CNI.l4+1.y3+PL.iñ2}} & Don't (you/s/d/p) tell us/p\\
		\noalign{\smallskip}\hline\noalign{\smallskip}
		1s → 3 & \shortstack{\emph{pi-ki-fi-l-chi} \\ \texttt{‑TV.pi+NEG.ki10+EDO.fi6+CNI.l4+IMP1SG.chi3}} & I don't tell him/her/them\\
		\noalign{\smallskip}\hline\noalign{\smallskip}
		1d → 3 & \shortstack{\emph{pi-ki-fi-l-y-u} \\ \texttt{‑TV.pi+NEG.ki10+EDO.fi6+CNI.l4+1.y3+DL.u2}} & Let us/d not tell him/her/them\\
		\noalign{\smallskip}\hline\noalign{\smallskip}
		1p → 3 & \shortstack{\emph{pi-ki-fi-l-y-iñ} \\ \texttt{‑TV.pi+NEG.ki10+EDO.fi6+CNI.l4+1.y3+PL.iñ2}} & Let us/p not tell him/her/them\\
		\noalign{\smallskip}\hline\noalign{\smallskip}
		2s → 3 & \shortstack{\emph{pi-ki-fi-l-nge} \\ \texttt{‑TV.pi+NEG.ki10+EDO.fi6+CNI.l4+IMP2SG.nge3}} & Don't (you/s) tell him/her/them\\
		\noalign{\smallskip}\hline\noalign{\smallskip}
		2d → 3 & \shortstack{\emph{pi-ki-fi-l-m-u} \\ \texttt{‑TV.pi+NEG.ki10+EDO.fi6+CNI.l4+2.m3+DL.u2}} & Don't (you/d) tell him/her/them\\
		\noalign{\smallskip}\hline\noalign{\smallskip}
		2p → 3 & \shortstack{\emph{pi-ki-fi-l-m-ün} \\ \texttt{‑TV.pi+NEG.ki10+EDO.fi6+CNI.l4+2.m3+PL.ün2}} & Don't (you/p) tell him/her/them\\
		\noalign{\smallskip}\hline\noalign{\smallskip}
		3 → 1s & \shortstack{\emph{pi-ki-e-l-chi-mu} \\ \texttt{-TV.pi+NEG.ki10+IDO.e6+CNI.l4+IMP1SG.chi3+DS3A.mew1}} & May he/she/they not tell me\\
		\noalign{\smallskip}\hline
	\end{tabular}}
\end{table}

\subsection{Flow chart diagrams of the FST analyser} \label{anx:13}

\begin{thebibliography}{}

\bibitem{RefB:01}
Beesley, K. and Karttunen, L., Finite State Morphology. CSLI Studies in Computational Linguistics. CSLI Publications, Stanford, U. S. A. (2003).

\bibitem{RefB:02}
Castro, R. and Rios, A., Allin Qillqay! A Free Online Web spell checking Service for Quechua, in Ugaz Burga, J. E., Gonzales Sánchez, S. R., and Torres Guerra, C., editors, Memoria - VI Congreso Internacional de Computación y Telecomunicaciones (COMTEL) 2014, Fondo Editorial de la Universidad Inca Garcilaso de la Vega, Lima, Peru, pages 23–30 (2014).

\bibitem{RefB:03}
Chandía, A. et al. CORLEXIM. Corpus lexicográfico del mapudüngun. <\href{http://corlexim.cl}{http://corlexim.cl}> [April, 2021]

\bibitem{RefB:04}
Chrupala, G. et al., Learning Morphology with Morfette. In
Calzolari, N., Choukri, K., Maegaard, B., Mariani, J., Odijk, J.,
Piperidis, S., and Tapias, D., editors, Proceedings of the Sixth International Conference on Language Resources and Evaluation (LREC
2008), European Language Resources Association (ELRA). Marrakech, Morocco, (2008).

\bibitem{RefB:05}
Corporación Nacional de Desarrollo Indígena. Gobierno de Chile, Temuco, Chile (2005).

\bibitem{RefB:06}
Febrés, A., Diccionario Araucano-Español ó sea Calepino Chileno-Hispano Por el P. Andrés Febrés de la Compañía de Jesus. Reproducido textualmente de la edición de Lima de 1765, por Juan M. Larsen. Con un Apéndice sobre las lenguas Quíchua, Aimará y Pampa. Buenos Aires: Juan A. Alsina (1882).

\bibitem{RefB:07}
Fernández-Garay, A. and Malvestitti, M., Formas no finitas del Mapudüngun en dos variedades de la Argentina. In IX Congreso de la Sociedad Argentina Lingüística: 1–14. Universidad Nacional de Córdoba. Córdoba, Argentina, (2002).

\bibitem{RefB:08}
Gasser, M. Computational Morphology and the Teaching of Indigenous Languages. Proceedings of the First Symposium on Teaching Indigenous Languages of Latin America. CLACS \& MLCP, Indiana University Bloomington \& Association for Teaching and Learning Indigenous Languages of Latin America (ATLILLA), (2011).

\bibitem{RefB:09}
Guevara, T. Las últimas familias i costumbres araucanas. Imprenta, litografía i encuadernación "Barcelona". Snatiago, Chile, (1913).

\bibitem{RefB:10}
Hulden, M., Fast approximate string matching with finite automata. Procesamiento del Lenguaje Natural, núm. 43. ISSN 1135-5948, pp. 57-64, (2009).

\bibitem{RefB:11}
Jurafsky, D., Definition of Minimum Edit Distance. CS 124: From Languages to Information. Stanford University. California, U. S. A. (2012).

\bibitem{RefB:12}
Lonkon, E., Morfología y Aspectos del Mapudüngun. Universidad Autónoma Metropolitana. Unidad Iztapalapa. México D. F., México, (2011).

\bibitem{RefB:13}
Lonkon, E., Políticas públicas de lengua y cultura aplicada al mapudüngun. El pueblo mapuche en el siglo XXI. Propuestas para un nuevo entendimiento entre culturas en Chile. Centro de Estudios Públicos. Santiago, Chile (2017).

\bibitem{RefB:14}
Mösbach, E., Vida y costubres de los indígena araucanos en la segunda mitad del siglo XIX. Imprenta Universitaria. Santiago, Chile (1936).

\bibitem{RefB:15}
Ragileo, A., Gramática del idioma Mapuche. Public domain, (1982).

\bibitem{RefB:16}
Ríos, A., Spell checking an agglutinative language: Quechua. In: 5th Language and Technology Conference: Human Language Technologies as a Challenge for Computer Science and Linguistics, Poznań, Poland, 25 November 2011 - 27 November 2011, 51-55, (2011).

\bibitem{RefB:17}
Ríos, A., A Basic Language Technology Toolkit for Quechua. Faculty of Arts of the University of Zurich. Zurich, Switzerland, (2015).

\bibitem{RefB:18}
Sadowsky, S. et al., Illustrations of the IPA: Mapudüngun. Journal of the International Phonetic Association 43(1). 87–96. doi: 10.1017/S0025100312000369, (2013).

\bibitem{RefB:19}
Salas, A., El mapuche o araucano. Centro de Estudios Públicos, 2nd edition. Santiago, Chile, (2006).

\bibitem{RefB:20}
Schmid, H. and Laws, F., Estimation of conditional probabilities with decision trees and an application to fine-grained POS tagging. In Proceedings of the 22 nd International Conference on Computational Linguistics, volume 1 of COLING ’08, pages 777–784. Association for Computational Linguistics. Stroudsburg, U. S. A., (2008).

\bibitem{RefB:21}
Smeets, I., A Grammar of Mapuche. Mouton de Gruyter. Berlin, Germany. New York, U. S. A., (2008).

\bibitem{RefB:22}
Sochil (Sociedad Chilena de Lingüística). Encuentro para la Unificación del Alfabeto Mapuche. Proposiciones y Acuerdos. Arturo Hernández Sallés, Coordinador del Encuentro. Temuco: Pontificia Universidad Católica de Chile, (1986).

\bibitem{RefB:23}
Sochil (Sociedad Chilena de Lingüística). Alfabeto Mapuche Unificado. Temuco: Pontificia Universidad Católica de Chile, (1988).

\bibitem{RefB:24}
Zúñiga, F., Mapudüngun. El habla mapuche. Centro de Estudios Públicos. Santiago, Chile, (2006).
\end{thebibliography}
\end{document}